\documentclass[10pt]{article} 
\PassOptionsToPackage{dvipsnames}{xcolor}
\usepackage[preprint]{tmlr}

\usepackage{amsmath,amsfonts,bm}

\def\eqref#1{equation~\ref{#1}}

\def\1{\bm{1}}

\DeclareMathAlphabet{\mathsfit}{\encodingdefault}{\sfdefault}{m}{sl}
\SetMathAlphabet{\mathsfit}{bold}{\encodingdefault}{\sfdefault}{bx}{n}

\usepackage{url}
\usepackage{graphicx}

\usepackage{hyperref}
\usepackage{cleveref}
\usepackage{tocloft}

\crefname{paracount}{paragraph}{paragraphs}
\Crefname{paracount}{Paragraph}{Paragraphs}

\usepackage{caption}
\title{Open Problems in Mechanistic Interpretability}

\author{\name Lee Sharkey$^*$ \aff Apollo Research
      \AND
      \name Bilal Chughtai$^*$ \aff Apollo Research
      \AND
      \name Joshua Batson \aff Anthropic
      \AND
      \name Jack Lindsey \aff Anthropic
      \AND
      \name Jeff Wu \aff Anthropic$^\dagger$
      \AND
      \name Lucius Bushnaq \aff Apollo Research
      \AND
      \name Nicholas Goldowsky-Dill \aff Apollo Research
      \AND
      \name Stefan Heimersheim \aff Apollo Research
      \AND
      \name Alejandro Ortega \aff Apollo Research
      \AND
      \name Joseph Bloom \aff Decode Research
      \AND
      \name Stella Biderman \aff Eleuther AI
      \AND
      \name Adria Garriga-Alonso \aff FAR AI
      \AND
      \name Arthur Conmy \aff Google DeepMind
      \AND
      \name Neel Nanda \aff Google DeepMind
      \AND
      \name Jessica Rumbelow \aff Leap Laboratories
      \AND
      \name Martin Wattenberg \aff Harvard University
      \AND
      \name Nandi Schoots \aff King's College London and Imperial College London
      \AND
      \name Joseph Miller \aff MATS
      \AND
      \name Eric J. Michaud \aff MIT
      \AND
      \name Stephen Casper \aff MIT
      \AND
      \name Max Tegmark \aff MIT
      \AND
      \name William Saunders \aff METR
      \AND
      \name David Bau \aff Northeastern University
      \AND
      \name Eric Todd \aff Northeastern University
      \AND
      \name Atticus Geiger \aff Pr(AI)$^2$r group
      \AND
      \name Mor Geva \aff Tel Aviv University
      \AND
      \name Jesse Hoogland \aff Timaeus
      \AND
      \name Daniel Murfet \aff University of Melbourne
      \AND
      \name Tom McGrath \aff Goodfire
      }
      
      \def\month{MM}  
      \def\year{YYYY} 
      \def\openreview{\url{https://openreview.net/forum?id=XXXX}} 
      
\footnotetext{$^*$Correspondence to: lee@apolloresearch.ai and brchughtaii@gmail.com}
\footnotetext{$^\dagger$Work done prior to joining Anthropic.}

\begin{document}

\maketitle

\newpage

\vspace*{\fill}
\begin{center}
\begin{abstract}
Mechanistic interpretability aims to understand the computational mechanisms underlying neural networks' capabilities in order to accomplish concrete scientific and engineering goals. Progress in this field thus promises to provide greater assurance over AI system behavior and shed light on exciting scientific questions about the nature of intelligence. Despite recent progress toward these goals, there are many open problems in the field that require solutions before many scientific and practical benefits can be realized: Our methods require both conceptual and practical improvements to reveal deeper insights; we must figure out how best to apply our methods in pursuit of specific goals; and the field must grapple with socio-technical challenges that influence and are influenced by our work. This forward-facing review discusses the current frontier of mechanistic interpretability and the open problems that the field may benefit from prioritizing. 
\end{abstract}
\end{center}
\vspace*{\fill}

\textit{This review collects the perspectives of its various authors and represents a synthesis of their views by Apollo Research on behalf of Schmidt Sciences. The perspectives presented here do not necessarily reflect the views of any individual author or the institutions with which they are affiliated.}

\newpage
\tableofcontents
\newpage

\section{Introduction}
\label{sec:introduction}

Recent progress in artificial intelligence (AI) has resulted in rapidly improved AI capabilities. These capabilities are not designed by humans. Instead, they are learned by deep neural networks \citep{Hinton_2006_EarlyDNN, LeCun_2015_DeepLearning}. Developers only need to design the training process; they do not need to -- and in almost all cases, do not -- understand the neural mechanisms underlying the capabilities learned by an AI system.

Although human understanding of these mechanisms is not necessary for AI capabilities, understanding them would enhance several \textit{human} abilities. For example, it would permit better human control over AI behavior and better monitoring during deployment. It would also facilitate trust in AI systems, allowing us to fully realize their potential benefits by enabling their deployment in safety-critical and ethically-sensitive settings.

Beyond the engineering benefits, understanding AI systems offers immense scientific opportunities. For the first time in history, we can create and study artificial minds with a level of access and control that is simply not possible in biological systems. What new laws of nature governing the mechanisms of minds might we discover from studying the internal workings of AI systems?

The scientific opportunities are not limited to the field of AI. If an AI can outperform tools designed by humans in a given scientific field, it suggests that the AI system is representing something about the world currently unknown to us. We can develop a deeper comprehension of the world by understanding those representations. What can we learn about protein folding from AIs that can successfully predict protein structure? What insights can we glean about disease from a radiographer that performs beyond human ability?

\textit{Mechanistic interpretability} might unlock these benefits. This field of study aims to understand neural networks' decision-making processes. Here, we define “Understanding a neural network's decision-making process” as the ability to use knowledge about the mechanisms underlying a network's decision-making process in order to successfully predict its behavior (even on arbitrary inputs) or to accomplish other practical goals with respect to the network. Such goals might include more precise control of the network's behavior, or improved network design. Interpretability promises greater assurances for AI systems through a better understanding of what neural networks have learned, thus enabling us to realize their potential benefits.

\subsection{The focus of this review: Open problems and the future of mechanistic interpretability}
\label{subsec:review-focus}

Several recent reviews of mechanistic interpretability research and related topics exist \citep{rauker2023transparentaisurveyinterpreting,Geiger_2022_CausalStructure,Bereska_2024_MechInterpReview, ferrando2024primerinnerworkingstransformerbased,rai2024practicalreviewmechanisticinterpretability,anwar2024foundationalchallengesassuringalignment,davies2024cognitiverevolutioninterpretabilityexplaining,mosbach2024insightsactionsimpactinterpretability, mueller2024questrightmediatorhistory}. Our review takes a more forward-looking stance. We discuss not only where the frontier is today, but also which directions we might benefit most from prioritizing in the future.

\subsubsection{Why `mechanistic' interpretability?}
\label{subsec:mechanistic}

The distinction between interpretability and mechanistic interpretability is not always clear and is therefore worth clarifying.
The motivations and methods used in interpretability work are often diverse \citep{Lipton_2016_InterpretabilityMotivations,doshivelez2017rigorousscienceinterpretablemachine,Jacovi_2023_ExplainableAiLiterature}. As a result, there are many ways in which interpretability research might be categorized. Prior categorizations of interpretability include causal vs. correlational methods, supervised vs. unsupervised methods, bottom-up vs. top-down methods, among others \citep{geiger2021causalabstractionsneuralnetworks, mueller2024questrightmediatorhistory,Bereska_2024_MechInterpReview,Belinkov_2022_ProbingClassifiers,zou2023representationengineeringtopdownapproach, davies2024cognitiverevolutioninterpretabilityexplaining}. This review focuses specifically on \textit{mechanistic} interpretability. But what distinguishes `mechanistic interpretability' from interpretability in general? It has been noted that the term is used in a number of (sometimes inconsistent) ways \citep{saphra2024mechanistic}. In this review, we use the term `mechanistic interpretability' in a technical sense, referring specifically to work that investigates the mechanisms underlying neural network generalization. Mechanistic interpretability represents one of three threads of interpretability research, each with distinct but sometimes overlapping motivations, which roughly reflects the changing aims of interpretability work over time.

The first thread aims to build AI systems that are interpretable by design. Much early interpretability work focused on explaining the sensitivity of machine learning models to inputs and training data. This work typically used small-to-medium sized models designed to be easily interpretable, such as decision trees \citep{Breiman_1984_Classification_Regression_Trees, Hu_2019_OptimalSparseDecisionTrees}, linear models \citep{Roweis_1999_LinearModel}, and generalized additive models \citep{Hastie_1986_AdditiveModel,Agarwal_2020_Additive_Model}. These models could be used alongside attribution methods such as influence functions \citep{Hampel_1986_Influence_Function,Koh_2017_influence_function} and Shapley values \citep{Shapley_1997_Shapley_Values, Lundberg_2017_SHAP}, which were common techniques used to characterize model decision boundaries with respect to inputs. “Interpretability by design” continues to be an active research area, including architectures such as Concept-Bottleneck Models \citep{Koh_2020_ConceptBottleneckModels}, Backpack Language Models \citep{Hewitt_2023_BackpackLanguageModels},
Kolmogorov-Arnold Networks \citep{Liu_2024_KolmogorovArtificialNetwork}, and sparse decision trees \citep{Xin_2022_SparseDecisionTrees}.

With the rise of larger-scale nonlinear neural networks \citep{Krizhevsky_2012_ImageNet, He_2015_ImageDNN}, another thread grew in importance, driven primarily by the question: Why did my model make this particular decision? However, one challenge in interpreting larger networks was finding attribution methods that could scale to large networks \citep{Zeiler_2013_VisualizingConvolutionalNetworks}. In response, a number of local attribution methods were developed, including grad-CAM \citep{Selvaraju_2019_GradCAM}, integrated gradients \citep{Sundarajan_2017_IntegratedGradients}, and masking-based causal attribution \citep{Fong_2017_MaskingBasedCausalAttribution}, SHAP \citep{Lundberg_2017_SHAP}, LIME \citep{Ribiero_2016_LIME}, and many other methods, including backprop-based visualization methods \citep{Simonyan_2014_BackpropVisualisation, Nguyen_2016_FeatureVisualization}.

Inspired by, for example, Inception \citep{szegedy2015inception} and GPT-3 \citep{Brown_2020_GPT3}, another thread emerged as models became capable of more profound generalization. Focused on the broader subject of generalization, it was driven by the question: How did my model solve this general class of problems? Due to its emphasis on the mechanisms underlying neural network generalization, the work in this category is commonly referred to as `mechanistic interpretability' (in addition to other, more cultural reasons, according to \citep{saphra2024mechanistic}). This kind of interpretability work is driven by a fundamental hypothesis in deep learning that generalization arises from shared computation \citep{LeCun_2015_DeepLearning}. Early work in this area, such as feature visualization \citep{olah2017feature}, or network dissection \citep{Bau_2017_NetworkDissection}, sought “global” explanations for model generalization by investigating the roles of model components across a class of decisions. More recent work in this area looks at “circuits” of components \citep{Wang_2022_Circuits}, generalizable patterns of information flow \citep{Geva_2023_InformationFlow}, representation subspaces
\citep{geiger2024findingalignmentsinterpretablecausal,zou2023representationengineeringtopdownapproach} and probes (or self-supervised searches via sparse dictionary learning) for representation vectors that carry information that generalizes across many instances for a particular task or set of tasks 
\citep{cunningham2023sparseautoencodershighlyinterpretable,Bricken_2023_dictionary,Todd_2024_Probe,Tigges_2023_Probe}.

\subsection{Types of open problems}
\label{subsec:types-of-open-problems}

The field of mechanistic interpretability ultimately aims to achieve concrete scientific and engineering goals. For instance, we would like to be able to:
\begin{itemize}
    \item Monitor AI systems for signs of cognition related to dangerous behavior (\Cref{subsec:monitoring-and-auditing});
    \item Modify internal mechanisms and edit model parameters to adapt their behavior to better suit our needs (\Cref{subsec:better-control});
    \item Predict how models will act in unseen situations or predict when a model might learn specific abilities (\Cref{subsec:better-predictions});
    \item Improve model inference, training, and mechanisms to better suit our preferences (\Cref{subsec:capabilities});
    \item Extract latent knowledge from AI systems so we can better model the world (\Cref{subsec:microscope-ai}).
\end{itemize}

Despite recent hopeful signs of progress, mechanistic interpretability still has considerable distance to cover before achieving satisfactory progress toward most of its scientific and engineering goals.

To achieve these goals, the field not only needs greater application of current state-of-the-art mechanistic interpretability methods, but also requires the development of improved techniques. The first major section (\Cref{sec:open-problems-methods-and-foundations}) therefore discusses open problems related to the methods and foundations of mechanistic interpretability.

We then explore key axes of research progress that will determine how far we can advance toward the goals of mechanistic interpretability (\Cref{subsec:axes}). \Cref{sec:open-problems-applications} outlines how applications of interpretability methods have made progress toward the field's goals, and, for each goal, discusses the specific axes along which progress is likely needed to achieve specific objectives.

Finally, we note that the goals, applications, and methods of mechanistic interpretability do not exist in a vacuum. Like any scientific field, they lie within a broader societal context. The final section of this review examines open socio-technical problems in mechanistic interpretability (\Cref{sec:open-problems-sociotechnical}). It discusses current initiatives and possible pathways to translate technical progress into levers for AI governance, alongside consequential social and philosophical challenges faced by the field.

\section{Open problems in mechanistic interpretability methods and foundations}
\label{sec:open-problems-methods-and-foundations}

One way of thinking about what neural networks do internally is that they learn parameters that implement neural algorithms. These neural algorithms take data as input and, through a series of steps, transform their internal activations to produce an output. Different parts of the network learn different steps in the algorithm; mechanistic interpretability aims to describe the function of different network components. 

Broadly speaking, there are two methodological approaches to achieving this. The first approach, often called `reverse engineering', is to decompose the network into components and then attempt to identify the function of those components (\Cref{subsec:reverse-engineering}). This approach “identifies the roles of network components”. Conversely, the second approach, sometimes referred to as `concept-based interpretability', proposes a set of concepts that might be used by the network and then looks for components that appear to correspond to those concepts (\Cref{subsec:concept-based-interpretability}). This approach thus “identifies network components for given roles”.

\begin{figure}[!t]
    \centering
    \includegraphics[width=\textwidth]{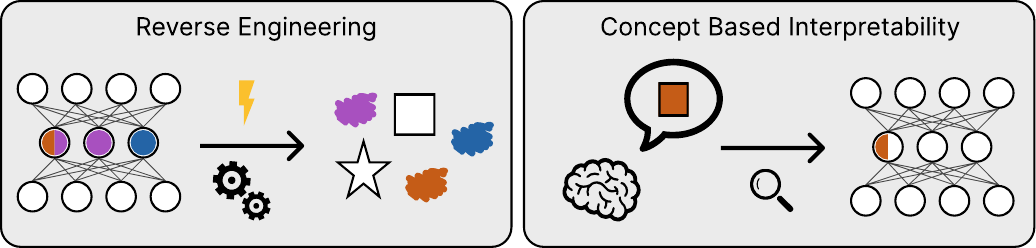}
    \caption{Two approaches to neural network interpretability. (Left) Reverse Engineering is characterized by decomposing networks into functional components and describing how those components interact to produce the network's behavior. It thus aims to `identify the roles of network components' (\Cref{subsec:reverse-engineering}). (Right) Concept-based interpretability on the other hand attempts to discover human concepts within neural network internals. It thus aims to `identify the network components for given roles' (\Cref{subsec:concept-based-interpretability}). 
    }
    \label{fig:reverse_engineering_vs_concept_based}
\end{figure}

In this section, we will examine both approaches (`Reverse engineering' and `Concept-based interpretability') (\Cref{fig:reverse_engineering_vs_concept_based}), discussing their methods and open problems. We will also touch on open problems that cut across either approach, including proceduralizing the mechanistic interpretability pipeline (\Cref{subsec:proceduralizing}) and its uses in automating interpretability research (\Cref{subsec:automation}).

\subsection{Reverse engineering: Identifying the roles of network components}
\label{subsec:reverse-engineering}

\subsubsection{Reverse engineering is necessary because AI and humans use different representations and perform different tasks}
\label{subsubsec:reverse-engineering-necessary}

Large language models produce text that closely resembles human writing, and it is tempting to assume that they generate it through cognitive processes similar to those of human writers\footnote{However, even if this assumption were true, understanding LLMs would remain challenging, as we also lack a mechanistic understanding of the cognitive processes involved in human text creation!}. However, humans and AI models often solve problems in different ways. For example, a model that is only 1\% the size of GPT-3 outperforms humans on next-token prediction tasks \citep{shlegeris2024languagemodelsbetterhumans}. Inversely, even state-of-the-art multimodal LLMs struggle with tasks that a four-year-old could easily master, such as learning causal properties of new objects involving simple lights and shapes \citep{kosoy2023comparingmachineschildrenusing}. An even clearer sign that humans and AI are using different representational processes are cases where humans cannot solve a problem at all, as in the case of predicting a protein structure from sequence \citep{jumper2021highly,lin2023evolutionary}.

Even when both humans and AI exhibit comparable levels of competence on a given task, they may use different heuristics. For example, research shows that image models tend to rely more heavily on textural features (such as recognizing elephants by their hide rather than their shape \citep{geirhos2022imagenettrainedcnnsbiasedtexture}) or rely on dataset correlations (as when identifying fish by the fingers that proud fisherman use to hold them \citep{brendel2019approximatingcnnsbagoflocalfeaturesmodels}) to a greater degree than people do. Even simple algorithmic tasks, like modular addition, which humans might solve with simple carries, were solved by a small transformer model by learning a Fourier transform strategy that researchers only understood in retrospect \citep{nanda2023progressmeasuresgrokkingmechanistic}.

To grasp the potentially alien cognition of these models, we must develop methods to uncover and understand the previously unknown concepts and mechanisms implemented within them. In other words, we must be able to reverse engineer these models \citep{Olah_2022}.

\begin{figure}[!t]
    \centering
    \includegraphics[width=\textwidth]{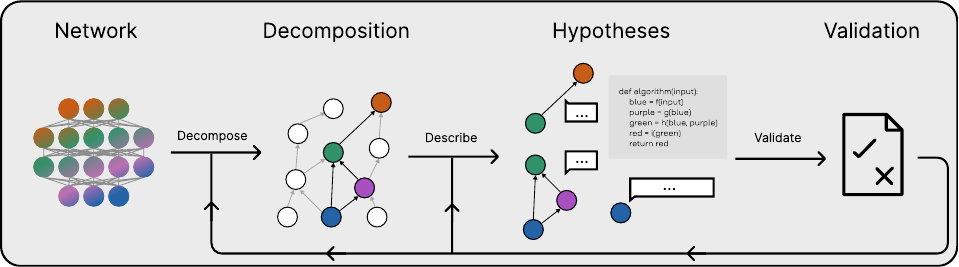}
    \caption{The steps of reverse engineering neural networks. (1) Decomposing a network into simpler components. This decomposition might not necessarily use architecturally-defined bases, such as individual neurons or layers (\Cref{subsec:reverse-engineering-step-1}). (2) Hypothesizing about the functional roles of some or all components (\Cref{subsec:reverse-engineering-step-2}). (3) Validating whether our hypotheses are correct, creating a cycle in which we iteratively refine our decompositions and hypotheses to improve our understanding of the network (\Cref{subsec:reverse-engineering-step-3}).
    }
    \label{fig:pipeline}
\end{figure}

Reverse engineering generally involves three steps, whether it is an engine, a piece of software, or a neural network (\Cref{fig:pipeline}):
\begin{enumerate}
    \item \textbf{Decomposition}: Breaking down the object of study into simpler parts (\Cref{subsec:reverse-engineering-step-1});
    \item \textbf{Description of components}: Formulating hypotheses about the functional role of component parts and how they interact (a process that can be called `interpretation') (\Cref{subsec:reverse-engineering-step-2});
    \item \textbf{Validation of descriptions}: Testing if our hypotheses are correct (\Cref{subsec:reverse-engineering-step-3}).
\end{enumerate}

If our hypotheses are invalidated, we must either improve our decomposition or improve our hypotheses regarding the functional roles of its components. We will systematically examine each step, analyzing current methods and their respective shortcomings.

\subsubsection{Reverse engineering step 1: Neural network decomposition}
\label{subsec:reverse-engineering-step-1}

In mechanistic interpretability, our aim is to decompose a neural network and study its parts in isolation in order to explain how neural networks generalize. This aim raises the question of how best to carve a neural network “at its joints” for the purposes of interpretability.

\parasection{Networks do not naturally decompose into architectural components.}
\label{par:decomposition}

The naive approach to decompose neural networks involves breaking them down into their architectural components, such as individual neurons, attention heads, or layers.

Some early attempts to interpret deep artificial neural networks studied the responses or weight structure of individual neurons or single convolutional filters \citep{Erhan_2009_visualising,le2012buildinghighlevelfeaturesusing,Krizhevsky_2012_ImageNet,szegedy2014intriguingpropertiesneuralnetworks,simonyan2014deepinsideconvolutionalnetworks, zhou2015objectdetectorsemergedeep,srivastava_2014,yosinski2015understandingneuralnetworksdeep,Mordvintsev_2015,Olah__2017_FeatureVisualisation,Bau_2017_NetworkDissection,dalvi2018grainsanddesertanalyzing, olah2020an,cammarata2020curve}. These efforts paid homage to the `Neuron Doctrine' in neuroscience, which posits that individual neurons are the structural and functional unit of the nervous system \citep{cajal1924estructura,sherrington1906observations}. However, researchers discovered that single neurons are `polysemantic' -- they seem to respond to multiple kinds of features in both artificial \citep{wei2015understandingintraclassknowledgeinside,nguyen2016multifacetedfeaturevisualizationuncovering,Olah__2017_FeatureVisualisation} and biological networks \citep{churchland2007temporal, rigotti2013importance,mante2013context,raposo2014category}. These observations support earlier theoretical work that suggested representations used by neural networks do not necessarily align with the activation of individual neurons \citep{hinton1981shape}.

Interpreting individual attention heads does not fare better than interpreting individual neurons, as attention heads also exhibit polysemanticity \citep{janiak2023polysemantic}. More broadly, research suggests that studying the attention patterns of models can often be misleading \citep{jain2019attentionexplanation, pruthi2020learningdeceiveattentionbasedexplanations}.

Some work even suggests that representations in language models might span multiple layers \citep{yun2023transformervisualizationdictionarylearning, lindsey2024crosscoders}. This chimes with work that edits or intervenes on individual layers, which indicates that this level is too coarse-grained to robustly carve the network at its joints \citep{meng2022locating,Wang_2022_Circuits}.

If natural architectural components, such as individual neurons, attention heads, or layers, do not provide a natural way to decompose neural network representations, then what does?

\parasection{Decomposition by dimensionality reduction methods.}
\label{par:dim-reduction}

If individual neurons are not the right decomposition, perhaps groups or patterns of neurons are. Many decomposition methods attempt to identify activation vectors that correspond to the basic unit of neural network computation. One common approach is to provide models with a range of unlabeled inputs, collect the resulting hidden activations, and then apply unsupervised dimensionality reduction techniques to these hidden activations. The hope is that structure in the hidden activations corresponds to the structure of neural computation. Commonly used dimensionality reduction methods include Principal Component Analysis or Singular Value Decomposition \citep{Tigges_2023_Probe,marks2024geometrytruthemergentlinear,huang2024ravelevaluatinginterpretabilitymethods,bushnaq2024localinteractionbasisidentifying} and non-negative matrix factorization \citep{olah2018the,voss2021visualizing,cammarata2020curve}, though these techniques are no longer predominant methods used for mechanistically decomposing language models \citep{friedman2024interpretabilityillusionsgeneralizationsimplified}.

\parasection{Decomposition by sparse dictionary learning (SDL).}
\label{par:sdl}

\begin{figure}[!t]
    \centering
    \includegraphics[width=0.8\textwidth]{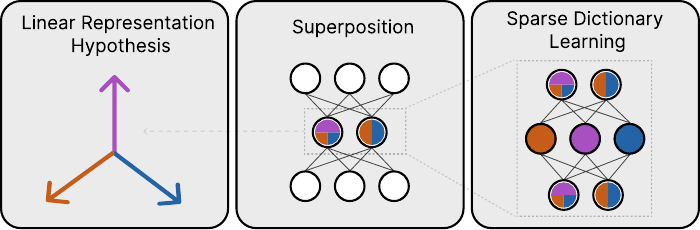}
    \caption{Three ideas underlying the sparse dictionary learning (SDL) paradigm in mechanistic interpretability. (Left) The linear representation hypothesis states that the map from `concepts' to neural activations is linear. (Middle) Superposition is the hypothesis that models represent many more concepts than they have dimensions by representing them both sparsely and linearly in activation spaces. (Right) SDL attempts to recover an overcomplete basis of concepts represented in superposition in activation space.
    }
    \label{fig:sdl}
\end{figure}

According to the `superposition hypothesis', neural networks are capable of representing more features than they have dimensions, as long as each feature activates sparsely \citep{elhage2021mathematical} (\Cref{fig:sdl}). This is a key reason that dimensionality reduction methods are not considered state-of-the-art, because they cannot identify more directions than there are activation dimensions. The superposition hypothesis, coupled with the failure of dimensionality reduction methods to overcome it, motivated the search for methods that can identify more directions than dimensions. Recent work has explored the use of sparse dictionary learning (SDL) to this end \citep{elhage2021mathematical,Sharkey_2022,cunningham2023sparseautoencodershighlyinterpretable,Bricken_2023_dictionary}.

Currently, SDL is the most popular set of unsupervised decomposition methods in mechanistic interpretability. SDL encapsulates a family of methods, including Sparse Autoencoders (SAEs) \citep{gao2024scalingevaluatingsparseautoencoders,templeton2024scaling, rajamanoharan2024improvingdictionarylearninggated, makelov2024principledevaluationssparseautoencoders, kissane2024interpretingattentionlayeroutputs, braun2024identifyingfunctionallyimportantfeatures}, Transcoders \citep{dunefsky2024transcodersinterpretablellmfeature}, and Crosscoders \citep{lindsey2024crosscoders}.

In SDL, hidden activations are typically passed to a small neural network consisting of only two layers, which correspond to an encoder and decoder respectively, with a wide hidden space. The encoder activations represent how active each `latent' \footnote{The term latent is often used instead of the word feature to refer to SDL dictionary elements, since the term `feature' is often used to refer to multiple different ideas \citep{smith2024latent}} is, and the decoder matrix corresponds to a dictionary of latent directions. We want to train the dictionary elements to align with `feature directions' in the network's hidden activations. Since we assume that individual features are sparsely present in the activations, the encoder activations are trained to be sparsely activating. In the case of SAEs, the output is trained to either reconstruct the input or, in the case of transcoders \citep{dunefsky2024transcodersinterpretablellmfeature}, to reconstruct the activations of the next layer. Crosscoders \citep{lindsey2024crosscoders} permit a wider class of inputs and outputs, potentially reconstructing the activations of many layers simultaneously. Since the encoder is nonlinear, it is thought to be able to learn to activate a latent only if a feature is `active' in the hidden activations and remain `off' if it is not.

\begin{figure}[!t]
    \centering
    \includegraphics[width=\textwidth]{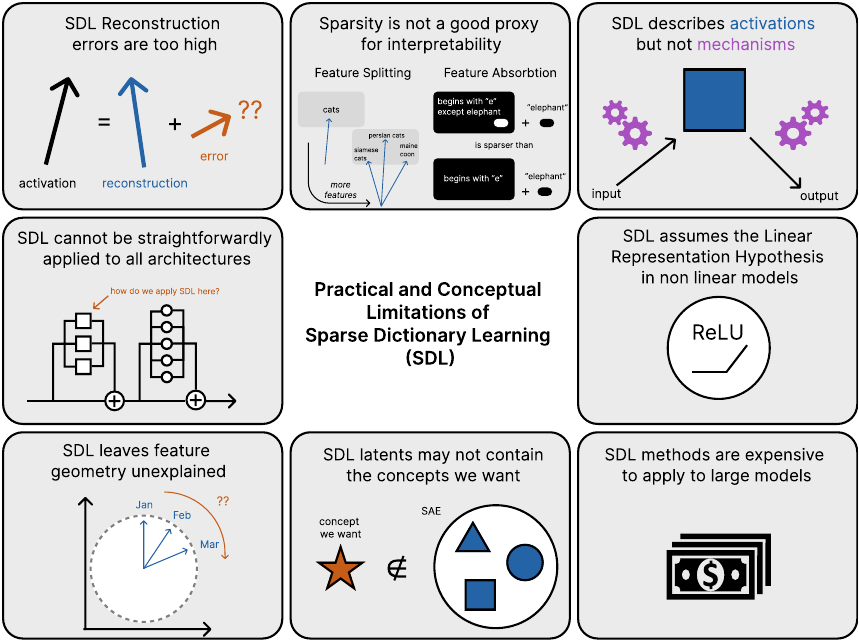}
    \caption{Sparse dictionary learning has a number of practical and conceptual limitations that cause issues when using it to reverse engineer neural networks (\Cref{par:sdl}).}
    \label{fig:sdl-problems}
\end{figure}

Although SDL is considered a leading decomposition method for mechanistic interpretability, it has substantial practical and conceptual limitations (\Cref{fig:sdl-problems}).

\textbf{SDL reconstruction errors are too high}: Large errors in SDL reconstruction raise the question whether SDL methods can reconstruct the hidden representation sufficiently well such that the latents learned by SDL are adequately faithful to the models being interpreted. To measure this, the model's true hidden activations can be replaced with sparse dictionary reconstructions, then subsequently evaluating the extent to which the model's performance decreases. In practice, the error results in significant performance reductions. When a sparse dictionary with 16 million latents was inserted into GPT-4, the language modeling loss was equivalent to a model with only 10\% of GPT-4's pretraining compute \citep{gao2024scalingevaluatingsparseautoencoders}. Similarly, \citet{makelov2024principledevaluationssparseautoencoders} found that using reconstructions from sparse autoencoders decreased GPT-2 small performance by 10\% when trained on task-specific data, and 40\% when trained on the full distribution.

Making sparse dictionaries much larger and sparser to reduce errors is a feasible but computationally expensive approach. Furthermore, in the limit, this results in merely assigning one dictionary latent per datapoint, which is clearly less interpretable. One partial solution is using SDL methods with `error nodes' \citep{marks2024sparsefeaturecircuitsdiscovering}, to account for the discrepancies between the original and reconstructed activations. However, while the sparse autoencoders identified interpretable latents, the error nodes contain `everything else', making them an inadequate solution to the problem. \citet{engels2024decomposingdarkmattersparse} found that these reconstruction errors are not purely random, as much of the direction of the error and its norm can be linearly predicted from the initial activation vector. This suggests that current SDL methods may systematically fail to capture certain structured aspects of model representations, but also implies potential solutions. To overcome this issue, it may be necessary to improve SDL training methods, or develop entirely new methods of network decomposition.

\textbf{SDL methods are expensive to apply to large models}: SDL involves training a small neural network for every layer of the AI model that we want to interpret. Typically, the sparse dictionaries have more parameters at that layer than the original model does, and consequently will probably be relatively expensive to train compared to the original model. \footnote{The actual relative cost is unclear since there are no public attempts to apply SDL to every vector space in a model, although some work applies SDL to various layers \citep{marks2024sparsefeaturecircuitsdiscovering,gao2024scalingevaluatingsparseautoencoders,braun2024identifyingfunctionallyimportantfeatures,lieberum2024gemmascopeopensparse, Bloom_Lin_2024,cunningham2023sparseautoencodershighlyinterpretable}} As AI models become larger, scaling costs of SDL also increase, although it remains unclear whether relative scaling costs are sub- or supra-linear. For cost effectiveness, it may be important to develop intrinsically decomposable methods for training models, while remaining at (or close to) state-of-the-art performance (\Cref{par:intrinsic-interpretability}). This approach will help avoid incurring the expense of both training and decomposing AI models.

\textbf{SDL assumes the linear representation hypothesis in nonlinear models}: Other problems with SDL arise from the assumptions on which SDL is based. One such assumption is the linear representation hypothesis. The linear representation hypothesis observes that though neural networks are nonlinear functions and could potentially use highly nonlinear representations \citep{black2022interpretingneuralnetworkspolytope,engels2024decomposingdarkmattersparse, kirch2024featurespromptsjailbreakllms}, they tend to use representations that exhibit strikingly linear behavior \citep{Smolensky_1986, mikolov-etal-2013-linguistic,olah2020zoom,elhage2021mathematical,park2024linearrepresentationhypothesisgeometry, guerner2024geometricnotioncausalprobing,gurnee2024languagemodelsrepresentspace}. The hypothesis formalizes this phenomenon by stating that high level concepts are linearly represented as directions (vectors) in neural network embeddings. Its two core claims are (i) that the composition of multiple concepts can be represented as the addition of their corresponding feature vectors, and (ii) that the intensity of a concept is represented by the scale of its corresponding feature vector \citep{olah2024linear}. Earlier work \citep{elhage2021mathematical} defined the linear representation hypothesis with the additional assumption of one-dimensional features, but this definition has since been refuted \citep{Yedidia_2023,chughtai2024understanding, engels2024languagemodelfeatureslinear} and clarified \citep{olah2024linear}. Another recently proposed criterion is that the intensity of concepts should be retrievable by a linear function of the embeddings, up to a small error \citep{hani2024mathematicalmodels, olah2024linear}.

A weak version of the linear representation hypothesis states that some concepts are linearly represented, while a strong version may assert that all concepts are \citep{Smith_2024_strongfeature}. Some works have shown that the strong version is false for some models \citep{black2022interpretingneuralnetworkspolytope, csordás2024recurrentneuralnetworkslearn}. The weaker version of the linear representation hypothesis is supported by the successes of linear probes (\Cref{subsubsec:concept-based-probes}), activation steering (\Cref{subsec:better-control}), and the success of sparse autoencoders in finding seemingly interpretable latents (\Cref{par:sdl}).

\textbf{Sparsity is not a good proxy for interpretability}: Another of SDL's key assumptions is that feature activations are sparse. Therefore, SDL methods optimize their latents accordingly to be sparsely activating, with the implicit assumption that sparser decompositions are more interpretable than denser ones. However, this assumption may not necessarily hold. The (related) problems of feature splitting \citep{Bricken_2023_dictionary}, feature absorption \citep{chanin2024absorptionstudyingfeaturesplitting} and composition \citep{till2024sparse} suggest that with sufficient optimization pressure, sparsity as a proxy for interpretability breaks down (though note that it is debated whether feature splitting is actually a problem). Other proxies, such as minimum description length, might be better optimization targets than sparsity \textit{per se} \citep{ayonrinde2024interpretabilitycompressionreconsideringsae}. It also may be the case that no proxy metric is sufficient.

\textbf{SDL leaves feature geometry unexplained}: SDL decomposes networks into single directions in the activation space, which is a reasonable approach only if we consider feature activations akin to `a bag of features' \citep{Harris_1954_distributionalstructure} without any internal structure. However, the geometric arrangement of features in relation to each other seems to reflect semantic and functional structure \citep{engels2024languagemodelfeatureslinear,gurnee2024languagemodelsrepresentspace, park2024geometrycategoricalhierarchicalconcepts,bussman2024metasaes}. Understanding the geometric structure of a network means understanding the position of one feature relative to (potentially) many others. Comprehending this may be necessary if we want to know why and how a network treats certain features similarly or differently. If understanding the global geometry (all-to-all relationships) of features is essential to understand neural networks, this might pose a fundamental problem for current approaches to mechanistic interpretability \citep{hani2024mathematicalmodels, mendel2024sae}. However, if only local geometric relationships between features need to be understood, understanding networks with a `bag of features' approach may be more feasible.

\textbf{SDL cannot straightforwardly be applied to all architectures}: SDL was originally developed to identify features that may be represented in superposition across many neurons within a single layer. However, representations may be spread over other network architecture components besides neurons. In transformers, for example, representations may be spread across separate attention heads \citep{jermyn2023attention, janiak2023polysemantic}, and even across different layers \citep{yun2023transformervisualizationdictionarylearning, lindsey2024crosscoders}. However, it is not immediately obvious how to decompose representations distributed across attention heads with SDL \citep{mathwin2024gated, wynroe2024decomposing}. Nor is it straightforward to translate cross-layer distributed representations into causal descriptions of neural network mechanisms \citep{lindsey2024crosscoders}.

\textbf{SDL decomposes the input and output activations of network mechanisms, but not the mechanisms themselves}: Ultimately, the aim of mechanistic interpretability is to understand the mechanisms learned by neural networks. The parameters of the network, along with its nonlinearities and other architectural components, implement these mechanisms, which are applied to the input's hidden activations. SDL identifies directions in activation space. Being activations, they only interact with the network's mechanisms, but are not the mechanisms themselves. Describing the mechanisms directly remains unresolved with SDL. Gaining insights about the network's mechanisms from SDL latents requires further post hoc analysis, which can be labor intensive, computationally expensive, or data set dependent \citep{cunningham2023sparseautoencodershighlyinterpretable,Bricken_2023_dictionary, riggs2023finding,marks2024sparsefeaturecircuitsdiscovering}. This is an instance of a more broad problem with current mechanistic interpretability work; we primarily focus on understanding neural network activations, with little attention paid to how this structure in activations is computed via weights \citep{chughtai2025activation}.

\textbf{SDL latents may not contain the concepts needed for downstream use cases}: When using SAEs for practical tasks, there sometimes exists a single latent or small set of latents representing some concept of interest for task performance (and not representing much else). For instance, \citet{kantamneni2024sae} found a single latent whose activation pattern was more accurate than official dataset labels on the NLP task GLUE CoLA \citep{warstadt2019neuralnetworkacceptabilityjudgments}. More often than not though, a sparse set of latents that encode some useful concept of interest do not exist. It is unclear what causes this problem. One hypothesis is that the concept we want isn't how the model `thinks' about the concept, and the SAE is working as intended. Alternatively, the SAE training distribution could be too narrow, resulting in the SAE not being incentivised to learn the important latents. \citep{kissane2024saes} found that SAEs trained on pretraining data generally do not have good latents for the concept of `refusing' harmful user requests, while SAEs trained on chat formatted data do. Or, the SAE might not have a large enough dictionary size to learn all concepts of interest. Many more hypotheses are plausible. A complicating factor in using SDL to identify the learned mechanisms of neural networks is that the latents identified by depend on the data set used to train them. This is an undesirable property for a decomposition method that was initially hoped to be capable of identifying the fundamental units of computation in neural networks \citep{kissane2024saes}.

\parasection{Current decomposition methods lack solid theoretical foundations.}
\label{par:theoretical-foundations}

Given the practical and conceptual issues with SDL, there is broad agreement that the question of how to correctly decompose networks into atomic units remains a central problem, evidenced by the large amount of effort focused on the direction in recent years. After investing considerable effort in SDL approaches, it is apparent that improved conceptual clarity beyond the idea of superposition \citep{elhage2021mathematical} is needed to advance neural network decomposition.

One of the most significant open questions is the absence of clarity around the nature of features, despite being the central focus of SDL's identification efforts. Satisfying formal definitions are elusive and conceptual foundations are not yet established. However, even without foundations, progress in mechanistic interpretability is possible — even confused concepts can be pragmatically useful \citep{henighan2024caloric}.

Without solid conceptual foundations, it remains unclear whether the superposition hypothesis, which underpins the SDL paradigm, is fundamentally valid or merely pragmatically useful \citep{henighan2024caloric,templeton2024scaling}. If it is the latter, there may be better methods than SDL for carving neural networks at their joints. Such methods may take into account feature geometry or take a more dynamic, developmental view of how mechanistic structure emerges in the training process \citep{hoogland2024developmentallandscapeincontextlearning, wang2024loss}. Moreover, although there is much emphasis on how models represent features in superposition, there is comparatively less work examining how they might perform computation on them natively in superposition. Further conceptual work into this problem may suggest new methodologies for decomposing networks, or provide bounds on the number of features we should expect models to be capable of learning \citep{hani2024mathematicalmodels, adler2024complexityneuralcomputationsuperposition, bushnaq2024circuits}.

Mechanistic interpretability should ideally be built on more formal foundations. Some work attempts to ground mechanistic interpretability in the formalisms of causality \citep{Geiger_2023_CausalAbstractionRepresentationSubspace}. However, this approach has not yet yielded canonical causal mediators \citep{mueller2024questrightmediatorhistory} that may form a basis for decomposition methods. How should we go about finding them? Given that our field's objective is to understand the learned structures that underlie networks' generalization behaviors, exploring theories about why neural networks generalize appears to be a promising avenue. But theories that attempt to characterize why neural networks generalize, such as the spline theory of neural networks \citep{balestriero_2018_spline}, theories of neural networks' simplicity bias \citep{vallepérez2019deeplearninggeneralizesparameterfunction}, deep learning theory involving the neural tangent kernel \citep{jacot2020neuraltangentkernelconvergence,Roberts_2022_principles}, or singular learning theory \citep{Watanabe_2009, Wei_2023_deeplearningsingular} have either not yet yielded mathematical objects that can be easily used for interpretability, or simply have not been successfully linked to approaches for interpreting neural networks. Establishing these connections would make significant progress toward carving neural networks at their joints to facilitate mechanistic interpretability. And if we can carve trained networks at their joints, it may suggest ways to train networks such that they come `pre-carved'. Thus, better theoretical foundations may also be important for developing models that are intrinsically decomposable by design, which we discuss next.

\parasection{Intrinsic interpretability: Building more easily decomposable models}
\label{par:intrinsic-interpretability}

The current strategy of training a model solely for performance and then interpreting it post hoc may not be optimal if our goal is a model that is both interpretable and performant. To this end, it may be beneficial to prioritize interpretability during model training, for which there are several plausible approaches. 

Instead of post-hoc decomposing trained network activations into discrete codes (\Cref{par:sdl}), network activations could be forced to use a discrete code from the outset, as in \citet{tamkin2023codebookfeaturessparsediscrete}. MLPs could also be trained with sparser activation functions, such as TopK \citep{makhzani2013ksparseautoencoders,bills2023language} or SoLU \citep{elhage2022solu}. Similar approaches could be potentially used to restrict attention superposition \citep{jermyn2023attention} by limiting the number of heads attending to any query-key pair. Several approaches, such as `mixture of experts' \citep{shazeer2017outrageouslylargeneuralnetworks,switchtransformers, he2024mixturemillionexperts}, use sparsely activating components — with a large enough number of experts, and with sufficient activation sparsity, experts may become individually interpretable \citep{warstadt2019neuralnetworkacceptabilityjudgments}. Many attempts to incentivize interpretable activations directly so far have not been competitively performant, and have also allowed `superposition to sneak through', mitigating benefits.

We can also target weight sparsity directly during training, including approaches such as L0 regularization \citep{louizos2018learningsparseneuralnetworks} or pruning \citep{Mozer_1988_skeletonization,frankle2019lotterytickethypothesisfinding, mocanu2018scalable}. \citet{Han_2015_weightsandconnections} achieve sparse weights by using magnitude pruning followed by finetuning. This highlights a general strategy of finetuning with interpretability in mind (also used by \citet{tamkin2023codebookfeaturessparsediscrete}), avoiding the potentially excessive cost of training from scratch. Another related strategy is targeting modularity \citep{Kirsch_2018_modularnetworks, andreas2017neuralmodulenetworks}. For example, brain-inspired modular training \citep{liu2023seeingbelievingbraininspiredmodular} trains for modularity by embedding neurons in a geometric space and encouraging geometrically local connections.

Existing research implements other approaches that can simplify linearity-reliant circuit analysis \citep{elhage2021mathematical}. For example, there is work removing the layer norm operations \citep{heimersheim2024removegpt2slayernormfinetuning}, using input-switched affine transformations for recurrence \citep{foerster2017inputswitchedaffinenetworks}. Other studies leverage architectures that are mathematically analyzable in other ways than linearity, such bilinear activations in MLPs \citep{sharkey2023technicalnotebilinearlayers, pearce2024weightbaseddecompositioncasebilinear}.

After decomposing a network into components, whether through post hoc decomposition or by using intrinsically decomposable models, our task remains unfinished. We must provide an interpretation of the functional role of each component (Step 2 -- \Cref{subsec:reverse-engineering-step-2}) and validate that interpretation (Step 3 -- \Cref{subsec:reverse-engineering-step-3}). In the next section, we will discuss common methods of interpretation and their shortcomings.

\subsubsection{Reverse engineering step 2: Describing the functional role of components}
\label{subsec:reverse-engineering-step-2}

After decomposing networks into parts, the next step of reverse engineering is to “describe” the functional role of these components. This step is best thought of as generating hypothesized `interpretations' or `explanations'. These explanations form candidate descriptions of the functional role of a given component, and should not be taken to be definitive conclusions before they are thoroughly validated (see \Cref{subsec:reverse-engineering-step-3}).

\begin{figure}[h]
    \centering
    \includegraphics[width=0.6\textwidth]{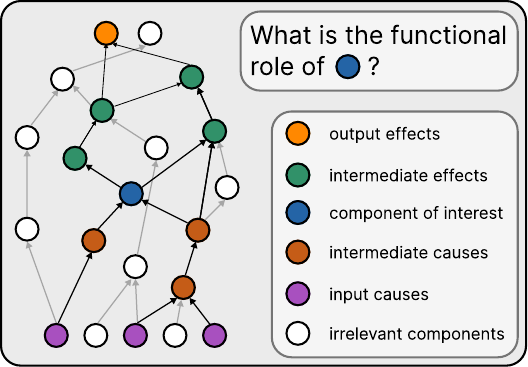}
    \caption{To study the functional role of \textcolor{NavyBlue}{the blue component} numerous approaches are possible. We could study its causes: \textcolor{Purple}{the purple input components} or \textcolor{Bittersweet}{the red intermediate components} via e.g. feature synthesis \citep{Olah__2017_FeatureVisualisation, olah2020zoom}, maximum activating examples \citep{Olah__2017_FeatureVisualisation, Bricken_2023_dictionary}, or attributions \citep{Sundarajan_2017_IntegratedGradients}. Or we could study its effects: \textcolor{orange}{the orange output components} or \textcolor{OliveGreen}{the green intermediate components} via e.g. the logit lens \citep{Nostalgebraist_2020}, activation steering \citep{turner2024steeringlanguagemodelsactivation}, or attributions \citep{marks2024sparsefeaturecircuitsdiscovering}.}
    \label{fig:descriptions}
\end{figure}

Descriptions of the functional role of neural network components can either indicate (1) the cause of a component's activation or (2) what occurs after that component has been activated, or -- preferably -- both (\Cref{fig:descriptions}). In this section, we will discuss the existing set of tools available to mechanistic interpretability researchers to describe the functional role of network components.

\parasection{Explanations for what causes components to activate}
\label{par:activation-causes}

Explanations of the causes of component activation can use three broad categories of methods, each with several problems: (1) Highly activating data set examples; (2) Attribution methods; and (3) Feature synthesis.

\textbf{Highly activating data set examples}. The simplest method is to use highly activating data set examples (sometimes called `exemplar representations' \citep{hernandez2022natural}). These are inputs on which a particular component is strongly activated. For a given component, analyzing apparent commonalities in the inputs suggests hypotheses for what causes that component to activate. This step may be carried out by humans or AI systems (\Cref{subsec:automation}).

Despite being widely used, this approach has several substantial issues. The first issue is that the method relies on human prior beliefs, which may lead interpreters to project their human understanding onto models that may, in fact, be using unfamiliar concepts. This bias could lead us to identify concepts in the model that do not truly explain the mode's functioning \citep{Freiesleben_2023_weneedtotalk, Donnelly_2019_sentimentneuron, GALE_2020_ObjectDetectors}.

Another issue with this approach is the potential for `interpretability illusions'. \citet{bolukbasi2021interpretabilityillusionbert} show that bias in data sets can create misleading explanations even when top activating examples are selected from real data sets, as opposed to synthetically created data. Depending on the data set from which the examples were drawn, human annotators identified dramatically different meanings for given directions in the activation space of BERT \citep{devlin2019bertpretrainingdeepbidirectional}.

A third issue is that this approach often yields plausible explanations for arbitrarily chosen directions in the activation space \citep{szegedy2014intriguingpropertiesneuralnetworks}. This means that plausible explanations based on highly activating data set examples cannot be solely relied upon to identify the basic units of computations in neural networks — other methods are needed to accurately identify them. Furthermore, it is possible to develop adversarial models that deliberately yield misleading feature visualizations \citep{geirhos2024donttrusteyesunreliability}.

Many of the issues with highly activating data set examples stem from the fact that they merely provide correlational explanations for the activation of a network component, rather than causal explanations. To identify causal explanations, other methods, such as attribution methods, are necessary.

\textbf{Attribution methods are necessary for causal explanations but are often difficult to interpret}. Attribution methods \citep{Simonyan_2014_BackpropVisualisation, Nguyen_2016_FeatureVisualization, Selvaraju_2019_GradCAM, Sundarajan_2017_IntegratedGradients, Fong_2017_MaskingBasedCausalAttribution, Lundberg_2017_SHAP,Ribiero_2016_LIME} are intended to measure the causal importance of upstream variables (such as inputs) on downstream variables (such as a network component). They are often gradient-based \citep{Mozer_1988_skeletonization, Simonyan_2014_BackpropVisualisation, Nguyen_2016_FeatureVisualization, Selvaraju_2019_GradCAM, Sundarajan_2017_IntegratedGradients, wang2024gradientbasedfeatureattribution} or sampling-, perturbation-, or ablation-based \citep{Fong_2017_MaskingBasedCausalAttribution, vig2020causalmediationanalysisinterpreting, geiger2020neuralnaturallanguageinference, Ghorbani_2020,meng2022locating, Chan_Garriga-alonso_Goldowsky-Dill_2022,NeelNanda_2023a}. However, on a theoretical level, many gradient-based methods identify only a first-order approximation of the ideal attribution, which is sometimes a poor approximation \citep{watson2022conceptual}. \citet{adebayo2020sanitycheckssaliencymaps} revealed more practical implications, demonstrating that some gradient-based methods identify attributions that are independent both of the model and of the data generating process. Furthermore, an adversary can train a model or perturb an input to reveal any attribution map \citep{Dombrowski_2019_advances, Ghorbani_Abid_Zou_2019, Heo_2019_Advances, Kindermans2019saliencymethods, Slack_2020_foolinglime, zhang2019interpretabledeeplearning}\footnote{However, \citet{Freiesleben_2023_weneedtotalk} argue that adversarial examples do not truly undermine saliency maps as these are highly dissimilar to real interpretability challenges.}. Perturbation methods present certain issues, such as taking models off their training distribution and eliciting unusual behavior, among other theoretical complications \citep{feng-etal-2018-pathologies, molnar2021generalpitfallsmodelagnosticinterpretation, hooker2021unrestrictedpermutationforcesextrapolation, molnar2024model, Freiesleben_2023_weneedtotalk, Slac_2021_advances}. Developing efficient and accurate attribution methods thus remains an open problem.

\textbf{Feature synthesis}. Feature synthesis is a strategy that combines highly activating data set examples and gradient-based attribution methods 
\citep{Erhan_2009_visualising, szegedy2014intriguingpropertiesneuralnetworks,Olah__2017_FeatureVisualisation}. This approach attempts to synthesize inputs that maximize the activation of a component subject to some regularization, such as consistency with a generative model \citep{nguyen2016synthesizingpreferredinputsneurons, nguyen2017plugplaygenerative} or total variation distance \citep{mahendran2014understandingdeepimagerepresentations}. However, criticisms of feature synthesis methods suggest that natural dataset examples may serve interpretation better \citep{zimmermann2021how, borowski2021exemplarynaturalimagesexplain} or show that current methods struggle to identify trojans \citep{casper2023redteamingdeepneural}.

\parasection{Explanations for the downstream effects of components}
\label{par:downstream-effects}

Alternatively, the functional role of a component can be described through its downstream effects.

\textbf{Studying the direct effect}. The logit lens \citep{Nostalgebraist_2020} applies the models unembedding matrix to an intermediate residual stream representations, converting it into a distribution over the model's output vocabulary. Direct logit attribution is a generalization of this technique, applying to any appropriately sized vector space in the model, e.g. the output of MLP layers \citep{geva2021transformerfeedforwardlayerskeyvalue,geva2022transformerfeedforwardlayersbuild,geva2022lmdebuggerinteractivetoolinspection,dar-etal-2023-analyzing}, attention blocks, gradients \citep{katz2024backwardlensprojectinglanguage} of these, and SDL decoder weights \citep{Bricken_2023_dictionary}. Adding a trainable affine or linear transformation before unembedding, as also referred to as the tuned lens, improves decoding accuracy \citep{belrose2023elicitinglatentpredictionstransformers,yom-din-etal-2024-jump}, at the cost of less faithfully representing when the model has completed computation.

In the language of causality \citep{Pearl_2009}, unembedding a residual stream vector measures the direct effect of that vector on the output \citep{mcgrath2023hydraeffectemergentselfrepair}. However, the logit lens cannot measure the indirect effect, the effects resulting from the influence that the embedding has on the hidden activations of subsequent layers. Other methods, such as causal interventions (see below), are necessary to measure the indirect effect. In the future, it may be possible to extend logit-lens-like approaches to not only project the effects of network components on directions in output vocabulary space, but also on intermediate downstream components.

\textbf{Causal Interventions}. Causal interventions typically substitute (“patch”) the value of some network component, usually an activation vector, with a different value during a forward pass, observing the resulting effect on the model. There exist a related set of techniques: ablation (for the special case of zero-ing or otherwise attempting to delete activations entirely), activation patching, causal mediation analysis, causal tracing, and interchange intervention \citep{vig2020causalmediationanalysisinterpreting, geiger2020neuralnaturallanguageinference, meng2023locatingeditingfactualassociations, Chan_Garriga-alonso_Goldowsky-Dill_2022, NeelNanda_2023a}. \footnote{For instance, by patching activations from the corrupt prompt “the capital of Italy is” into the clean prompt “the capital of France is”, we can observe the effect on the output (“Rome” vs.“Paris”). This tells us which component values are relevant for the differing output between the two prompts, but not the information that remains consistent (e.g. the fact that the answer is a city).}

More surgical patches are also sometimes also made on edges between network components. This approach, known as “path” patching, allows us to isolate the effect of one particular component on only one other component, instead of on the entire rest of the network \citep{goldowskydill2023localizingmodelbehaviorpath}. Causal scrubbing \citep{Chan_Garriga-alonso_Goldowsky-Dill_2022} is a generalization of path patching that allows for testing hypotheses concerning any given connection between a set of network components.

Causal intervention methods can also generate supervision signal to identify subspaces of interest. For instance, distributed alignment search \citep{geiger2024findingalignmentsinterpretablecausal} learns a linear subspace that represents a particular concept, using data with interventions on that concept as supervision \citep{geiger2024findingalignmentsinterpretablecausal, guerner2024geometricnotioncausalprobing, wu2024interpretabilityscaleidentifyingcausal}. Other work learns masks over network components to remove irrelevant components \citep{de-cao-etal-2020-decisions, csordás2021neuralnetsmodularinspecting, davies2023discoveringvariablebindingcircuitry}.

A causal intervention typically requires a forward pass of the model. This may make performing one for every network component in large models, long contexts, or when using finer-grained components such as sparse autoencoder latents prohibitively expensive. Faster alternatives that work well in practice \citep{marks2024sparsefeaturecircuitsdiscovering,templeton2024scaling} include gradient-based approximations to activation patching, such as attribution patching \citep{NeelNanda_2023a, syed2023attributionpatchingoutperformsautomated}, AtP* \citep{kramár2024atpefficientscalablemethod}, and integrated gradients \citep{Sundarajan_2017_IntegratedGradients}.

\textbf{Observing the effects of components on sequential behavior}. Another way to study the effects of model components is to patch in activated components and observe their effect on model behavior. Steering is one example of this \citep{panickssery2024steeringllama2contrastive,turner2024steeringlanguagemodelsactivation}, where components (activation vectors) are activated in order to influence the network's behavior, often in interpretable ways. To determine the functional role of an activation vector, a related method is to have a language model itself decode the activations \citep{chen2024selfieselfinterpretationlargelanguage, Mwatkins_2023, ghandeharioun2024patchscopesunifyingframeworkinspecting, huang2024inversionviewgeneralpurposemethodreading, kharlapenko2024self}. These approaches, known as “patchscopes”, patch activations from one forward pass of a model into a different forward pass (perhaps in a different model). The context of the new forward pass is designed to elicit relevant information from the activations of the original forward pass.

A related sequential behavior based technique is simply to read the chain-of-thought \citep{wei2023chainofthoughtpromptingelicitsreasoning, kojima2023largelanguagemodelszeroshot} produced by a language model's output. While this could be considered an `interpretability technique' in the sense that it aims to explain model decisions, it does not use model internals in those explanations, at least not directly. Recent research demonstrates that chains of thought are not entirely faithful to the model's underlying decision-making process \citep{agarwal2024faithfulnessvsplausibilityunreliability, atanasova2023faithfulnesstestsnaturallanguage, Turpin2023_cot, lanham2023measuringfaithfulnesschainofthoughtreasoning, Ye_2022_unreliability}. A promising future direction for interpretability research may be to incorporate model internals into chain-of-thought training, which could incentivize faithfulness. Another possibility is building monitors based on model internals to improve transparency in chain-of-thought faithfulness.

\subsubsection{Reverse engineering step 3: Validation of descriptions}
\label{subsec:reverse-engineering-step-3}

Initial descriptions of network components' functional roles should be treated as hypotheses that first require validation to ensure that they are reasonable.

Conflating hypotheses with conclusions has regrettably been commonplace in mechanistic interpretability research, making validation an important area for the field to improve \citep{madsen2024interpretabilityneedsnewparadigm, stander2024grokkinggroupmultiplicationcosets}. Unfortunately, it is often hard to distinguish faithful explanations of neural network components from merely plausible ones. Numerous examples of model interpretations fail sanity checks \citep{adebayo2020sanitycheckssaliencymaps, leavitt2020falsifiableinterpretabilityresearch, miller2024transformercircuitfaithfulnessmetrics}; “interpretability illusions” in which seemingly convincing interpretations of a model later turned out to be false \citep{bolukbasi2021interpretabilityillusionbert,makelov2023subspacelookingforinterpretability}; or instances where different approaches to explaining the same phenomenon yielded different interpretations (e.g. \citet{Chan_2022_InductionHeads} or \citet{chughtai_2023_toyModel} vs. \citet{stander2024grokkinggroupmultiplicationcosets} vs.  \citet{wu2024unifyingverifyingmechanisticinterpretations}). Several other instances were previously cited in this review (\Cref{subsec:reverse-engineering-step-2}). Hypotheses in interpretability require extensive validation beyond what appearances might imply.

Validating a hypothesis involves posing a simple question: Does the hypothesis make good predictions about the neural network's behavior? Testing the hypothesis often requires multiple approaches \citep{mueller2024questrightmediatorhistory}. To validate descriptions, many approaches simply apply a different description method than the one used to generate the initial hypothesis (\Cref{subsec:reverse-engineering-step-2}). If one description method yields a different description from another, it invalidates the hypothesis, which necessitates returning to an earlier step in the reverse engineering cycle (\Cref{fig:pipeline}). However, methods for validating descriptions are not limited to other component description methods. Hypothesis validation may take many forms, including the following:

\textbf{Predicting activations and counterfactuals}: By using natural language explanations of a given network component's function, it is possible to predict the component's activation levels on different inputs. This analysis can be carried out by humans or by AI systems (\Cref{subsec:automation}), as in \citep{hernandez2022natural,bills2023language,shaham2024multimodalautomatedinterpretabilityagent, Belrose_2024_autointerp}. In essence, our interpretations should enable us to successfully predict counterfactual scenarios in neural networks. For instance, if we ablate or activate particular network components, we should be able to predict specific downstream effects on other components.

\textbf{Predicting and explaining unusual failures or adversarial examples}: Good explanations of neural network behavior should help us identify and explain cases where that behavior fails to produce expected outcomes. For instance, \citet{hilton2020understanding} validated their methods by explaining cases where a deep reinforcement learning agent's neural network failed to achieve maximum reward and also explained specific hallucinations exhibited by the network. Another approach is to use use an interpretability approach to handcraft an input for a network that functioned as an adversarial example \citep{carter2019activation, casper2023diagnosticsdeepneuralnetworks, Mu_2020_Advances, hernandez2022natural}.

\textbf{Handcrafting a network that reconstructs a network behavior}: If our explanations for network behavior are sufficient, we should be able to use them to build replacement parts for the original network. \citet{cammarata2020curve} validated their interpretation of a curve detector's function in a convolutional neural network by substituting its parts with simple handcrafted replacements.

\textbf{Testing on ground truth}: If the weights of a toy neural network were handcrafted by humans, it is possible to obtain a ground truth explanation for how it works. This proves useful for testing explanations produced by interpretability methods. For example, \citet{conmy2023automatedcircuitdiscoverymechanistic} validated a tool's ability to attribute model behaviors to internal components by running it on a simple model that implemented a known algorithm. (See also \Cref{par:benchmarks}).

\textbf{Using the hypothesis to achieve particular engineering goals}: 
Another way to test explanations is to assess their utility in downstream applications \citep{doshivelez2017rigorousscienceinterpretablemachine, casper2023redteamingdeepneural}. For example, \citet{templeton2024scaling} discussed examples where manually editing a large language model based on an interpretation led to predictable high-level changes in its behavior. Meanwhile, \cite{marks2024sparsefeaturecircuitsdiscovering} showed how an interpretability tool could assist humans with debugging a classifier in a toy task. \citet{farrell2024applyingsparseautoencodersunlearn} use unlearning (\Cref{subsec:better-control}) to demonstrate that learned SDL latents don't quite match human concepts, and might not be optimal for particular downstream use cases, highlighting potential issues with SDL.

\textbf{Using the hypothesis to achieve specific engineering goals competitively}: Achieving not only useful, but competitive methods sets an even higher standard. For interpretability tools, the highest evaluation criteria require fair comparisons against relevant baselines on real-world tasks instead of cherry-picking them. However, the practice of conducting evaluations using non-cherry-picked tasks remains relatively uncommon. Although attempts have been made to use techniques in the current interpretability toolkit in such evaluations, they have not proven to be consistently useful \citep{adebayo2020debuggingtestsmodelexplanations,denain2023auditingvisualizationstransparencymethods,Caspar_2023_RedTeamingDNNFeatureSynthesis,hase2023doeslocalizationinformediting, durmus2024steering}. Some of the most promising research directions are to use interpretability methods to achieve things that would be hard or impossible to achieve without them \citep{schut2023bridginghumanaiknowledgegap}. Unless interpretability methods demonstrate that they are competitive with alternative approaches to achieve engineering goals, then the act of demonstrating their usefulness may lead to a bias toward developing methods that only perform well in best-case scenarios and on simple tasks, rather than those that can handle worst-case scenarios and practical challenges.

Interpretability researchers have historically faced challenges in adequately validating their hypotheses due to the high costs in terms of time and cognitive labor. In the following section, we explore two potential solutions that could simplify the validation process: model organisms and interpretability benchmarks.

\parasection{`Model organisms' facilitate hypothesis validation.}
\label{par:model-organisms}

Although interpretability is sometimes motivated by achieving engineering goals, it is often also approached through the perspective of the natural sciences \citep{olah2020zoom}. In certain natural sciences, such as genomics and neuroscience, it is common for researchers to investigate a few extensively studied species known as `model organisms' or 'model systems'. By conducting in-depth studies on a select group of organisms, like \textit{E. coli}, fruit flies, mice, and macaque monkeys, researchers can leverage the insights and tools gained from those organisms and apply them to other species. For example, imaging specific types of neural activity in mice is more tractable due to existing hypotheses about which proteins should be fluorescently labeled in order to identify specific types of neurons. The use of model organisms allows for cross-checking results with previous work, enabling stronger validation of hypotheses.

Currently, interpretability researchers lack consensus on which networks should serve as model organisms. Essentially, what should the \textit{Drosophila melanogaster} of mechanistic interpretability be? In mechanistic interpretability, an ideal model organism should be open source, easy and cheap to use, representative of a broad range of systems and phenomena, have a replicable training process with open source training data, and have multiple instances with different random seeds, among other criteria \citep{Sharkey_2022_CurrentThemes}. Thus far, researchers have mostly used model organisms that possess only some of these criteria, such as a transformer than can perform modular addition \citep{nanda2023progressmeasuresgrokkingmechanistic} or GPT-2 \citep{radford2018language}.

Model organisms not only support cross-validation of hypotheses, but also facilitate the progressive construction of experimental infrastructure by providing a reliable foundation for experiment design. This simplifies the process of rigorous hypothesis testing, thus helping prevent oversimplification and `interpretability illusions'.

Studying solely model organisms, instead of more directly pursuing engineering goals, risks merely making true statements about neural network structure, rather generating insights that are of immediate practical benefit. For mechanistic interpretability to make the fastest and most substantial progress toward engineering goals, both scientific and engineering wins should be pursued in parallel.

Furthermore, certain choices made while studying model organisms risk steering the field in suboptimal directions. For instance, interpretability research is often motivated by the engineering goal of understanding state-of-the-art models thoroughly enough to make assurances of their safety \citep{Bereska_2024_MechInterpReview,tegmark2023provablysafesystemspath, dalrymple2024guaranteedsafeaiframework}. However, limiting its focus by studying small toy models (e.g. \citet{nanda2023progressmeasuresgrokkingmechanistic}) or how larger models accomplish select subtasks \citep{arditi2024refusallanguagemodelsmediated}, risks incentivizing research and methods that fail to generalize to more safety-relevant real-world settings.

\parasection{Validating interpretability methods using benchmarks.}
\label{par:benchmarks}

Beyond validating individual hypotheses, we may wish to validate entire interpretability methods. Benchmarking is a proven approach to making incremental improvements in other areas of machine learning, with several approaches to benchmarking interpretability methods being developed in recent years.

One desideratum for interpretability benchmarks is to evaluate interpretations against ground truth explanations \citep{Freiesleben_2023_weneedtotalk, Zhou_2022_featureattribution}. Benchmarks can be established using models with known ground truth explanations. Such models can be created by compiling simple programs into weights of models that exactly implement the known program \citep{lindner2023tracrcompiledtransformerslaboratory, Weiss_2021_transformer, thurnherr2024tracrbench,gupta2024interpbenchsemisynthetictransformersevaluating}.  Alternatively, predetermined explanations can be enforced at training time in conventional models using Interchange Intervention Training \citep{gupta2024interpbenchsemisynthetictransformersevaluating,Geiger_2022_CausalStructure}. Other interpretability benchmarks that evaluate specific steps in the interpretability pipelines also exist, such as model decomposition \citep{huang2024ravelevaluatinginterpretabilitymethods,makelov2024principledevaluationssparseautoencoders}, generating descriptions of network component functions \citep{schwettmann2023findfunctiondescriptionbenchmark} , or testing natural language explanations \citep{huang2023rigorouslyassessingnaturallanguage}.

\subsection{Concept-based interpretability: Identifying components for given roles}
\label{subsec:concept-based-interpretability}

\subsubsection{Concept-based probes}
\label{subsubsec:concept-based-probes}

When attempting to localize a human-interpretable concept within the network, an intuitive approach is to `probe' for it \citep{kohn-2015-whats, gupta-etal-2015-distributional, alain2018understandingintermediatelayersusing, ettinger-etal-2016-probing}. A concept-based probe is a classifier trained to predict a concept from the hidden representation of another model \citep{hupkes2018visualisationdiagnosticclassifiersreveal}. Probing requires a labeling function that assigns classification labels to input data, indicating the `value' of the concept on that data (Note: A binary value indicating the presence or absence of a concept is a special case of this approach). Once the labels are assigned, a probe, which is a simple parameterized model, is trained to predict concept labels based on hidden activations. If the probe is a linear model, then we have localized the concept as a vector in latent space.

Probes were first introduced in NLP \citep{kohn-2015-whats,gupta-etal-2015-distributional} and have since been extensively explored in the field \citep{conneau2018cramsinglevectorprobing,tenney2019bertrediscoversclassicalnlp,rogers2020primerbertologyknowbert,gurnee2023findingneuronshaystackcase,peters2018dissectingcontextualwordembeddings,burns2024discoveringlatentknowledgelanguage,marks2024geometrytruthemergentlinear}. They have also been applied in vision \citep{alain2018understandingintermediatelayersusing, kim_2018_activationvectors} and deep reinforcement learning \citep{McGrath_2022_alphazero, forde2023where}. Probing also includes concept activation vectors \citep{Kim_2018_Activation}, information-theoretic probing \citep{voita-titov-2020-information}, and structural probing \citep{hewitt-manning-2019-structural}.

Although relatively simple to implement, probing has two main challenges \citep{ravichander2021probingprobingparadigmdoes,belinkov-2022-probing}: (1) The need for carefully chosen data for well-defined concepts, and (2) probes detect correlations instead of causal variables in hidden activations.

\subsubsection{Probes need carefully chosen data for well-defined concepts}
\label{subsubsec:probes-data}

Concept-based probing requires a labeling function that assigns labels to input data. Obtaining a labeling function is not always trivial and may require substantial human effort to define a data set for a single concept. Moreover, it is only possible to identify concepts that we have defined precisely enough to create high-quality data. This limitation implies that concept-based probing can only identify concepts that we were already looking for, rather than reveal unexpected features in the network. Another approach, known as Contrast-Consistent Search (CCS), probes not for single concepts but an axis in activation space that corresponds to positive or negative propositions by enforcing probabilistic consistency conditions \citep{burns2024discoveringlatentknowledgelanguage}. Despite being more unsupervised than standard concept-based probing, even CCS requires the construction of data sets with clear positive and negative cases. A potential path forward for concept-based decomposition is to develop methods that automatically develop data sets for probing and concept localization \citep{shaham2024multimodalautomatedinterpretabilityagent} (\Cref{subsec:automation}).

\subsubsection{Probes detect correlations, rather than causal variables, in hidden activations}
\label{subsubsec:probes-spurious-correlations}

Probes are tools for a correlational analysis, measuring if hidden activations serve as signals for a given concept. Indeed, from an information-theoretic point of view, an arbitrarily powerful probe measures the mutual information between a hidden representation and a concept \citep{hewitt2019designinginterpretingprobescontrol, pimentel-etal-2020-information}. 

However, training a probe to associate a concept with specific hidden activations does not necessarily imply that those activations causally mediate how that concept is used by the network, or even if the network uses the concept at all \citep{ravichander2021probingprobingparadigmdoes, geiger2021causalabstractionsneuralnetworks, Elazar_2021, belinkov-2022-probing}. Probes can be successfully trained on hidden activations that lack any causal connection to the output, only localizing correlated hidden activation vectors. For this reason, probing should be used only to generate hypotheses about which network components might be causally linked to a concept. Confirming such hypotheses requires further investigation using causal interventions or other probing methods.

To improve the causal relevance of probe vectors, one approach is to use counterfactual data, which involves intervening on the concept of interest (for instance, observing the resulting output if the dog in an image was changed to a cat)  \citep{Elazar_2021,mueller2024missedcausesambiguouseffects, geiger2021causalabstractionsneuralnetworks}. Methods include distributed alignment search \citep{geiger2024findingalignmentsinterpretablecausal, wu2024interpretabilityscaleidentifyingcausal, huang2024ravelevaluatinginterpretabilitymethods}, causal probing \citep{guerner2024geometricnotioncausalprobing}, using attribution methods to measure the effect of concept vectors on network predictions \citep{kim_2018_activationvectors}, and various concept erasure methods \citep{ravfogel-etal-2020-null, ravfogel2024linearadversarialconcepterasure,Elazar_2021, belrose2023leaceperfectlinearconcept, belrose2023leaceperfectlinearconcept}. While these methods can identify causal mediators of concepts in hidden representations, they require more specialized data than probes do. 

At times, it might be acceptable for probes to identify merely correlated hidden activations if the correlations generalize to the test distribution. However, probing approaches face a considerable risk of not only discovering correlated features, but also spurious correlations due to the high dimensionality of hidden activations. Validating probes is therefore essential to avoid overfitting. This includes evaluating them on out-of-distribution test data that varies along task-specific dimensions to ensure that general purpose features have been found. One question that remains unanswered is how to use regularization to achieve good probe generalization.

\subsubsection{Concept-based intrinsic interpretability}
\label{subsubsec:concept-based-intrinsic-interpretability}

Although probes begin with a trained network and search for specific concepts within it, it is also feasible to leverage the concepts in the network training process itself, such as in the case of concept bottleneck models \citep{Koh_2020_ConceptBottleneckModels}. This is beneficial as it is more likely that the components to which concepts are assigned are causally relevant. Instead of specific concepts, networks can also be trained to use particular causal structures \citep{Geiger_2022_CausalStructure}. While it may not be possible to prespecify all relevant concepts or structures, integrating this approach with methods for disentangling concepts could prove useful \citep{chen2019isolatingsourcesdisentanglementvariational}. \citet{cloud2024gradientroutingmaskinggradients} incentivize modularity via applying data-dependent, weighted masks to gradients during backpropagation.

\subsection{Proceduralizing mechanistic interpretability into circuit discovery pipelines: A case study}
\label{subsec:proceduralizing}

How should we codify the process of mechanistic interpretability to yield the deepest possible insights? To form a complete pipeline by combining various methods, several methodological choices regarding decomposition, description, and validation, must be made. Circuit discovery has emerged as a prominent pipeline in recent mechanistic interpretability research \citep{Wang_2022_Circuits, hanna2023doesgpt2computegreaterthan, heimersheim2023circuit}. Its objective is to describe how a neural network performs a task of interest while making specific choices for network decomposition, component description, and hypothesis validation. In this section, we look at the typical choices in each step in greater depth and discuss how this popular pipeline could be improved.

The `circuit discovery' pipeline takes the following steps:

\begin{enumerate}

\item   \textbf{Task Definition}. For a given model we want to study, we select a task that the model can perform, and a dataset on which the network performs that task. This is a concept-based step, since the definition of the task was based on how human researchers define a task distribution.

\item   \textbf{Decomposition}. During the decomposition step, it is common to think of the neural network as a directed acyclic graph (DAG), where activations are represented by nodes and the “abstract weights” between them represented by edges. Most work thus far has selected architectural components (\Cref{subsec:reverse-engineering-step-1}), such as attention heads and MLP layers, to be the nodes. However, more recent work has also used SDL latents for nodes \citep{marks2024sparsefeaturecircuitsdiscovering}

\item   \textbf{An initial description step: Identify task-relevant vs. -irrelevant subgraphs.} The circuit discovery procedure then identifies task-relevant nodes and edges. Typically, causal interventions are used, drawing samples from some “clean” and “counterfactual” data sets. Circuit discovery methods are generally based on iterative activation patching \citep{Wang_2022_Circuits,Chan_Garriga-alonso_Goldowsky-Dill_2022,lieberum2023doescircuitanalysisinterpretability} or integrated gradients \citep{marks2024sparsefeaturecircuitsdiscovering}.

\item   \textbf{An iterative description-validation loop}. After obtaining a task-relevant subgraph, the next step involves describing the function of each node or edge individually. This step is less formulaic than previous steps. Researchers rely on their intuition, attempting to create testable hypotheses for the function of a component or edge of the circuit, and then design custom experiments to validate or invalidate their hypothesis. Only after several iterations of hypothesis testing through experimentation are researchers finally satisfied with their explanation. In research papers, this loop is rarely made explicit, as only the final description is presented. However, \citet{Chan_2022_InductionHeads} detail this process for understanding the induction task, and \citet{nanda2023othello}provides another description of such a loop (building on work by \citet{li2024emergentworldrepresentationsexploring}).

\item   \textbf{Final Validation}. Circuits are commonly evaluated based on three attributes \citep{Wang_2022_Circuits}: \textit{faithfulness}, which refers to how closely the circuit approximates the entire network's behavior, \textit{minimality}, which assesses if nodes in the subgraph are unnecessary, and \textit{completeness}, which determines whether any nodes not included in the subgraph are important for task behavior. Additional ad hoc validation methodologies also exist. For example, \citet{Wang_2022_Circuits} generate adversarial examples for their task based on their mechanistic understanding, while \citet{shi2024hypothesistestingcircuithypothesis} devise a suite of formal statistical hypothesis tests for circuit efficacy.
\end{enumerate}

This circuit discovery procedure has yielded valuable insight, but falls short using current methods. The pipeline has several issues:

\textbf{Task definition is concept-based}. Defining circuits has thus far been with respect to tasks defined by humans. \citet{miller2024transformercircuitfaithfulnessmetrics} demonstrate that the within-task variance of model performance across the distribution of data points in a task is large, implying that the circuit provides a good approximation of the average case performance on the dataset, but a poor one for any individual data point. This suggests that the process of first selecting a task and then studying how the model performs it may not be an effective approach to achieve “reverse engineering” -style interpretability. Thus, it might be worth learning the task decomposition instead \citep{kaarel2024starting}.

\textbf{Network decomposition methods are flawed}. Perhaps most importantly, prior circuit discovery work has attempted to decompose models in either architectural bases \citep{Wang_2022_Circuits,conmy2023automatedcircuitdiscoverymechanistic} or sparse autoencoder latents \citep{marks2024sparsefeaturecircuitsdiscovering,cunningham2023sparseautoencodershighlyinterpretable}, which are imperfect ways to decompose neural networks for mechanistic interpretability (\Cref{subsec:reverse-engineering-step-1}). Future work could locate circuits in improved decompositions or simultaneously learn both network decompositions and circuits.

\textbf{Circuit faithfulness is low}. Simple early circuits were found to be unfaithful \citep{Chan_2022_InductionHeads, chan2023causal}. \citet{miller2024transformercircuitfaithfulnessmetrics} show that existing measures of faithfulness depend on the causal intervention implementation used, and further demonstrate that such metrics are misleading when applied to several complex end-to-end circuits. \citet{makelov2023subspacelookingforinterpretability} argue that subspace activation patching via distributed alignment search may lead to interpretability illusion mechanisms, although these findings are contested by \citet{wu2024replymakelovetal}.

\textbf{Scalable methods are only approximate}. Identifying relevant components through individual interventions is costly when there are many components. Attribution patching \citet{syed2023attributionpatchingoutperformsautomated} was designed to identify potential relevant candidates for further testing through intervention, which becomes more important as the number of components expands significantly through sparse dictionary learning \citep{marks2024sparsefeaturecircuitsdiscovering}. However, attribution patching uses gradients, which only yield a first-order approximation of the effect of ablating components \citep{wu2024replymakelovetal, molchanov2017pruning}, leaving it unclear whether this method and any improvements on it \citep{kramár2024atpefficientscalablemethod} produce adequate approximations.

\textbf{Circuit discovery algorithms struggle with backup and negative behavior}. Additional challenges for circuit analysis arise from the effects of “backup” and “negative” behavior \citep{Wang_2022_Circuits,mcgrath2023hydraeffectemergentselfrepair, mcdougall2023copysuppressioncomprehensivelyunderstanding}, which actively suppress task performance and are thus not captured by maximizing task performance metrics. Despite this, they remain important factors to consider; \citet{mueller2024missedcausesambiguouseffects} provides further discussion of these issues.

\textbf{Streetlight Interpretability}: The tasks studied so far have been deliberately selected to be simple to define and study mechanistically \citep{Wang_2022_Circuits}. This gives a misleading impression of the level of difficulty involved in implementing circuit discovery for any arbitrary task that a network implements. Indeed, attempts to study arbitrary circuits have proceeded less successfully \citep{nanda2023fact}.

Solving issues with current mechanistic interpretability pipelines remains an open challenge that promises significant benefit. Upon establishing reasonable procedures, automating the overall pipeline will become more feasible. However, some individual steps in mechanistic interpretability can already be fruitfully automated, as discussed in the next section (\Cref{subsec:automation}).

\subsection{Automating steps in mechanistic interpretability research}
\label{subsec:automation}

Historically, mechanistic interpretability research has required considerable manual researcher effort, though it typically studies models that are smaller than those at the frontier. To make interpretability useful for downstream use cases, scalable approaches are crucial. In this section, we discuss \textit{automated interpretability} methods. We will explore previous cases where manual tasks in mechanistic interpretability have been successfully automated and address open problems in further automation.

\textbf{Automating feature description and validation}. A task that is amenable to automation is `describing the functional role of model components' (\Cref{subsec:reverse-engineering-step-2}). With the increasing sophistication of language models, researchers have generated descriptions of the functional role of neurons in image models \citep{hernandez2022natural}, neurons in language models \citep{bills2023language}, and sparse autoencoder latents in language models \citep{cunningham2023sparseautoencodershighlyinterpretable,Bricken_2023_dictionary, Belrose_2024_autointerp} using highly activating data set examples. These interpretations are validated by assessing how effectively a human or model can use them to predict the activation of a feature in a given data set example, or predict where a feature is active within a single image or text excerpt. The success of these predictions can be used as a quantitative measurement of `interpretability'. This was previously used to measure progress toward a decomposition method that carves networks at the joints of their generalization structure, assuming that such a decomposition would be maximally interpretable (\Cref{subsec:reverse-engineering-step-1}). While imperfect, these methods for interpretation hypothesis generation and validation might be improved by automating the generation of inputs to test the interpretation hypotheses by ensuring that generations activate the interpreted feature \citep{huang2023rigorouslyassessingnaturallanguage}, or defining more rigorous statistical tests \citep{bloom2024understanding}. Automatic component labeling could expand in the future to include descriptions of feature effects, relationships between features \citep{bussman2024metasaes}, or how components interact during runtime produce behavior.

\textbf{Automating circuit discovery pipelines}. An approach called Automated Circuit DisCovery' (ACDC) automates part of the pipeline discussed in \Cref{subsec:proceduralizing} to identify computational subgraphs involved in particular tasks \citep{conmy2023automatedcircuitdiscoverymechanistic}. Several works have since improved upon and accelerated this process \citep{syed2023attributionpatchingoutperformsautomated,kramár2024atpefficientscalablemethod, marks2024sparsefeaturecircuitsdiscovering}. Note that ACDC-like approaches in general assist in identifying relevant subgraphs for a pre-defined task, but do not automate important subsequent steps, such as describing the functional role of subgraph components.

While significant progress has been made toward automating steps of mechanistic interpretability pipelines, fully automating current pipelines would not yield satisfactory explanations of model behavior\footnote{For one attempt at this using leading decomposition and description methods, see \citet{marks2024sparsefeaturecircuitsdiscovering}}. Further methodological progress is required for fully automated neural network interpretability to be capable of generating the quality of interpretations necessary to achieve our goals.

\section{Open problems in applications of mechanistic interpretability}
\label{sec:open-problems-applications}

\begin{figure}[h]
    \centering
    \includegraphics[width=0.8\textwidth]{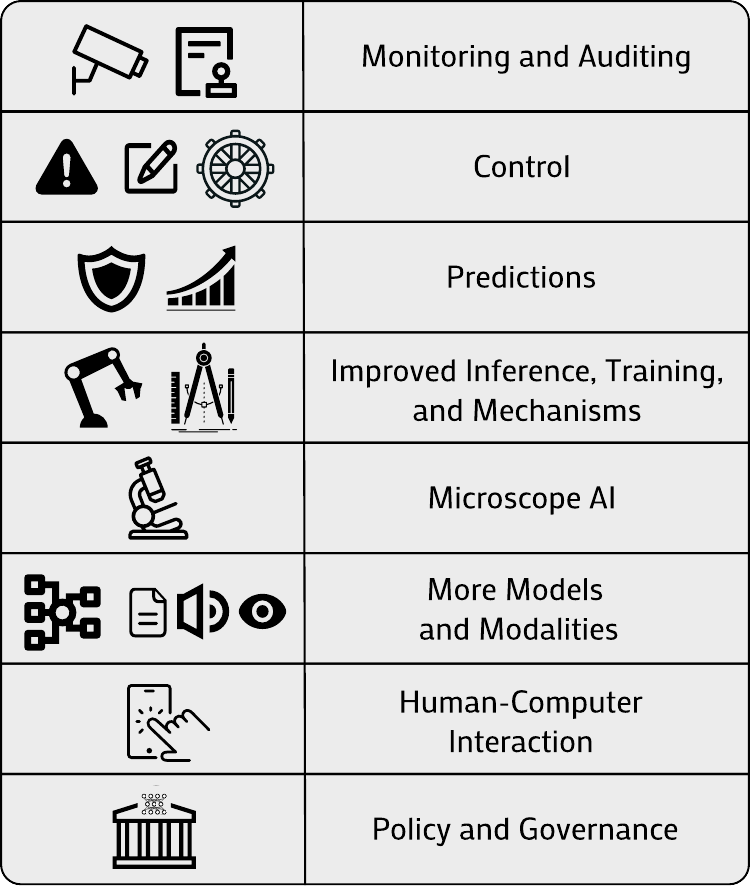}
    \caption{A summary of problem areas for applications of mechanistic interpretability.}
    \label{fig:applications}
\end{figure}

Ultimately, we need mechanistic interpretability methods that enable us to solve concrete scientific and engineering problems (\Cref{fig:applications}). While predicting the impact of fundamental science in advance is difficult, having concrete goals in mind during research is usually beneficial. We want mechanistic interpretability methods to help us achieve various outcomes, such as monitoring and auditing AI systems more effectively (\Cref{subsec:monitoring-and-auditing}), controlling of AI system behavior more precisely (\Cref{subsec:better-control}), predicting AI system outcomes more accurately (\Cref{subsec:better-predictions}), enhancing AI system capabilities (\Cref{subsec:capabilities}), and extracting knowledge from AI systems (\Cref{subsec:microscope-ai}). We should also anticipate that mechanistic interpretability will uncover “unknown problems” present in systems, revealing that the true realm of challenges and possibilities is greater than what we currently perceive it to be.

As highlighted in the previous section, progress in mechanistic interpretability methods is multifaceted. Each axis of methodological advancement leads to varying degrees of progress toward different goals. Before we discuss open questions in its applications, we identify distinct axes of methodological progress that lead to different amounts of progress toward different goals.

\subsection{Axes of mechanistic interpretability progress.}
\label{subsec:axes}

\textbf{Decomposition vs. description of network components}: Improvements in network decomposition versus component description methods offer varying benefits for different goals. Decomposition methods vary in their efficacy at carving networks at the joints of their generalization structure, while description methods can yield descriptions that vary in depth. Deeper descriptions of a component are typically more causal or mechanistic, whereas shallower descriptions may rely more on correlations and only connect to inputs or outputs without referencing intermediate causes or effects (\Cref{fig:descriptions}. Deeper descriptions thus attempt to explain more about how the component interacts with other components within the network's algorithm. Certain goals can be achieved with minimal or no progress in decomposition or description, while others may demand substantial progress.

\textbf{Extent of network decomposition or description}: The extent of network decomposition or description needed may vary depending on the goal. Certain goals only require an understanding of specific network components (such as an individual features or a circuit), while others might require enumerating or understanding of larger circuits or the entire model (as in `enumerative safety').

\textbf{Extent of task distribution analyzed}: The scope of task distribution analysis also depends on the intended goal. For instance, monitoring a model for a single kind of behavior might only require decomposing or understanding the model only over a narrow task distribution, while others, such as formal verification, might demand understanding over the entire distribution of tasks.

\textbf{Mechanistic understanding post vs. during training}: Understanding the mechanisms of a fixed model might suffice for some goals, but more ambitious goals might require an understanding of not only the models' mechanisms, but also how they change during the learning process.

In this section, we'll discuss how mechanistic interpretability has been used or could be leveraged to further the field's various goals. We'll assess the progress made thus far, and identify the advancements along different axes of methodological progress that will be most crucial to success.

\subsection{Using mechanistic interpretability for better monitoring and auditing of AI systems for potentially unsafe cognition}
\label{subsec:monitoring-and-auditing}

\subsubsection{Mechanistic interpretability-based evaluations could help us detect unsafe or unethical AI cognition}
\label{par:interp-evals}

Currently, we rely on “black box” evaluations to understand a model's capabilities, but studying input-output behavior alone may not reveal all dangerous behaviors. Such behaviors include deceiving users \citep{park2023aideceptionsurveyexamples, ward2023honesty, scheurer2024largelanguagemodelsstrategically, meinke2024frontiermodelscapableincontext}; for instance, by intentionally underperforming on evaluations (“sandbagging”, \citet{vanderweij2024aisandbagginglanguagemodels}), leveraging situational awareness \citep{laine2024me}; or giving dishonest responses tailored to match the user’s beliefs (“sycophancy”; \citet{sharma2023understandingsycophancylanguagemodels}). Interpretability techniques could be used to uncover the mechanisms underlying these potentially harmful behaviors and thus help to detect and characterize them. This becomes increasingly important as the capabilities of models increases, especially when using training methods that incentivize models to feign particular properties for the purpose of passing evaluation. 

Using interpretability methods to identify internal signs of concern (also known as “white-box” evaluations \citep{Casper_2024} or “understanding-based” evaluations \citep{hubinger2023towards}) is therefore an important problem. White-box evaluation methods could serve as tools to detect potential biases that arise when models learn to use spurious correlations \citep{gandelsman2024interpretingclipsimagerepresentation, Caspar_2022_featurelevel, Abid_2022_counterfactualExplanations}. However, human judgment might be required to determine which features are `supposed' to be relevant to the task \citep{marks2024sparsefeaturecircuitsdiscovering, Kim_2018_Activation, Goyal_2022}.

Despite current shortcomings in decomposition and description, white-box evaluations are likely feasible today. Even shallow, correlation-based descriptions could signal potentially concerning cognition. For example, developing new methods that reliably distinguish between features that merely recognize deceptive behavior vs. mechanisms that cause deceptive behavior may be challenging. However, a correlation-based method that flags both can facilitate catching the latter. To be useful, it may not even be necessary to decompose or describe the entire network; having descriptions for components that are used on concerning subdistributions of model behavior might suffice. For instance, imperfect interpretability methods may help evaluators develop hypotheses about how models will behave, thus guiding further inquiry. Meanwhile, recent work proposes incorporating SDL (\Cref{par:sdl}) into safety cases for advanced AI systems. By monitoring internal representations, it could aid in detecting potential sabotage or deceptive behavior before deployment \citep{grosse2024three}. While such approaches show promise, they have difficulty in validating whether learned features capture all concerning patterns of reasoning reliably.

Evaluations for unsafe cognition may be a particularly important use case as it plays well to the comparative advantages of mechanistic interpretability relative to the other areas of machine learning. The majority of other areas of machine learning already focus on controlling or steering the behavior of AI systems to alter input-output behavior. It is therefore unclear that this is to mechanistic interpretability's comparative advantage. On the other hand, interpretability is perhaps the only research area that attempts to understand the mechanisms of model cognition. This implies that it might be particularly fruitful for interpretability researchers to tackle problems that become easier to solve through improving such understanding: auditing for unsafe cognition, debugging unexpected behavior, and monitoring systems in deployment.

\parasection{Enabling real time monitoring of AI systems for potentially unsafe cognition}
\label{par:monitoring}

Beyond white-box evaluations, interpretability has further applications in monitoring. For instance, internals could be used to passively monitor the system during deployment, much like content moderation systems currently in use today. Alternatively, internals could be used to flag when a model takes an action for abnormal reasons even in absence of satisfactory descriptions (\Cref{subsec:reverse-engineering-step-2}), known as “mechanistic anomaly detection” \citep{Christiano_2022_MAD,Johnston_2024_MAD}, which may be a sign of suspicious behavior. Mechanistic anomaly detection primarily requires progress in decomposition methods (\Cref{subsec:reverse-engineering-step-1}) as it is necessary to be confident about what constitutes an individual mechanism within the network. Current SDL methods identify active latents, but not active mechanisms, which are implemented by network parameters. It may not be necessary to have deep descriptions of the function of individual mechanisms as long as anomalies can be detected.

\parasection{Improving our ability to red-team AI systems and elicit unsafe outputs}
\label{par:red-teaming}

Beyond white-box evaluations, leveraging interpretability could improve our ability to conduct adversarial attacks, red-team, or jailbreak AI systems \citep{Casper_2024}. This process is beneficial as it exhibits failure modes models may display in the wild, when facing adversarial pressure from wide deployment or malicious actors, thereby enabling developers to effectively preempt and address them. Furthermore, it may form a significant element of safety cases \citep{clymer2024safetycasesjustifysafety, balesni2024evaluationsbasedsafetycasesai,grosse2024three,goemans2024safetycasetemplatefrontier} for AI systems, by providing assurances of form: We tried hard to red-team the system, yet failed to exhibit concerning behavior despite having \textit{more} affordances than users may have. One reasonable assumption is that developers may have white-box access to models, while users may not. Although many existing red-teaming methods \citep{perez2022redteaminglanguagemodels,zou2023universaltransferableadversarialattacks} already require gradient access, we could additionally leverage interpretability insights to accelerate red-teaming. For instance, \citet{arditi2024refusallanguagemodelsmediated} discovered a universal “refusal direction” in chat-finetuned language models, which is causally important for models engaging in the behavior of refusing harmful requests. \citet{lin2024understandingjailbreakattacksllms} used this to red-team models by optimizing for inputs that minimize the projection of the residual stream onto this direction during the forward pass. This approach may be more efficient than optimizing over the whole model, as previous methods did \citep{zou2023universaltransferableadversarialattacks}. Mechanistic interpretability techniques also promise to improve our ability to attribute model outputs to their corresponding inputs \cref{par:activation-causes}. This could enhance human red-teamers' ability to find key input features responsible for bad behavior in models, thus speeding up iteration cycles. Though such approaches are possible today with only feature-based understanding, they might improve with more crisp mechanism-based understanding.

\subsubsection{Using mechanistic interpretability for better control of AI system behavior}
\label{subsec:better-control}

Ensuring the safe deployment of AI first requires effective control over their behavior. Currently, the techniques used for this purpose are mostly unrelated to interpretability (e.g. \citet{christiano2023deepreinforcementlearninghuman}; \citet{rafailov2024directpreferenceoptimizationlanguage}; inter alia), but sometimes inspired by it (\citet{panickssery2024steeringllama2contrastive}, \citet{zou2024improvingalignmentrobustnesscircuit}, \citet{kirch2024featurespromptsjailbreakllms}; inter alia). Mechanistic interpretability could assist in interpreting  \citep{lee2024mechanisticunderstandingalignmentalgorithms} and improving \citep{conmy2024activation} these control methods, or in developing new ones. In this section, we outline interpretability-inspired control methods, and envision future possibilities with further progress in interpretability methods.

One new control method derived from mechanistic interpretability insights is \textbf{activation steering} (a.k.a. activation addition). A fixed activation vector, hypothesized to linearly represent a model concept, is added to an intermediate activation of a model at inference time \citep{2024inferencetimeinterventionelicitingtruthful, turner2024steeringlanguagemodelsactivation,zou2023representationengineeringtopdownapproach,panickssery2024steeringllama2contrastive}. \citet{turner2024steeringlanguagemodelsactivation} introduced activation steering, directly inspired by the Linear Representation Hypothesis (discussed in \Cref{par:sdl}). This technique results from the hypothesis, and its success can be thought of as evidence for the hypothesis. Moderate success can be achieved in steering using basic decomposition and description methods. Mechanistic interpretability decomposition methods enable the steering of models toward a narrower range of behaviors with fewer side effects \citep{chalnev2024improvingsteeringvectorstargeting}. Advancements in mechanistic interpretability methods are likely to result in improved steering capabilities, such as activating entire mechanisms instead of individual features.

\textbf{Machine unlearning} was originally defined as the problem of scrubbing the influence of particular data points on a trained machine learning model \citep{Cao_2015_MachineUnlearning}. In the context of modern generative models, machine unlearning is more broadly defined as removing particular undesirable knowledge or capabilities ('unlearning targets') from models, while preserving model performance on tasks involving non-targets \citep{liu2024rethinkingmachineunlearninglarge}. Targets for unlearning that are of particular interest include sensitive private or copyrighted data, model biases \citep{liu2024machineunlearninggenerativeai}, and hazardous knowledge that could be misused by malicious actors \citep{li2024wmdp}; for instance, information regarding the creation of bioweapons. A better understanding of how knowledge or capabilities are implemented within model internals can help in the development of new techniques for machine unlearning \citep{belrose2023leaceperfectlinearconcept,zou2024improvingalignmentrobustnesscircuit, guo2024robust,pochinkov2024dissectinglanguagemodelsmachine, ashuach2024revsunlearningsensitiveinformation}, as well as to better evaluate unlearning efficacy through white-box, non-behavioral, techniques \citep{lynch2024methodsevaluaterobustunlearning,deeb2024unlearningmethodsremoveinformation,hong2024intrinsicevaluationunlearningusing}. Thus far, mechanistic interpretability methods that modify intermediate activations (but not weights) for unlearning have yet to yield competitive results \citep{farrell2024applyingsparseautoencodersunlearn}.

Unlearning falls under the broader aim of \textbf{model (knowledge) editing}, which seeks to make precise modifications to a machine learning model that incorporates specific knowledge with desirable generalization properties, while minimizing the impact on other knowledge \citep{wang2023knowledgeeditinglargelanguage}. By attempting to carve neural networks at their joints, mechanistic interpretability could improve our ability to make interventions on knowledge with few side effects. \citet{meng2023locatingeditingfactualassociations} make initial progress toward interpretability-based model editing with their ROME technique. However, \citet{Thibodeau_2022} and \citet{hase2023doeslocalizationinformediting} highlight flaws in the technique, indicating that mechanistic interpretability has not yet found appropriate model components to intervene on (\Cref{subsec:reverse-engineering-step-1}). With better comprehension of neural networks, we should anticipate more surgical model editing techniques in the future.

Editing any given capability or piece of knowledge presents a greater challenge than deleting them. Meaningful progress in unlearning and editing methods may depend on improved network decomposition methods, as it would require isolating the individual mechanisms that correspond to specific knowledge or capabilities. Progress in mechanistic interpretability may elucidate the structure of knowledge and capabilities in AI models, leading to a better understanding of what kinds of model edits possibilities are realistic in future. Knowledge and capabilities could, in fact, be part of large mechanisms that overlap with each other, making it challenging to isolate them into discrete components. For mechanistic interpretability to effectively guide editing, strong description methods will be necessary to understand how to modify specific targets without affecting others.

Finally, mechanistic interpretability may provide tools to rigorously \textbf{understand how finetuning alters models}. This may assist in debugging instances in which finetuning leads to undesired and spurious effects \citep{casper2023openproblemsfundamentallimitations}. Recent work \citep{jain2024mechanisticallyanalyzingeffectsfinetuning, prakash2024finetuningenhancesexistingmechanisms, lee2024mechanisticunderstandingalignmentalgorithms} suggests that existing finetuning methodologies primarily make shallow edits to existing model representations and circuitry. Importantly, this suggests that harmlessness training (which trains models to refuse to answer harmful requests) may be cheaply undone. Empirical evidence supports this claim, both with further finetuning \citep{gade2024badllamacheaplyremovingsafety,lermen2024lorafinetuningefficientlyundoes}, as well as with causal interventions on the forward pass \citep{arditi2024refusallanguagemodelsmediated}. Separately, localizing knowledge and capabilities within models may improve the sample efficiency of finetuning, by selectively modifying only relevant parameters (as in, e.g. \citet{wu2024reftrepresentationfinetuninglanguage}). Further advancement in tools for comparing feature-level differences between models (such as \citet{lindsey2024crosscoders}) may accelerate our ability to debug finetuning or other control methods \citep{bricken2024stage}. Mechanistic interpretability work has thus yielded several insights into how finetuning changes models and how to improve it, and may provide further insights and improvements in the future. 

\subsection{Using mechanistic interpretability for better predictions about AI systems}
\label{subsec:better-predictions}

Accurately predicting model behavior in new scenarios or regimes is difficult (arguably impossible) without understanding model internals. Interpretability could hopefully facilitate two kinds of predictions:
\begin{itemize}
    \item Predicting model behavior in novel situations
    \item Predicting capabilities that arise during training or finetuning
\end{itemize}

\subsubsection{Predicting behavior in novel situations}
\label{subsubsec:predicting-behavior-in-novel-situations}

In order to determine whether an AI system may potentially underperform poorly or pose a safety risk in new situations, the ability to predict its behavior in untested settings is imperative. A model's behavior, which may only become apparent in unforeseen circumstances, cannot be fully captured by its performance on a finite set of behavioral evaluations.

By understanding the mechanisms of jailbreaking, we can anticipate the means through which a user might bypass existing safeguards \citep{lee2024mechanisticunderstandingalignmentalgorithms,arditi2024refusallanguagemodelsmediated}. Similarly, if models have “trojans”, backdoors \citep{hubinger2024sleeperagentstrainingdeceptive}, adversarial examples, or biases, comprehending a model's internal mechanisms could improve our ability to predict when models will display undesirable behavior, even if these scenarios were not encountered during standard training or behavioral evaluations. \citet{casper2023redteamingdeepneural} benchmark feature synthesis tools through their ability to aid developers in identifying trojans, while interpretability assisted in identifying cases of adversarial examples 
\citep{gandelsman2024interpretingsecondordereffectsneurons, Mu_2020_Advances,Wang_2022_Circuits, Kissane_Conmy_Nanda_2024}, and SDL was used to uncover biases based on spurious correlations in an LLM-based classifier \citep{marks2024sparsefeaturecircuitsdiscovering}. Beyond specific failures, interpretability methods can also be used to gain a broader understanding of network behavior. For example, prior work identified signatures in model internals that predict a model’s likelihood of hallucinating \citep{yu2024mechanisticunderstandingmitigationlanguage} or its knowledge of particular facts \citep{gottesman2024estimatingknowledgelargelanguage}.

Generally, being able to predict an AI system's behavior in advance is more challenging -- but also more desirable -- than merely being able to monitor its behavior and cognition. Mechanistic interpretability could allow us to make a certain type of claim, namely, “there exists no mechanisms that would cause the model to deliberately behave undesirably” \citep{olah2023interpretability}. For a strong version of this claim, substantial progress in both decomposition and description methods is necessary. However, weaker versions of the claim, addressing specific undesirable behaviors, might be more feasible with near-term methods. For instance, if it is possible to decompose networks and identify all components, even basic description methods might let us recognize that there are no mechanisms relating to bioweapons or illicit substances within the network, thus letting us predict that models are probably not capable of instructing users how to fabricate bioweapons or illicit substances (a possibility sometimes referred to as “enumerative safety” \citep{elhage2021mathematical, olah2023interpretability}). As AI systems become increasingly agentic, claims about even more general behavior may be possible. Understanding their values or goals (or, less anthropomorphically, `the internal mechanisms that determine their action plans and actions') should enable us to better predict their behavior across a broad range of contexts \citep{colognese2023high}.

When deploying AI in high-stakes scenarios, rigorous and reliable predictions are necessary, much like those demanded of safety-critical software applications. Sometimes, such software is formally verified, thereby ensuring certain safety-critical aspects of its behavior are guaranteed, since its compliance with specific properties is mathematically proven. In the context of mechanistic interpretability, the equivalent would be formal verification of AI systems \citep{dalrymple2024guaranteedsafeaiframework,tegmark2023provablysafesystemspath, critch2020airesearchconsiderationshuman}: mathematically proving that an AI system's behavior will satisfy a desired property on any input in a given distribution. Formal verification of AI systems remains an unresolved issue at present. The level of understanding of AI necessary to enable formal verification of large, general AI systems for nontrivial properties is well beyond the current capabilities of mechanistic interpretability. However, some recent studies using toy models provide a glimpse into what solving formal verification of AI might look like. Approaches inspired by mechanistic interpretability have been used to prove accuracy bounds on a single-layer transformer trained on a synthetic task, albeit with great difficulty \citep{gross2024compactproofsmodelperformance}. Program synthesis through mechanistic analysis offers an alternative approach by converting simple trained neural networks into more interpretable, controllable, and verifiable programs \citep{michaud2024openingaiblackbox}. 

Several open questions remain about the tractability of scaling these approaches from toy models to frontier systems. For instance, for program synthesis, it is uncertain to what extent computations within real-world neural networks can be reduced to operations that can be cleanly represented in symbolic code, or what the total length of such code would be. Ensuring the safety of agentic systems with formal guarantees is further complicated by the need to model a system's interactions in an arbitrarily complex environment that might not be formalizable \citep{seshia2020verifiedartificialintelligence,Wongpiromsarn_2023,dalrymple2024guaranteedsafeaiframework}.

\subsubsection{Predicting capabilities that arise during training or finetuning}
\label{subsubsec:predicting-capabilities}

The most competitive methods of AI development systems result in uninterpretable systems that often fail in ways that surprise their developers \citep{openai2024gpt4technicalreport, geminiteam2024geminifamilyhighlycapable, Anthropic}. Applying mechanistic interpretability to alleviate this issue is a key area for future research.

Improved mechanistic understanding of model training could enhance the ability to predict when certain capabilities will appear. For instance, it has been observed that new model capabilities can emerge as a function of scale (\citet{wei2022emergentabilitieslargelanguage}, though also see \citet{schaeffer2023emergentabilitieslargelanguage}). Evidence suggests that new capabilities may be learned in a somewhat discrete \citet{michaud2023the} or stagewise \citep{wang2024loss} fashion, and that in synthetic data settings, the emergence of new capabilities coincides with abrupt changes in the trajectory of model parameters \citep{park2024emergencehiddencapabilitiesexploring}.

Other work shows a correlation between in-context learning capabilities and the emergence of induction heads, an attention-based circuit mechanism \citep{olsson2022incontextlearninginductionheads}. By connecting these threads of research, the long-term hope for mechanistic interpretability research is to link small-scale mechanistic structure to larger-scale structure, such as the evolving shape of the loss landscape during model scaling \citep{olah2023interpretability}. To make progress toward this goal, research needs to move beyond simply improving `decomposition' (\Cref{subsec:reverse-engineering-step-1}) or `description' (\Cref{subsec:reverse-engineering-step-2}) quality and instead be capable of describing the dynamic changes in the mechanistic structure of networks throughout the learning process.

We may also want to link the emergence of capabilities to specific properties of the training data set. Through mechanistic interpretability, we can create data sets that facilitate training models to demonstrate desirable attributes and predict their behavior. By attributing model outputs to specific training examples, influence functions have been applied to LLMs \citep{Koh_2017_influence_function,grosse2023studyinglargelanguagemodel} to predict limitations in their generalization abilities, such as a lack of robustness when the order of certain phrases was flipped \citep{berglund2024reversalcursellmstrained}. Other work examines how data set composition shapes the emergence of in-context and weights-based learning \citep{reddy2024the}.

A related problem of interest involves predicting which model capabilities, that may not be present in a given model, can be “elicited” with sufficiently advanced prompting or finetuning strategies \citep{greenblatt2024stresstestingcapabilityelicitationpasswordlocked}. \citet{prakash2024finetuningenhancesexistingmechanisms} find evidence that finetuning improves capabilities primarily by enhancing existing circuits, rather than developing fundamentally new mechanisms. Relatedly, \citep{jain2024mechanisticallyanalyzingeffectsfinetuning} and \citet{lee2024mechanisticunderstandingalignmentalgorithms} show that finetuning can mask capabilities present in a base model in a way that can easily be reversed via simple changes to the model. Improved mechanistic understanding of finetuning could help reveal capabilities obscured in this fashion. Since capabilities are behaviors that often span multiple sequential steps, it may be necessary to have mechanistic interpretability methods that examine mechanisms spanning multiple time steps. However, current mechanistic interpretability research is primarily focused on understanding mechanisms involved in predictions at a single time step.

\subsection{Using mechanistic interpretability to improve our ability to perform inference, improve training, and make better use of learned mechanisms}
\label{subsec:capabilities}

A mechanistic understanding of AI models could be leveraged to improve their utility, from faster inference and better training, to enhancing and manipulating representations.

By understanding the internal generation process of AI models, we could accelerate their inference. For example, it could help identify which parts of the computation could be skipped without changing the model's final output \citep{voita-2019-analyzing, din2024jumpconclusionsshortcuttingtransformers, voita-etal-2024-neurons, gromov2024unreasonableineffectivenessdeeperlayers}. The ability to inspect the information or functions implemented in a model could facilitate the development of more effective distillation methods by recognizing gaps that should be distilled \citep{gottesman2024estimatingknowledgelargelanguage} and discovering novel ways to distill them \citep{zhang2024instillinginductivebiasessubnetworks}.

Another aspect that mechanistic interpretability could enhance is model training. Interpreting how the model processes specific examples and using this information to influence its predictions \citep{Koh_2017_influence_function,grosse2023studyinglargelanguagemodel} may inform the selection of better training data to improve the model's capabilities in desired ways. Moreover, better monitoring of the training process can be achieved by correlating certain drops in training loss with capability gains \citep{olsson2022incontextlearninginductionheads, wang2024grokkedtransformersimplicitreasoners} or identifying a general order in which specific capabilities emerge during training \citep{michaud2023the}. In addition, identifying the contributing components of a given task could help to devise novel, parameter-efficient training methods. Finally, being able to decompose networks into their functional components presents possibilities to build components that lend themselves to learning computational structures that we better understand \citep{crowson2022vqganclipopendomainimage,fu2023hungry}.

Mechanistic interpretability has the potential to not only accelerate AI inference and training, but also enhance its utility. Intervening in the model’s computation has the potential to remove unwanted bugs in its reasoning abilities, and achieve better balance between its knowledge recall process and latent reasoning \citep{yu-etal-2023-characterizing,jin-etal-2024-cutting,biran2024hoppinglateexploringlimitations,balesni2024twohopcursellmstrained}. More broadly, understanding the inner workings of different models could lead to better recombination of what they have learned, such as combining model parameters \citep{wortsman_2022_ModelSoups} and transferring representations across models \citep{ghandeharioun2024patchscopes, csiszarik2021similarity}.

\subsection{Using mechanistic interpretability for `microscope AI'}
\label{subsec:microscope-ai}

Current approaches for knowledge discovery from data involve statistical or causal analysis, dimensionality reduction, or using machine learning models that are inherently interpretable. These techniques can be valuable, but are influenced by human priors, typically assume linear relationships between variables, and cannot handle massive multimodal data. On the other hand, neural networks can do these things. Deep learning models are capable of encoding complex, non-linear relationships and extracting meaningful features from massive data sets without human intervention. Historically, these abilities had limited scientific value, as without methods to interpret these models, we could not understand the patterns they found. However, with ongoing advancements in interpretability research, this is beginning to change.

Applying interpretability for knowledge discovery is sometimes called \textbf{microscope AI}. This approach involves training a neural network to model a data set, then applying interpretability techniques to the model to gain insight into any (potentially novel) predictors it discovers. In this way, the superhuman pattern matching skills of deep neural networks can serve as a tool to parse complex data sets.

Current methods allow for versions of microscope AI, depending on the kind of insights that we want to learn. Some examples of these applications include extracting novel chess concepts from AlphaZero and teaching them to top grandmasters \citep{schut2023bridginghumanaiknowledgegap} using a CNN trained on defendant mugshots and judge decisions to reveal how facial features affect judgments \citep{Ludwig_2023}, transforming psychology articles into a causal graph with an LLM to enable link prediction and produce expert-level hypotheses \citep{tong2024automating}, and analyzing a CNN to learn previously unknown morphological features for predicting immune cell protein expression \citep{cooper2022lymphocyte}, among several other studies \citep{obrien2023machine,hicks2021explaining,Narayanaswamy_2020,korot2021predicting}. As interpretability methods improve along various axes, deeper insights in and across more domains will become possible.

Currently, the majority of scientists are unable to access microscope AI due to the need for specialized expertise in machine learning, interpretability, and domain knowledge to recognize significant new patterns, a combination of skills that is rare in many fields. This may change as interpretability research becomes more widely adopted in the sciences and as interpretability becomes increasingly automated and accessible.

\subsection{Mechanistic interpretability on a broader range of models and model families}
\label{subsec:broader-range-of-models}

The vast majority of mechanistic interpretability research to date has focused on just three model families: CNN-based image models (e.g. \cite{Erhan_2009_visualising, nguyen2016multifacetedfeaturevisualizationuncovering,olah2020zoom}), BERT-based text models (e.g. \cite{devlin2019bertpretrainingdeepbidirectional, rogers2020primerbertologyknowbert}) 
and GPT-based text models (e.g. \cite{elhage2021mathematical, Wang_2022_Circuits,nanda2023progressmeasuresgrokkingmechanistic}). The degree of generalizability of these findings to other models and contexts is currently a somewhat open question. Given that future frontier models may use architectures that differ from the current state-of-the-art, and are expected to be multimodal by default, interpretability researchers may need to expand the range of models and modalities that they study and try to identify universal approaches that can be applied to all of them.

Assessing how well interpretability methods apply to architectures beyond those for which they were developed, and whether we can develop techniques that generalize effectively across architectures remain open questions. This is especially important due to the recent success of other competitive architectures as alternatives to CNNs and transformers. Notable alternatives include diffusion models \citep{pmlr-v37-sohl-dickstein15, rombach2022highresolutionimagesynthesislatent} and Vision Transformers \citep{dosovitskiy2021imageworth16x16words} for image generation/classification and RWKV \citep{peng2023rwkvreinventingrnnstransformer} and later state space models (SSMs) \citep{gu2024mambalineartimesequencemodeling} for language modeling. Recent studies show that certain methods transfer from CNNs to SSMs  \citep{paulo2024doestransformerinterpretabilitytransfer}, and from transformers to some SSMs e.g. (\citet{meng2022locating} vs. \citet{sharma2024locatingeditingfactualassociations}) and (\citet{Wang_2022_Circuits} vs. \citet{ensign2024investigatingindirectobjectidentification}).

Beyond the transferability of interpretability methods, a related open question concerns the transferability of conclusions across model families. The overwhelming majority of mechanistic interpretability research focuses on the transformer model family and therefore does not distinguish between observations that are model-specific and those that are not. Consequently, we may be overlooking valuable insights that could be obtained by comparing the results across multiple model families. These insights may, for example, evince or refute the `universality hypothesis', which states that \citep{li2016convergentlearningdifferentneural,olah2020zoom} different neural networks learn similar features and circuits to one another.

\subsection{Human computer interaction with model internals}
\label{subsec:human-computer-interaction}

As we saw in \Cref{subsec:better-predictions}, the ability to control and understand neural networks is tightly linked. Thus, mechanistic interpretability has great potential to facilitate new types of human-AI interaction. Systems amenable to human comprehension and control would allow diverse users to intuitively manipulate and interact with them based on their preferences, greatly broadening their utility. As a starting point, AI engineers who build and test AI systems have an obvious interest in working with the internal workings of neural nets. If experts could visualize and interact with internal representations, it would unlock obvious benefits for scientific research.

However, interactive tooling has a much broader constituency, including policy-focused AI auditors, as well as end users. Consider an auditor looking for bias or safety issues in a neural net. With a way to probe the network directly instead of relying on testing behavior, the auditor can be much more successful in discovering potential low-probability but high-stakes errors. For end users, transparency into a network facilitates appropriate calibration of trust. One recent idea proposes a dashboard that can show, in real time, the internal features that influence a chatbot's answers during a text chat \citep{chen2024designingdashboardtransparencycontrol,zou2023representationengineeringtopdownapproach, viégas2023modelusermodelexploring}. Such a dashboard might help users spot AI errors, or display warnings if safety-relevant features activate. More generally, results from mechanistic interpretability could be used to blend direct-manipulation interfaces with text interfaces, providing users with a richer palette of controls \citep{carter2017using}.
\section{Open socio-technical problems in mechanistic interpretability}
\label{sec:open-problems-sociotechnical}

Effective practical application of mechanistic interpretability brings both technical challenges and complex social ramifications. It could enable us to act on AI policy and governance, presenting a valuable opportunity to implement regulatory standards and social ideals through technical means (\Cref{subsec:technical-progress}). Such consequential impacts inevitably come with important social and philosophical considerations, which likewise require rigorous inquiry if we are to fully realize the potential benefits of AI (\Cref{subsec:social-philosophical}).

\subsection{Translating technical progress in mechanistic interpretability into levers for AI policy and governance}
\label{subsec:technical-progress}

Current frontier AI governance efforts rarely specify concrete ways in which a mechanistic understanding of AI models might be used to help implementation. For example, OpenAI's Preparedness Framework commits to mitigating biological risks posed by its AI systems \citep{openai2023preparedness}, but the framework lacks details on specific measures that might be taken. Progress in interpretability could potentially enable the removal of any knowledge from the model which could aid users in creating biological weapons \citep{li2024wmdp}. However, it remains uncertain how technical progress will translate into better AI governance, largely due to the numerous open technical problems in mechanistic interpretability. However, there are several promising routes toward better levers for AI policy and governance.

These avenues include assisting companies and governments to identify risks through evaluations and enhancing forecasts about new AI developments; more thorough oversight of AI systems in deployment; simplifying how AI systems operate within existing liability law through clearer explanations of AI decisions; enabling governments to establish risk mitigation regulations and companies to commit to concrete mitigation commitments; and protecting copyright law.

An understanding of model internals can help AI labs assess the risks from frontier models \citep{Chang_2024_SurveyLLM, shevlane2023modelevaluationextremerisks,Casper_2024} and thus better fulfill their obligations under the EU AI Act to “perform model evaluation ... with a view to identifying and mitigating systemic risk” \citep{eu2024ai}. More specifically, a mechanistic understanding of models could help evaluators elicit dangerous capabilities via improved finetuning \citep{ukAISI2024early}, guide their adversarial red-team attempts \citep{tong2024massproducingfailuresmultimodalsystems}, and ensure that AI systems are not intentionally underperforming evaluations \citep{vanderweij2024aisandbagginglanguagemodels}. Advancements in mechanistic interpretability could also assist companies and governments in anticipating when or if AI models will obtain specific dangerous capabilities. Improved forecasting capabilities could enhance threat modeling by reducing the “reasonable disagreement amongst experts over which risks to prioritize” \citep{anthropic_rsp}. For instance, it may help build consensus on whether language models are just stochastic parrots \citep{bender2021dangers} or if they have coherent world models \citep{li2024emergentworldrepresentationsexploring}. We also might be able to use interpretability to build evidence supporting or refuting different models of catastrophic threat, such as determining the validity of mesa-optimization \citep{hubinger2021riskslearnedoptimizationadvanced} or inner alignment concerns \citep{carlsmith2023schemingaisaisfake}. It would also give additional time for companies to prepare adequate risk mitigation measures, \citep{openai2023preparedness}, and for governments to establish appropriate guidance or regulation \citep{ukgov2022ai}.

The EU AI Act mandates that developers of General Purpose AI models with systemic risk have obligations to report incidents involving their system to the AI office. Interpretability tools have the potential to continuously monitor AI inference and detect incidents that require reporting. Compared to incidents in other domains (for instance, nuclear security), AI systems allow us to log all inputs and system states, even those that may lead to catastrophic harm. Access to a small number of such data points may greatly improve our ability to mitigate similar future failures \citep{greenblatt2024catching}. For example, interpretability could be used to investigate the critical “features” of the input that led to the AI incident. This could improve our ability to further red-team the system (\Cref{subsec:better-predictions}) and generate more similar data points that could result in similar incidents. This, in turn, may help us reduce the likelihood of future incidents \citep{chan2024visibilityaiagents}. The incident could be utilized more directly in the construction of test-time monitors to detect similar future incidents (see \citet{roger2023coup}). More speculatively, interpretability could be used to verify companies' compliance with domestic regulation \citep{obrien2023deploymentcorrectionsincidentresponse}, or to authenticate states' compliance with future international treaties regarding the use of AI \citep{aarne2024secure}.

Leveraging a mechanistic understanding of model internals could also make decision rationales for AI model outputs more easily obtainable. This could aid in enforcing citizens' rights under the EU General Data Protection Regulation “to obtain an explanation of the decision reached” by a system “based solely on automated processing” \citep{GDPR_2016, gilpin2019explainingexplanationssociety}. Model editing tools could also resolve problems regarding the copyright status of existing generative models \citep{grynbaum2023times}. According to the US Copyright Act \citep{USCopyright2022}, for any copyrighted work, an artifact “from which the work can be perceived, reproduced, or otherwise communicated…with the aid of a machine or device” is considered a copy of the work \citep{lee2024talkinboutaigeneration}. Interpretability tools could help detect and remove memorized works that can be reproduced verbatim by generative models.

\subsection{Social and philosophical problems in mechanistic interpretability}
\label{subsec:social-philosophical}

The ability to interpret advanced AI systems holds immense potential to advance the science of AI and increase our ability to control it \citep{critch2020airesearchconsiderationshuman,tegmark2023provablysafesystemspath,dalrymple2024guaranteedsafeaiframework}. In this paper, we provide an overview of various interpretability tools that offer novel insights. Nonetheless, interpretability research has thus far produced few tools that are used to make state-of-the-art systems safer in the real world \citep{rauker2023transparentaisurveyinterpreting}. Modern AI systems are still generally trained, evaluated, monitored, and debugged using techniques that do not rely on understanding their internal workings.

The absence of paradigmatic clarity is a major socio-technical factor for this. Questions such as which goals the field of interpretability should pursue, how success should be graded, and how we should define interpretability warrant more thoughtful answers than exist at present. In interpretable AI research, the motivations and methods employed are often described as “diverse and occasionally discordant” \citep{Lipton_2016_InterpretabilityMotivations}. At times, the objective of AI interpretability research is articulated as advancing a fundamental “understanding” or “uncovering the true essence” \citep{christensen2015peering} of what is happening inside black-box models. The intrinsic validity of this paradigm deserves philosophical inquiry. 

However, from an engineer's perspective, pursuing “understanding” without a practical downstream application misses an engineer's objective. Proponents of this view may contend that quantifiable benchmarks linked to concrete practical goals are accurate measures of the success of interpretability. This motivation is concrete and useful, but some have criticized interpretability research as artificially limiting the solution space to engineering problems. When interpretability tools are studied with motivations such as fairness or safety, \citet{krishnan2020against} argues that, “Since addressing these problems need not involve something that looks like an `interpretation' (etc.) of an algorithm, the focus on interpretability artificially constrains the solution space by characterizing one possible solution as the problem itself.” Thus, some argue that interpretability research has failed to produce competitive techniques \citep{rauker2023transparentaisurveyinterpreting,Caspar_2023_RedTeamingDNNFeatureSynthesis}, omitted non-interpretability baselines \citep{rudin2019stop, krishnan2020against}, and graded interpretability tools on their own curve \citep{doshivelez2017rigorousscienceinterpretablemachine, MILLER_2019_Explanation, rauker2023transparentaisurveyinterpreting}. In all safety-relevant applications of mechanistic interpretability, it is important to assess the usefulness of interpretability against alternative methodologies. Failing to do so or misrepresenting these comparisons can lead to follow-up work that rests on false assumptions \citep{lipton2018troublingtrendsmachinelearning, leech2024questionablepracticesmachinelearning}.This is particularly problematic when it impacts the efforts of critical safety work.

Despite these concerns, it is not always imperative for mechanistic interpretability to strictly outperform uninterpretable baselines: It may still be helpful to develop methods that offer interpretability-based advantages, along with the benefits of more competitive uninterpretable methods (e.g. \citet{panickssery2024steeringllama2contrastive}). While interpretability-based methods do not currently outperform black-box baselines, if they perform in a similar ballpark, further progress in mechanistic interpretability could soon lead to better methods, especially in problems which we believe might disproportionately benefit from improved mechanistic understanding (\Cref{sec:open-problems-applications}).

Another potential reason for lack of clarity is that models are often studied in a vacuum: \citet{Smart_2024} emphasize that the usefulness or correctness of model interpretations can depend on the broader context of the model's development or deployment. For example, it may be hard to identify representations of fairness within models, since understanding this requires an understanding of broader contexts. The same data may lead to different conclusions under different definitions of fairness.

Finally, the interpretability community must exercise caution in their communication to minimize potential abuses of the results of its work. Unfortunately, selective transparency can be used to actively mislead \citep{Ananny_2018_LimitationsTransparency}. Furthermore, interpretability is at risk of being used for purposes that might serve corporate interests at the potential expense of safety. The field of AI interpretability is highly influenced by research teams in the private sector. On one hand, industry resources and research contributions have significantly advanced interpretability research. On the other hand, compared to academia, corporations publish selectively, often have financial conflicts of interest, and may provide limited transparency. Meanwhile, the recent definite — but ultimately modest — progress in mechanistic interpretability has been used to lobby against specific AI regulation by falsely claiming that, “Although advocates for AI safety guidelines often allude to the `black box' nature of AI models, where the logic behind their conclusions is not transparent, recent advancements in the AI sector have resolved this issue, thereby ensuring the integrity of open-source code models.” \citep{a16z_2023_lobbying}.

\section{Conclusion}
\label{sec:conclusion}

While mechanistic interpretability has made meaningful progress in both methods and applications, significant challenges remain before we can achieve many of the field's ambitious goals.

The path forward requires progress along multiple axes. We would benefit from stronger theoretical foundations for decomposing neural networks at the joints of their generalization structure. Current methods like sparse dictionary learning, while promising, face both practical limitations in scaling to larger models and deeper conceptual challenges regarding their underlying assumptions. We must also develop more robust methods for validating our interpretations of model behavior, moving beyond correlation-based descriptions to capture true causal mechanisms. Additionally, we need better techniques to understand how mechanisms evolve during training and how they interact to produce complex behaviors.

These methodological advances could unlock several promising applications. Improved interpretability methods could enable more effective monitoring of potential risks, better control over model behavior, and more accurate predictions of capabilities. For AI capabilities, mechanistic understanding could lead to more efficient architectures, better training procedures, and more targeted ways to enhance model performance. In various scientific domains, microscope AI approaches could help extract valuable insights from model internals. To achieve these diverse goals, the field must ensure a focus on generating insights that have real-world utility. This will involve establishing better benchmarks and comparing interpretability-based approaches to non-interpretability baselines. However, fastest progress will likely come from the field pursuing both scientific and engineering goals simultaneously, rather than one at the expense of the other.

The practical impact of progress in mechanistic interpretability extends beyond technical achievements. Interpretability tools could provide crucial mechanisms for governance and oversight. They could help verify compliance with safety standards, detect potential risks before deployment, and provide clearer attribution of model decisions. However, realizing these benefits will require careful attention to the risks of potential misuse and of giving false assurance about AI safety.

Looking toward the future, many expect current AI capabilities to be only a foretaste of what is to come. As AI capabilities advance, the need for a mechanistic understanding of their decision-making processes becomes increasingly urgent. While the black-box nature of AI models remains unresolved, the untapped potential of mechanistic interpretability is what makes it such an exciting research area, and highlights the importance of solving its many open research problems.


\subsubsection*{Author Contributions}
\textbf{LS} managed the project and made major contributions to planning, writing, and editing content, as well as coordinating other contributors and synthesizing their perspectives where necessary. \textbf{BC} substantially contributed to the framing, writing and editing of the final manuscript, and made all of the figures. All authors (\textbf{LS}, \textbf{BC}, \textbf{JBn}, \textbf{JL}, \textbf{JW}, \textbf{LB}, \textbf{NGD}, \textbf{SH}, \textbf{AO}, \textbf{JBm}, \textbf{SB}, \textbf{AGA}, \textbf{AC}, \textbf{NN}, \textbf{MW}, \textbf{NS}, \textbf{JM}, \textbf{EJM}, \textbf{SC}, \textbf{MT}, \textbf{WS}, \textbf{DB}, \textbf{ET}, \textbf{AG}, \textbf{MG}, \textbf{JH}, \textbf{DM}, \textbf{TM}) contributed to the initial writing and editing of various sections and giving feedback on versions of the manuscript.

We also greatly thank Fayth Tan for assistance with editing and Schmidt Sciences for their feedback at various stages of the project and for funding the work.


\bibliographystyle{tmlr}
\bibliography{main}

\begin{thebibliography}{442}
\providecommand{\natexlab}[1]{#1}
\providecommand{\url}[1]{\texttt{#1}}
\expandafter\ifx\csname urlstyle\endcsname\relax
  \providecommand{\doi}[1]{doi: #1}\else
  \providecommand{\doi}{doi: \begingroup \urlstyle{rm}\Url}\fi

\bibitem[GDP(2016)]{GDPR_2016}
Regulation (eu) 2016/679 of the european parliament and of the council of 27 april 2016 on the protection of natural persons with regard to the processing of personal data and on the free movement of such data, and repealing directive 95/46/ec (general data protection regulation) (text with eea relevance), 2016.
\newblock URL \url{https://eur-lex.europa.eu/legal-content/EN/TXT/?uri=CELEX%3A32016R0679}.

\bibitem[Aarne et~al.(2024)Aarne, Fist, and Withers]{aarne2024secure}
Onni Aarne, Tim Fist, and Caleb Withers.
\newblock Secure, governable chips.
\newblock Report, {Center for a New American Security}, 2024.
\newblock URL \url{https://www.cnas.org/publications/reports/secure-governable-chips}.

\bibitem[Abid et~al.(2022)Abid, Yuksekgonul, and Zou]{Abid_2022_counterfactualExplanations}
Abubakar Abid, Mert Yuksekgonul, and James Zou.
\newblock Meaningfully debugging model mistakes using conceptual counterfactual explanations.
\newblock In Kamalika Chaudhuri, Stefanie Jegelka, Le~Song, Csaba Szepesvari, Gang Niu, and Sivan Sabato (eds.), \emph{Proceedings of the 39th International Conference on Machine Learning}, volume 162 of \emph{Proceedings of Machine Learning Research}, pp.\  66--88. PMLR, 17--23 Jul 2022.
\newblock URL \url{https://proceedings.mlr.press/v162/abid22a.html}.

\bibitem[Adebayo et~al.(2018)Adebayo, Gilmer, Muelly, Goodfellow, Hardt, and Kim]{adebayo2020sanitycheckssaliencymaps}
Julius Adebayo, Justin Gilmer, Michael Muelly, Ian Goodfellow, Moritz Hardt, and Been Kim.
\newblock Sanity checks for saliency maps.
\newblock In S.~Bengio, H.~Wallach, H.~Larochelle, K.~Grauman, N.~Cesa-Bianchi, and R.~Garnett (eds.), \emph{Advances in Neural Information Processing Systems}, volume~31. Curran Associates, Inc., 2018.
\newblock URL \url{https://proceedings.neurips.cc/paper_files/paper/2018/file/294a8ed24b1ad22ec2e7efea049b8737-Paper.pdf}.

\bibitem[Adebayo et~al.(2020)Adebayo, Muelly, Liccardi, and Kim]{adebayo2020debuggingtestsmodelexplanations}
Julius Adebayo, Michael Muelly, Ilaria Liccardi, and Been Kim.
\newblock Debugging tests for model explanations.
\newblock In H.~Larochelle, M.~Ranzato, R.~Hadsell, M.F. Balcan, and H.~Lin (eds.), \emph{Advances in Neural Information Processing Systems}, volume~33, pp.\  700--712. Curran Associates, Inc., 2020.
\newblock URL \url{https://proceedings.neurips.cc/paper_files/paper/2020/file/075b051ec3d22dac7b33f788da631fd4-Paper.pdf}.

\bibitem[Adler \& Shavit(2024)Adler and Shavit]{adler2024complexityneuralcomputationsuperposition}
Micah Adler and Nir Shavit.
\newblock On the complexity of neural computation in superposition, 2024.
\newblock URL \url{https://arxiv.org/abs/2409.15318}.

\bibitem[Agarwal et~al.(2024)Agarwal, Tanneru, and Lakkaraju]{agarwal2024faithfulnessvsplausibilityunreliability}
Chirag Agarwal, Sree~Harsha Tanneru, and Himabindu Lakkaraju.
\newblock Faithfulness vs. plausibility: On the (un)reliability of explanations from large language models, 2024.
\newblock URL \url{https://arxiv.org/abs/2402.04614}.

\bibitem[Agarwal et~al.(2021)Agarwal, Melnick, Frosst, Zhang, Lengerich, Caruana, and Hinton]{Agarwal_2020_Additive_Model}
Rishabh Agarwal, Levi Melnick, Nicholas Frosst, Xuezhou Zhang, Ben Lengerich, Rich Caruana, and Geoffrey~E Hinton.
\newblock Neural additive models: Interpretable machine learning with neural nets.
\newblock In M.~Ranzato, A.~Beygelzimer, Y.~Dauphin, P.S. Liang, and J.~Wortman Vaughan (eds.), \emph{Advances in Neural Information Processing Systems}, volume~34, pp.\  4699--4711. Curran Associates, Inc., 2021.
\newblock URL \url{https://proceedings.neurips.cc/paper_files/paper/2021/file/251bd0442dfcc53b5a761e050f8022b8-Paper.pdf}.

\bibitem[Alain \& Bengio(2017)Alain and Bengio]{alain2018understandingintermediatelayersusing}
Guillaume Alain and Yoshua Bengio.
\newblock Understanding intermediate layers using linear classifier probes, 2017.
\newblock URL \url{https://openreview.net/forum?id=ryF7rTqgl}.

\bibitem[Ananny \& Crawford(2018)Ananny and Crawford]{Ananny_2018_LimitationsTransparency}
Mike Ananny and Kate Crawford.
\newblock Seeing without knowing: Limitations of the transparency ideal and its application to algorithmic accountability.
\newblock \emph{New Media \& Society}, 20\penalty0 (3):\penalty0 973--989, 2018.
\newblock \doi{10.1177/1461444816676645}.
\newblock URL \url{https://doi.org/10.1177/1461444816676645}.

\bibitem[Andreas et~al.(2016)Andreas, Rohrbach, Darrell, and Klein]{andreas2017neuralmodulenetworks}
Jacob Andreas, Marcus Rohrbach, Trevor Darrell, and Dan Klein.
\newblock Neural module networks.
\newblock In \emph{Proceedings of the IEEE conference on computer vision and pattern recognition}, pp.\  39--48, 2016.

\bibitem[{Andreessen Horowitz}(2023)]{a16z_2023_lobbying}
{Andreessen Horowitz}.
\newblock Written evidence to the {UK} {House} of {Lords} {Communications} and {Digital} {Select} {Committee} inquiry: {Large} language models ({LLM0114}), 2023.
\newblock URL \url{https://committees.parliament.uk/writtenevidence/127070/pdf/}.
\newblock Written evidence submitted to the UK Parliament.

\bibitem[Anthropic(2024)]{Anthropic}
Anthropic.
\newblock Giving claude a role with a system prompt, 2024.
\newblock URL \url{https://docs.anthropic.com/en/docs/build-with-claude/prompt-engineering/system-prompts}.

\bibitem[{Anthropic}(2024)]{anthropic_rsp}
{Anthropic}.
\newblock Responsible scaling policy.
\newblock Policy document, {Anthropic}, 2024.
\newblock URL \url{https://assets.anthropic.com/m/24a47b00f10301cd/original/Anthropic-Responsible-Scaling-Policy-2024-10-15.pdf}.

\bibitem[Anwar et~al.(2024)Anwar, Saparov, Rando, Paleka, Turpin, Hase, Lubana, Jenner, Casper, Sourbut, Edelman, Zhang, G{\"u}nther, Korinek, Hernandez-Orallo, Hammond, Bigelow, Pan, Langosco, Korbak, Zhang, Zhong, hEigeartaigh, Recchia, Corsi, Chan, Anderljung, Edwards, Petrov, de~Witt, Motwani, Bengio, Chen, Torr, Albanie, Maharaj, Foerster, Tram{\`e}r, He, Kasirzadeh, Choi, and Krueger]{anwar2024foundationalchallengesassuringalignment}
Usman Anwar, Abulhair Saparov, Javier Rando, Daniel Paleka, Miles Turpin, Peter Hase, Ekdeep~Singh Lubana, Erik Jenner, Stephen Casper, Oliver Sourbut, Benjamin~L. Edelman, Zhaowei Zhang, Mario G{\"u}nther, Anton Korinek, Jose Hernandez-Orallo, Lewis Hammond, Eric~J Bigelow, Alexander Pan, Lauro Langosco, Tomasz Korbak, Heidi~Chenyu Zhang, Ruiqi Zhong, Sean~O hEigeartaigh, Gabriel Recchia, Giulio Corsi, Alan Chan, Markus Anderljung, Lilian Edwards, Aleksandar Petrov, Christian~Schroeder de~Witt, Sumeet~Ramesh Motwani, Yoshua Bengio, Danqi Chen, Philip Torr, Samuel Albanie, Tegan Maharaj, Jakob~Nicolaus Foerster, Florian Tram{\`e}r, He~He, Atoosa Kasirzadeh, Yejin Choi, and David Krueger.
\newblock Foundational challenges in assuring alignment and safety of large language models.
\newblock \emph{Transactions on Machine Learning Research}, 2024.
\newblock ISSN 2835-8856.
\newblock URL \url{https://openreview.net/forum?id=oVTkOs8Pka}.
\newblock Survey Certification, Expert Certification.

\bibitem[Arditi et~al.(2024)Arditi, Obeso, Syed, Paleka, Rimsky, Gurnee, and Nanda]{arditi2024refusallanguagemodelsmediated}
Andy Arditi, Oscar~Balcells Obeso, Aaquib Syed, Daniel Paleka, Nina Rimsky, Wes Gurnee, and Neel Nanda.
\newblock Refusal in language models is mediated by a single direction.
\newblock In \emph{The Thirty-eighth Annual Conference on Neural Information Processing Systems}, 2024.
\newblock URL \url{https://openreview.net/forum?id=pH3XAQME6c}.

\bibitem[Ashuach et~al.(2024)Ashuach, Tutek, and Belinkov]{ashuach2024revsunlearningsensitiveinformation}
Tomer Ashuach, Martin Tutek, and Yonatan Belinkov.
\newblock Revs: Unlearning sensitive information in language models via rank editing in the vocabulary space, 2024.
\newblock URL \url{https://arxiv.org/abs/2406.09325}.

\bibitem[Atanasova et~al.(2023)Atanasova, Camburu, Lioma, Lukasiewicz, Simonsen, and Augenstein]{atanasova2023faithfulnesstestsnaturallanguage}
Pepa Atanasova, Oana-Maria Camburu, Christina Lioma, Thomas Lukasiewicz, Jakob~Grue Simonsen, and Isabelle Augenstein.
\newblock Faithfulness tests for natural language explanations.
\newblock In \emph{ACL}, 2023.
\newblock URL \url{https://arxiv.org/abs/2305.18029}.

\bibitem[Ayonrinde et~al.(2024)Ayonrinde, Pearce, and Sharkey]{ayonrinde2024interpretabilitycompressionreconsideringsae}
Kola Ayonrinde, Michael~T. Pearce, and Lee Sharkey.
\newblock Interpretability as compression: Reconsidering sae explanations of neural activations with mdl-saes, 2024.
\newblock URL \url{https://arxiv.org/abs/2410.11179}.

\bibitem[Balesni et~al.(2024{\natexlab{a}})Balesni, Hobbhahn, Lindner, Meinke, Korbak, Clymer, Shlegeris, Scheurer, Stix, Shah, Goldowsky-Dill, Braun, Chughtai, Evans, Kokotajlo, and Bushnaq]{balesni2024evaluationsbasedsafetycasesai}
Mikita Balesni, Marius Hobbhahn, David Lindner, Alexander Meinke, Tomek Korbak, Joshua Clymer, Buck Shlegeris, Jérémy Scheurer, Charlotte Stix, Rusheb Shah, Nicholas Goldowsky-Dill, Dan Braun, Bilal Chughtai, Owain Evans, Daniel Kokotajlo, and Lucius Bushnaq.
\newblock Towards evaluations-based safety cases for ai scheming, 2024{\natexlab{a}}.
\newblock URL \url{https://arxiv.org/abs/2411.03336}.

\bibitem[Balesni et~al.(2024{\natexlab{b}})Balesni, Korbak, and Evans]{balesni2024twohopcursellmstrained}
Mikita Balesni, Tomek Korbak, and Owain Evans.
\newblock The two-hop curse: Llms trained on a->b, b->c fail to learn a-->c, 2024{\natexlab{b}}.
\newblock URL \url{https://arxiv.org/abs/2411.16353}.

\bibitem[Balestriero \& richard baraniuk(2018)Balestriero and richard baraniuk]{balestriero_2018_spline}
Randall Balestriero and richard baraniuk.
\newblock A spline theory of deep learning.
\newblock In Jennifer Dy and Andreas Krause (eds.), \emph{Proceedings of the 35th International Conference on Machine Learning}, volume~80 of \emph{Proceedings of Machine Learning Research}, pp.\  374--383. PMLR, 10--15 Jul 2018.
\newblock URL \url{https://proceedings.mlr.press/v80/balestriero18b.html}.

\bibitem[Bau et~al.(2020)Bau, Zhu, Strobelt, Lapedriza, Zhou, and Torralba]{Bau_2017_NetworkDissection}
David Bau, Jun-Yan Zhu, Hendrik Strobelt, Agata Lapedriza, Bolei Zhou, and Antonio Torralba.
\newblock Understanding the role of individual units in a deep neural network.
\newblock \emph{Proceedings of the National Academy of Sciences}, 2020.
\newblock ISSN 0027-8424.
\newblock \doi{10.1073/pnas.1907375117}.
\newblock URL \url{https://www.pnas.org/content/early/2020/08/31/1907375117}.

\bibitem[Belinkov(2022{\natexlab{a}})]{Belinkov_2022_ProbingClassifiers}
Yonatan Belinkov.
\newblock Probing classifiers: Promises, shortcomings, and advances.
\newblock \emph{Computational Linguistics}, 48\penalty0 (1):\penalty0 207--219, March 2022{\natexlab{a}}.
\newblock \doi{10.1162/coli_a_00422}.
\newblock URL \url{https://aclanthology.org/2022.cl-1.7}.

\bibitem[Belinkov(2022{\natexlab{b}})]{belinkov-2022-probing}
Yonatan Belinkov.
\newblock Probing classifiers: Promises, shortcomings, and advances.
\newblock \emph{Computational Linguistics}, 48\penalty0 (1):\penalty0 207--219, March 2022{\natexlab{b}}.
\newblock \doi{10.1162/coli_a_00422}.
\newblock URL \url{https://aclanthology.org/2022.cl-1.7}.

\bibitem[Belrose et~al.(2023{\natexlab{a}})Belrose, Furman, Smith, Halawi, Ostrovsky, McKinney, Biderman, and Steinhardt]{belrose2023elicitinglatentpredictionstransformers}
Nora Belrose, Zach Furman, Logan Smith, Danny Halawi, Igor Ostrovsky, Lev McKinney, Stella Biderman, and Jacob Steinhardt.
\newblock Eliciting latent predictions from transformers with the tuned lens, 2023{\natexlab{a}}.
\newblock URL \url{https://arxiv.org/abs/2303.08112}.

\bibitem[Belrose et~al.(2023{\natexlab{b}})Belrose, Schneider-Joseph, Ravfogel, Cotterell, Raff, and Biderman]{belrose2023leaceperfectlinearconcept}
Nora Belrose, David Schneider-Joseph, Shauli Ravfogel, Ryan Cotterell, Edward Raff, and Stella Biderman.
\newblock Leace: Perfect linear concept erasure in closed form.
\newblock In A.~Oh, T.~Naumann, A.~Globerson, K.~Saenko, M.~Hardt, and S.~Levine (eds.), \emph{Advances in Neural Information Processing Systems}, volume~36, pp.\  66044--66063. Curran Associates, Inc., 2023{\natexlab{b}}.
\newblock URL \url{https://proceedings.neurips.cc/paper_files/paper/2023/file/d066d21c619d0a78c5b557fa3291a8f4-Paper-Conference.pdf}.

\bibitem[Bender et~al.(2021)Bender, Gebru, McMillan-Major, and Shmitchell]{bender2021dangers}
Emily~M. Bender, Timnit Gebru, Angelina McMillan-Major, and Shmargaret Shmitchell.
\newblock On the dangers of stochastic parrots: Can language models be too big?
\newblock In \emph{Proceedings of the 2021 ACM Conference on Fairness, Accountability, and Transparency}, FAccT '21, pp.\  610--623, New York, NY, USA, 2021. Association for Computing Machinery.
\newblock ISBN 9781450383097.
\newblock \doi{10.1145/3442188.3445922}.
\newblock URL \url{https://doi.org/10.1145/3442188.3445922}.

\bibitem[Bereska \& Gavves(2024)Bereska and Gavves]{Bereska_2024_MechInterpReview}
Leonard Bereska and Stratis Gavves.
\newblock Mechanistic interpretability for {AI} safety - a review.
\newblock \emph{Transactions on Machine Learning Research}, 2024.
\newblock ISSN 2835-8856.
\newblock URL \url{https://openreview.net/forum?id=ePUVetPKu6}.
\newblock Survey Certification, Expert Certification.

\bibitem[Berglund et~al.(2024)Berglund, Tong, Kaufmann, Balesni, Stickland, Korbak, and Evans]{berglund2024reversalcursellmstrained}
Lukas Berglund, Meg Tong, Maximilian Kaufmann, Mikita Balesni, Asa~Cooper Stickland, Tomasz Korbak, and Owain Evans.
\newblock The reversal curse: {LLM}s trained on {\textquotedblleft}a is b{\textquotedblright} fail to learn {\textquotedblleft}b is a{\textquotedblright}.
\newblock In \emph{The Twelfth International Conference on Learning Representations}, 2024.
\newblock URL \url{https://openreview.net/forum?id=GPKTIktA0k}.

\bibitem[Bills et~al.(2023)Bills, Cammarata, Mossing, Tillman, Gao, Goh, Sutskever, Leike, Wu, and Saunders]{bills2023language}
Steven Bills, Nick Cammarata, Dan Mossing, Henk Tillman, Leo Gao, Gabriel Goh, Ilya Sutskever, Jan Leike, Jeff Wu, and William Saunders.
\newblock Language models can explain neurons in language models.
\newblock \url{https://openaipublic.blob.core.windows.net/neuron-explainer/paper/index.html}, 2023.

\bibitem[Biran et~al.(2024)Biran, Gottesman, Yang, Geva, and Globerson]{biran2024hoppinglateexploringlimitations}
Eden Biran, Daniela Gottesman, Sohee Yang, Mor Geva, and Amir Globerson.
\newblock Hopping too late: Exploring the limitations of large language models on multi-hop queries, 2024.
\newblock URL \url{https://arxiv.org/abs/2406.12775}.

\bibitem[Black et~al.(2022)Black, Sharkey, Grinsztajn, Winsor, Braun, Merizian, Parker, Guevara, Millidge, Alfour, and Leahy]{black2022interpretingneuralnetworkspolytope}
Sid Black, Lee Sharkey, Leo Grinsztajn, Eric Winsor, Dan Braun, Jacob Merizian, Kip Parker, Carlos~Ramón Guevara, Beren Millidge, Gabriel Alfour, and Connor Leahy.
\newblock Interpreting neural networks through the polytope lens, 2022.
\newblock URL \url{https://arxiv.org/abs/2211.12312}.

\bibitem[Bloom \& Lin(2024{\natexlab{a}})Bloom and Lin]{Bloom_Lin_2024}
Joseph Bloom and Johnny Lin.
\newblock Understanding sae features with the logit lens - ai alignment forum, Mar 2024{\natexlab{a}}.
\newblock URL \url{https://www.alignmentforum.org/posts/qykrYY6rXXM7EEs8Q/understanding-sae-features-with-the-logit-lens}.

\bibitem[Bloom \& Lin(2024{\natexlab{b}})Bloom and Lin]{bloom2024understanding}
Joseph~Isaac Bloom and Johnny Lin.
\newblock Understanding {SAE} features with the logit lens.
\newblock \emph{Alignment Forum}, March 2024{\natexlab{b}}.
\newblock URL \url{https://www.alignmentforum.org/posts/qykrYY6rXXM7EEs8Q/understanding-sae-features-with-the-logit-lens}.

\bibitem[Bolukbasi et~al.(2021)Bolukbasi, Pearce, Yuan, Coenen, Reif, Viégas, and Wattenberg]{bolukbasi2021interpretabilityillusionbert}
Tolga Bolukbasi, Adam Pearce, Ann Yuan, Andy Coenen, Emily Reif, Fernanda Viégas, and Martin Wattenberg.
\newblock An interpretability illusion for bert, 2021.
\newblock URL \url{https://arxiv.org/abs/2104.07143}.

\bibitem[Borowski et~al.(2021)Borowski, Zimmermann, Schepers, Geirhos, Wallis, Bethge, and Brendel]{borowski2021exemplarynaturalimagesexplain}
Judy Borowski, Roland~Simon Zimmermann, Judith Schepers, Robert Geirhos, Thomas S.~A. Wallis, Matthias Bethge, and Wieland Brendel.
\newblock Exemplary natural images explain {\{}cnn{\}} activations better than state-of-the-art feature visualization.
\newblock In \emph{International Conference on Learning Representations}, 2021.
\newblock URL \url{https://openreview.net/forum?id=QO9-y8also-}.

\bibitem[Braun et~al.(2024)Braun, Taylor, Goldowsky-Dill, and Sharkey]{braun2024identifyingfunctionallyimportantfeatures}
Dan Braun, Jordan Taylor, Nicholas Goldowsky-Dill, and Lee Sharkey.
\newblock Identifying functionally important features with end-to-end sparse dictionary learning.
\newblock In \emph{The Thirty-eighth Annual Conference on Neural Information Processing Systems}, 2024.
\newblock URL \url{https://openreview.net/forum?id=7txPaUpUnc}.

\bibitem[Breiman(1984)]{Breiman_1984_Classification_Regression_Trees}
Leo Breiman.
\newblock \emph{Classification and regression trees}.
\newblock Routledge, 1984.

\bibitem[Brendel \& Bethge(2019)Brendel and Bethge]{brendel2019approximatingcnnsbagoflocalfeaturesmodels}
Wieland Brendel and Matthias Bethge.
\newblock Approximating {CNN}s with bag-of-local-features models works surprisingly well on imagenet.
\newblock In \emph{International Conference on Learning Representations}, 2019.
\newblock URL \url{https://openreview.net/forum?id=SkfMWhAqYQ}.

\bibitem[Bricken(2023)]{Bricken_2023_dictionary}
Trenton Bricken, Oct 2023.
\newblock URL \url{https://transformer-circuits.pub/2023/monosemantic-features/index.html}.

\bibitem[Bricken et~al.(2024)Bricken, Mishra-Sharma, Marcus, Jermyn, Olah, Rivoire, and Henighan]{bricken2024stage}
Trenton Bricken, Siddharth Mishra-Sharma, Jonathan Marcus, Adam Jermyn, Christopher Olah, Kelley Rivoire, and Thomas Henighan.
\newblock Stage-wise model diffing.
\newblock \emph{Transformer Circuits}, 2024.
\newblock URL \url{https://transformer-circuits.pub/2024/model-diffing/index.html}.

\bibitem[Brown et~al.(2020)Brown, Mann, Ryder, Subbiah, Kaplan, Dhariwal, Neelakantan, Shyam, Sastry, Askell, Agarwal, Herbert-Voss, Krueger, Henighan, Child, Ramesh, Ziegler, Wu, Winter, Hesse, Chen, Sigler, Litwin, Gray, Chess, Clark, Berner, McCandlish, Radford, Sutskever, and Amodei]{Brown_2020_GPT3}
Tom~B. Brown, Benjamin Mann, Nick Ryder, Melanie Subbiah, Jared Kaplan, Prafulla Dhariwal, Arvind Neelakantan, Pranav Shyam, Girish Sastry, Amanda Askell, Sandhini Agarwal, Ariel Herbert-Voss, Gretchen Krueger, Tom Henighan, Rewon Child, Aditya Ramesh, Daniel~M. Ziegler, Jeffrey Wu, Clemens Winter, Christopher Hesse, Mark Chen, Eric Sigler, Mateusz Litwin, Scott Gray, Benjamin Chess, Jack Clark, Christopher Berner, Sam McCandlish, Alec Radford, Ilya Sutskever, and Dario Amodei.
\newblock Language models are few-shot learners, 2020.
\newblock URL \url{https://arxiv.org/abs/2005.14165}.

\bibitem[Burns et~al.(2023)Burns, Ye, Klein, and Steinhardt]{burns2024discoveringlatentknowledgelanguage}
Collin Burns, Haotian Ye, Dan Klein, and Jacob Steinhardt.
\newblock Discovering latent knowledge in language models without supervision.
\newblock In \emph{The Eleventh International Conference on Learning Representations}, 2023.
\newblock URL \url{https://openreview.net/forum?id=ETKGuby0hcs}.

\bibitem[Bushnaq \& Mendel(2024)Bushnaq and Mendel]{bushnaq2024circuits}
Lucius Bushnaq and Jake Mendel.
\newblock Circuits in superposition: Compressing many small neural networks into one.
\newblock \emph{Alignment Forum}, oct 2024.
\newblock URL \url{https://www.alignmentforum.org/posts/roE7SHjFWEoMcGZKd/circuits-in-superposition-compressing-many-small-neural}.

\bibitem[Bushnaq et~al.(2024)Bushnaq, Heimersheim, Goldowsky-Dill, Braun, Mendel, Hänni, Griffin, Stöhler, Wache, and Hobbhahn]{bushnaq2024localinteractionbasisidentifying}
Lucius Bushnaq, Stefan Heimersheim, Nicholas Goldowsky-Dill, Dan Braun, Jake Mendel, Kaarel Hänni, Avery Griffin, Jörn Stöhler, Magdalena Wache, and Marius Hobbhahn.
\newblock The local interaction basis: Identifying computationally-relevant and sparsely interacting features in neural networks, 2024.
\newblock URL \url{https://arxiv.org/abs/2405.10928}.

\bibitem[Bussman et~al.(2024)Bussman, Pearce, Leask, Bloom, Sharkey, and Nanda]{bussman2024metasaes}
Bart Bussman, Michael Pearce, Patrick Leask, Joseph Bloom, Lee Sharkey, and Neel Nanda.
\newblock Showing sae latents are not atomic using meta-saes, Aug 2024.
\newblock URL \url{https://www.lesswrong.com/posts/TMAmHh4DdMr4nCSr5/showing-sae-latents-are-not-atomic-using-meta-saes}.

\bibitem[Cajal(1924)]{cajal1924estructura}
S.R. Cajal.
\newblock \emph{Estructura de los centros nerviosos de las aves (1888)}.
\newblock Jim{\'e}nez y Molina, 1924.
\newblock URL \url{https://books.google.co.uk/books?id=SXr8sgEACAAJ}.

\bibitem[Cammarata et~al.(2020)Cammarata, Goh, Carter, Schubert, Petrov, and Olah]{cammarata2020curve}
Nick Cammarata, Gabriel Goh, Shan Carter, Ludwig Schubert, Michael Petrov, and Chris Olah.
\newblock Curve detectors.
\newblock \emph{Distill}, 2020.
\newblock \doi{10.23915/distill.00024.003}.
\newblock https://distill.pub/2020/circuits/curve-detectors.

\bibitem[Cao \& Yang(2015)Cao and Yang]{Cao_2015_MachineUnlearning}
Yinzhi Cao and Junfeng Yang.
\newblock Towards making systems forget with machine unlearning.
\newblock In \emph{2015 IEEE Symposium on Security and Privacy}, pp.\  463--480, 2015.
\newblock \doi{10.1109/SP.2015.35}.

\bibitem[Carlsmith(2023)]{carlsmith2023schemingaisaisfake}
Joe Carlsmith.
\newblock Scheming ais: Will ais fake alignment during training in order to get power?, 2023.
\newblock URL \url{https://arxiv.org/abs/2311.08379}.

\bibitem[Carter \& Nielsen(2017)Carter and Nielsen]{carter2017using}
Shan Carter and Michael Nielsen.
\newblock Using artificial intelligence to augment human intelligence.
\newblock \emph{Distill}, 2017.
\newblock \doi{10.23915/distill.00009}.
\newblock https://distill.pub/2017/aia.

\bibitem[Carter et~al.(2019)Carter, Armstrong, Schubert, Johnson, and Olah]{carter2019activation}
Shan Carter, Zan Armstrong, Ludwig Schubert, Ian Johnson, and Chris Olah.
\newblock Activation atlas.
\newblock \emph{Distill}, 2019.
\newblock \doi{10.23915/distill.00015}.
\newblock https://distill.pub/2019/activation-atlas.

\bibitem[Casper et~al.(2022{\natexlab{a}})Casper, Nadeau, Hadfield-Menell, and Kreiman]{Caspar_2022_featurelevel}
Stephen Casper, Max Nadeau, Dylan Hadfield-Menell, and Gabriel Kreiman.
\newblock Robust feature-level adversaries are interpretability tools.
\newblock In S.~Koyejo, S.~Mohamed, A.~Agarwal, D.~Belgrave, K.~Cho, and A.~Oh (eds.), \emph{Advances in Neural Information Processing Systems}, volume~35, pp.\  33093--33106. Curran Associates, Inc., 2022{\natexlab{a}}.
\newblock URL \url{https://proceedings.neurips.cc/paper_files/paper/2022/file/d616a353c711f11c722e3f28d2d9e956-Paper-Conference.pdf}.

\bibitem[Casper et~al.(2022{\natexlab{b}})Casper, Nadeau, Hadfield-Menell, and Kreiman]{Caspar_2023_RedTeamingDNNFeatureSynthesis}
Stephen Casper, Max Nadeau, Dylan Hadfield-Menell, and Gabriel Kreiman.
\newblock Robust feature-level adversaries are interpretability tools.
\newblock In S.~Koyejo, S.~Mohamed, A.~Agarwal, D.~Belgrave, K.~Cho, and A.~Oh (eds.), \emph{Advances in Neural Information Processing Systems}, volume~35, pp.\  33093--33106. Curran Associates, Inc., 2022{\natexlab{b}}.
\newblock URL \url{https://proceedings.neurips.cc/paper_files/paper/2022/file/d616a353c711f11c722e3f28d2d9e956-Paper-Conference.pdf}.

\bibitem[Casper et~al.(2023{\natexlab{a}})Casper, Bu, Li, Li, Zhang, Hariharan, and Hadfield-Menell]{casper2023redteamingdeepneural}
Stephen Casper, Tong Bu, Yuxiao Li, Jiawei Li, Kevin Zhang, Kaivalya Hariharan, and Dylan Hadfield-Menell.
\newblock Red teaming deep neural networks with feature synthesis tools.
\newblock In \emph{Thirty-seventh Conference on Neural Information Processing Systems}, 2023{\natexlab{a}}.
\newblock URL \url{https://openreview.net/forum?id=Od6CHhPM7I}.

\bibitem[Casper et~al.(2023{\natexlab{b}})Casper, Davies, Shi, Gilbert, Scheurer, Rando, Freedman, Korbak, Lindner, Freire, Wang, Marks, Segerie, Carroll, Peng, Christoffersen, Damani, Slocum, Anwar, Siththaranjan, Nadeau, Michaud, Pfau, Krasheninnikov, Chen, Langosco, Hase, Biyik, Dragan, Krueger, Sadigh, and Hadfield-Menell]{casper2023openproblemsfundamentallimitations}
Stephen Casper, Xander Davies, Claudia Shi, Thomas~Krendl Gilbert, J{\'e}r{\'e}my Scheurer, Javier Rando, Rachel Freedman, Tomek Korbak, David Lindner, Pedro Freire, Tony~Tong Wang, Samuel Marks, Charbel-Raphael Segerie, Micah Carroll, Andi Peng, Phillip~J.K. Christoffersen, Mehul Damani, Stewart Slocum, Usman Anwar, Anand Siththaranjan, Max Nadeau, Eric~J Michaud, Jacob Pfau, Dmitrii Krasheninnikov, Xin Chen, Lauro Langosco, Peter Hase, Erdem Biyik, Anca Dragan, David Krueger, Dorsa Sadigh, and Dylan Hadfield-Menell.
\newblock Open problems and fundamental limitations of reinforcement learning from human feedback.
\newblock \emph{Transactions on Machine Learning Research}, 2023{\natexlab{b}}.
\newblock ISSN 2835-8856.
\newblock URL \url{https://openreview.net/forum?id=bx24KpJ4Eb}.
\newblock Survey Certification, Featured Certification.

\bibitem[Casper et~al.(2023{\natexlab{c}})Casper, Hariharan, and Hadfield-Menell]{casper2023diagnosticsdeepneuralnetworks}
Stephen Casper, Kaivalya Hariharan, and Dylan Hadfield-Menell.
\newblock Diagnostics for deep neural networks with automated copy/paste attacks, 2023{\natexlab{c}}.
\newblock URL \url{https://arxiv.org/abs/2211.10024}.

\bibitem[Casper et~al.(2024)Casper, Ezell, Siegmann, Kolt, Curtis, Bucknall, Haupt, Wei, Scheurer, Hobbhahn, Sharkey, Krishna, Von~Hagen, Alberti, Chan, Sun, Gerovitch, Bau, Tegmark, Krueger, and Hadfield-Menell]{Casper_2024}
Stephen Casper, Carson Ezell, Charlotte Siegmann, Noam Kolt, Taylor~Lynn Curtis, Benjamin Bucknall, Andreas Haupt, Kevin Wei, Jérémy Scheurer, Marius Hobbhahn, Lee Sharkey, Satyapriya Krishna, Marvin Von~Hagen, Silas Alberti, Alan Chan, Qinyi Sun, Michael Gerovitch, David Bau, Max Tegmark, David Krueger, and Dylan Hadfield-Menell.
\newblock Black-box access is insufficient for rigorous ai audits.
\newblock In \emph{The 2024 ACM Conference on Fairness, Accountability, and Transparency}, volume~35 of \emph{FAccT '24}, pp.\  2254--2272. ACM, June 2024.
\newblock \doi{10.1145/3630106.3659037}.
\newblock URL \url{http://dx.doi.org/10.1145/3630106.3659037}.

\bibitem[Chalnev et~al.(2024)Chalnev, Siu, and Conmy]{chalnev2024improvingsteeringvectorstargeting}
Sviatoslav Chalnev, Matthew Siu, and Arthur Conmy.
\newblock Improving steering vectors by targeting sparse autoencoder features, 2024.
\newblock URL \url{https://arxiv.org/abs/2411.02193}.

\bibitem[Chan et~al.(2024)Chan, Ezell, Kaufmann, Wei, Hammond, Bradley, Bluemke, Rajkumar, Krueger, Kolt, Heim, and Anderljung]{chan2024visibilityaiagents}
Alan Chan, Carson Ezell, Max Kaufmann, Kevin Wei, Lewis Hammond, Herbie Bradley, Emma Bluemke, Nitarshan Rajkumar, David Krueger, Noam Kolt, Lennart Heim, and Markus Anderljung.
\newblock Visibility into ai agents.
\newblock In \emph{Proceedings of the 2024 ACM Conference on Fairness, Accountability, and Transparency}, FAccT '24, pp.\  958–973, New York, NY, USA, 2024. Association for Computing Machinery.
\newblock ISBN 9798400704505.
\newblock \doi{10.1145/3630106.3658948}.
\newblock URL \url{https://doi.org/10.1145/3630106.3658948}.

\bibitem[Chan et~al.(2022{\natexlab{a}})Chan, Garriga-alonso, and Goldowsky-Dill]{Chan_Garriga-alonso_Goldowsky-Dill_2022}
Lawrence Chan, Adrià Garriga-alonso, and Nicholas Goldowsky-Dill.
\newblock Causal scrubbing: A method for rigorously testing interpretability hypotheses [redwood research] - ai alignment forum, Oct 2022{\natexlab{a}}.
\newblock URL \url{https://www.alignmentforum.org/posts/JvZhhzycHu2Yd57RN/causal-scrubbing-a-method-for-rigorously-testing}.

\bibitem[Chan et~al.(2022{\natexlab{b}})Chan, Garriga-alonso, Goldowsky-Dill, Greenblatt, Lin, Nitishinskaya, Radhakrishnan, Shlegeris, and Thomas]{Chan_2022_InductionHeads}
Lawrence Chan, Adrià Garriga-alonso, Nicholas Goldowsky-Dill, Ryan Greenblatt, Tao Lin, Jenny Nitishinskaya, Ansh Radhakrishnan, Buck Shlegeris, and Nate Thomas.
\newblock Causal scrubbing: Results on induction heads - ai alignment forum, Dec 2022{\natexlab{b}}.
\newblock URL \url{https://www.alignmentforum.org/posts/j6s9H9SHrEhEfuJnq/causal-scrubbing-results-on-induction-heads}.

\bibitem[Chan et~al.(2023)Chan, Garriga-Alonso, Goldowsky-Dill, Greenblatt, Lin, Nitishinskaya, Radhakrishnan, Shlegeris, and Thomas]{chan2023causal}
Lawrence Chan, Adri{`a} Garriga-Alonso, Nicholas Goldowsky-Dill, Ryan Greenblatt, Tao Lin, Jenny Nitishinskaya, Ansh Radhakrishnan, Buck Shlegeris, and Nate Thomas.
\newblock Causal scrubbing: results on a paren balance checker.
\newblock \emph{Alignment Forum}, 2023.
\newblock URL \url{https://www.alignmentforum.org/s/h95ayYYwMebGEYN5y/p/kjudfaQazMmC74SbF}.

\bibitem[Chang et~al.(2024)Chang, Wang, Wang, Wu, Yang, Zhu, Chen, Yi, Wang, Wang, Ye, Zhang, Chang, Yu, Yang, and Xie]{Chang_2024_SurveyLLM}
Yupeng Chang, Xu~Wang, Jindong Wang, Yuan Wu, Linyi Yang, Kaijie Zhu, Hao Chen, Xiaoyuan Yi, Cunxiang Wang, Yidong Wang, Wei Ye, Yue Zhang, Yi~Chang, Philip~S. Yu, Qiang Yang, and Xing Xie.
\newblock A survey on evaluation of large language models.
\newblock \emph{ACM Trans. Intell. Syst. Technol.}, 15\penalty0 (3), mar 2024.
\newblock ISSN 2157-6904.
\newblock \doi{10.1145/3641289}.
\newblock URL \url{https://doi.org/10.1145/3641289}.

\bibitem[Chanin et~al.(2024)Chanin, Wilken-Smith, Dulka, Bhatnagar, and Bloom]{chanin2024absorptionstudyingfeaturesplitting}
David Chanin, James Wilken-Smith, Tomáš Dulka, Hardik Bhatnagar, and Joseph Bloom.
\newblock A is for absorption: Studying feature splitting and absorption in sparse autoencoders, 2024.
\newblock URL \url{https://arxiv.org/abs/2409.14507}.

\bibitem[Chen et~al.(2025)Chen, Vondrick, and Mao]{chen2024selfieselfinterpretationlargelanguage}
Haozhe Chen, Carl Vondrick, and Chengzhi Mao.
\newblock Selfie: self-interpretation of large language model embeddings.
\newblock In \emph{Proceedings of the 41st International Conference on Machine Learning}, ICML'24. JMLR.org, 2025.

\bibitem[Chen et~al.(2018)Chen, Li, Grosse, and Duvenaud]{chen2019isolatingsourcesdisentanglementvariational}
Ricky T.~Q. Chen, Xuechen Li, Roger~B Grosse, and David~K Duvenaud.
\newblock Isolating sources of disentanglement in variational autoencoders.
\newblock In S.~Bengio, H.~Wallach, H.~Larochelle, K.~Grauman, N.~Cesa-Bianchi, and R.~Garnett (eds.), \emph{Advances in Neural Information Processing Systems}, volume~31. Curran Associates, Inc., 2018.
\newblock URL \url{https://proceedings.neurips.cc/paper_files/paper/2018/file/1ee3dfcd8a0645a25a35977997223d22-Paper.pdf}.

\bibitem[Chen et~al.(2024)Chen, Wu, DePodesta, Yeh, Li, Marin, Patel, Riecke, Raval, Seow, Wattenberg, and Viégas]{chen2024designingdashboardtransparencycontrol}
Yida Chen, Aoyu Wu, Trevor DePodesta, Catherine Yeh, Kenneth Li, Nicholas~Castillo Marin, Oam Patel, Jan Riecke, Shivam Raval, Olivia Seow, Martin Wattenberg, and Fernanda Viégas.
\newblock Designing a dashboard for transparency and control of conversational ai, 2024.
\newblock URL \url{https://arxiv.org/abs/2406.07882}.

\bibitem[Christensen \& Cheney(2015)Christensen and Cheney]{christensen2015peering}
Lars~Th{\o}ger Christensen and George Cheney.
\newblock Peering into transparency: Challenging ideals, proxies, and organizational practices.
\newblock \emph{Communication Theory}, 25\penalty0 (1):\penalty0 70--90, February 2015.
\newblock \doi{10.1111/comt.12052}.
\newblock URL \url{https://doi.org/10.1111/comt.12052}.

\bibitem[Christiano(2022)]{Christiano_2022_MAD}
Paul~F Christiano.
\newblock Can we efficiently distinguish different mechanisms?, Dec 2022.
\newblock URL \url{https://www.alignmentforum.org/posts/JLyWP2Y9LAruR2gi9/can-we-efficiently-distinguish-different-mechanisms}.

\bibitem[Christiano et~al.(2017)Christiano, Leike, Brown, Martic, Legg, and Amodei]{christiano2023deepreinforcementlearninghuman}
Paul~F Christiano, Jan Leike, Tom Brown, Miljan Martic, Shane Legg, and Dario Amodei.
\newblock Deep reinforcement learning from human preferences.
\newblock In I.~Guyon, U.~Von Luxburg, S.~Bengio, H.~Wallach, R.~Fergus, S.~Vishwanathan, and R.~Garnett (eds.), \emph{Advances in Neural Information Processing Systems}, volume~30. Curran Associates, Inc., 2017.
\newblock URL \url{https://proceedings.neurips.cc/paper_files/paper/2017/file/d5e2c0adad503c91f91df240d0cd4e49-Paper.pdf}.

\bibitem[Chughtai \& Bushnaq(2025)Chughtai and Bushnaq]{chughtai2025activation}
Bilal Chughtai and Lucius Bushnaq.
\newblock Activation space interpretability may be doomed.
\newblock \emph{LessWrong}, 1 2025.
\newblock URL \url{https://www.lesswrong.com/posts/gYfpPbww3wQRaxAFD/activation-space-interpretability-may-be-doomed}.

\bibitem[Chughtai \& Lau(2024)Chughtai and Lau]{chughtai2024understanding}
Bilal Chughtai and Yeu-Tong Lau.
\newblock Understanding positional features in layer 0 {SAE}s.
\newblock \emph{LessWrong}, 2024.
\newblock URL \url{https://www.lesswrong.com/posts/ctGeJGHg9pbc8memF/understanding-positional-features-in-layer-0-saes}.

\bibitem[Chughtai et~al.(2023)Chughtai, Chan, and Nanda]{chughtai_2023_toyModel}
Bilal Chughtai, Lawrence Chan, and Neel Nanda.
\newblock A toy model of universality: Reverse engineering how networks learn group operations.
\newblock In Andreas Krause, Emma Brunskill, Kyunghyun Cho, Barbara Engelhardt, Sivan Sabato, and Jonathan Scarlett (eds.), \emph{Proceedings of the 40th International Conference on Machine Learning}, volume 202 of \emph{Proceedings of Machine Learning Research}, pp.\  6243--6267. PMLR, 23--29 Jul 2023.
\newblock URL \url{https://proceedings.mlr.press/v202/chughtai23a.html}.

\bibitem[Churchland \& Shenoy(2007)Churchland and Shenoy]{churchland2007temporal}
Mark~M. Churchland and Krishna~V. Shenoy.
\newblock Temporal complexity and heterogeneity of single-neuron activity in premotor and motor cortex.
\newblock \emph{Journal of Neurophysiology}, 97\penalty0 (6):\penalty0 4235--4257, 6 2007.
\newblock \doi{10.1152/jn.00095.2007}.
\newblock URL \url{https://journals.physiology.org/doi/full/10.1152/jn.00095.2007}.
\newblock PubMed ID: 17376854.

\bibitem[Cloud et~al.(2024)Cloud, Goldman-Wetzler, Wybitul, Miller, and Turner]{cloud2024gradientroutingmaskinggradients}
Alex Cloud, Jacob Goldman-Wetzler, Evžen Wybitul, Joseph Miller, and Alexander~Matt Turner.
\newblock Gradient routing: Masking gradients to localize computation in neural networks, 2024.
\newblock URL \url{https://arxiv.org/abs/2410.04332}.

\bibitem[Clymer et~al.(2024)Clymer, Gabrieli, Krueger, and Larsen]{clymer2024safetycasesjustifysafety}
Joshua Clymer, Nick Gabrieli, David Krueger, and Thomas Larsen.
\newblock Safety cases: How to justify the safety of advanced ai systems, 2024.
\newblock URL \url{https://arxiv.org/abs/2403.10462}.

\bibitem[Colognese \& Jose(2023)Colognese and Jose]{colognese2023high}
Paul Colognese and Arun Jose.
\newblock High-level interpretability: detecting an {AI}'s objectives.
\newblock \emph{Alignment Forum}, September 2023.
\newblock URL \url{https://www.alignmentforum.org/posts/tFYGdq9ivjA3rdaS2/high-level-interpretability-detecting-an-ai-s-objectives}.

\bibitem[Conmy \& Nanda(2024)Conmy and Nanda]{conmy2024activation}
Arthur Conmy and Neel Nanda.
\newblock Activation steering with {SAEs}.
\newblock \emph{Alignment Forum}, 2024.
\newblock URL \url{https://www.alignmentforum.org/posts/C5KAZQib3bzzpeyrg/full-post-progress-update-1-from-the-gdm-mech-interp-team#Activation_Steering_with_SAEs}.

\bibitem[Conmy et~al.(2023)Conmy, Mavor-Parker, Lynch, Heimersheim, and Garriga-Alonso]{conmy2023automatedcircuitdiscoverymechanistic}
Arthur Conmy, Augustine Mavor-Parker, Aengus Lynch, Stefan Heimersheim, and Adri\`{a} Garriga-Alonso.
\newblock Towards automated circuit discovery for mechanistic interpretability.
\newblock In A.~Oh, T.~Naumann, A.~Globerson, K.~Saenko, M.~Hardt, and S.~Levine (eds.), \emph{Advances in Neural Information Processing Systems}, volume~36, pp.\  16318--16352. Curran Associates, Inc., 2023.
\newblock URL \url{https://proceedings.neurips.cc/paper_files/paper/2023/file/34e1dbe95d34d7ebaf99b9bcaeb5b2be-Paper-Conference.pdf}.

\bibitem[Conneau et~al.(2018)Conneau, Kruszewski, Lample, Barrault, and Baroni]{conneau2018cramsinglevectorprobing}
Alexis Conneau, German Kruszewski, Guillaume Lample, Lo{\"i}c Barrault, and Marco Baroni.
\newblock What you can cram into a single {\$}{\&}!{\#}* vector: Probing sentence embeddings for linguistic properties.
\newblock In Iryna Gurevych and Yusuke Miyao (eds.), \emph{Proceedings of the 56th Annual Meeting of the Association for Computational Linguistics (Volume 1: Long Papers)}, pp.\  2126--2136, Melbourne, Australia, July 2018. Association for Computational Linguistics.
\newblock \doi{10.18653/v1/P18-1198}.
\newblock URL \url{https://aclanthology.org/P18-1198/}.

\bibitem[Cooper et~al.(2022)Cooper, Um, Arandjelović, and Harrison]{cooper2022lymphocyte}
Jack Cooper, In~Hwa Um, Ognjen Arandjelović, and David~J Harrison.
\newblock Lymphocyte classification from hoechst stained slides with deep learning.
\newblock \emph{Cancers}, 14\penalty0 (23):\penalty0 5957, Dec 2022.
\newblock \doi{10.3390/cancers14235957}.
\newblock URL \url{https://doi.org/10.3390/cancers14235957}.
\newblock PMID: 36497439; PMCID: PMC9738034.

\bibitem[{Council of the European Union}(2024)]{eu2024ai}
{Council of the European Union}.
\newblock Proposal for a {Regulation} of the {European Parliament} and of the {Council} laying down harmonised rules on artificial intelligence ({Artificial Intelligence Act}) and amending certain {Union} legislative acts, 2024.
\newblock URL \url{https://data.consilium.europa.eu/doc/document/ST-5662-2024-INIT/en/pdf}.

\bibitem[Critch \& Krueger(2020)Critch and Krueger]{critch2020airesearchconsiderationshuman}
Andrew Critch and David Krueger.
\newblock Ai research considerations for human existential safety (arches), 2020.
\newblock URL \url{https://arxiv.org/abs/2006.04948}.

\bibitem[Crowson et~al.(2022)Crowson, Biderman, Kornis, Stander, Hallahan, Castricato, and Raff]{crowson2022vqganclipopendomainimage}
Katherine Crowson, Stella Biderman, Daniel Kornis, Dashiell Stander, Eric Hallahan, Louis Castricato, and Edward Raff.
\newblock Vqgan-clip: Open domain image generation and editing with natural language guidance.
\newblock In Shai Avidan, Gabriel Brostow, Moustapha Ciss{\'e}, Giovanni~Maria Farinella, and Tal Hassner (eds.), \emph{Computer Vision -- ECCV 2022}, pp.\  88--105, Cham, 2022. Springer Nature Switzerland.
\newblock ISBN 978-3-031-19836-6.

\bibitem[Csisz{\'a}rik et~al.(2021)Csisz{\'a}rik, K{\H{o}}r{\"o}si-Szab{\'o}, Matszangosz, Papp, and Varga]{csiszarik2021similarity}
Adri{\'a}n Csisz{\'a}rik, P{\'e}ter K{\H{o}}r{\"o}si-Szab{\'o}, {\'A}kos~K. Matszangosz, Gergely Papp, and D{\'a}niel Varga.
\newblock Similarity and matching of neural network representations.
\newblock In A.~Beygelzimer, Y.~Dauphin, P.~Liang, and J.~Wortman Vaughan (eds.), \emph{Advances in Neural Information Processing Systems}, 2021.
\newblock URL \url{https://openreview.net/forum?id=aedFIIRRfXr}.

\bibitem[Csord{\'a}s et~al.(2021)Csord{\'a}s, van Steenkiste, and Schmidhuber]{csordás2021neuralnetsmodularinspecting}
R{\'o}bert Csord{\'a}s, Sjoerd van Steenkiste, and J{\"u}rgen Schmidhuber.
\newblock Are neural nets modular? inspecting functional modularity through differentiable weight masks.
\newblock In \emph{International Conference on Learning Representations}, 2021.
\newblock URL \url{https://openreview.net/forum?id=7uVcpu-gMD}.

\bibitem[Csordás et~al.(2024)Csordás, Potts, Manning, and Geiger]{csordás2024recurrentneuralnetworkslearn}
Róbert Csordás, Christopher Potts, Christopher~D. Manning, and Atticus Geiger.
\newblock Recurrent neural networks learn to store and generate sequences using non-linear representations, 2024.
\newblock URL \url{https://arxiv.org/abs/2408.10920}.

\bibitem[Dalrymple et~al.(2024)Dalrymple, Skalse, Bengio, Russell, Tegmark, Seshia, Omohundro, Szegedy, Goldhaber, Ammann, Abate, Halpern, Barrett, Zhao, Zhi-Xuan, Wing, and Tenenbaum]{dalrymple2024guaranteedsafeaiframework}
David Dalrymple, Joar Skalse, Yoshua Bengio, Stuart Russell, Max Tegmark, Sanjit Seshia, Steve Omohundro, Christian Szegedy, Ben Goldhaber, Nora Ammann, Alessandro Abate, Joe Halpern, Clark Barrett, Ding Zhao, Tan Zhi-Xuan, Jeannette Wing, and Joshua Tenenbaum.
\newblock Towards guaranteed safe ai: A framework for ensuring robust and reliable ai systems, 2024.
\newblock URL \url{https://arxiv.org/abs/2405.06624}.

\bibitem[Dalvi et~al.(2019)Dalvi, Durrani, Sajjad, Belinkov, Bau, and Glass]{dalvi2018grainsanddesertanalyzing}
Fahim Dalvi, Nadir Durrani, Hassan Sajjad, Yonatan Belinkov, Anthony Bau, and James Glass.
\newblock What is one grain of sand in the desert? analyzing individual neurons in deep nlp models.
\newblock In \emph{Proceedings of the Thirty-Third AAAI Conference on Artificial Intelligence and Thirty-First Innovative Applications of Artificial Intelligence Conference and Ninth AAAI Symposium on Educational Advances in Artificial Intelligence}, AAAI'19/IAAI'19/EAAI'19. AAAI Press, 2019.
\newblock ISBN 978-1-57735-809-1.
\newblock \doi{10.1609/aaai.v33i01.33016309}.
\newblock URL \url{https://doi.org/10.1609/aaai.v33i01.33016309}.

\bibitem[Dar et~al.(2023)Dar, Geva, Gupta, and Berant]{dar-etal-2023-analyzing}
Guy Dar, Mor Geva, Ankit Gupta, and Jonathan Berant.
\newblock Analyzing transformers in embedding space.
\newblock In Anna Rogers, Jordan Boyd-Graber, and Naoaki Okazaki (eds.), \emph{Proceedings of the 61st Annual Meeting of the Association for Computational Linguistics (Volume 1: Long Papers)}, pp.\  16124--16170, Toronto, Canada, July 2023. Association for Computational Linguistics.
\newblock \doi{10.18653/v1/2023.acl-long.893}.
\newblock URL \url{https://aclanthology.org/2023.acl-long.893}.

\bibitem[Davies \& Khakzar(2024)Davies and Khakzar]{davies2024cognitiverevolutioninterpretabilityexplaining}
Adam Davies and Ashkan Khakzar.
\newblock The cognitive revolution in interpretability: From explaining behavior to interpreting representations and algorithms, 2024.
\newblock URL \url{https://arxiv.org/abs/2408.05859}.

\bibitem[Davies et~al.(2023)Davies, Nadeau, Prakash, Shaham, and Bau]{davies2023discoveringvariablebindingcircuitry}
Xander Davies, Max Nadeau, Nikhil Prakash, Tamar~Rott Shaham, and David Bau.
\newblock Discovering variable binding circuitry with desiderata, 2023.
\newblock URL \url{https://arxiv.org/abs/2307.03637}.

\bibitem[De~Cao et~al.(2020)De~Cao, Schlichtkrull, Aziz, and Titov]{de-cao-etal-2020-decisions}
Nicola De~Cao, Michael~Sejr Schlichtkrull, Wilker Aziz, and Ivan Titov.
\newblock How do decisions emerge across layers in neural models? interpretation with differentiable masking.
\newblock In Bonnie Webber, Trevor Cohn, Yulan He, and Yang Liu (eds.), \emph{Proceedings of the 2020 Conference on Empirical Methods in Natural Language Processing (EMNLP)}, pp.\  3243--3255, Online, November 2020. Association for Computational Linguistics.
\newblock \doi{10.18653/v1/2020.emnlp-main.262}.
\newblock URL \url{https://aclanthology.org/2020.emnlp-main.262}.

\bibitem[Deeb \& Roger(2024)Deeb and Roger]{deeb2024unlearningmethodsremoveinformation}
Aghyad Deeb and Fabien Roger.
\newblock Do unlearning methods remove information from language model weights?, 2024.
\newblock URL \url{https://arxiv.org/abs/2410.08827}.

\bibitem[Denain \& Steinhardt(2023)Denain and Steinhardt]{denain2023auditingvisualizationstransparencymethods}
Jean-Stanislas Denain and Jacob Steinhardt.
\newblock Auditing visualizations: Transparency methods struggle to detect anomalous behavior, 2023.
\newblock URL \url{https://arxiv.org/abs/2206.13498}.

\bibitem[Devlin(2018)]{devlin2019bertpretrainingdeepbidirectional}
Jacob Devlin.
\newblock Bert: Pre-training of deep bidirectional transformers for language understanding.
\newblock \emph{arXiv preprint arXiv:1810.04805}, 2018.

\bibitem[Din et~al.(2024)Din, Karidi, Choshen, and Geva]{din2024jumpconclusionsshortcuttingtransformers}
Alexander~Yom Din, Taelin Karidi, Leshem Choshen, and Mor Geva.
\newblock Jump to conclusions: Short-cutting transformers with linear transformations, 2024.
\newblock URL \url{https://arxiv.org/abs/2303.09435}.

\bibitem[Dombrowski et~al.(2019)Dombrowski, Alber, Anders, Ackermann, M\"{u}ller, and Kessel]{Dombrowski_2019_advances}
Ann-Kathrin Dombrowski, Maximillian Alber, Christopher Anders, Marcel Ackermann, Klaus-Robert M\"{u}ller, and Pan Kessel.
\newblock Explanations can be manipulated and geometry is to blame.
\newblock In H.~Wallach, H.~Larochelle, A.~Beygelzimer, F.~d\textquotesingle Alch\'{e}-Buc, E.~Fox, and R.~Garnett (eds.), \emph{Advances in Neural Information Processing Systems}, volume~32. Curran Associates, Inc., 2019.
\newblock URL \url{https://proceedings.neurips.cc/paper_files/paper/2019/file/bb836c01cdc9120a9c984c525e4b1a4a-Paper.pdf}.

\bibitem[Donnelly \& Roegiest(2019)Donnelly and Roegiest]{Donnelly_2019_sentimentneuron}
Jonathan Donnelly and Adam Roegiest.
\newblock On interpretability and feature representations: An analysis of the sentiment neuron.
\newblock In Leif Azzopardi, Benno Stein, Norbert Fuhr, Philipp Mayr, Claudia Hauff, and Djoerd Hiemstra (eds.), \emph{Advances in Information Retrieval}, pp.\  795--802, Cham, 2019. Springer International Publishing.
\newblock ISBN 978-3-030-15712-8.

\bibitem[Doshi-Velez \& Kim(2017)Doshi-Velez and Kim]{doshivelez2017rigorousscienceinterpretablemachine}
Finale Doshi-Velez and Been Kim.
\newblock Towards a rigorous science of interpretable machine learning, 2017.
\newblock URL \url{https://arxiv.org/abs/1702.08608}.

\bibitem[Dosovitskiy et~al.(2021)Dosovitskiy, Beyer, Kolesnikov, Weissenborn, Zhai, Unterthiner, Dehghani, Minderer, Heigold, Gelly, Uszkoreit, and Houlsby]{dosovitskiy2021imageworth16x16words}
Alexey Dosovitskiy, Lucas Beyer, Alexander Kolesnikov, Dirk Weissenborn, Xiaohua Zhai, Thomas Unterthiner, Mostafa Dehghani, Matthias Minderer, Georg Heigold, Sylvain Gelly, Jakob Uszkoreit, and Neil Houlsby.
\newblock An image is worth 16x16 words: Transformers for image recognition at scale.
\newblock In \emph{International Conference on Learning Representations}, 2021.
\newblock URL \url{https://openreview.net/forum?id=YicbFdNTTy}.

\bibitem[Dunefsky et~al.(2024)Dunefsky, Chlenski, and Nanda]{dunefsky2024transcodersinterpretablellmfeature}
Jacob Dunefsky, Philippe Chlenski, and Neel Nanda.
\newblock Transcoders find interpretable {LLM} feature circuits.
\newblock In \emph{The Thirty-eighth Annual Conference on Neural Information Processing Systems}, 2024.
\newblock URL \url{https://openreview.net/forum?id=J6zHcScAo0}.

\bibitem[Durmus et~al.(2024)Durmus, Tamkin, Clark, Wei, Marcus, Batson, Handa, Lovitt, Tong, McCain, Rausch, Huang, Bowman, Ritchie, Henighan, and Ganguli]{durmus2024steering}
Esin Durmus, Alex Tamkin, Jack Clark, Jerry Wei, Jonathan Marcus, Joshua Batson, Kunal Handa, Liane Lovitt, Meg Tong, Miles McCain, Oliver Rausch, Saffron Huang, Sam Bowman, Stuart Ritchie, Tom Henighan, and Deep Ganguli.
\newblock Evaluating feature steering: A case study in mitigating social biases, 2024.
\newblock URL \url{https://anthropic.com/research/evaluating-feature-steering}.

\bibitem[Elazar et~al.(2021)Elazar, Ravfogel, Jacovi, and Goldberg]{Elazar_2021}
Yanai Elazar, Shauli Ravfogel, Alon Jacovi, and Yoav Goldberg.
\newblock {Amnesic Probing: Behavioral Explanation with Amnesic Counterfactuals}.
\newblock \emph{Transactions of the Association for Computational Linguistics}, 9:\penalty0 160--175, 03 2021.
\newblock ISSN 2307-387X.
\newblock \doi{10.1162/tacl_a_00359}.
\newblock URL \url{https://doi.org/10.1162/tacl\_a\_00359}.

\bibitem[Elhage et~al.(2021)Elhage, Nanda, Olsson, Henighan, Joseph, Mann, Askell, Bai, Chen, Conerly, DasSarma, Drain, Ganguli, Hatfield-Dodds, Hernandez, Jones, Kernion, Lovitt, Ndousse, Amodei, Brown, Clark, Kaplan, McCandlish, and Olah]{elhage2021mathematical}
Nelson Elhage, Neel Nanda, Catherine Olsson, Tom Henighan, Nicholas Joseph, Ben Mann, Amanda Askell, Yuntao Bai, Anna Chen, Tom Conerly, Nova DasSarma, Dawn Drain, Deep Ganguli, Zac Hatfield-Dodds, Danny Hernandez, Andy Jones, Jackson Kernion, Liane Lovitt, Kamal Ndousse, Dario Amodei, Tom Brown, Jack Clark, Jared Kaplan, Sam McCandlish, and Chris Olah.
\newblock A mathematical framework for transformer circuits.
\newblock \emph{Transformer Circuits Thread}, 2021.
\newblock https://transformer-circuits.pub/2021/framework/index.html.

\bibitem[Elhage et~al.(2022)Elhage, Hume, Olsson, Nanda, Henighan, Johnston, ElShowk, Joseph, DasSarma, Mann, Hernandez, Askell, Ndousse, Jones, Drain, Chen, Bai, Ganguli, Lovitt, Hatfield-Dodds, Kernion, Conerly, Kravec, Fort, Kadavath, Jacobson, Tran-Johnson, Kaplan, Clark, Brown, McCandlish, Amodei, and Olah]{elhage2022solu}
Nelson Elhage, Tristan Hume, Catherine Olsson, Neel Nanda, Tom Henighan, Scott Johnston, Sheer ElShowk, Nicholas Joseph, Nova DasSarma, Ben Mann, Danny Hernandez, Amanda Askell, Kamal Ndousse, Andy Jones, Dawn Drain, Anna Chen, Yuntao Bai, Deep Ganguli, Liane Lovitt, Zac Hatfield-Dodds, Jackson Kernion, Tom Conerly, Shauna Kravec, Stanislav Fort, Saurav Kadavath, Josh Jacobson, Eli Tran-Johnson, Jared Kaplan, Jack Clark, Tom Brown, Sam McCandlish, Dario Amodei, and Christopher Olah.
\newblock Softmax linear units.
\newblock \emph{Transformer Circuits Thread}, 2022.
\newblock https://transformer-circuits.pub/2022/solu/index.html.

\bibitem[Engels et~al.(2024{\natexlab{a}})Engels, Liao, Michaud, Gurnee, and Tegmark]{engels2024languagemodelfeatureslinear}
Joshua Engels, Isaac Liao, Eric~J. Michaud, Wes Gurnee, and Max Tegmark.
\newblock Not all language model features are linear, 2024{\natexlab{a}}.
\newblock URL \url{https://arxiv.org/abs/2405.14860}.

\bibitem[Engels et~al.(2024{\natexlab{b}})Engels, Riggs, and Tegmark]{engels2024decomposingdarkmattersparse}
Joshua Engels, Logan Riggs, and Max Tegmark.
\newblock Decomposing the dark matter of sparse autoencoders, 2024{\natexlab{b}}.
\newblock URL \url{https://arxiv.org/abs/2410.14670}.

\bibitem[Ensign \& Garriga-Alonso(2024)Ensign and Garriga-Alonso]{ensign2024investigatingindirectobjectidentification}
Danielle Ensign and Adrià Garriga-Alonso.
\newblock Investigating the indirect object identification circuit in mamba, 2024.
\newblock URL \url{https://arxiv.org/abs/2407.14008}.

\bibitem[Erhan et~al.(2009)Erhan, Bengio, Courville, and Vincent]{Erhan_2009_visualising}
Dumitru Erhan, Y.~Bengio, Aaron Courville, and Pascal Vincent.
\newblock Visualizing higher-layer features of a deep network.
\newblock \emph{Technical Report, University of Montreal}, 01 2009.

\bibitem[Ettinger et~al.(2016)Ettinger, Elgohary, and Resnik]{ettinger-etal-2016-probing}
Allyson Ettinger, Ahmed Elgohary, and Philip Resnik.
\newblock Probing for semantic evidence of composition by means of simple classification tasks.
\newblock In \emph{Proceedings of the 1st Workshop on Evaluating Vector-Space Representations for {NLP}}, pp.\  134--139, Berlin, Germany, August 2016. Association for Computational Linguistics.
\newblock \doi{10.18653/v1/W16-2524}.
\newblock URL \url{https://aclanthology.org/W16-2524}.

\bibitem[Farrell et~al.(2024)Farrell, Lau, and Conmy]{farrell2024applyingsparseautoencodersunlearn}
Eoin Farrell, Yeu-Tong Lau, and Arthur Conmy.
\newblock Applying sparse autoencoders to unlearn knowledge in language models, 2024.
\newblock URL \url{https://arxiv.org/abs/2410.19278}.

\bibitem[Fedus et~al.(2022)Fedus, Zoph, and Shazeer]{switchtransformers}
William Fedus, Barret Zoph, and Noam Shazeer.
\newblock Switch transformers: Scaling to trillion parameter models with simple and efficient sparsity.
\newblock \emph{Journal of Machine Learning Research}, 23\penalty0 (120):\penalty0 1--39, 2022.
\newblock URL \url{http://jmlr.org/papers/v23/21-0998.html}.

\bibitem[Feng et~al.(2018)Feng, Wallace, Grissom~II, Iyyer, Rodriguez, and Boyd-Graber]{feng-etal-2018-pathologies}
Shi Feng, Eric Wallace, Alvin Grissom~II, Mohit Iyyer, Pedro Rodriguez, and Jordan Boyd-Graber.
\newblock Pathologies of neural models make interpretations difficult.
\newblock In Ellen Riloff, David Chiang, Julia Hockenmaier, and Jun{'}ichi Tsujii (eds.), \emph{Proceedings of the 2018 Conference on Empirical Methods in Natural Language Processing}, pp.\  3719--3728, Brussels, Belgium, October-November 2018. Association for Computational Linguistics.
\newblock \doi{10.18653/v1/D18-1407}.
\newblock URL \url{https://aclanthology.org/D18-1407}.

\bibitem[Ferrando et~al.(2024)Ferrando, Sarti, Bisazza, and Costa-jussà]{ferrando2024primerinnerworkingstransformerbased}
Javier Ferrando, Gabriele Sarti, Arianna Bisazza, and Marta~R. Costa-jussà.
\newblock A primer on the inner workings of transformer-based language models, 2024.
\newblock URL \url{https://arxiv.org/abs/2405.00208}.

\bibitem[Foerster et~al.(2017)Foerster, Gilmer, Sohl-Dickstein, Chorowski, and Sussillo]{foerster2017inputswitchedaffinenetworks}
Jakob~N. Foerster, Justin Gilmer, Jascha Sohl-Dickstein, Jan Chorowski, and David Sussillo.
\newblock Input switched affine networks: an rnn architecture designed for interpretability.
\newblock In \emph{Proceedings of the 34th International Conference on Machine Learning - Volume 70}, ICML'17, pp.\  1136–1145. JMLR.org, 2017.

\bibitem[Fong \& Vedaldi(2017)Fong and Vedaldi]{Fong_2017_MaskingBasedCausalAttribution}
Ruth~C. Fong and Andrea Vedaldi.
\newblock Interpretable explanations of black boxes by meaningful perturbation.
\newblock In \emph{2017 IEEE International Conference on Computer Vision (ICCV)}. IEEE, October 2017.
\newblock \doi{10.1109/iccv.2017.371}.
\newblock URL \url{http://dx.doi.org/10.1109/ICCV.2017.371}.

\bibitem[Forde et~al.(2023)Forde, Lovering, Konidaris, Pavlick, and Littman]{forde2023where}
Jessica~Zosa Forde, Charles Lovering, George Konidaris, Ellie Pavlick, and Michael~L. Littman.
\newblock Where, when \& which concepts does alphazero learn? lessons from the game of hex.
\newblock Unpublished manuscript, 2023.

\bibitem[Frankle \& Carbin(2019)Frankle and Carbin]{frankle2019lotterytickethypothesisfinding}
Jonathan Frankle and Michael Carbin.
\newblock The lottery ticket hypothesis: Finding sparse, trainable neural networks.
\newblock In \emph{International Conference on Learning Representations}, 2019.
\newblock URL \url{https://openreview.net/forum?id=rJl-b3RcF7}.

\bibitem[Freiesleben \& K{\"o}nig(2023)Freiesleben and K{\"o}nig]{Freiesleben_2023_weneedtotalk}
Timo Freiesleben and Gunnar K{\"o}nig.
\newblock Dear xai community, we need to talk!
\newblock In Luca Longo (ed.), \emph{Explainable Artificial Intelligence}, pp.\  48--65, Cham, 2023. Springer Nature Switzerland.
\newblock ISBN 978-3-031-44064-9.

\bibitem[Friedman et~al.(2024)Friedman, Lampinen, Dixon, Chen, and Ghandeharioun]{friedman2024interpretabilityillusionsgeneralizationsimplified}
Dan Friedman, Andrew~Kyle Lampinen, Lucas Dixon, Danqi Chen, and Asma Ghandeharioun.
\newblock Interpretability illusions in the generalization of simplified models, 2024.
\newblock URL \url{https://openreview.net/forum?id=v675Iyu0ta}.

\bibitem[Fu et~al.(2023)Fu, Dao, Saab, Thomas, Rudra, and Re]{fu2023hungry}
Daniel~Y Fu, Tri Dao, Khaled~Kamal Saab, Armin~W Thomas, Atri Rudra, and Christopher Re.
\newblock Hungry hungry hippos: Towards language modeling with state space models.
\newblock In \emph{The Eleventh International Conference on Learning Representations}, 2023.
\newblock URL \url{https://openreview.net/forum?id=COZDy0WYGg}.

\bibitem[Gade et~al.(2024)Gade, Lermen, Rogers-Smith, and Ladish]{gade2024badllamacheaplyremovingsafety}
Pranav Gade, Simon Lermen, Charlie Rogers-Smith, and Jeffrey Ladish.
\newblock Badllama: cheaply removing safety fine-tuning from llama 2-chat 13b, 2024.
\newblock URL \url{https://arxiv.org/abs/2311.00117}.

\bibitem[Gale et~al.(2020)Gale, Martin, Blything, Nguyen, and Bowers]{GALE_2020_ObjectDetectors}
Ella~M. Gale, Nicholas Martin, Ryan Blything, Anh Nguyen, and Jeffrey~S. Bowers.
\newblock Are there any ‘object detectors’ in the hidden layers of cnns trained to identify objects or scenes?
\newblock \emph{Vision Research}, 176:\penalty0 60--71, 2020.
\newblock ISSN 0042-6989.
\newblock \doi{https://doi.org/10.1016/j.visres.2020.06.007}.
\newblock URL \url{https://www.sciencedirect.com/science/article/pii/S0042698920301140}.

\bibitem[Gandelsman et~al.(2024{\natexlab{a}})Gandelsman, Efros, and Steinhardt]{gandelsman2024interpretingclipsimagerepresentation}
Yossi Gandelsman, Alexei~A Efros, and Jacob Steinhardt.
\newblock Interpreting {CLIP}'s image representation via text-based decomposition.
\newblock In \emph{The Twelfth International Conference on Learning Representations}, 2024{\natexlab{a}}.
\newblock URL \url{https://openreview.net/forum?id=5Ca9sSzuDp}.

\bibitem[Gandelsman et~al.(2024{\natexlab{b}})Gandelsman, Efros, and Steinhardt]{gandelsman2024interpretingsecondordereffectsneurons}
Yossi Gandelsman, Alexei~A. Efros, and Jacob Steinhardt.
\newblock Interpreting the second-order effects of neurons in clip, 2024{\natexlab{b}}.
\newblock URL \url{https://arxiv.org/abs/2406.04341}.

\bibitem[Gao et~al.(2024)Gao, la~Tour, Tillman, Goh, Troll, Radford, Sutskever, Leike, and Wu]{gao2024scalingevaluatingsparseautoencoders}
Leo Gao, Tom~Dupré la~Tour, Henk Tillman, Gabriel Goh, Rajan Troll, Alec Radford, Ilya Sutskever, Jan Leike, and Jeffrey Wu.
\newblock Scaling and evaluating sparse autoencoders, 2024.
\newblock URL \url{https://arxiv.org/abs/2406.04093}.

\bibitem[Geiger et~al.(2020)Geiger, Richardson, and Potts]{geiger2020neuralnaturallanguageinference}
Atticus Geiger, Kyle Richardson, and Christopher Potts.
\newblock Neural natural language inference models partially embed theories of lexical entailment and negation.
\newblock In Afra Alishahi, Yonatan Belinkov, Grzegorz Chrupa{\l}a, Dieuwke Hupkes, Yuval Pinter, and Hassan Sajjad (eds.), \emph{Proceedings of the Third BlackboxNLP Workshop on Analyzing and Interpreting Neural Networks for NLP}, pp.\  163--173, Online, November 2020. Association for Computational Linguistics.
\newblock \doi{10.18653/v1/2020.blackboxnlp-1.16}.
\newblock URL \url{https://aclanthology.org/2020.blackboxnlp-1.16/}.

\bibitem[Geiger et~al.(2022)Geiger, Wu, Lu, Rozner, Kreiss, Icard, Goodman, and Potts]{Geiger_2022_CausalStructure}
Atticus Geiger, Zhengxuan Wu, Hanson Lu, Josh Rozner, Elisa Kreiss, Thomas Icard, Noah Goodman, and Christopher Potts.
\newblock Inducing causal structure for interpretable neural networks.
\newblock In Kamalika Chaudhuri, Stefanie Jegelka, Le~Song, Csaba Szepesvari, Gang Niu, and Sivan Sabato (eds.), \emph{Proceedings of the 39th International Conference on Machine Learning}, volume 162 of \emph{Proceedings of Machine Learning Research}, pp.\  7324--7338. PMLR, 17--23 Jul 2022.
\newblock URL \url{https://proceedings.mlr.press/v162/geiger22a.html}.

\bibitem[Geiger et~al.(2024{\natexlab{a}})Geiger, Ibeling, Zur, Chaudhary, Chauhan, Huang, Arora, Wu, Goodman, Potts, and Icard]{Geiger_2023_CausalAbstractionRepresentationSubspace}
Atticus Geiger, Duligur Ibeling, Amir Zur, Maheep Chaudhary, Sonakshi Chauhan, Jing Huang, Aryaman Arora, Zhengxuan Wu, Noah Goodman, Christopher Potts, and Thomas Icard.
\newblock Causal abstraction: A theoretical foundation for mechanistic interpretability.
\newblock In \emph{Machine Learning Research}, 2024{\natexlab{a}}.
\newblock URL \url{https://arxiv.org/abs/2301.04709}.

\bibitem[Geiger et~al.(2024{\natexlab{b}})Geiger, Lu, Icard, and Potts]{geiger2021causalabstractionsneuralnetworks}
Atticus Geiger, Hanson Lu, Thomas Icard, and Christopher Potts.
\newblock Causal abstractions of neural networks.
\newblock In \emph{Proceedings of the 35th International Conference on Neural Information Processing Systems}, NIPS '21, Red Hook, NY, USA, 2024{\natexlab{b}}. Curran Associates Inc.
\newblock ISBN 9781713845393.

\bibitem[Geiger et~al.(2024{\natexlab{c}})Geiger, Wu, Potts, Icard, and Goodman]{geiger2024findingalignmentsinterpretablecausal}
Atticus Geiger, Zhengxuan Wu, Christopher Potts, Thomas Icard, and Noah Goodman.
\newblock Finding alignments between interpretable causal variables and distributed neural representations.
\newblock In Francesco Locatello and Vanessa Didelez (eds.), \emph{Proceedings of the Third Conference on Causal Learning and Reasoning}, volume 236 of \emph{Proceedings of Machine Learning Research}, pp.\  160--187. PMLR, 01--03 Apr 2024{\natexlab{c}}.
\newblock URL \url{https://proceedings.mlr.press/v236/geiger24a.html}.

\bibitem[Geirhos et~al.(2019)Geirhos, Rubisch, Michaelis, Bethge, Wichmann, and Brendel]{geirhos2022imagenettrainedcnnsbiasedtexture}
Robert Geirhos, Patricia Rubisch, Claudio Michaelis, Matthias Bethge, Felix~A. Wichmann, and Wieland Brendel.
\newblock Imagenet-trained {CNN}s are biased towards texture; increasing shape bias improves accuracy and robustness.
\newblock In \emph{International Conference on Learning Representations}, 2019.
\newblock URL \url{https://openreview.net/forum?id=Bygh9j09KX}.

\bibitem[Geirhos et~al.(2025)Geirhos, Zimmermann, Bilodeau, Brendel, and Kim]{geirhos2024donttrusteyesunreliability}
Robert Geirhos, Roland~S. Zimmermann, Blair Bilodeau, Wieland Brendel, and Been Kim.
\newblock Don't trust your eyes: on the (un)reliability of feature visualizations.
\newblock In \emph{Proceedings of the 41st International Conference on Machine Learning}, ICML'24. JMLR.org, 2025.

\bibitem[Geva et~al.(2021)Geva, Schuster, Berant, and Levy]{geva2021transformerfeedforwardlayerskeyvalue}
Mor Geva, Roei Schuster, Jonathan Berant, and Omer Levy.
\newblock Transformer feed-forward layers are key-value memories.
\newblock In Marie-Francine Moens, Xuanjing Huang, Lucia Specia, and Scott Wen-tau Yih (eds.), \emph{Proceedings of the 2021 Conference on Empirical Methods in Natural Language Processing}, pp.\  5484--5495, Online and Punta Cana, Dominican Republic, November 2021. Association for Computational Linguistics.
\newblock \doi{10.18653/v1/2021.emnlp-main.446}.
\newblock URL \url{https://aclanthology.org/2021.emnlp-main.446/}.

\bibitem[Geva et~al.(2022{\natexlab{a}})Geva, Caciularu, Dar, Roit, Sadde, Shlain, Tamir, and Goldberg]{geva2022lmdebuggerinteractivetoolinspection}
Mor Geva, Avi Caciularu, Guy Dar, Paul Roit, Shoval Sadde, Micah Shlain, Bar Tamir, and Yoav Goldberg.
\newblock {LM}-debugger: An interactive tool for inspection and intervention in transformer-based language models.
\newblock In Wanxiang Che and Ekaterina Shutova (eds.), \emph{Proceedings of the 2022 Conference on Empirical Methods in Natural Language Processing: System Demonstrations}, pp.\  12--21, Abu Dhabi, UAE, December 2022{\natexlab{a}}. Association for Computational Linguistics.
\newblock \doi{10.18653/v1/2022.emnlp-demos.2}.
\newblock URL \url{https://aclanthology.org/2022.emnlp-demos.2/}.

\bibitem[Geva et~al.(2022{\natexlab{b}})Geva, Caciularu, Wang, and Goldberg]{geva2022transformerfeedforwardlayersbuild}
Mor Geva, Avi Caciularu, Kevin Wang, and Yoav Goldberg.
\newblock Transformer feed-forward layers build predictions by promoting concepts in the vocabulary space.
\newblock In Yoav Goldberg, Zornitsa Kozareva, and Yue Zhang (eds.), \emph{Proceedings of the 2022 Conference on Empirical Methods in Natural Language Processing}, pp.\  30--45, Abu Dhabi, United Arab Emirates, December 2022{\natexlab{b}}. Association for Computational Linguistics.
\newblock \doi{10.18653/v1/2022.emnlp-main.3}.
\newblock URL \url{https://aclanthology.org/2022.emnlp-main.3/}.

\bibitem[Geva et~al.(2023)Geva, Bastings, Filippova, and Globerson]{Geva_2023_InformationFlow}
Mor Geva, Jasmijn Bastings, Katja Filippova, and Amir Globerson.
\newblock Dissecting recall of factual associations in auto-regressive language models.
\newblock In Houda Bouamor, Juan Pino, and Kalika Bali (eds.), \emph{Proceedings of the 2023 Conference on Empirical Methods in Natural Language Processing}, pp.\  12216--12235, Singapore, December 2023. Association for Computational Linguistics.
\newblock \doi{10.18653/v1/2023.emnlp-main.751}.
\newblock URL \url{https://aclanthology.org/2023.emnlp-main.751/}.

\bibitem[Ghandeharioun et~al.(2024)Ghandeharioun, Caciularu, Pearce, Dixon, and Geva]{ghandeharioun2024patchscopesunifyingframeworkinspecting}
Asma Ghandeharioun, Avi Caciularu, Adam Pearce, Lucas Dixon, and Mor Geva.
\newblock Patchscopes: A unifying framework for inspecting hidden representations of language models.
\newblock In \emph{ICML}, 2024.
\newblock URL \url{https://arxiv.org/abs/2401.06102}.

\bibitem[Ghandeharioun et~al.(2025)Ghandeharioun, Caciularu, Pearce, Dixon, and Geva]{ghandeharioun2024patchscopes}
Asma Ghandeharioun, Avi Caciularu, Adam Pearce, Lucas Dixon, and Mor Geva.
\newblock Patchscopes: a unifying framework for inspecting hidden representations of language models.
\newblock In \emph{Proceedings of the 41st International Conference on Machine Learning}, ICML'24. JMLR.org, 2025.

\bibitem[Ghorbani \& Zou(2020)Ghorbani and Zou]{Ghorbani_2020}
Amirata Ghorbani and James~Y Zou.
\newblock Neuron shapley: Discovering the responsible neurons.
\newblock In H.~Larochelle, M.~Ranzato, R.~Hadsell, M.F. Balcan, and H.~Lin (eds.), \emph{Advances in Neural Information Processing Systems}, volume~33, pp.\  5922--5932. Curran Associates, Inc., 2020.
\newblock URL \url{https://proceedings.neurips.cc/paper_files/paper/2020/file/41c542dfe6e4fc3deb251d64cf6ed2e4-Paper.pdf}.

\bibitem[Ghorbani et~al.(2019)Ghorbani, Abid, and Zou]{Ghorbani_Abid_Zou_2019}
Amirata Ghorbani, Abubakar Abid, and James Zou.
\newblock Interpretation of neural networks is fragile.
\newblock \emph{Proceedings of the AAAI Conference on Artificial Intelligence}, 33\penalty0 (01):\penalty0 3681--3688, Jul. 2019.
\newblock \doi{10.1609/aaai.v33i01.33013681}.
\newblock URL \url{https://ojs.aaai.org/index.php/AAAI/article/view/4252}.

\bibitem[Gilpin et~al.(2019)Gilpin, Testart, Fruchter, and Adebayo]{gilpin2019explainingexplanationssociety}
Leilani~H. Gilpin, Cecilia Testart, Nathaniel Fruchter, and Julius Adebayo.
\newblock Explaining explanations to society.
\newblock In \emph{NeurIPS Workshop on Ethical, Social and Governance Issues in AI}, 2019.
\newblock URL \url{https://arxiv.org/abs/1901.06560}.

\bibitem[Goemans et~al.(2024)Goemans, Buhl, Schuett, Korbak, Wang, Hilton, and Irving]{goemans2024safetycasetemplatefrontier}
Arthur Goemans, Marie~Davidsen Buhl, Jonas Schuett, Tomek Korbak, Jessica Wang, Benjamin Hilton, and Geoffrey Irving.
\newblock Safety case template for frontier ai: A cyber inability argument, 2024.
\newblock URL \url{https://arxiv.org/abs/2411.08088}.

\bibitem[Goldowsky-Dill et~al.(2023)Goldowsky-Dill, MacLeod, Sato, and Arora]{goldowskydill2023localizingmodelbehaviorpath}
Nicholas Goldowsky-Dill, Chris MacLeod, Lucas Sato, and Aryaman Arora.
\newblock Localizing model behavior with path patching, 2023.
\newblock URL \url{https://arxiv.org/abs/2304.05969}.

\bibitem[Gottesman \& Geva(2024)Gottesman and Geva]{gottesman2024estimatingknowledgelargelanguage}
Daniela Gottesman and Mor Geva.
\newblock Estimating knowledge in large language models without generating a single token.
\newblock In Yaser Al-Onaizan, Mohit Bansal, and Yun-Nung Chen (eds.), \emph{Proceedings of the 2024 Conference on Empirical Methods in Natural Language Processing}, pp.\  3994--4019, Miami, Florida, USA, November 2024. Association for Computational Linguistics.
\newblock \doi{10.18653/v1/2024.emnlp-main.232}.
\newblock URL \url{https://aclanthology.org/2024.emnlp-main.232/}.

\bibitem[Goyal et~al.(2022)Goyal, Soriano, Hazirbas, Sagun, and Usunier]{Goyal_2022}
Priya Goyal, Adriana~Romero Soriano, Caner Hazirbas, Levent Sagun, and Nicolas Usunier.
\newblock Fairness indicators for systematic assessments of visual feature extractors.
\newblock In \emph{Proceedings of the 2022 ACM Conference on Fairness, Accountability, and Transparency}, FAccT '22, pp.\  70–88, New York, NY, USA, 2022. Association for Computing Machinery.
\newblock ISBN 9781450393522.
\newblock \doi{10.1145/3531146.3533074}.
\newblock URL \url{https://doi.org/10.1145/3531146.3533074}.

\bibitem[Greenblatt \& Shlegeris(2024)Greenblatt and Shlegeris]{greenblatt2024catching}
Ryan Greenblatt and Buck Shlegeris.
\newblock Catching {AIs} red-handed.
\newblock \emph{Alignment Forum}, January 2024.
\newblock URL \url{https://www.alignmentforum.org/posts/i2nmBfCXnadeGmhzW/catching-ais-red-handed}.

\bibitem[Greenblatt et~al.(2024)Greenblatt, Roger, Krasheninnikov, and Krueger]{greenblatt2024stresstestingcapabilityelicitationpasswordlocked}
Ryan Greenblatt, Fabien Roger, Dmitrii Krasheninnikov, and David Krueger.
\newblock Stress-testing capability elicitation with password-locked models.
\newblock In \emph{The Thirty-eighth Annual Conference on Neural Information Processing Systems}, 2024.
\newblock URL \url{https://openreview.net/forum?id=zzOOqD6R1b}.

\bibitem[Gromov et~al.(2024)Gromov, Tirumala, Shapourian, Glorioso, and Roberts]{gromov2024unreasonableineffectivenessdeeperlayers}
Andrey Gromov, Kushal Tirumala, Hassan Shapourian, Paolo Glorioso, and Daniel~A. Roberts.
\newblock The unreasonable ineffectiveness of the deeper layers, 2024.
\newblock URL \url{https://arxiv.org/abs/2403.17887}.

\bibitem[Gross et~al.(2024)Gross, Agrawal, Kwa, Ong, Yip, Gibson, Noubir, and Chan]{gross2024compactproofsmodelperformance}
Jason Gross, Rajashree Agrawal, Thomas Kwa, Euan Ong, Chun~Hei Yip, Alex Gibson, Soufiane Noubir, and Lawrence Chan.
\newblock Compact proofs of model performance via mechanistic interpretability.
\newblock In \emph{ICML Workshop on Mechanistic Interpretability}, 2024.
\newblock URL \url{https://arxiv.org/abs/2406.11779}.

\bibitem[Grosse(2024)]{grosse2024three}
Roger Grosse.
\newblock Three sketches of {ASL}-4 safety case components, November 2024.
\newblock URL \url{https://alignment.anthropic.com/2024/safety-cases/}.

\bibitem[Grosse et~al.(2023)Grosse, Bae, Anil, Elhage, Tamkin, Tajdini, Steiner, Li, Durmus, Perez, Hubinger, Lukošiūtė, Nguyen, Joseph, McCandlish, Kaplan, and Bowman]{grosse2023studyinglargelanguagemodel}
Roger Grosse, Juhan Bae, Cem Anil, Nelson Elhage, Alex Tamkin, Amirhossein Tajdini, Benoit Steiner, Dustin Li, Esin Durmus, Ethan Perez, Evan Hubinger, Kamilė Lukošiūtė, Karina Nguyen, Nicholas Joseph, Sam McCandlish, Jared Kaplan, and Samuel~R. Bowman.
\newblock Studying large language model generalization with influence functions, 2023.
\newblock URL \url{https://arxiv.org/abs/2308.03296}.

\bibitem[Grynbaum \& Mac(2023)Grynbaum and Mac]{grynbaum2023times}
Michael~M. Grynbaum and Ryan Mac.
\newblock The {Times} sues {OpenAI} and {Microsoft} over {A.I.} use of copyrighted work.
\newblock \emph{The New York Times}, December 2023.
\newblock URL \url{https://www.nytimes.com/2023/12/27/business/media/new-york-times-open-ai-microsoft-lawsuit.html}.

\bibitem[Gu \& Dao(2024)Gu and Dao]{gu2024mambalineartimesequencemodeling}
Albert Gu and Tri Dao.
\newblock Mamba: Linear-time sequence modeling with selective state spaces, 2024.
\newblock URL \url{https://arxiv.org/abs/2312.00752}.

\bibitem[Guerner et~al.(2024)Guerner, Svete, Liu, Warstadt, and Cotterell]{guerner2024geometricnotioncausalprobing}
Clément Guerner, Anej Svete, Tianyu Liu, Alexander Warstadt, and Ryan Cotterell.
\newblock A geometric notion of causal probing, 2024.
\newblock URL \url{https://arxiv.org/abs/2307.15054}.

\bibitem[Guo et~al.(2024)Guo, Syed, Sheshadri, Ewart, and Dziugaite]{guo2024robust}
Phillip~Huang Guo, Aaquib Syed, Abhay Sheshadri, Aidan Ewart, and Gintare~Karolina Dziugaite.
\newblock Robust unlearning via mechanistic localizations.
\newblock In \emph{ICML 2024 Workshop on Mechanistic Interpretability}, 2024.
\newblock URL \url{https://openreview.net/forum?id=06pNzrEjnH}.

\bibitem[Gupta et~al.(2015)Gupta, Boleda, Baroni, and Pad{\'o}]{gupta-etal-2015-distributional}
Abhijeet Gupta, Gemma Boleda, Marco Baroni, and Sebastian Pad{\'o}.
\newblock Distributional vectors encode referential attributes.
\newblock In Llu{\'\i}s M{\`a}rquez, Chris Callison-Burch, and Jian Su (eds.), \emph{Proceedings of the 2015 Conference on Empirical Methods in Natural Language Processing}, pp.\  12--21, Lisbon, Portugal, September 2015. Association for Computational Linguistics.
\newblock \doi{10.18653/v1/D15-1002}.
\newblock URL \url{https://aclanthology.org/D15-1002}.

\bibitem[Gupta et~al.(2024)Gupta, Arcuschin, Kwa, and Garriga-Alonso]{gupta2024interpbenchsemisynthetictransformersevaluating}
Rohan Gupta, Iv{\'a}n Arcuschin, Thomas Kwa, and Adri{\`a} Garriga-Alonso.
\newblock Interpbench: Semi-synthetic transformers for evaluating mechanistic interpretability techniques.
\newblock In \emph{The Thirty-eight Conference on Neural Information Processing Systems Datasets and Benchmarks Track}, 2024.
\newblock URL \url{https://openreview.net/forum?id=R9gR9MPuD5}.

\bibitem[Gurnee \& Tegmark(2024)Gurnee and Tegmark]{gurnee2024languagemodelsrepresentspace}
Wes Gurnee and Max Tegmark.
\newblock Language models represent space and time.
\newblock In \emph{The Twelfth International Conference on Learning Representations}, 2024.
\newblock URL \url{https://openreview.net/forum?id=jE8xbmvFin}.

\bibitem[Gurnee et~al.(2023)Gurnee, Nanda, Pauly, Harvey, Troitskii, and Bertsimas]{gurnee2023findingneuronshaystackcase}
Wes Gurnee, Neel Nanda, Matthew Pauly, Katherine Harvey, Dmitrii Troitskii, and Dimitris Bertsimas.
\newblock Finding neurons in a haystack: Case studies with sparse probing.
\newblock \emph{Transactions on Machine Learning Research}, 2023.
\newblock ISSN 2835-8856.
\newblock URL \url{https://openreview.net/forum?id=JYs1R9IMJr}.

\bibitem[Haani et~al.(2024)Haani, Popper, and Mendel]{kaarel2024starting}
Kaarel Haani, Rio Popper, and Jake Mendel.
\newblock A starting point for making sense of task structure (in machine learning).
\newblock \emph{LessWrong}, February 2024.
\newblock URL \url{https://www.lesswrong.com/posts/exp4JGPJu46g6sdRp/a-starting-point-for-making-sense-of-task-structure-in}.

\bibitem[Hampel(1974)]{Hampel_1986_Influence_Function}
Frank~R. Hampel.
\newblock The influence curve and its role in robust estimation.
\newblock \emph{Journal of the American Statistical Association}, 69\penalty0 (346):\penalty0 383--393, 1974.
\newblock \doi{10.1080/01621459.1974.10482962}.

\bibitem[Han et~al.(2015)Han, Pool, Tran, and Dally]{Han_2015_weightsandconnections}
Song Han, Jeff Pool, John Tran, and William~J. Dally.
\newblock Learning both weights and connections for efficient neural networks.
\newblock In \emph{Proceedings of the 28th International Conference on Neural Information Processing Systems - Volume 1}, NIPS'15, pp.\  1135–1143, Cambridge, MA, USA, 2015. MIT Press.

\bibitem[Hanna et~al.(2023)Hanna, Liu, and Variengien]{hanna2023doesgpt2computegreaterthan}
Michael Hanna, Ollie Liu, and Alexandre Variengien.
\newblock How does {GPT}-2 compute greater-than?: Interpreting mathematical abilities in a pre-trained language model.
\newblock In \emph{Thirty-seventh Conference on Neural Information Processing Systems}, 2023.
\newblock URL \url{https://openreview.net/forum?id=p4PckNQR8k}.

\bibitem[H{\"a}nni et~al.(2024)H{\"a}nni, Mendel, Vaintrob, and Chan]{hani2024mathematicalmodels}
Kaarel H{\"a}nni, Jake Mendel, Dmitry Vaintrob, and Lawrence Chan.
\newblock Mathematical models of computation in superposition.
\newblock In \emph{ICML 2024 Workshop on Mechanistic Interpretability}, 2024.
\newblock URL \url{https://openreview.net/forum?id=OcVJP8kClR}.

\bibitem[Harris(1954)]{Harris_1954_distributionalstructure}
Zellig~S. Harris.
\newblock Distributional structure.
\newblock \emph{WORD}, 10\penalty0 (2-3):\penalty0 146--162, 1954.
\newblock \doi{10.1080/00437956.1954.11659520}.
\newblock URL \url{https://doi.org/10.1080/00437956.1954.11659520}.

\bibitem[Hase et~al.(2023)Hase, Bansal, Kim, and Ghandeharioun]{hase2023doeslocalizationinformediting}
Peter Hase, Mohit Bansal, Been Kim, and Asma Ghandeharioun.
\newblock Does localization inform editing? surprising differences in causality-based localization vs. knowledge editing in language models.
\newblock In \emph{Thirty-seventh Conference on Neural Information Processing Systems}, 2023.
\newblock URL \url{https://openreview.net/forum?id=EldbUlZtbd}.

\bibitem[Hastie \& Tibshirani(1986)Hastie and Tibshirani]{Hastie_1986_AdditiveModel}
Trevor Hastie and Robert Tibshirani.
\newblock {Generalized Additive Models}.
\newblock \emph{Statistical Science}, 1\penalty0 (3):\penalty0 297 -- 310, 1986.
\newblock \doi{10.1214/ss/1177013604}.
\newblock URL \url{https://doi.org/10.1214/ss/1177013604}.

\bibitem[He et~al.(2016)He, Zhang, Ren, and Sun]{He_2015_ImageDNN}
Kaiming He, Xiangyu Zhang, Shaoqing Ren, and Jian Sun.
\newblock Deep residual learning for image recognition.
\newblock In \emph{2016 IEEE Conference on Computer Vision and Pattern Recognition (CVPR)}, pp.\  770--778, 2016.
\newblock \doi{10.1109/CVPR.2016.90}.

\bibitem[He(2024)]{he2024mixturemillionexperts}
Xu~Owen He.
\newblock Mixture of a million experts, 2024.
\newblock URL \url{https://arxiv.org/abs/2407.04153}.

\bibitem[Heimersheim(2024)]{heimersheim2024removegpt2slayernormfinetuning}
Stefan Heimersheim.
\newblock You can remove gpt2's layernorm by fine-tuning, 2024.
\newblock URL \url{https://arxiv.org/abs/2409.13710}.

\bibitem[Heimersheim \& Janiak(2023)Heimersheim and Janiak]{heimersheim2023circuit}
Stefan Heimersheim and Jett Janiak.
\newblock A circuit for {Python} docstrings in a 4-layer attention-only transformer.
\newblock \emph{Alignment Forum}, February 2023.
\newblock URL \url{https://www.alignmentforum.org/posts/u6KXXmKFbXfWzoAXn/}.

\bibitem[Henighan(2024)]{henighan2024caloric}
Tom Henighan.
\newblock Caloric and the utility of incorrect theories.
\newblock \emph{Transformer Circuits}, 2024.
\newblock URL \url{https://transformer-circuits.pub/2024/april-update/index.html#caloric-theory}.

\bibitem[Heo et~al.(2019)Heo, Joo, and Moon]{Heo_2019_Advances}
Juyeon Heo, Sunghwan Joo, and Taesup Moon.
\newblock Fooling neural network interpretations via adversarial model manipulation.
\newblock In H.~Wallach, H.~Larochelle, A.~Beygelzimer, F.~d\textquotesingle Alch\'{e}-Buc, E.~Fox, and R.~Garnett (eds.), \emph{Advances in Neural Information Processing Systems}, volume~32. Curran Associates, Inc., 2019.
\newblock URL \url{https://proceedings.neurips.cc/paper_files/paper/2019/file/7fea637fd6d02b8f0adf6f7dc36aed93-Paper.pdf}.

\bibitem[Hernandez et~al.(2022)Hernandez, Schwettmann, Bau, Bagashvili, Torralba, and Andreas]{hernandez2022natural}
Evan Hernandez, Sarah Schwettmann, David Bau, Teona Bagashvili, Antonio Torralba, and Jacob Andreas.
\newblock Natural language descriptions of deep features.
\newblock In \emph{International Conference on Learning Representations}, 2022.
\newblock URL \url{https://openreview.net/forum?id=NudBMY-tzDr}.

\bibitem[Hewitt \& Liang(2019)Hewitt and Liang]{hewitt2019designinginterpretingprobescontrol}
John Hewitt and Percy Liang.
\newblock Designing and interpreting probes with control tasks.
\newblock In \emph{EMNLP}, 2019.
\newblock URL \url{https://arxiv.org/abs/1909.03368}.

\bibitem[Hewitt \& Manning(2019)Hewitt and Manning]{hewitt-manning-2019-structural}
John Hewitt and Christopher~D. Manning.
\newblock {A} structural probe for finding syntax in word representations.
\newblock In Jill Burstein, Christy Doran, and Thamar Solorio (eds.), \emph{Proceedings of the 2019 Conference of the North {A}merican Chapter of the Association for Computational Linguistics: Human Language Technologies, Volume 1 (Long and Short Papers)}, pp.\  4129--4138, Minneapolis, Minnesota, June 2019. Association for Computational Linguistics.
\newblock \doi{10.18653/v1/N19-1419}.
\newblock URL \url{https://aclanthology.org/N19-1419}.

\bibitem[Hewitt et~al.(2023)Hewitt, Thickstun, Manning, and Liang]{Hewitt_2023_BackpackLanguageModels}
John Hewitt, John Thickstun, Christopher Manning, and Percy Liang.
\newblock Backpack language models.
\newblock In Anna Rogers, Jordan Boyd-Graber, and Naoaki Okazaki (eds.), \emph{Proceedings of the 61st Annual Meeting of the Association for Computational Linguistics (Volume 1: Long Papers)}, pp.\  9103--9125, Toronto, Canada, July 2023. Association for Computational Linguistics.
\newblock \doi{10.18653/v1/2023.acl-long.506}.
\newblock URL \url{https://aclanthology.org/2023.acl-long.506/}.

\bibitem[Hicks et~al.(2021)Hicks, Isaksen, Thambawita, et~al.]{hicks2021explaining}
Steven~A. Hicks, Jonas~L. Isaksen, Vajira Thambawita, et~al.
\newblock Explaining deep neural networks for knowledge discovery in electrocardiogram analysis.
\newblock \emph{Scientific Reports}, 11:\penalty0 10949, 2021.
\newblock \doi{10.1038/s41598-021-90285-5}.
\newblock URL \url{https://doi.org/10.1038/s41598-021-90285-5}.

\bibitem[Hilton et~al.(2020)Hilton, Cammarata, Carter, Goh, and Olah]{hilton2020understanding}
Jacob Hilton, Nick Cammarata, Shan Carter, Gabriel Goh, and Chris Olah.
\newblock Understanding rl vision.
\newblock \emph{Distill}, 2020.
\newblock \doi{10.23915/distill.00029}.
\newblock https://distill.pub/2020/understanding-rl-vision.

\bibitem[Hinton(1981)]{hinton1981shape}
Geoffrey Hinton.
\newblock Shape representation in parallel systems.
\newblock \emph{Proceedings of teh Seventh International Joint Conference on Artificial Intelligence}, 1981.
\newblock URL \url{https://www.cs.toronto.edu/~hinton/absps/shape81.pdf}.

\bibitem[Hinton et~al.(2006)Hinton, Osindero, and Teh]{Hinton_2006_EarlyDNN}
Geoffrey~E. Hinton, Simon Osindero, and Yee-Whye Teh.
\newblock {A Fast Learning Algorithm for Deep Belief Nets}.
\newblock \emph{Neural Computation}, 18\penalty0 (7):\penalty0 1527--1554, 07 2006.
\newblock ISSN 0899-7667.
\newblock \doi{10.1162/neco.2006.18.7.1527}.
\newblock URL \url{https://doi.org/10.1162/neco.2006.18.7.1527}.

\bibitem[Hollinsworth et~al.(2024)Hollinsworth, Tigges, Geiger, and Nanda]{Tigges_2023_Probe}
Oskar~John Hollinsworth, Curt Tigges, Atticus Geiger, and Neel Nanda.
\newblock Language models linearly represent sentiment.
\newblock In Yonatan Belinkov, Najoung Kim, Jaap Jumelet, Hosein Mohebbi, Aaron Mueller, and Hanjie Chen (eds.), \emph{Proceedings of the 7th BlackboxNLP Workshop: Analyzing and Interpreting Neural Networks for NLP}, pp.\  58--87, Miami, Florida, US, November 2024. Association for Computational Linguistics.
\newblock \doi{10.18653/v1/2024.blackboxnlp-1.5}.
\newblock URL \url{https://aclanthology.org/2024.blackboxnlp-1.5/}.

\bibitem[Hong et~al.(2024)Hong, Yu, Yang, Ravfogel, and Geva]{hong2024intrinsicevaluationunlearningusing}
Yihuai Hong, Lei Yu, Haiqin Yang, Shauli Ravfogel, and Mor Geva.
\newblock Intrinsic evaluation of unlearning using parametric knowledge traces, 2024.
\newblock URL \url{https://arxiv.org/abs/2406.11614}.

\bibitem[Hoogland et~al.(2024)Hoogland, Wang, Farrugia-Roberts, Carroll, Wei, and Murfet]{hoogland2024developmentallandscapeincontextlearning}
Jesse Hoogland, George Wang, Matthew Farrugia-Roberts, Liam Carroll, Susan Wei, and Daniel Murfet.
\newblock The developmental landscape of in-context learning, 2024.
\newblock URL \url{https://arxiv.org/abs/2402.02364}.

\bibitem[Hooker et~al.(2021)Hooker, Mentch, and Zhou]{hooker2021unrestrictedpermutationforcesextrapolation}
Giles Hooker, Lucas Mentch, and Siyu Zhou.
\newblock Unrestricted permutation forces extrapolation: Variable importance requires at least one more model, or there is no free variable importance, 2021.
\newblock URL \url{https://arxiv.org/abs/1905.03151}.

\bibitem[Hu et~al.(2019)Hu, Rudin, and Seltzer]{Hu_2019_OptimalSparseDecisionTrees}
Xiyang Hu, Cynthia Rudin, and Margo~I. Seltzer.
\newblock Optimal sparse decision trees.
\newblock In \emph{Neural Information Processing Systems}, 2019.
\newblock URL \url{https://api.semanticscholar.org/CorpusID:139104649}.

\bibitem[Huang et~al.(2023)Huang, Geiger, D{'}Oosterlinck, Wu, and Potts]{huang2023rigorouslyassessingnaturallanguage}
Jing Huang, Atticus Geiger, Karel D{'}Oosterlinck, Zhengxuan Wu, and Christopher Potts.
\newblock Rigorously assessing natural language explanations of neurons.
\newblock In Yonatan Belinkov, Sophie Hao, Jaap Jumelet, Najoung Kim, Arya McCarthy, and Hosein Mohebbi (eds.), \emph{Proceedings of the 6th BlackboxNLP Workshop: Analyzing and Interpreting Neural Networks for NLP}, pp.\  317--331, Singapore, December 2023. Association for Computational Linguistics.
\newblock \doi{10.18653/v1/2023.blackboxnlp-1.24}.
\newblock URL \url{https://aclanthology.org/2023.blackboxnlp-1.24/}.

\bibitem[Huang et~al.(2024{\natexlab{a}})Huang, Wu, Potts, Geva, and Geiger]{huang2024ravelevaluatinginterpretabilitymethods}
Jing Huang, Zhengxuan Wu, Christopher Potts, Mor Geva, and Atticus Geiger.
\newblock {RAVEL}: Evaluating interpretability methods on disentangling language model representations.
\newblock In Lun-Wei Ku, Andre Martins, and Vivek Srikumar (eds.), \emph{Proceedings of the 62nd Annual Meeting of the Association for Computational Linguistics (Volume 1: Long Papers)}, pp.\  8669--8687, Bangkok, Thailand, August 2024{\natexlab{a}}. Association for Computational Linguistics.
\newblock \doi{10.18653/v1/2024.acl-long.470}.
\newblock URL \url{https://aclanthology.org/2024.acl-long.470/}.

\bibitem[Huang et~al.(2024{\natexlab{b}})Huang, Panwar, Goyal, and Hahn]{huang2024inversionviewgeneralpurposemethodreading}
Xinting Huang, Madhur Panwar, Navin Goyal, and Michael Hahn.
\newblock Inversionview: A general-purpose method for reading information from neural activations.
\newblock In \emph{ICML Interpretability Workshop}, 2024{\natexlab{b}}.
\newblock URL \url{https://arxiv.org/abs/2405.17653}.

\bibitem[Huben et~al.(2024)Huben, Cunningham, Smith, Ewart, and Sharkey]{cunningham2023sparseautoencodershighlyinterpretable}
Robert Huben, Hoagy Cunningham, Logan~Riggs Smith, Aidan Ewart, and Lee Sharkey.
\newblock Sparse autoencoders find highly interpretable features in language models.
\newblock In \emph{The Twelfth International Conference on Learning Representations}, 2024.
\newblock URL \url{https://openreview.net/forum?id=F76bwRSLeK}.

\bibitem[Hubinger(2023)]{hubinger2023towards}
Evan Hubinger.
\newblock Towards understanding-based safety evaluations.
\newblock \emph{Alignment Forum}, March 2023.
\newblock URL \url{https://www.alignmentforum.org/posts/uqAdqrvxqGqeBHjTP/towards-understanding-based-safety-evaluations}.

\bibitem[Hubinger et~al.(2021)Hubinger, van Merwijk, Mikulik, Skalse, and Garrabrant]{hubinger2021riskslearnedoptimizationadvanced}
Evan Hubinger, Chris van Merwijk, Vladimir Mikulik, Joar Skalse, and Scott Garrabrant.
\newblock Risks from learned optimization in advanced machine learning systems, 2021.
\newblock URL \url{https://arxiv.org/abs/1906.01820}.

\bibitem[Hubinger et~al.(2024)Hubinger, Denison, Mu, Lambert, Tong, MacDiarmid, Lanham, Ziegler, Maxwell, Cheng, Jermyn, Askell, Radhakrishnan, Anil, Duvenaud, Ganguli, Barez, Clark, Ndousse, Sachan, Sellitto, Sharma, DasSarma, Grosse, Kravec, Bai, Witten, Favaro, Brauner, Karnofsky, Christiano, Bowman, Graham, Kaplan, Mindermann, Greenblatt, Shlegeris, Schiefer, and Perez]{hubinger2024sleeperagentstrainingdeceptive}
Evan Hubinger, Carson Denison, Jesse Mu, Mike Lambert, Meg Tong, Monte MacDiarmid, Tamera Lanham, Daniel~M. Ziegler, Tim Maxwell, Newton Cheng, Adam Jermyn, Amanda Askell, Ansh Radhakrishnan, Cem Anil, David Duvenaud, Deep Ganguli, Fazl Barez, Jack Clark, Kamal Ndousse, Kshitij Sachan, Michael Sellitto, Mrinank Sharma, Nova DasSarma, Roger Grosse, Shauna Kravec, Yuntao Bai, Zachary Witten, Marina Favaro, Jan Brauner, Holden Karnofsky, Paul Christiano, Samuel~R. Bowman, Logan Graham, Jared Kaplan, Sören Mindermann, Ryan Greenblatt, Buck Shlegeris, Nicholas Schiefer, and Ethan Perez.
\newblock Sleeper agents: Training deceptive llms that persist through safety training, 2024.
\newblock URL \url{https://arxiv.org/abs/2401.05566}.

\bibitem[Hupkes \& Zuidema(2018)Hupkes and Zuidema]{hupkes2018visualisationdiagnosticclassifiersreveal}
Dieuwke Hupkes and Willem Zuidema.
\newblock Visualisation and 'diagnostic classifiers' reveal how recurrent and recursive neural networks process hierarchical structure (extended abstract).
\newblock In \emph{Proceedings of the Twenty-Seventh International Joint Conference on Artificial Intelligence, {IJCAI-18}}, pp.\  5617--5621. International Joint Conferences on Artificial Intelligence Organization, 7 2018.
\newblock \doi{10.24963/ijcai.2018/796}.
\newblock URL \url{https://doi.org/10.24963/ijcai.2018/796}.

\bibitem[Jacot et~al.(2018)Jacot, Gabriel, and Hongler]{jacot2020neuraltangentkernelconvergence}
Arthur Jacot, Franck Gabriel, and Clement Hongler.
\newblock Neural tangent kernel: Convergence and generalization in neural networks.
\newblock In S.~Bengio, H.~Wallach, H.~Larochelle, K.~Grauman, N.~Cesa-Bianchi, and R.~Garnett (eds.), \emph{Advances in Neural Information Processing Systems}, volume~31. Curran Associates, Inc., 2018.
\newblock URL \url{https://proceedings.neurips.cc/paper_files/paper/2018/file/5a4be1fa34e62bb8a6ec6b91d2462f5a-Paper.pdf}.

\bibitem[Jacovi(2023)]{Jacovi_2023_ExplainableAiLiterature}
Alon Jacovi.
\newblock Trends in explainable ai (xai) literature, 2023.
\newblock URL \url{https://arxiv.org/abs/2301.05433}.

\bibitem[Jain et~al.(2024)Jain, Kirk, Lubana, Dick, Tanaka, Rockt{\"a}schel, Grefenstette, and Krueger]{jain2024mechanisticallyanalyzingeffectsfinetuning}
Samyak Jain, Robert Kirk, Ekdeep~Singh Lubana, Robert~P. Dick, Hidenori Tanaka, Tim Rockt{\"a}schel, Edward Grefenstette, and David Krueger.
\newblock Mechanistically analyzing the effects of fine-tuning on procedurally defined tasks.
\newblock In \emph{The Twelfth International Conference on Learning Representations}, 2024.
\newblock URL \url{https://openreview.net/forum?id=A0HKeKl4Nl}.

\bibitem[Jain \& Wallace(2019)Jain and Wallace]{jain2019attentionexplanation}
Sarthak Jain and Byron~C. Wallace.
\newblock {A}ttention is not {E}xplanation.
\newblock In Jill Burstein, Christy Doran, and Thamar Solorio (eds.), \emph{Proceedings of the 2019 Conference of the North {A}merican Chapter of the Association for Computational Linguistics: Human Language Technologies, Volume 1 (Long and Short Papers)}, pp.\  3543--3556, Minneapolis, Minne@inproceedings{katz-etal-2024-backward, title = "Backward Lens: Projecting Language Model Gradients into the Vocabulary Space", author = "Katz, Shahar and Belinkov, Yonatan and Geva, Mor and Wolf, Lior", editor = "Al-Onaizan, Yaser and Bansal, Mohit and Chen, Yun-Nung", booktitle = "Proceedings of the 2024 Conference on Empirical Methods in Natural Language Processing", month = nov, year = "2024", address = "Miami, Florida, USA", publisher = "Association for Computational Linguistics", url = "https://aclanthology.org/2024.emnlp-main.142/", doi = "10.18653/v1/2024.emnlp-main.142", pages = "2390--2422", abstract = "Understanding how Transformer-based Language Models (LMs) learn and recall information is a key goal of the
  deep learning community. Recent interpretability methods project weights and hidden states obtained from the forward pass to the models' vocabularies, helping to uncover how information flows within LMs. In this work, we extend this methodology to LMs' backward pass and gradients. We first prove that a gradient matrix can be cast as a low-rank linear combination of its forward and backward passes' inputs. We then develop methods to project these gradients into vocabulary items and explore the mechanics of how new information is stored in the LMs' neurons." }sota, June 2019. Association for Computational Linguistics.
\newblock \doi{10.18653/v1/N19-1357}.
\newblock URL \url{https://aclanthology.org/N19-1357/}.

\bibitem[Janiak et~al.(2023)Janiak, Mathwin, and Heimersheim]{janiak2023polysemantic}
Jett Janiak, Chris Mathwin, and Stefan Heimersheim.
\newblock Polysemantic attention head in a 4-layer transformer.
\newblock LessWrong Blog, 2023.
\newblock URL \url{https://www.lesswrong.com/posts/nuJFTS5iiJKT5G5yh/polysemantic-attention-head-in-a-4-layer-transformer}.

\bibitem[Jermyn et~al.(2023)Jermyn, Olah, and Henighan]{jermyn2023attention}
Adam Jermyn, Chris Olah, and Tom Henighan.
\newblock Attention head superposition.
\newblock \emph{Transformer Circuits}, 2023.
\newblock URL \url{https://transformer-circuits.pub/2023/may-update/index.html#attention-superposition}.

\bibitem[Jin et~al.(2024)Jin, Cao, Yuan, Chen, Xu, Li, Jiang, Liu, and Zhao]{jin-etal-2024-cutting}
Zhuoran Jin, Pengfei Cao, Hongbang Yuan, Yubo Chen, Jiexin Xu, Huaijun Li, Xiaojian Jiang, Kang Liu, and Jun Zhao.
\newblock Cutting off the head ends the conflict: A mechanism for interpreting and mitigating knowledge conflicts in language models.
\newblock In Lun-Wei Ku, Andre Martins, and Vivek Srikumar (eds.), \emph{Findings of the Association for Computational Linguistics ACL 2024}, pp.\  1193--1215, Bangkok, Thailand and virtual meeting, August 2024. Association for Computational Linguistics.
\newblock URL \url{https://aclanthology.org/2024.findings-acl.70}.

\bibitem[Johnston et~al.(2024)Johnston, Chakraborty, and Belrose]{Johnston_2024_MAD}
David Johnston, Arkajyoti Chakraborty, and Nora Belrose.
\newblock Mechanistic anomaly detection research update, Aug 2024.
\newblock URL \url{https://blog.eleuther.ai/mad_research_update/}.

\bibitem[Juang et~al.(2024)Juang, Paulo, Drori, and Belrose]{Belrose_2024_autointerp}
Caden Juang, Gonccedilalo Paulo, Jacob Drori, and Nora Belrose.
\newblock Open source automated interpretability for sparse autoencoder features, Jul 2024.
\newblock URL \url{https://blog.eleuther.ai/autointerp/}.

\bibitem[Jumper et~al.(2021)Jumper, Evans, Pritzel, Green, Figurnov, Ronneberger, Tunyasuvunakool, Bates, {\v{Z}}{\'\i}dek, Potapenko, Bridgland, Meyer, Kohl, Ballard, Cowie, Romera-Paredes, Nikolov, Jain, Adler, Back, Petersen, Reiman, Clancy, Zielinski, Steinegger, Pacholska, Berghammer, Bodenstein, Silver, Vinyals, Senior, Kavukcuoglu, Kohli, and Hassabis]{jumper2021highly}
John Jumper, Richard Evans, Alexander Pritzel, Tim Green, Michael Figurnov, Olaf Ronneberger, Kathryn Tunyasuvunakool, Russ Bates, Augustin {\v{Z}}{\'\i}dek, Anna Potapenko, Alex Bridgland, Clemens Meyer, Simon A~A Kohl, Andrew~J Ballard, Andrew Cowie, Bernardino Romera-Paredes, Stanislav Nikolov, Rishub Jain, Jonas Adler, Trevor Back, Stig Petersen, David Reiman, Ellen Clancy, Michal Zielinski, Martin Steinegger, Michalina Pacholska, Tamas Berghammer, Sebastian Bodenstein, David Silver, Oriol Vinyals, Andrew~W Senior, Koray Kavukcuoglu, Pushmeet Kohli, and Demis Hassabis.
\newblock Highly accurate protein structure prediction with {AlphaFold}.
\newblock \emph{Nature}, 596:\penalty0 583--589, 2021.
\newblock \doi{10.1038/s41586-021-03819-2}.
\newblock URL \url{https://doi.org/10.1038/s41586-021-03819-2}.

\bibitem[Kantamneni et~al.(2024)Kantamneni, Engels, Rajamanoharan, and Nanda]{kantamneni2024sae}
Subhash Kantamneni, Josh Engels, Senthooran Rajamanoharan, and Neel Nanda.
\newblock {SAE} probing: What is it good for? absolutely something!
\newblock \emph{AI Alignment Forum}, nov 2024.
\newblock URL \url{https://www.lesswrong.com/posts/NMLq8yoTecAF44KX9/sae-probing-what-is-it-good-for-absolutely-something}.

\bibitem[Katz et~al.(2024)Katz, Belinkov, Geva, and Wolf]{katz2024backwardlensprojectinglanguage}
Shahar Katz, Yonatan Belinkov, Mor Geva, and Lior Wolf.
\newblock Backward lens: Projecting language model gradients into the vocabulary space.
\newblock In Yaser Al-Onaizan, Mohit Bansal, and Yun-Nung Chen (eds.), \emph{Proceedings of the 2024 Conference on Empirical Methods in Natural Language Processing}, pp.\  2390--2422, Miami, Florida, USA, November 2024. Association for Computational Linguistics.
\newblock \doi{10.18653/v1/2024.emnlp-main.142}.
\newblock URL \url{https://aclanthology.org/2024.emnlp-main.142/}.

\bibitem[Kharlapenko et~al.(2024)Kharlapenko, neverix, Nanda, and Conmy]{kharlapenko2024self}
Dmitrii Kharlapenko, neverix, Neel Nanda, and Arthur Conmy.
\newblock Self-explaining {SAE} features.
\newblock \emph{Alignment Forum}, August 2024.
\newblock URL \url{https://www.alignmentforum.org/posts/8ev6coxChSWcxDy8/self-explaining-sae-features}.

\bibitem[Kim et~al.(2018{\natexlab{a}})Kim, Wattenberg, Gilmer, Cai, Wexler, Viegas, and sayres]{Kim_2018_Activation}
Been Kim, Martin Wattenberg, Justin Gilmer, Carrie Cai, James Wexler, Fernanda Viegas, and Rory sayres.
\newblock Interpretability beyond feature attribution: Quantitative testing with concept activation vectors ({TCAV}).
\newblock In Jennifer Dy and Andreas Krause (eds.), \emph{Proceedings of the 35th International Conference on Machine Learning}, volume~80 of \emph{Proceedings of Machine Learning Research}, pp.\  2668--2677. PMLR, 10--15 Jul 2018{\natexlab{a}}.
\newblock URL \url{https://proceedings.mlr.press/v80/kim18d.html}.

\bibitem[Kim et~al.(2018{\natexlab{b}})Kim, Wattenberg, Gilmer, Cai, Wexler, Viegas, and sayres]{kim_2018_activationvectors}
Been Kim, Martin Wattenberg, Justin Gilmer, Carrie Cai, James Wexler, Fernanda Viegas, and Rory sayres.
\newblock Interpretability beyond feature attribution: Quantitative testing with concept activation vectors ({TCAV}).
\newblock In Jennifer Dy and Andreas Krause (eds.), \emph{Proceedings of the 35th International Conference on Machine Learning}, volume~80 of \emph{Proceedings of Machine Learning Research}, pp.\  2668--2677. PMLR, 10--15 Jul 2018{\natexlab{b}}.
\newblock URL \url{https://proceedings.mlr.press/v80/kim18d.html}.

\bibitem[Kindermans et~al.(2019)Kindermans, Hooker, Adebayo, Alber, Sch{\"u}tt, D{\"a}hne, Erhan, and Kim]{Kindermans2019saliencymethods}
Pieter-Jan Kindermans, Sara Hooker, Julius Adebayo, Maximilian Alber, Kristof~T. Sch{\"u}tt, Sven D{\"a}hne, Dumitru Erhan, and Been Kim.
\newblock \emph{The (Un)reliability of Saliency Methods}, pp.\  267--280.
\newblock Springer International Publishing, Cham, 2019.
\newblock ISBN 978-3-030-28954-6.
\newblock \doi{10.1007/978-3-030-28954-6_14}.
\newblock URL \url{https://doi.org/10.1007/978-3-030-28954-6_14}.

\bibitem[Kirch et~al.(2024)Kirch, Field, and Casper]{kirch2024featurespromptsjailbreakllms}
Nathalie~Maria Kirch, Severin Field, and Stephen Casper.
\newblock What features in prompts jailbreak llms? investigating the mechanisms behind attacks, 2024.
\newblock URL \url{https://arxiv.org/abs/2411.03343}.

\bibitem[Kirsch et~al.(2018)Kirsch, Kunze, and Barber]{Kirsch_2018_modularnetworks}
Louis Kirsch, Julius Kunze, and David Barber.
\newblock Modular networks: learning to decompose neural computation.
\newblock In \emph{Proceedings of the 32nd International Conference on Neural Information Processing Systems}, NIPS'18, pp.\  2414–2423, Red Hook, NY, USA, 2018. Curran Associates Inc.

\bibitem[Kissane et~al.(2024{\natexlab{a}})Kissane, Conmy, and Nanda]{Kissane_Conmy_Nanda_2024}
Connor Kissane, Arthur Conmy, and Neel Nanda.
\newblock Attention output saes improve circuit analysis - ai alignment forum, Jun 2024{\natexlab{a}}.
\newblock URL \url{https://www.alignmentforum.org/posts/EGvtgB7ctifzxZg6v/attention-output-saes-improve-circuit-analysis}.

\bibitem[Kissane et~al.(2024{\natexlab{b}})Kissane, Krzyzanowski, Bloom, Conmy, and Nanda]{kissane2024interpretingattentionlayeroutputs}
Connor Kissane, Robert Krzyzanowski, Joseph~Isaac Bloom, Arthur Conmy, and Neel Nanda.
\newblock Interpreting attention layer outputs with sparse autoencoders, 2024{\natexlab{b}}.
\newblock URL \url{https://arxiv.org/abs/2406.17759}.

\bibitem[Kissane et~al.(2024{\natexlab{c}})Kissane, Krzyzanowski, Nanda, and Conmy]{kissane2024saes}
Connor Kissane, Robert Krzyzanowski, Neel Nanda, and Arthur Conmy.
\newblock {SAE}s are highly dataset dependent: a case study on the refusal direction.
\newblock \emph{Alignment Forum}, nov 2024{\natexlab{c}}.
\newblock URL \url{https://www.alignmentforum.org/posts/rtp6n7Z23uJpEH7od/saes-are-highly-dataset-dependent-a-case-study-on-the}.

\bibitem[Koh \& Liang(2017)Koh and Liang]{Koh_2017_influence_function}
Pang~Wei Koh and Percy Liang.
\newblock Understanding black-box predictions via influence functions.
\newblock In Doina Precup and Yee~Whye Teh (eds.), \emph{Proceedings of the 34th International Conference on Machine Learning}, volume~70 of \emph{Proceedings of Machine Learning Research}, pp.\  1885--1894. PMLR, 06--11 Aug 2017.
\newblock URL \url{https://proceedings.mlr.press/v70/koh17a.html}.

\bibitem[Koh et~al.(2020)Koh, Nguyen, Tang, Mussmann, Pierson, Kim, and Liang]{Koh_2020_ConceptBottleneckModels}
Pang~Wei Koh, Thao Nguyen, Yew~Siang Tang, Stephen Mussmann, Emma Pierson, Been Kim, and Percy Liang.
\newblock Concept bottleneck models.
\newblock In Hal~Daumé III and Aarti Singh (eds.), \emph{Proceedings of the 37th International Conference on Machine Learning}, volume 119 of \emph{Proceedings of Machine Learning Research}, pp.\  5338--5348. PMLR, 13--18 Jul 2020.
\newblock URL \url{https://proceedings.mlr.press/v119/koh20a.html}.

\bibitem[K{\"o}hn(2015)]{kohn-2015-whats}
Arne K{\"o}hn.
\newblock What{'}s in an embedding? analyzing word embeddings through multilingual evaluation.
\newblock In Llu{\'\i}s M{\`a}rquez, Chris Callison-Burch, and Jian Su (eds.), \emph{Proceedings of the 2015 Conference on Empirical Methods in Natural Language Processing}, pp.\  2067--2073, Lisbon, Portugal, September 2015. Association for Computational Linguistics.
\newblock \doi{10.18653/v1/D15-1246}.
\newblock URL \url{https://aclanthology.org/D15-1246}.

\bibitem[Kojima et~al.(2024)Kojima, Gu, Reid, Matsuo, and Iwasawa]{kojima2023largelanguagemodelszeroshot}
Takeshi Kojima, Shixiang~Shane Gu, Machel Reid, Yutaka Matsuo, and Yusuke Iwasawa.
\newblock Large language models are zero-shot reasoners.
\newblock In \emph{Proceedings of the 36th International Conference on Neural Information Processing Systems}, NIPS '22, Red Hook, NY, USA, 2024. Curran Associates Inc.
\newblock ISBN 9781713871088.

\bibitem[Korot et~al.(2021)Korot, Pontikos, Liu, et~al.]{korot2021predicting}
Edward Korot, Nikolas Pontikos, Xiaoxuan Liu, et~al.
\newblock Predicting sex from retinal fundus photographs using automated deep learning.
\newblock \emph{Scientific Reports}, 11:\penalty0 10286, 2021.
\newblock \doi{10.1038/s41598-021-89743-x}.
\newblock URL \url{https://doi.org/10.1038/s41598-021-89743-x}.

\bibitem[Kosoy et~al.(2023)Kosoy, Reagan, Lai, Gopnik, and Cobb]{kosoy2023comparingmachineschildrenusing}
Eliza Kosoy, Emily~Rose Reagan, Leslie Lai, Alison Gopnik, and Danielle~Krettek Cobb.
\newblock Comparing machines and children: Using developmental psychology experiments to assess the strengths and weaknesses of lamda responses, 2023.
\newblock URL \url{https://arxiv.org/abs/2305.11243}.

\bibitem[Kramár et~al.(2024)Kramár, Lieberum, Shah, and Nanda]{kramár2024atpefficientscalablemethod}
János Kramár, Tom Lieberum, Rohin Shah, and Neel Nanda.
\newblock Atp*: An efficient and scalable method for localizing llm behaviour to components, 2024.
\newblock URL \url{https://arxiv.org/abs/2403.00745}.

\bibitem[Krishnan(2020)]{krishnan2020against}
Mukund Krishnan.
\newblock Against interpretability: a critical examination of the interpretability problem in machine learning.
\newblock \emph{Philosophy \& Technology}, 33\penalty0 (3):\penalty0 487--502, 2020.
\newblock \doi{10.1007/s13347-019-00372-9}.
\newblock URL \url{https://doi.org/10.1007/s13347-019-00372-9}.

\bibitem[Krizhevsky et~al.(2012)Krizhevsky, Sutskever, and Hinton]{Krizhevsky_2012_ImageNet}
Alex Krizhevsky, Ilya Sutskever, and Geoffrey~E. Hinton.
\newblock Imagenet classification with deep convolutional neural networks.
\newblock In \emph{NeurIPS}, pp.\  84–90, may 2012.
\newblock \doi{10.1145/3065386}.

\bibitem[Laine et~al.(2024)Laine, Chughtai, Betley, Hariharan, Balesni, Scheurer, Hobbhahn, Meinke, and Evans]{laine2024me}
Rudolf Laine, Bilal Chughtai, Jan Betley, Kaivalya Hariharan, Mikita Balesni, J{\'e}r{\'e}my Scheurer, Marius Hobbhahn, Alexander Meinke, and Owain Evans.
\newblock Me, myself, and {AI}: The situational awareness dataset ({SAD}) for {LLM}s.
\newblock In \emph{The Thirty-eight Conference on Neural Information Processing Systems Datasets and Benchmarks Track}, 2024.
\newblock URL \url{https://openreview.net/forum?id=UnWhcpIyUC}.

\bibitem[Lanham et~al.(2023)Lanham, Chen, Radhakrishnan, Steiner, Denison, Hernandez, Li, Durmus, Hubinger, Kernion, Lukošiūtė, Nguyen, Cheng, Joseph, Schiefer, Rausch, Larson, McCandlish, Kundu, Kadavath, Yang, Henighan, Maxwell, Telleen-Lawton, Hume, Hatfield-Dodds, Kaplan, Brauner, Bowman, and Perez]{lanham2023measuringfaithfulnesschainofthoughtreasoning}
Tamera Lanham, Anna Chen, Ansh Radhakrishnan, Benoit Steiner, Carson Denison, Danny Hernandez, Dustin Li, Esin Durmus, Evan Hubinger, Jackson Kernion, Kamilė Lukošiūtė, Karina Nguyen, Newton Cheng, Nicholas Joseph, Nicholas Schiefer, Oliver Rausch, Robin Larson, Sam McCandlish, Sandipan Kundu, Saurav Kadavath, Shannon Yang, Thomas Henighan, Timothy Maxwell, Timothy Telleen-Lawton, Tristan Hume, Zac Hatfield-Dodds, Jared Kaplan, Jan Brauner, Samuel~R. Bowman, and Ethan Perez.
\newblock Measuring faithfulness in chain-of-thought reasoning, 2023.
\newblock URL \url{https://arxiv.org/abs/2307.13702}.

\bibitem[Le et~al.(2012)Le, Ranzato, Monga, Devin, Chen, Corrado, Dean, and Ng]{le2012buildinghighlevelfeaturesusing}
Quoc~V. Le, Marc'Aurelio Ranzato, Rajat Monga, Matthieu Devin, Kai Chen, Greg~S. Corrado, Jeff Dean, and Andrew~Y. Ng.
\newblock Building high-level features using large scale unsupervised learning.
\newblock In \emph{Proceedings of the 29th International Coference on International Conference on Machine Learning}, ICML'12, pp.\  507–514, Madison, WI, USA, 2012. Omnipress.
\newblock ISBN 9781450312851.

\bibitem[Leavitt \& Morcos(2020)Leavitt and Morcos]{leavitt2020falsifiableinterpretabilityresearch}
Matthew~L. Leavitt and Ari Morcos.
\newblock Towards falsifiable interpretability research, 2020.
\newblock URL \url{https://arxiv.org/abs/2010.12016}.

\bibitem[LeCun et~al.(2015)LeCun, Bengio, and Hinton]{LeCun_2015_DeepLearning}
Yann LeCun, Yoshua Bengio, and Geoffrey Hinton.
\newblock Deep learning, May 2015.
\newblock URL \url{https://www.nature.com/articles/nature14539}.

\bibitem[Lee et~al.(2025)Lee, Bai, Pres, Wattenberg, Kummerfeld, and Mihalcea]{lee2024mechanisticunderstandingalignmentalgorithms}
Andrew Lee, Xiaoyan Bai, Itamar Pres, Martin Wattenberg, Jonathan~K. Kummerfeld, and Rada Mihalcea.
\newblock A mechanistic understanding of alignment algorithms: a case study on dpo and toxicity.
\newblock In \emph{Proceedings of the 41st International Conference on Machine Learning}, ICML'24. JMLR.org, 2025.

\bibitem[Lee et~al.(2024)Lee, Cooper, and Grimmelmann]{lee2024talkinboutaigeneration}
Katherine Lee, A.~Feder Cooper, and James Grimmelmann.
\newblock Talkin' 'bout ai generation: Copyright and the generative-ai supply chain.
\newblock \emph{Journal of the Copyright Society of the USA}, 2024.
\newblock URL \url{https://arxiv.org/abs/2309.08133}.

\bibitem[Leech et~al.(2024)Leech, Vazquez, Yagudin, Kupper, and Aitchison]{leech2024questionablepracticesmachinelearning}
Gavin Leech, Juan~J. Vazquez, Misha Yagudin, Niclas Kupper, and Laurence Aitchison.
\newblock Questionable practices in machine learning, 2024.
\newblock URL \url{https://arxiv.org/abs/2407.12220}.

\bibitem[Lermen et~al.(2024)Lermen, Rogers-Smith, and Ladish]{lermen2024lorafinetuningefficientlyundoes}
Simon Lermen, Charlie Rogers-Smith, and Jeffrey Ladish.
\newblock Lora fine-tuning efficiently undoes safety training in llama 2-chat 70b, 2024.
\newblock URL \url{https://arxiv.org/abs/2310.20624}.

\bibitem[Li et~al.(2023)Li, Hopkins, Bau, Vi{\'e}gas, Pfister, and Wattenberg]{li2024emergentworldrepresentationsexploring}
Kenneth Li, Aspen~K Hopkins, David Bau, Fernanda Vi{\'e}gas, Hanspeter Pfister, and Martin Wattenberg.
\newblock Emergent world representations: Exploring a sequence model trained on a synthetic task.
\newblock In \emph{The Eleventh International Conference on Learning Representations}, 2023.
\newblock URL \url{https://openreview.net/forum?id=DeG07_TcZvT}.

\bibitem[Li et~al.(2024{\natexlab{a}})Li, Patel, Viégas, Pfister, and Wattenberg]{2024inferencetimeinterventionelicitingtruthful}
Kenneth Li, Oam Patel, Fernanda Viégas, Hanspeter Pfister, and Martin Wattenberg.
\newblock Inference-time intervention: Eliciting truthful answers from a language model.
\newblock In \emph{NeurIPS}, 2024{\natexlab{a}}.
\newblock URL \url{https://arxiv.org/abs/2306.03341}.

\bibitem[Li et~al.(2024{\natexlab{b}})Li, Pan, Gopal, Yue, Berrios, Gatti, Li, Dombrowski, Goel, Phan, Mukobi, Helm-Burger, Lababidi, Justen, Liu, Chen, Barrass, Zhang, Zhu, Tamirisa, Bharathi, Khoja, Zhao, Herbert-Voss, Breuer, Marks, Patel, Zou, Mazeika, Wang, Oswal, Liu, Hunt, Tienken-Harder, Shih, Talley, Guan, Kaplan, Steneker, Campbell, Jokubaitis, Levinson, Wang, Qian, Karmakar, Basart, Fitz, Levine, Kumaraguru, Tupakula, Varadharajan, Shoshitaishvili, Ba, Esvelt, Wang, and Hendrycks]{li2024wmdp}
Nathaniel Li, Alexander Pan, Anjali Gopal, Summer Yue, Daniel Berrios, Alice Gatti, Justin~D. Li, Ann-Kathrin Dombrowski, Shashwat Goel, Long Phan, Gabriel Mukobi, Nathan Helm-Burger, Rassin Lababidi, Lennart Justen, Andrew~B. Liu, Michael Chen, Isabelle Barrass, Oliver Zhang, Xiaoyuan Zhu, Rishub Tamirisa, Bhrugu Bharathi, Adam Khoja, Zhenqi Zhao, Ariel Herbert-Voss, Cort~B. Breuer, Samuel Marks, Oam Patel, Andy Zou, Mantas Mazeika, Zifan Wang, Palash Oswal, Weiran Liu, Adam~A. Hunt, Justin Tienken-Harder, Kevin~Y. Shih, Kemper Talley, John Guan, Russell Kaplan, Ian Steneker, David Campbell, Brad Jokubaitis, Alex Levinson, Jean Wang, William Qian, Kallol~Krishna Karmakar, Steven Basart, Stephen Fitz, Mindy Levine, Ponnurangam Kumaraguru, Uday Tupakula, Vijay Varadharajan, Yan Shoshitaishvili, Jimmy Ba, Kevin~M. Esvelt, Alexandr Wang, and Dan Hendrycks.
\newblock The wmdp benchmark: Measuring and reducing malicious use with unlearning, 2024{\natexlab{b}}.

\bibitem[Li et~al.(2015)Li, Yosinski, Clune, Lipson, and Hopcroft]{li2016convergentlearningdifferentneural}
Yixuan Li, Jason Yosinski, Jeff Clune, Hod Lipson, and John Hopcroft.
\newblock Convergent learning: Do different neural networks learn the same representations?
\newblock In Dmitry Storcheus, Afshin Rostamizadeh, and Sanjiv Kumar (eds.), \emph{Proceedings of the 1st International Workshop on Feature Extraction: Modern Questions and Challenges at NIPS 2015}, volume~44 of \emph{Proceedings of Machine Learning Research}, pp.\  196--212, Montreal, Canada, 11 Dec 2015. PMLR.
\newblock URL \url{https://proceedings.mlr.press/v44/li15convergent.html}.

\bibitem[Lieberum et~al.(2023)Lieberum, Rahtz, Kramár, Nanda, Irving, Shah, and Mikulik]{lieberum2023doescircuitanalysisinterpretability}
Tom Lieberum, Matthew Rahtz, János Kramár, Neel Nanda, Geoffrey Irving, Rohin Shah, and Vladimir Mikulik.
\newblock Does circuit analysis interpretability scale? evidence from multiple choice capabilities in chinchilla, 2023.
\newblock URL \url{https://arxiv.org/abs/2307.09458}.

\bibitem[Lieberum et~al.(2024)Lieberum, Rajamanoharan, Conmy, Smith, Sonnerat, Varma, Kramár, Dragan, Shah, and Nanda]{lieberum2024gemmascopeopensparse}
Tom Lieberum, Senthooran Rajamanoharan, Arthur Conmy, Lewis Smith, Nicolas Sonnerat, Vikrant Varma, János Kramár, Anca Dragan, Rohin Shah, and Neel Nanda.
\newblock Gemma scope: Open sparse autoencoders everywhere all at once on gemma 2, 2024.
\newblock URL \url{https://arxiv.org/abs/2408.05147}.

\bibitem[Lin et~al.(2024)Lin, He, Xu, Xing, Yamada, Liu, and Tang]{lin2024understandingjailbreakattacksllms}
Yuping Lin, Pengfei He, Han Xu, Yue Xing, Makoto Yamada, Hui Liu, and Jiliang Tang.
\newblock Towards understanding jailbreak attacks in {LLM}s: A representation space analysis.
\newblock In Yaser Al-Onaizan, Mohit Bansal, and Yun-Nung Chen (eds.), \emph{Proceedings of the 2024 Conference on Empirical Methods in Natural Language Processing}, pp.\  7067--7085, Miami, Florida, USA, November 2024. Association for Computational Linguistics.
\newblock \doi{10.18653/v1/2024.emnlp-main.401}.
\newblock URL \url{https://aclanthology.org/2024.emnlp-main.401/}.

\bibitem[Lin et~al.(2023)Lin, Akin, Rao, Hie, Zhu, Lu, Smetanin, Verkuil, Kabeli, Rives, et~al.]{lin2023evolutionary}
Zeming Lin, Halil Akin, Roshan Rao, Brian Hie, Zhongkai Zhu, Wenting Lu, Nikita Smetanin, Robert Verkuil, Ori Kabeli, Alexander Rives, et~al.
\newblock Evolutionary-scale prediction of atomic-level protein structure with a language model.
\newblock \emph{Science}, 379\penalty0 (6637):\penalty0 1123--1130, 3 2023.
\newblock \doi{10.1126/science.ade2574}.

\bibitem[Lindner et~al.(2023)Lindner, Kramar, Farquhar, Rahtz, McGrath, and Mikulik]{lindner2023tracrcompiledtransformerslaboratory}
David Lindner, Janos Kramar, Sebastian Farquhar, Matthew Rahtz, Thomas McGrath, and Vladimir Mikulik.
\newblock Tracr: Compiled transformers as a laboratory for interpretability.
\newblock In \emph{Thirty-seventh Conference on Neural Information Processing Systems}, 2023.
\newblock URL \url{https://openreview.net/forum?id=tbbId8u7nP}.

\bibitem[Lindsey et~al.(2024)Lindsey, Templeton, Marcus, Conerly, Batson, and Olah]{lindsey2024crosscoders}
Jack Lindsey, Adly Templeton, Jonathan Marcus, Thomas Conerly, Joshua Batson, and Christopher Olah.
\newblock Sparse crosscoders for cross-layer features and model diffing.
\newblock \emph{Transformer Circuits}, 2024.
\newblock URL \url{https://transformer-circuits.pub/2024/crosscoders/index.html}.
\newblock * Equal contribution.

\bibitem[Lipton(2018)]{Lipton_2016_InterpretabilityMotivations}
Zachary~C. Lipton.
\newblock The mythos of model interpretability: In machine learning, the concept of interpretability is both important and slippery.
\newblock \emph{Queue}, 16\penalty0 (3):\penalty0 31–57, June 2018.
\newblock ISSN 1542-7730.
\newblock \doi{10.1145/3236386.3241340}.
\newblock URL \url{https://doi.org/10.1145/3236386.3241340}.

\bibitem[Lipton \& Steinhardt(2019)Lipton and Steinhardt]{lipton2018troublingtrendsmachinelearning}
Zachary~C. Lipton and Jacob Steinhardt.
\newblock Troubling trends in machine learning scholarship: Some ml papers suffer from flaws that could mislead the public and stymie future research.
\newblock \emph{Queue}, 17\penalty0 (1):\penalty0 45–77, February 2019.
\newblock ISSN 1542-7730.
\newblock \doi{10.1145/3317287.3328534}.
\newblock URL \url{https://doi.org/10.1145/3317287.3328534}.

\bibitem[Liu et~al.(2024{\natexlab{a}})Liu, Yao, Jia, Casper, Baracaldo, Hase, Yao, Liu, Xu, Li, Varshney, Bansal, Koyejo, and Liu]{liu2024rethinkingmachineunlearninglarge}
Sijia Liu, Yuanshun Yao, Jinghan Jia, Stephen Casper, Nathalie Baracaldo, Peter Hase, Yuguang Yao, Chris~Yuhao Liu, Xiaojun Xu, Hang Li, Kush~R. Varshney, Mohit Bansal, Sanmi Koyejo, and Yang Liu.
\newblock Rethinking machine unlearning for large language models, 2024{\natexlab{a}}.
\newblock URL \url{https://arxiv.org/abs/2402.08787}.

\bibitem[Liu et~al.(2024{\natexlab{b}})Liu, Dou, Tan, Tian, and Jiang]{liu2024machineunlearninggenerativeai}
Zheyuan Liu, Guangyao Dou, Zhaoxuan Tan, Yijun Tian, and Meng Jiang.
\newblock Machine unlearning in generative ai: A survey, 2024{\natexlab{b}}.
\newblock URL \url{https://arxiv.org/abs/2407.20516}.

\bibitem[Liu et~al.(2023)Liu, Gan, and Tegmark]{liu2023seeingbelievingbraininspiredmodular}
Ziming Liu, Eric Gan, and Max Tegmark.
\newblock Seeing is believing: Brain-inspired modular training for mechanistic interpretability, 2023.
\newblock URL \url{https://arxiv.org/abs/2305.08746}.

\bibitem[Liu et~al.(2024{\natexlab{c}})Liu, Wang, Vaidya, Ruehle, Halverson, Soljačić, Hou, and Tegmark]{Liu_2024_KolmogorovArtificialNetwork}
Ziming Liu, Yixuan Wang, Sachin Vaidya, Fabian Ruehle, James Halverson, Marin Soljačić, Thomas~Y. Hou, and Max Tegmark.
\newblock Kan: Kolmogorov-arnold networks, 2024{\natexlab{c}}.
\newblock URL \url{https://arxiv.org/abs/2404.19756}.

\bibitem[Louizos et~al.(2018)Louizos, Welling, and Kingma]{louizos2018learningsparseneuralnetworks}
Christos Louizos, Max Welling, and Diederik~P. Kingma.
\newblock Learning sparse neural networks through $l_0$ regularization.
\newblock In \emph{International Conference on Learning Representations}, 2018.
\newblock URL \url{https://openreview.net/forum?id=H1Y8hhg0b}.

\bibitem[Ludwig \& Mullainathan(2023)Ludwig and Mullainathan]{Ludwig_2023}
Jens Ludwig and Sendhil Mullainathan.
\newblock Machine learning as a tool for hypothesis generation.
\newblock Working Paper 31017, National Bureau of Economic Research, March 2023.
\newblock URL \url{http://www.nber.org/papers/w31017}.

\bibitem[Lundberg \& Lee(2017)Lundberg and Lee]{Lundberg_2017_SHAP}
Scott~M. Lundberg and Su-In Lee.
\newblock A unified approach to interpreting model predictions.
\newblock In \emph{Proceedings of the 31st International Conference on Neural Information Processing Systems}, NIPS'17, pp.\  4768–4777, Red Hook, NY, USA, 2017. Curran Associates Inc.
\newblock ISBN 9781510860964.

\bibitem[Lynch et~al.(2024)Lynch, Guo, Ewart, Casper, and Hadfield-Menell]{lynch2024methodsevaluaterobustunlearning}
Aengus Lynch, Phillip Guo, Aidan Ewart, Stephen Casper, and Dylan Hadfield-Menell.
\newblock Eight methods to evaluate robust unlearning in llms, 2024.
\newblock URL \url{https://arxiv.org/abs/2402.16835}.

\bibitem[Madsen et~al.(2024)Madsen, Lakkaraju, Reddy, and Chandar]{madsen2024interpretabilityneedsnewparadigm}
Andreas Madsen, Himabindu Lakkaraju, Siva Reddy, and Sarath Chandar.
\newblock Interpretability needs a new paradigm, 2024.
\newblock URL \url{https://arxiv.org/abs/2405.05386}.

\bibitem[Mahendran \& Vedaldi(2014)Mahendran and Vedaldi]{mahendran2014understandingdeepimagerepresentations}
Aravindh Mahendran and Andrea Vedaldi.
\newblock Understanding deep image representations by inverting them, 2014.
\newblock URL \url{https://arxiv.org/abs/1412.0035}.

\bibitem[Makelov et~al.(2023)Makelov, Lange, and Nanda]{makelov2023subspacelookingforinterpretability}
Aleksandar Makelov, Georg Lange, and Neel Nanda.
\newblock Is this the subspace you are looking for? an interpretability illusion for subspace activation patching.
\newblock In \emph{NeurIPS Workshop on Attributing Model Behavior at Scale}, 2023.
\newblock URL \url{https://arxiv.org/abs/2311.17030}.

\bibitem[Makelov et~al.(2024)Makelov, Lange, and Nanda]{makelov2024principledevaluationssparseautoencoders}
Aleksandar Makelov, Georg Lange, and Neel Nanda.
\newblock Towards principled evaluations of sparse autoencoders for interpretability and control.
\newblock In \emph{ICLR 2024 Workshop on Secure and Trustworthy Large Language Models}, 2024.
\newblock URL \url{https://openreview.net/forum?id=MHIX9H8aYF}.

\bibitem[Makhzani \& Frey(2013)Makhzani and Frey]{makhzani2013ksparseautoencoders}
Alireza Makhzani and Brendan Frey.
\newblock k-sparse autoencoders, 2013.
\newblock URL \url{https://arxiv.org/abs/1312.5663}.

\bibitem[Mante et~al.(2013)Mante, Sussillo, Shenoy, and Newsome]{mante2013context}
Valerio Mante, David Sussillo, Krishna~V. Shenoy, and William~T. Newsome.
\newblock Context-dependent computation by recurrent dynamics in prefrontal cortex.
\newblock \emph{Nature}, 503:\penalty0 78--84, 2013.
\newblock \doi{10.1038/nature12742}.
\newblock URL \url{https://doi.org/10.1038/nature12742}.

\bibitem[Marks \& Tegmark(2024)Marks and Tegmark]{marks2024geometrytruthemergentlinear}
Samuel Marks and Max Tegmark.
\newblock The geometry of truth: Emergent linear structure in large language model representations of true/false datasets.
\newblock In \emph{First Conference on Language Modeling}, 2024.
\newblock URL \url{https://openreview.net/forum?id=aajyHYjjsk}.

\bibitem[Marks et~al.(2024)Marks, Rager, Michaud, Belinkov, Bau, and Mueller]{marks2024sparsefeaturecircuitsdiscovering}
Samuel Marks, Can Rager, Eric~J. Michaud, Yonatan Belinkov, David Bau, and Aaron Mueller.
\newblock Sparse feature circuits: Discovering and editing interpretable causal graphs in language models, 2024.
\newblock URL \url{https://arxiv.org/abs/2403.19647}.

\bibitem[Mathwin et~al.(2024)Mathwin, Akar, and Sharkey]{mathwin2024gated}
Chris Mathwin, Dennis Akar, and Lee Sharkey.
\newblock Gated attention blocks: Preliminary progress toward removing attention head superposition.
\newblock \emph{LessWrong}, apr 2024.
\newblock URL \url{https://www.lesswrong.com/posts/kzc3qNMsP2xJcxhGn/gated-attention-blocks-preliminary-progress-toward-removing-1}.

\bibitem[McDougall et~al.(2024)McDougall, Conmy, Rushing, McGrath, and Nanda]{mcdougall2023copysuppressioncomprehensivelyunderstanding}
Callum~Stuart McDougall, Arthur Conmy, Cody Rushing, Thomas McGrath, and Neel Nanda.
\newblock Copy suppression: Comprehensively understanding a motif in language model attention heads.
\newblock In \emph{The 7th BlackboxNLP Workshop}, 2024.
\newblock URL \url{https://openreview.net/forum?id=5Hd6813x3U}.

\bibitem[McGrath et~al.(2022)McGrath, Kapishnikov, Tomašev, Pearce, Wattenberg, Hassabis, Kim, Paquet, and Kramnik]{McGrath_2022_alphazero}
Thomas McGrath, Andrei Kapishnikov, Nenad Tomašev, Adam Pearce, Martin Wattenberg, Demis Hassabis, Been Kim, Ulrich Paquet, and Vladimir Kramnik.
\newblock Acquisition of chess knowledge in alphazero.
\newblock \emph{Proceedings of the National Academy of Sciences}, 119\penalty0 (47):\penalty0 e2206625119, 2022.
\newblock \doi{10.1073/pnas.2206625119}.
\newblock URL \url{https://www.pnas.org/doi/abs/10.1073/pnas.2206625119}.

\bibitem[McGrath et~al.(2023)McGrath, Rahtz, Kramar, Mikulik, and Legg]{mcgrath2023hydraeffectemergentselfrepair}
Thomas McGrath, Matthew Rahtz, Janos Kramar, Vladimir Mikulik, and Shane Legg.
\newblock The hydra effect: Emergent self-repair in language model computations, 2023.
\newblock URL \url{https://arxiv.org/abs/2307.15771}.

\bibitem[Meinke et~al.(2024)Meinke, Schoen, Scheurer, Balesni, Shah, and Hobbhahn]{meinke2024frontiermodelscapableincontext}
Alexander Meinke, Bronson Schoen, Jérémy Scheurer, Mikita Balesni, Rusheb Shah, and Marius Hobbhahn.
\newblock Frontier models are capable of in-context scheming, 2024.
\newblock URL \url{https://arxiv.org/abs/2412.04984}.

\bibitem[Mendel(2024)]{mendel2024sae}
Jake Mendel.
\newblock {SAE} feature geometry is outside the superposition hypothesis.
\newblock \emph{Alignment Forum}, 2024.
\newblock URL \url{https://www.alignmentforum.org/posts/MFBTjb2qf3ziWmzz6/sae-feature-geometry-is-outside-the-superposition-hypothesis}.

\bibitem[Meng et~al.(2022{\natexlab{a}})Meng, Bau, Andonian, and Belinkov]{meng2023locatingeditingfactualassociations}
Kevin Meng, David Bau, Alex Andonian, and Yonatan Belinkov.
\newblock Locating and editing factual associations in {GPT}.
\newblock \emph{Advances in Neural Information Processing Systems}, 36, 2022{\natexlab{a}}.
\newblock arXiv:2202.05262.

\bibitem[Meng et~al.(2022{\natexlab{b}})Meng, Bau, Andonian, and Belinkov]{meng2022locating}
Kevin Meng, David Bau, Alex~J Andonian, and Yonatan Belinkov.
\newblock Locating and editing factual associations in {GPT}.
\newblock In Alice~H. Oh, Alekh Agarwal, Danielle Belgrave, and Kyunghyun Cho (eds.), \emph{Advances in Neural Information Processing Systems}, 2022{\natexlab{b}}.
\newblock URL \url{https://openreview.net/forum?id=-h6WAS6eE4}.

\bibitem[Michaud et~al.(2023)Michaud, Liu, Girit, and Tegmark]{michaud2023the}
Eric~J Michaud, Ziming Liu, Uzay Girit, and Max Tegmark.
\newblock The quantization model of neural scaling.
\newblock In \emph{Thirty-seventh Conference on Neural Information Processing Systems}, 2023.
\newblock URL \url{https://openreview.net/forum?id=3tbTw2ga8K}.

\bibitem[Michaud et~al.(2024)Michaud, Liao, Lad, Liu, Mudide, Loughridge, Guo, Kheirkhah, Vukelić, and Tegmark]{michaud2024openingaiblackbox}
Eric~J. Michaud, Isaac Liao, Vedang Lad, Ziming Liu, Anish Mudide, Chloe Loughridge, Zifan~Carl Guo, Tara~Rezaei Kheirkhah, Mateja Vukelić, and Max Tegmark.
\newblock Opening the ai black box: program synthesis via mechanistic interpretability, 2024.
\newblock URL \url{https://arxiv.org/abs/2402.05110}.

\bibitem[Mikolov et~al.(2013)Mikolov, Yih, and Zweig]{mikolov-etal-2013-linguistic}
Tomas Mikolov, Wen-tau Yih, and Geoffrey Zweig.
\newblock Linguistic regularities in continuous space word representations.
\newblock In Lucy Vanderwende, Hal Daum{\'e}~III, and Katrin Kirchhoff (eds.), \emph{Proceedings of the 2013 Conference of the North {A}merican Chapter of the Association for Computational Linguistics: Human Language Technologies}, pp.\  746--751, Atlanta, Georgia, June 2013. Association for Computational Linguistics.
\newblock URL \url{https://aclanthology.org/N13-1090}.

\bibitem[Miller et~al.(2024)Miller, Chughtai, and Saunders]{miller2024transformercircuitfaithfulnessmetrics}
Joseph Miller, Bilal Chughtai, and William Saunders.
\newblock Transformer circuit evaluation metrics are not robust.
\newblock In \emph{First Conference on Language Modeling}, 2024.
\newblock URL \url{https://openreview.net/forum?id=zSf8PJyQb2}.

\bibitem[Miller(2019)]{MILLER_2019_Explanation}
Tim Miller.
\newblock Explanation in artificial intelligence: Insights from the social sciences.
\newblock \emph{Artificial Intelligence}, 267:\penalty0 1--38, 2019.
\newblock ISSN 0004-3702.
\newblock \doi{https://doi.org/10.1016/j.artint.2018.07.007}.
\newblock URL \url{https://www.sciencedirect.com/science/article/pii/S0004370218305988}.

\bibitem[Mocanu et~al.(2018)Mocanu, Mocanu, Stone, Nguyen, Gibescu, and Liotta]{mocanu2018scalable}
Decebal~Constantin Mocanu, Elena Mocanu, Peter Stone, Phuong~H. Nguyen, Madeleine Gibescu, and Antonio Liotta.
\newblock Scalable training of artificial neural networks with adaptive sparse connectivity inspired by network science.
\newblock \emph{Nature Communications}, 9:\penalty0 2383, 2018.
\newblock \doi{10.1038/s41467-018-04316-3}.
\newblock URL \url{https://doi.org/10.1038/s41467-018-04316-3}.

\bibitem[Molchanov et~al.(2017)Molchanov, Tyree, Karras, Aila, and Kautz]{molchanov2017pruning}
Pavlo Molchanov, Stephen Tyree, Tero Karras, Timo Aila, and Jan Kautz.
\newblock Pruning convolutional neural networks for resource efficient inference.
\newblock In \emph{International Conference on Learning Representations}, 2017.
\newblock URL \url{https://openreview.net/forum?id=SJGCiw5gl}.

\bibitem[Molnar et~al.(2021)Molnar, König, Herbinger, Freiesleben, Dandl, Scholbeck, Casalicchio, Grosse-Wentrup, and Bischl]{molnar2021generalpitfallsmodelagnosticinterpretation}
Christoph Molnar, Gunnar König, Julia Herbinger, Timo Freiesleben, Susanne Dandl, Christian~A. Scholbeck, Giuseppe Casalicchio, Moritz Grosse-Wentrup, and Bernd Bischl.
\newblock General pitfalls of model-agnostic interpretation methods for machine learning models, 2021.
\newblock URL \url{https://arxiv.org/abs/2007.04131}.

\bibitem[Molnar et~al.(2024)Molnar, K{\"o}nig, Bischl, and Casalicchio]{molnar2024model}
Christoph Molnar, Gunnar K{\"o}nig, Bernd Bischl, and Giuseppe Casalicchio.
\newblock Model-agnostic feature importance and effects with dependent features: a conditional subgroup approach.
\newblock \emph{Data Mining and Knowledge Discovery}, 38:\penalty0 2903--2941, 2024.
\newblock \doi{10.1007/s10618-022-00901-9}.
\newblock URL \url{https://doi.org/10.1007/s10618-022-00901-9}.

\bibitem[Mordvintsev(2015)]{Mordvintsev_2015}
Mordvintsev.
\newblock Inceptionism: Going deeper into neural networks, 2015.
\newblock URL \url{https://research.google/blog/inceptionism-going-deeper-into-neural-networks/}.

\bibitem[Mosbach et~al.(2024)Mosbach, Gautam, Vergara-Browne, Klakow, and Geva]{mosbach2024insightsactionsimpactinterpretability}
Marius Mosbach, Vagrant Gautam, Tomás Vergara-Browne, Dietrich Klakow, and Mor Geva.
\newblock From insights to actions: The impact of interpretability and analysis research on nlp, 2024.
\newblock URL \url{https://arxiv.org/abs/2406.12618}.

\bibitem[Mozer \& Smolensky(1988)Mozer and Smolensky]{Mozer_1988_skeletonization}
Michael~C Mozer and Paul Smolensky.
\newblock Skeletonization: A technique for trimming the fat from a network via relevance assessment.
\newblock In D.~Touretzky (ed.), \emph{Advances in Neural Information Processing Systems}, volume~1. Morgan-Kaufmann, 1988.
\newblock URL \url{https://proceedings.neurips.cc/paper_files/paper/1988/file/07e1cd7dca89a1678042477183b7ac3f-Paper.pdf}.

\bibitem[Mu \& Andreas(2020)Mu and Andreas]{Mu_2020_Advances}
Jesse Mu and Jacob Andreas.
\newblock Compositional explanations of neurons.
\newblock In H.~Larochelle, M.~Ranzato, R.~Hadsell, M.F. Balcan, and H.~Lin (eds.), \emph{Advances in Neural Information Processing Systems}, volume~33, pp.\  17153--17163. Curran Associates, Inc., 2020.
\newblock URL \url{https://proceedings.neurips.cc/paper_files/paper/2020/file/c74956ffb38ba48ed6ce977af6727275-Paper.pdf}.

\bibitem[Mueller(2024)]{mueller2024missedcausesambiguouseffects}
Aaron Mueller.
\newblock Missed causes and ambiguous effects: Counterfactuals pose challenges for interpreting neural networks, 2024.
\newblock URL \url{https://arxiv.org/abs/2407.04690}.

\bibitem[Mueller et~al.(2024)Mueller, Brinkmann, Li, Marks, Pal, Prakash, Rager, Sankaranarayanan, Sharma, Sun, Todd, Bau, and Belinkov]{mueller2024questrightmediatorhistory}
Aaron Mueller, Jannik Brinkmann, Millicent Li, Samuel Marks, Koyena Pal, Nikhil Prakash, Can Rager, Aruna Sankaranarayanan, Arnab~Sen Sharma, Jiuding Sun, Eric Todd, David Bau, and Yonatan Belinkov.
\newblock The quest for the right mediator: A history, survey, and theoretical grounding of causal interpretability, 2024.
\newblock URL \url{https://arxiv.org/abs/2408.01416}.

\bibitem[Nanda(2023{\natexlab{a}})]{NeelNanda_2023a}
Neel Nanda.
\newblock Attribution patching: Activation patching at industrial scale, Mar 2023{\natexlab{a}}.
\newblock URL \url{https://www.neelnanda.io/mechanistic-interpretability/attribution-patching}.

\bibitem[Nanda(2023{\natexlab{b}})]{nanda2023othello}
Neel Nanda.
\newblock Othello-{GPT}: Reflections on the research process.
\newblock \emph{Alignment Forum}, March 2023{\natexlab{b}}.
\newblock URL \url{https://www.alignmentforum.org/posts/TAz44Lb9n9yf52pv8/othello-gpt-reflections-on-the-research-process}.

\bibitem[Nanda et~al.(2023{\natexlab{a}})Nanda, Chan, Lieberum, Smith, and Steinhardt]{nanda2023progressmeasuresgrokkingmechanistic}
Neel Nanda, Lawrence Chan, Tom Lieberum, Jess Smith, and Jacob Steinhardt.
\newblock Progress measures for grokking via mechanistic interpretability.
\newblock In \emph{The Eleventh International Conference on Learning Representations}, 2023{\natexlab{a}}.
\newblock URL \url{https://openreview.net/forum?id=9XFSbDPmdW}.

\bibitem[Nanda et~al.(2023{\natexlab{b}})Nanda, Rajamanoharan, Kram{'a}r, and Shah]{nanda2023fact}
Neel Nanda, Senthooran Rajamanoharan, J{'a}nos Kram{'a}r, and Rohin Shah.
\newblock Fact finding: Attempting to reverse-engineer factual recall on the neuron level ({Post} 1).
\newblock \emph{Alignment Forum}, December 2023{\natexlab{b}}.
\newblock URL \url{https://www.alignmentforum.org/posts/iGuwZTHWb6DFY3sKB/fact-finding-attempting-to-reverse-engineer-factual-recall}.

\bibitem[Narayanaswamy et~al.(2020)Narayanaswamy, Venugopalan, Webster, Peng, Corrado, Ruamviboonsuk, Bavishi, Brenner, Nelson, and Varadarajan]{Narayanaswamy_2020}
Arunachalam Narayanaswamy, Subhashini Venugopalan, Dale~R. Webster, Lily Peng, Greg~S. Corrado, Paisan Ruamviboonsuk, Pinal Bavishi, Michael Brenner, Philip~C. Nelson, and Avinash~V. Varadarajan.
\newblock Scientific discovery by generating counterfactuals using image translation.
\newblock In Anne~L. Martel, Purang Abolmaesumi, Danail Stoyanov, Diana Mateus, Maria~A. Zuluaga, S.~Kevin Zhou, Daniel Racoceanu, and Leo Joskowicz (eds.), \emph{Medical Image Computing and Computer Assisted Intervention -- MICCAI 2020}, pp.\  273--283, Cham, 2020. Springer International Publishing.
\newblock ISBN 978-3-030-59710-8.

\bibitem[Nguyen et~al.(2016{\natexlab{a}})Nguyen, Dosovitskiy, Yosinski, Brox, and Clune]{nguyen2016synthesizingpreferredinputsneurons}
Anh Nguyen, Alexey Dosovitskiy, Jason Yosinski, Thomas Brox, and Jeff Clune.
\newblock Synthesizing the preferred inputs for neurons in neural networks via deep generator networks.
\newblock In \emph{Proceedings of the 30th International Conference on Neural Information Processing Systems}, NIPS'16, pp.\  3395–3403, Red Hook, NY, USA, 2016{\natexlab{a}}. Curran Associates Inc.
\newblock ISBN 9781510838819.

\bibitem[Nguyen et~al.(2016{\natexlab{b}})Nguyen, Yosinski, and Clune]{Nguyen_2016_FeatureVisualization}
Anh Nguyen, Jason Yosinski, and Jeff Clune.
\newblock Multifaceted feature visualization: Uncovering the different types of features learned by each neuron in deep neural networks, 2016{\natexlab{b}}.
\newblock URL \url{https://arxiv.org/abs/1602.03616}.

\bibitem[Nguyen et~al.(2016{\natexlab{c}})Nguyen, Yosinski, and Clune]{nguyen2016multifacetedfeaturevisualizationuncovering}
Anh Nguyen, Jason Yosinski, and Jeff Clune.
\newblock Multifaceted feature visualization: Uncovering the different types of features learned by each neuron in deep neural networks, 2016{\natexlab{c}}.
\newblock URL \url{https://arxiv.org/abs/1602.03616}.

\bibitem[Nguyen et~al.(2017)Nguyen, Clune, Bengio, Dosovitskiy, and Yosinski]{nguyen2017plugplaygenerative}
Anh Nguyen, Jeff Clune, Yoshua Bengio, Alexey Dosovitskiy, and Jason Yosinski.
\newblock Plug and play generative networks: Conditional iterative generation of images in latent space, 2017.
\newblock URL \url{https://arxiv.org/abs/1612.00005}.

\bibitem[Nostalgebraist(2020)]{Nostalgebraist_2020}
Nostalgebraist.
\newblock Interpreting gpt: The logit lens - ai alignment forum, Aug 2020.
\newblock URL \url{https://www.alignmentforum.org/posts/AcKRB8wDpdaN6v6ru/interpreting-gpt-the-logit-lens}.

\bibitem[O'Brien et~al.(2023{\natexlab{a}})O'Brien, Ee, and Williams]{obrien2023deploymentcorrectionsincidentresponse}
Joe O'Brien, Shaun Ee, and Zoe Williams.
\newblock Deployment corrections: An incident response framework for frontier ai models, 2023{\natexlab{a}}.
\newblock URL \url{https://arxiv.org/abs/2310.00328}.

\bibitem[O'Brien et~al.(2023{\natexlab{b}})O'Brien, Stremmel, Pio-Lopez, McMillen, Rasmussen-Ivey, and Levin]{obrien2023machine}
Thomas O'Brien, James Stremmel, Lola Pio-Lopez, Paul McMillen, Cody Rasmussen-Ivey, and Michael Levin.
\newblock Machine learning for hypothesis generation in biology and medicine: Exploring the latent space of neuroscience and developmental bioelectricity.
\newblock Preprint, September 2023{\natexlab{b}}.
\newblock URL \url{https://doi.org/10.31219/osf.io/269e5}.

\bibitem[Olah(2023)]{olah2023interpretability}
Chris Olah.
\newblock Interpretability dreams.
\newblock \emph{Transformer Circuits}, May 2023.
\newblock URL \url{https://transformer-circuits.pub/2023/interpretability-dreams/index.html}.
\newblock An informal note on future goals for mechanistic interpretability.

\bibitem[Olah(2024)]{Olah_2022}
Chris Olah, 2024.
\newblock URL \url{https://transformer-circuits.pub/2022/mech-interp-essay/index.html}.

\bibitem[Olah \& Jermyn(2024)Olah and Jermyn]{olah2024linear}
Chris Olah and Adam Jermyn.
\newblock What is a linear representation? what is a multidimensional feature?
\newblock \emph{Transformer Circuits}, 2024.
\newblock URL \url{https://transformer-circuits.pub/2024/july-update/index.html#linear-representations}.

\bibitem[Olah et~al.(2017{\natexlab{a}})Olah, Mordvintsev, and Schubert]{Olah__2017_FeatureVisualisation}
Chris Olah, Alexander Mordvintsev, and Ludwig Schubert.
\newblock Feature visualization.
\newblock \emph{Distill}, 2\penalty0 (11), Nov 2017{\natexlab{a}}.
\newblock \doi{10.23915/distill.00007}.

\bibitem[Olah et~al.(2017{\natexlab{b}})Olah, Mordvintsev, and Schubert]{olah2017feature}
Chris Olah, Alexander Mordvintsev, and Ludwig Schubert.
\newblock Feature visualization.
\newblock \emph{Distill}, 2017{\natexlab{b}}.
\newblock \doi{10.23915/distill.00007}.
\newblock https://distill.pub/2017/feature-visualization.

\bibitem[Olah et~al.(2018)Olah, Satyanarayan, Johnson, Carter, Schubert, Ye, and Mordvintsev]{olah2018the}
Chris Olah, Arvind Satyanarayan, Ian Johnson, Shan Carter, Ludwig Schubert, Katherine Ye, and Alexander Mordvintsev.
\newblock The building blocks of interpretability.
\newblock \emph{Distill}, 2018.
\newblock \doi{10.23915/distill.00010}.
\newblock https://distill.pub/2018/building-blocks.

\bibitem[Olah et~al.(2020{\natexlab{a}})Olah, Cammarata, Schubert, Goh, Petrov, and Carter]{olah2020an}
Chris Olah, Nick Cammarata, Ludwig Schubert, Gabriel Goh, Michael Petrov, and Shan Carter.
\newblock An overview of early vision in inceptionv1.
\newblock \emph{Distill}, 2020{\natexlab{a}}.
\newblock \doi{10.23915/distill.00024.002}.
\newblock https://distill.pub/2020/circuits/early-vision.

\bibitem[Olah et~al.(2020{\natexlab{b}})Olah, Cammarata, Schubert, Goh, Petrov, and Carter]{olah2020zoom}
Chris Olah, Nick Cammarata, Ludwig Schubert, Gabriel Goh, Michael Petrov, and Shan Carter.
\newblock Zoom in: An introduction to circuits.
\newblock \emph{Distill}, 2020{\natexlab{b}}.
\newblock \doi{10.23915/distill.00024.001}.
\newblock https://distill.pub/2020/circuits/zoom-in.

\bibitem[Olsson et~al.(2022)Olsson, Elhage, Nanda, Joseph, DasSarma, Henighan, Mann, Askell, Bai, Chen, Conerly, Drain, Ganguli, Hatfield-Dodds, Hernandez, Johnston, Jones, Kernion, Lovitt, Ndousse, Amodei, Brown, Clark, Kaplan, McCandlish, and Olah]{olsson2022incontextlearninginductionheads}
Catherine Olsson, Nelson Elhage, Neel Nanda, Nicholas Joseph, Nova DasSarma, Tom Henighan, Ben Mann, Amanda Askell, Yuntao Bai, Anna Chen, Tom Conerly, Dawn Drain, Deep Ganguli, Zac Hatfield-Dodds, Danny Hernandez, Scott Johnston, Andy Jones, Jackson Kernion, Liane Lovitt, Kamal Ndousse, Dario Amodei, Tom Brown, Jack Clark, Jared Kaplan, Sam McCandlish, and Chris Olah.
\newblock In-context learning and induction heads, 2022.
\newblock URL \url{https://arxiv.org/abs/2209.11895}.

\bibitem[{OpenAI}(2023)]{openai2023preparedness}
{OpenAI}.
\newblock Open{AI} preparedness framework.
\newblock Framework, OpenAI, 2023.
\newblock URL \url{https://cdn.openai.com/openai-preparedness-framework-beta.pdf}.
\newblock Beta version.

\bibitem[OpenAI et~al.(2024)OpenAI, Achiam, Adler, Agarwal, Ahmad, Akkaya, Aleman, Almeida, Altenschmidt, Altman, Anadkat, Avila, Babuschkin, Balaji, Balcom, Baltescu, Bao, Bavarian, Belgum, Bello, Berdine, Bernadett-Shapiro, Berner, Bogdonoff, Boiko, Boyd, Brakman, Brockman, Brooks, Brundage, Button, Cai, Campbell, Cann, Carey, Carlson, Carmichael, Chan, Chang, Chantzis, Chen, Chen, Chen, Chen, Chen, Chess, Cho, Chu, Chung, Cummings, Currier, Dai, Decareaux, Degry, Deutsch, Deville, Dhar, Dohan, Dowling, Dunning, Ecoffet, Eleti, Eloundou, Farhi, Fedus, Felix, Fishman, Forte, Fulford, Gao, Georges, Gibson, Goel, Gogineni, Goh, Gontijo-Lopes, Gordon, Grafstein, Gray, Greene, Gross, Gu, Guo, Hallacy, Han, Harris, He, Heaton, Heidecke, Hesse, Hickey, Hickey, Hoeschele, Houghton, Hsu, Hu, Hu, Huizinga, Jain, Jain, Jang, Jiang, Jiang, Jin, Jin, Jomoto, Jonn, Jun, Kaftan, Łukasz Kaiser, Kamali, Kanitscheider, Keskar, Khan, Kilpatrick, Kim, Kim, Kim, Kirchner, Kiros, Knight, Kokotajlo, Łukasz Kondraciuk, Kondrich,
  Konstantinidis, Kosic, Krueger, Kuo, Lampe, Lan, Lee, Leike, Leung, Levy, Li, Lim, Lin, Lin, Litwin, Lopez, Lowe, Lue, Makanju, Malfacini, Manning, Markov, Markovski, Martin, Mayer, Mayne, McGrew, McKinney, McLeavey, McMillan, McNeil, Medina, Mehta, Menick, Metz, Mishchenko, Mishkin, Monaco, Morikawa, Mossing, Mu, Murati, Murk, Mély, Nair, Nakano, Nayak, Neelakantan, Ngo, Noh, Ouyang, O'Keefe, Pachocki, Paino, Palermo, Pantuliano, Parascandolo, Parish, Parparita, Passos, Pavlov, Peng, Perelman, de~Avila Belbute~Peres, Petrov, de~Oliveira~Pinto, Michael, Pokorny, Pokrass, Pong, Powell, Power, Power, Proehl, Puri, Radford, Rae, Ramesh, Raymond, Real, Rimbach, Ross, Rotsted, Roussez, Ryder, Saltarelli, Sanders, Santurkar, Sastry, Schmidt, Schnurr, Schulman, Selsam, Sheppard, Sherbakov, Shieh, Shoker, Shyam, Sidor, Sigler, Simens, Sitkin, Slama, Sohl, Sokolowsky, Song, Staudacher, Such, Summers, Sutskever, Tang, Tezak, Thompson, Tillet, Tootoonchian, Tseng, Tuggle, Turley, Tworek, Uribe, Vallone, Vijayvergiya,
  Voss, Wainwright, Wang, Wang, Wang, Ward, Wei, Weinmann, Welihinda, Welinder, Weng, Weng, Wiethoff, Willner, Winter, Wolrich, Wong, Workman, Wu, Wu, Wu, Xiao, Xu, Yoo, Yu, Yuan, Zaremba, Zellers, Zhang, Zhang, Zhao, Zheng, Zhuang, Zhuk, and Zoph]{openai2024gpt4technicalreport}
OpenAI, Josh Achiam, Steven Adler, Sandhini Agarwal, Lama Ahmad, Ilge Akkaya, Florencia~Leoni Aleman, Diogo Almeida, Janko Altenschmidt, Sam Altman, Shyamal Anadkat, Red Avila, Igor Babuschkin, Suchir Balaji, Valerie Balcom, Paul Baltescu, Haiming Bao, Mohammad Bavarian, Jeff Belgum, Irwan Bello, Jake Berdine, Gabriel Bernadett-Shapiro, Christopher Berner, Lenny Bogdonoff, Oleg Boiko, Madelaine Boyd, Anna-Luisa Brakman, Greg Brockman, Tim Brooks, Miles Brundage, Kevin Button, Trevor Cai, Rosie Campbell, Andrew Cann, Brittany Carey, Chelsea Carlson, Rory Carmichael, Brooke Chan, Che Chang, Fotis Chantzis, Derek Chen, Sully Chen, Ruby Chen, Jason Chen, Mark Chen, Ben Chess, Chester Cho, Casey Chu, Hyung~Won Chung, Dave Cummings, Jeremiah Currier, Yunxing Dai, Cory Decareaux, Thomas Degry, Noah Deutsch, Damien Deville, Arka Dhar, David Dohan, Steve Dowling, Sheila Dunning, Adrien Ecoffet, Atty Eleti, Tyna Eloundou, David Farhi, Liam Fedus, Niko Felix, Simón~Posada Fishman, Juston Forte, Isabella Fulford, Leo
  Gao, Elie Georges, Christian Gibson, Vik Goel, Tarun Gogineni, Gabriel Goh, Rapha Gontijo-Lopes, Jonathan Gordon, Morgan Grafstein, Scott Gray, Ryan Greene, Joshua Gross, Shixiang~Shane Gu, Yufei Guo, Chris Hallacy, Jesse Han, Jeff Harris, Yuchen He, Mike Heaton, Johannes Heidecke, Chris Hesse, Alan Hickey, Wade Hickey, Peter Hoeschele, Brandon Houghton, Kenny Hsu, Shengli Hu, Xin Hu, Joost Huizinga, Shantanu Jain, Shawn Jain, Joanne Jang, Angela Jiang, Roger Jiang, Haozhun Jin, Denny Jin, Shino Jomoto, Billie Jonn, Heewoo Jun, Tomer Kaftan, Łukasz Kaiser, Ali Kamali, Ingmar Kanitscheider, Nitish~Shirish Keskar, Tabarak Khan, Logan Kilpatrick, Jong~Wook Kim, Christina Kim, Yongjik Kim, Jan~Hendrik Kirchner, Jamie Kiros, Matt Knight, Daniel Kokotajlo, Łukasz Kondraciuk, Andrew Kondrich, Aris Konstantinidis, Kyle Kosic, Gretchen Krueger, Vishal Kuo, Michael Lampe, Ikai Lan, Teddy Lee, Jan Leike, Jade Leung, Daniel Levy, Chak~Ming Li, Rachel Lim, Molly Lin, Stephanie Lin, Mateusz Litwin, Theresa Lopez, Ryan
  Lowe, Patricia Lue, Anna Makanju, Kim Malfacini, Sam Manning, Todor Markov, Yaniv Markovski, Bianca Martin, Katie Mayer, Andrew Mayne, Bob McGrew, Scott~Mayer McKinney, Christine McLeavey, Paul McMillan, Jake McNeil, David Medina, Aalok Mehta, Jacob Menick, Luke Metz, Andrey Mishchenko, Pamela Mishkin, Vinnie Monaco, Evan Morikawa, Daniel Mossing, Tong Mu, Mira Murati, Oleg Murk, David Mély, Ashvin Nair, Reiichiro Nakano, Rajeev Nayak, Arvind Neelakantan, Richard Ngo, Hyeonwoo Noh, Long Ouyang, Cullen O'Keefe, Jakub Pachocki, Alex Paino, Joe Palermo, Ashley Pantuliano, Giambattista Parascandolo, Joel Parish, Emy Parparita, Alex Passos, Mikhail Pavlov, Andrew Peng, Adam Perelman, Filipe de~Avila Belbute~Peres, Michael Petrov, Henrique~Ponde de~Oliveira~Pinto, Michael, Pokorny, Michelle Pokrass, Vitchyr~H. Pong, Tolly Powell, Alethea Power, Boris Power, Elizabeth Proehl, Raul Puri, Alec Radford, Jack Rae, Aditya Ramesh, Cameron Raymond, Francis Real, Kendra Rimbach, Carl Ross, Bob Rotsted, Henri Roussez,
  Nick Ryder, Mario Saltarelli, Ted Sanders, Shibani Santurkar, Girish Sastry, Heather Schmidt, David Schnurr, John Schulman, Daniel Selsam, Kyla Sheppard, Toki Sherbakov, Jessica Shieh, Sarah Shoker, Pranav Shyam, Szymon Sidor, Eric Sigler, Maddie Simens, Jordan Sitkin, Katarina Slama, Ian Sohl, Benjamin Sokolowsky, Yang Song, Natalie Staudacher, Felipe~Petroski Such, Natalie Summers, Ilya Sutskever, Jie Tang, Nikolas Tezak, Madeleine~B. Thompson, Phil Tillet, Amin Tootoonchian, Elizabeth Tseng, Preston Tuggle, Nick Turley, Jerry Tworek, Juan Felipe~Cerón Uribe, Andrea Vallone, Arun Vijayvergiya, Chelsea Voss, Carroll Wainwright, Justin~Jay Wang, Alvin Wang, Ben Wang, Jonathan Ward, Jason Wei, CJ~Weinmann, Akila Welihinda, Peter Welinder, Jiayi Weng, Lilian Weng, Matt Wiethoff, Dave Willner, Clemens Winter, Samuel Wolrich, Hannah Wong, Lauren Workman, Sherwin Wu, Jeff Wu, Michael Wu, Kai Xiao, Tao Xu, Sarah Yoo, Kevin Yu, Qiming Yuan, Wojciech Zaremba, Rowan Zellers, Chong Zhang, Marvin Zhang, Shengjia
  Zhao, Tianhao Zheng, Juntang Zhuang, William Zhuk, and Barret Zoph.
\newblock Gpt-4 technical report, 2024.
\newblock URL \url{https://arxiv.org/abs/2303.08774}.

\bibitem[Park et~al.(2024{\natexlab{a}})Park, Okawa, Lee, Lubana, and Tanaka]{park2024emergencehiddencapabilitiesexploring}
Core~Francisco Park, Maya Okawa, Andrew Lee, Ekdeep~Singh Lubana, and Hidenori Tanaka.
\newblock Emergence of hidden capabilities: Exploring learning dynamics in concept space.
\newblock In \emph{The Thirty-eighth Annual Conference on Neural Information Processing Systems}, 2024{\natexlab{a}}.
\newblock URL \url{https://openreview.net/forum?id=owuEcT6BTl}.

\bibitem[Park et~al.(2023{\natexlab{a}})Park, Choe, and Veitch]{park2024linearrepresentationhypothesisgeometry}
Kiho Park, Yo~Joong Choe, and Victor Veitch.
\newblock The linear representation hypothesis and the geometry of large language models.
\newblock In \emph{Causal Representation Learning Workshop at NeurIPS 2023}, 2023{\natexlab{a}}.
\newblock URL \url{https://openreview.net/forum?id=T0PoOJg8cK}.

\bibitem[Park et~al.(2024{\natexlab{b}})Park, Choe, Jiang, and Veitch]{park2024geometrycategoricalhierarchicalconcepts}
Kiho Park, Yo~Joong Choe, Yibo Jiang, and Victor Veitch.
\newblock The geometry of categorical and hierarchical concepts in large language models.
\newblock In \emph{ICML 2024 Workshop on Mechanistic Interpretability}, 2024{\natexlab{b}}.
\newblock URL \url{https://openreview.net/forum?id=KXuYjuBzKo}.

\bibitem[Park et~al.(2023{\natexlab{b}})Park, Goldstein, O'Gara, Chen, and Hendrycks]{park2023aideceptionsurveyexamples}
Peter~S. Park, Simon Goldstein, Aidan O'Gara, Michael Chen, and Dan Hendrycks.
\newblock Ai deception: A survey of examples, risks, and potential solutions, 2023{\natexlab{b}}.
\newblock URL \url{https://arxiv.org/abs/2308.14752}.

\bibitem[Paulo et~al.(2024)Paulo, Marshall, and Belrose]{paulo2024doestransformerinterpretabilitytransfer}
Gonçalo Paulo, Thomas Marshall, and Nora Belrose.
\newblock Does transformer interpretability transfer to rnns?, 2024.
\newblock URL \url{https://arxiv.org/abs/2404.05971}.

\bibitem[Pearce et~al.(2024)Pearce, Dooms, and Rigg]{pearce2024weightbaseddecompositioncasebilinear}
Michael~T. Pearce, Thomas Dooms, and Alice Rigg.
\newblock Weight-based decomposition: A case for bilinear mlps, 2024.
\newblock URL \url{https://arxiv.org/abs/2406.03947}.

\bibitem[Pearl(2009)]{Pearl_2009}
Judea Pearl.
\newblock \emph{Causality}.
\newblock Cambridge University Press, 2 edition, 2009.

\bibitem[Peng et~al.(2023)Peng, Alcaide, Anthony, Albalak, Arcadinho, Biderman, Cao, Cheng, Chung, Derczynski, Du, Grella, Gv, He, Hou, Kazienko, Kocon, Kong, Koptyra, Lau, Lin, Mantri, Mom, Saito, Song, Tang, Wind, Wo{\'z}niak, Zhang, Zhou, Zhu, and Zhu]{peng2023rwkvreinventingrnnstransformer}
Bo~Peng, Eric Alcaide, Quentin Anthony, Alon Albalak, Samuel Arcadinho, Stella Biderman, Huanqi Cao, Xin Cheng, Michael Chung, Leon Derczynski, Xingjian Du, Matteo Grella, Kranthi Gv, Xuzheng He, Haowen Hou, Przemyslaw Kazienko, Jan Kocon, Jiaming Kong, Bart{\l}omiej Koptyra, Hayden Lau, Jiaju Lin, Krishna Sri~Ipsit Mantri, Ferdinand Mom, Atsushi Saito, Guangyu Song, Xiangru Tang, Johan Wind, Stanis{\l}aw Wo{\'z}niak, Zhenyuan Zhang, Qinghua Zhou, Jian Zhu, and Rui-Jie Zhu.
\newblock {RWKV}: Reinventing {RNN}s for the transformer era.
\newblock In Houda Bouamor, Juan Pino, and Kalika Bali (eds.), \emph{Findings of the Association for Computational Linguistics: EMNLP 2023}, pp.\  14048--14077, Singapore, December 2023. Association for Computational Linguistics.
\newblock \doi{10.18653/v1/2023.findings-emnlp.936}.
\newblock URL \url{https://aclanthology.org/2023.findings-emnlp.936/}.

\bibitem[Perez et~al.(2022)Perez, Huang, Song, Cai, Ring, Aslanides, Glaese, McAleese, and Irving]{perez2022redteaminglanguagemodels}
Ethan Perez, Saffron Huang, Francis Song, Trevor Cai, Roman Ring, John Aslanides, Amelia Glaese, Nat McAleese, and Geoffrey Irving.
\newblock Red teaming language models with language models.
\newblock In Yoav Goldberg, Zornitsa Kozareva, and Yue Zhang (eds.), \emph{Proceedings of the 2022 Conference on Empirical Methods in Natural Language Processing}, pp.\  3419--3448, Abu Dhabi, United Arab Emirates, December 2022. Association for Computational Linguistics.
\newblock \doi{10.18653/v1/2022.emnlp-main.225}.
\newblock URL \url{https://aclanthology.org/2022.emnlp-main.225/}.

\bibitem[Peters et~al.(2018)Peters, Neumann, Zettlemoyer, and Yih]{peters2018dissectingcontextualwordembeddings}
Matthew~E. Peters, Mark Neumann, Luke Zettlemoyer, and Wen-tau Yih.
\newblock Dissecting contextual word embeddings: Architecture and representation.
\newblock In Ellen Riloff, David Chiang, Julia Hockenmaier, and Jun{'}ichi Tsujii (eds.), \emph{Proceedings of the 2018 Conference on Empirical Methods in Natural Language Processing}, pp.\  1499--1509, Brussels, Belgium, October-November 2018. Association for Computational Linguistics.
\newblock \doi{10.18653/v1/D18-1179}.
\newblock URL \url{https://aclanthology.org/D18-1179/}.

\bibitem[Pimentel et~al.(2020)Pimentel, Valvoda, Maudslay, Zmigrod, Williams, and Cotterell]{pimentel-etal-2020-information}
Tiago Pimentel, Josef Valvoda, Rowan~Hall Maudslay, Ran Zmigrod, Adina Williams, and Ryan Cotterell.
\newblock Information-theoretic probing for linguistic structure.
\newblock In Dan Jurafsky, Joyce Chai, Natalie Schluter, and Joel Tetreault (eds.), \emph{Proceedings of the 58th Annual Meeting of the Association for Computational Linguistics}, pp.\  4609--4622, Online, July 2020. Association for Computational Linguistics.
\newblock \doi{10.18653/v1/2020.acl-main.420}.
\newblock URL \url{https://aclanthology.org/2020.acl-main.420}.

\bibitem[Pochinkov \& Schoots(2024)Pochinkov and Schoots]{pochinkov2024dissectinglanguagemodelsmachine}
Nicholas Pochinkov and Nandi Schoots.
\newblock Dissecting language models: Machine unlearning via selective pruning, 2024.
\newblock URL \url{https://arxiv.org/abs/2403.01267}.

\bibitem[Prakash et~al.(2024)Prakash, Shaham, Haklay, Belinkov, and Bau]{prakash2024finetuningenhancesexistingmechanisms}
Nikhil Prakash, Tamar~Rott Shaham, Tal Haklay, Yonatan Belinkov, and David Bau.
\newblock Fine-tuning enhances existing mechanisms: A case study on entity tracking.
\newblock In \emph{Proceedings of the 2024 International Conference on Learning Representations}, 2024.
\newblock arXiv:2402.14811.

\bibitem[Pruthi et~al.(2020)Pruthi, Gupta, Dhingra, Neubig, and Lipton]{pruthi2020learningdeceiveattentionbasedexplanations}
Danish Pruthi, Mansi Gupta, Bhuwan Dhingra, Graham Neubig, and Zachary~C. Lipton.
\newblock Learning to deceive with attention-based explanations.
\newblock In Dan Jurafsky, Joyce Chai, Natalie Schluter, and Joel Tetreault (eds.), \emph{Proceedings of the 58th Annual Meeting of the Association for Computational Linguistics}, pp.\  4782--4793, Online, July 2020. Association for Computational Linguistics.
\newblock \doi{10.18653/v1/2020.acl-main.432}.
\newblock URL \url{https://aclanthology.org/2020.acl-main.432/}.

\bibitem[Radford et~al.(2018)Radford, Wu, Child, Luan, Amodei, and Sutskever]{radford2018language}
Alec Radford, Jeffrey Wu, Rewon Child, David Luan, Dario Amodei, and Ilya Sutskever.
\newblock Language models are unsupervised multitask learners, 2018.
\newblock URL \url{https://cdn.openai.com/better-language-models/language_models_are_unsupervised_multitask_learners.pdf}.

\bibitem[Rafailov et~al.(2023)Rafailov, Sharma, Mitchell, Manning, Ermon, and Finn]{rafailov2024directpreferenceoptimizationlanguage}
Rafael Rafailov, Archit Sharma, Eric Mitchell, Christopher~D Manning, Stefano Ermon, and Chelsea Finn.
\newblock Direct preference optimization: Your language model is secretly a reward model.
\newblock In \emph{Thirty-seventh Conference on Neural Information Processing Systems}, 2023.
\newblock URL \url{https://openreview.net/forum?id=HPuSIXJaa9}.

\bibitem[Rai et~al.(2024)Rai, Zhou, Feng, Saparov, and Yao]{rai2024practicalreviewmechanisticinterpretability}
Daking Rai, Yilun Zhou, Shi Feng, Abulhair Saparov, and Ziyu Yao.
\newblock A practical review of mechanistic interpretability for transformer-based language models, 2024.
\newblock URL \url{https://arxiv.org/abs/2407.02646}.

\bibitem[Rajamanoharan et~al.(2024)Rajamanoharan, Conmy, Smith, Lieberum, Varma, Kramar, Shah, and Nanda]{rajamanoharan2024improvingdictionarylearninggated}
Senthooran Rajamanoharan, Arthur Conmy, Lewis Smith, Tom Lieberum, Vikrant Varma, Janos Kramar, Rohin Shah, and Neel Nanda.
\newblock Improving sparse decomposition of language model activations with gated sparse autoencoders.
\newblock In \emph{The Thirty-eighth Annual Conference on Neural Information Processing Systems}, 2024.
\newblock URL \url{https://openreview.net/forum?id=zLBlin2zvW}.

\bibitem[Raposo et~al.(2014)Raposo, Kaufman, and Churchland]{raposo2014category}
David Raposo, Matthew~T. Kaufman, and Anne~K. Churchland.
\newblock A category-free neural population supports evolving demands during decision-making.
\newblock \emph{Nature Neuroscience}, 17:\penalty0 1784--1792, 2014.
\newblock \doi{10.1038/nn.3865}.
\newblock URL \url{https://doi.org/10.1038/nn.3865}.

\bibitem[Rauker et~al.(2023)Rauker, Ho, Casper, and Hadfield-Menell]{rauker2023transparentaisurveyinterpreting}
Tilman Rauker, Anson Ho, Stephen Casper, and Dylan Hadfield-Menell.
\newblock { Toward Transparent AI: A Survey on Interpreting the Inner Structures of Deep Neural Networks }.
\newblock In \emph{2023 IEEE Conference on Secure and Trustworthy Machine Learning (SaTML)}, pp.\  464--483, Los Alamitos, CA, USA, February 2023. IEEE Computer Society.
\newblock \doi{10.1109/SaTML54575.2023.00039}.
\newblock URL \url{https://doi.ieeecomputersociety.org/10.1109/SaTML54575.2023.00039}.

\bibitem[Ravfogel et~al.(2020)Ravfogel, Elazar, Gonen, Twiton, and Goldberg]{ravfogel-etal-2020-null}
Shauli Ravfogel, Yanai Elazar, Hila Gonen, Michael Twiton, and Yoav Goldberg.
\newblock Null it out: Guarding protected attributes by iterative nullspace projection.
\newblock In Dan Jurafsky, Joyce Chai, Natalie Schluter, and Joel Tetreault (eds.), \emph{Proceedings of the 58th Annual Meeting of the Association for Computational Linguistics}, pp.\  7237--7256, Online, July 2020. Association for Computational Linguistics.
\newblock \doi{10.18653/v1/2020.acl-main.647}.
\newblock URL \url{https://aclanthology.org/2020.acl-main.647}.

\bibitem[Ravfogel et~al.(2022)Ravfogel, Twiton, Goldberg, and Cotterell]{ravfogel2024linearadversarialconcepterasure}
Shauli Ravfogel, Michael Twiton, Yoav Goldberg, and Ryan~D Cotterell.
\newblock Linear adversarial concept erasure.
\newblock In Kamalika Chaudhuri, Stefanie Jegelka, Le~Song, Csaba Szepesvari, Gang Niu, and Sivan Sabato (eds.), \emph{Proceedings of the 39th International Conference on Machine Learning}, volume 162 of \emph{Proceedings of Machine Learning Research}, pp.\  18400--18421. PMLR, 17--23 Jul 2022.
\newblock URL \url{https://proceedings.mlr.press/v162/ravfogel22a.html}.

\bibitem[Ravichander et~al.(2021)Ravichander, Belinkov, and Hovy]{ravichander2021probingprobingparadigmdoes}
Abhilasha Ravichander, Yonatan Belinkov, and Eduard Hovy.
\newblock Probing the probing paradigm: Does probing accuracy entail task relevance?
\newblock In Paola Merlo, Jorg Tiedemann, and Reut Tsarfaty (eds.), \emph{Proceedings of the 16th Conference of the European Chapter of the Association for Computational Linguistics: Main Volume}, pp.\  3363--3377, Online, April 2021. Association for Computational Linguistics.
\newblock \doi{10.18653/v1/2021.eacl-main.295}.
\newblock URL \url{https://aclanthology.org/2021.eacl-main.295/}.

\bibitem[Reddy(2024)]{reddy2024the}
Gautam Reddy.
\newblock The mechanistic basis of data dependence and abrupt learning in an in-context classification task.
\newblock In \emph{The Twelfth International Conference on Learning Representations}, 2024.
\newblock URL \url{https://openreview.net/forum?id=aN4Jf6Cx69}.

\bibitem[Ribeiro et~al.(2016)Ribeiro, Singh, and Guestrin]{Ribiero_2016_LIME}
Marco~Tulio Ribeiro, Sameer Singh, and Carlos Guestrin.
\newblock "why should i trust you?": Explaining the predictions of any classifier.
\newblock In \emph{Proceedings of the 22nd ACM SIGKDD International Conference on Knowledge Discovery and Data Mining}, KDD '16, pp.\  1135--1144, New York, NY, USA, 2016. Association for Computing Machinery.
\newblock ISBN 9781450342322.
\newblock \doi{10.1145/2939672.2939778}.
\newblock URL \url{https://doi.org/10.1145/2939672.2939778}.

\bibitem[Riggs et~al.(2023)Riggs, Mitchell, and Kaufman]{riggs2023finding}
Logan Riggs, Sam Mitchell, and Adam Kaufman.
\newblock Finding sparse linear connections between features in {LLM}s.
\newblock \emph{AI Alignment Forum}, dec 2023.
\newblock URL \url{https://www.lesswrong.com/posts/7fxusXdkMNmAhkAfc/finding-sparse-linear-connections-between-features-in-llms}.

\bibitem[Rigotti et~al.(2013)Rigotti, Barak, Warden, Wang, Daw, Miller, and Fusi]{rigotti2013importance}
Mattia Rigotti, Omri Barak, Melissa Warden, Xiao-Jing Wang, Nathaniel~D. Daw, Earl~K. Miller, and Stefano Fusi.
\newblock The importance of mixed selectivity in complex cognitive tasks.
\newblock \emph{Nature}, 497:\penalty0 585--590, 2013.
\newblock \doi{10.1038/nature12160}.
\newblock URL \url{https://doi.org/10.1038/nature12160}.

\bibitem[Rimsky et~al.(2024)Rimsky, Gabrieli, Schulz, Tong, Hubinger, and Turner]{panickssery2024steeringllama2contrastive}
Nina Rimsky, Nick Gabrieli, Julian Schulz, Meg Tong, Evan Hubinger, and Alexander Turner.
\newblock Steering llama 2 via contrastive activation addition.
\newblock In Lun-Wei Ku, Andre Martins, and Vivek Srikumar (eds.), \emph{Proceedings of the 62nd Annual Meeting of the Association for Computational Linguistics (Volume 1: Long Papers)}, pp.\  15504--15522, Bangkok, Thailand, August 2024. Association for Computational Linguistics.
\newblock \doi{10.18653/v1/2024.acl-long.828}.
\newblock URL \url{https://aclanthology.org/2024.acl-long.828/}.

\bibitem[Roberts et~al.(2022)Roberts, Yaida, and Hanin]{Roberts_2022_principles}
Daniel~A. Roberts, Sho Yaida, and Boris Hanin.
\newblock \emph{The Principles of Deep Learning Theory: An Effective Theory Approach to Understanding Neural Networks}.
\newblock Cambridge University Press, May 2022.
\newblock ISBN 9781316519332.
\newblock \doi{10.1017/9781009023405}.
\newblock URL \url{http://dx.doi.org/10.1017/9781009023405}.

\bibitem[Roger(2023)]{roger2023coup}
Fabien Roger.
\newblock Coup probes: Catching catastrophes with probes trained off-policy.
\newblock \emph{Alignment Forum}, November 2023.
\newblock URL \url{https://www.alignmentforum.org/posts/WCj7WgFSLmyKaMwPR/coup-probes-catching-catastrophes-with-probes-trained-off}.

\bibitem[Rogers et~al.(2020)Rogers, Kovaleva, and Rumshisky]{rogers2020primerbertologyknowbert}
Anna Rogers, Olga Kovaleva, and Anna Rumshisky.
\newblock A primer in {BERT}ology: What we know about how {BERT} works.
\newblock \emph{Transactions of the Association for Computational Linguistics}, 8:\penalty0 842--866, 2020.
\newblock \doi{10.1162/tacl_a_00349}.
\newblock URL \url{https://aclanthology.org/2020.tacl-1.54/}.

\bibitem[Rombach et~al.(2022)Rombach, Blattmann, Lorenz, Esser, and Ommer]{rombach2022highresolutionimagesynthesislatent}
Robin Rombach, Andreas Blattmann, Dominik Lorenz, Patrick Esser, and Bj{\"o}rn Ommer.
\newblock High-resolution image synthesis with latent diffusion models.
\newblock In \emph{Proceedings of the IEEE/CVF Conference on Computer Vision and Pattern Recognition}, pp.\  10684--10695, 2022.

\bibitem[Roweis \& Ghahramani(1999)Roweis and Ghahramani]{Roweis_1999_LinearModel}
Sam Roweis and Zoubin Ghahramani.
\newblock A unifying review of linear gaussian models.
\newblock \emph{Neural Computation}, 11\penalty0 (2):\penalty0 305--345, 1999.
\newblock \doi{10.1162/089976699300016674}.

\bibitem[Rudin(2019)]{rudin2019stop}
Cynthia Rudin.
\newblock Stop explaining black box machine learning models for high stakes decisions and use interpretable models instead.
\newblock \emph{Nature Machine Intelligence}, 1:\penalty0 206--215, 2019.
\newblock \doi{10.1038/s42256-019-0048-x}.

\bibitem[Saphra \& Wiegreffe(2024)Saphra and Wiegreffe]{saphra2024mechanistic}
Naomi Saphra and Sarah Wiegreffe.
\newblock Mechanistic?, 2024.
\newblock URL \url{https://arxiv.org/abs/2410.09087}.

\bibitem[Schaeffer et~al.(2023)Schaeffer, Miranda, and Koyejo]{schaeffer2023emergentabilitieslargelanguage}
Rylan Schaeffer, Brando Miranda, and Sanmi Koyejo.
\newblock Are emergent abilities of large language models a mirage?
\newblock In \emph{Thirty-seventh Conference on Neural Information Processing Systems}, 2023.
\newblock URL \url{https://openreview.net/forum?id=ITw9edRDlD}.

\bibitem[Scheurer et~al.(2024)Scheurer, Balesni, and Hobbhahn]{scheurer2024largelanguagemodelsstrategically}
Jérémy Scheurer, Mikita Balesni, and Marius Hobbhahn.
\newblock Large language models can strategically deceive their users when put under pressure, 2024.
\newblock URL \url{https://arxiv.org/abs/2311.07590}.

\bibitem[Schut et~al.(2023)Schut, Tomasev, McGrath, Hassabis, Paquet, and Kim]{schut2023bridginghumanaiknowledgegap}
Lisa Schut, Nenad Tomasev, Tom McGrath, Demis Hassabis, Ulrich Paquet, and Been Kim.
\newblock Bridging the human-ai knowledge gap: Concept discovery and transfer in alphazero, 2023.
\newblock URL \url{https://arxiv.org/abs/2310.16410}.

\bibitem[Schwettmann et~al.(2023)Schwettmann, Shaham, Materzynska, Chowdhury, Li, Andreas, Bau, and Torralba]{schwettmann2023findfunctiondescriptionbenchmark}
Sarah Schwettmann, Tamar~Rott Shaham, Joanna Materzynska, Neil Chowdhury, Shuang Li, Jacob Andreas, David Bau, and Antonio Torralba.
\newblock {FIND}: A function description benchmark for evaluating interpretability methods.
\newblock In \emph{Thirty-seventh Conference on Neural Information Processing Systems Datasets and Benchmarks Track}, 2023.
\newblock URL \url{https://openreview.net/forum?id=mkSDXjX6EM}.

\bibitem[Selvaraju et~al.(2019)Selvaraju, Cogswell, Das, Vedantam, Parikh, and Batra]{Selvaraju_2019_GradCAM}
Ramprasaath~R. Selvaraju, Michael Cogswell, Abhishek Das, Ramakrishna Vedantam, Devi Parikh, and Dhruv Batra.
\newblock Grad-cam: Visual explanations from deep networks via gradient-based localization.
\newblock \emph{International Journal of Computer Vision}, 128\penalty0 (2):\penalty0 336–359, Oct 2019.
\newblock \doi{10.1007/s11263-019-01228-7}.

\bibitem[Seshia et~al.(2022)Seshia, Sadigh, and Sastry]{seshia2020verifiedartificialintelligence}
Sanjit~A. Seshia, Dorsa Sadigh, and S.~Shankar Sastry.
\newblock Toward verified artificial intelligence.
\newblock \emph{Commun. ACM}, 65\penalty0 (7):\penalty0 46–55, June 2022.
\newblock ISSN 0001-0782.
\newblock \doi{10.1145/3503914}.
\newblock URL \url{https://doi.org/10.1145/3503914}.

\bibitem[Shaham et~al.(2025)Shaham, Schwettmann, Wang, Rajaram, Hernandez, Andreas, and Torralba]{shaham2024multimodalautomatedinterpretabilityagent}
Tamar~Rott Shaham, Sarah Schwettmann, Franklin Wang, Achyuta Rajaram, Evan Hernandez, Jacob Andreas, and Antonio Torralba.
\newblock A multimodal automated interpretability agent.
\newblock In \emph{Proceedings of the 41st International Conference on Machine Learning}, ICML'24. JMLR.org, 2025.

\bibitem[Shapley(1997)]{Shapley_1997_Shapley_Values}
L.~Shapley.
\newblock \emph{7. A Value for n-Person Games. Contributions to the Theory of Games II (1953) 307-317.}, pp.\  69--79.
\newblock Princeton University Press, Princeton, 1997.
\newblock ISBN 9781400829156.
\newblock \doi{doi:10.1515/9781400829156-012}.
\newblock URL \url{https://doi.org/10.1515/9781400829156-012}.

\bibitem[Sharkey(2023)]{sharkey2023technicalnotebilinearlayers}
Lee Sharkey.
\newblock A technical note on bilinear layers for interpretability, 2023.
\newblock URL \url{https://arxiv.org/abs/2305.03452}.

\bibitem[Sharkey et~al.(2022{\natexlab{a}})Sharkey, Black, and Millidge]{Sharkey_2022_CurrentThemes}
Lee Sharkey, Sid Black, and Beren Millidge.
\newblock Current themes in mechanistic interpretability research, Nov 2022{\natexlab{a}}.
\newblock URL \url{https://www.lesswrong.com/posts/Jgs7LQwmvErxR9BCC/current-themes-in-mechanistic-interpretability-research#Study_model_systems_in_depth}.

\bibitem[Sharkey et~al.(2022{\natexlab{b}})Sharkey, Braun, and Beren]{Sharkey_2022}
Lee Sharkey, Dan Braun, and Beren.
\newblock [interim research report] taking features out of superposition with sparse autoencoders - ai alignment forum, Dec 2022{\natexlab{b}}.
\newblock URL \url{https://www.alignmentforum.org/posts/z6QQJbtpkEAX3Aojj/interim-research-report-taking-features-out-of-superposition}.

\bibitem[Sharma et~al.(2024{\natexlab{a}})Sharma, Atkinson, and Bau]{sharma2024locatingeditingfactualassociations}
Arnab~Sen Sharma, David Atkinson, and David Bau.
\newblock Locating and editing factual associations in mamba.
\newblock In \emph{First Conference on Language Modeling}, 2024{\natexlab{a}}.
\newblock URL \url{https://openreview.net/forum?id=yoVRyrEgix}.

\bibitem[Sharma et~al.(2024{\natexlab{b}})Sharma, Tong, Korbak, Duvenaud, Askell, Bowman, DURMUS, Hatfield-Dodds, Johnston, Kravec, Maxwell, McCandlish, Ndousse, Rausch, Schiefer, Yan, Zhang, and Perez]{sharma2023understandingsycophancylanguagemodels}
Mrinank Sharma, Meg Tong, Tomasz Korbak, David Duvenaud, Amanda Askell, Samuel~R. Bowman, Esin DURMUS, Zac Hatfield-Dodds, Scott~R Johnston, Shauna~M Kravec, Timothy Maxwell, Sam McCandlish, Kamal Ndousse, Oliver Rausch, Nicholas Schiefer, Da~Yan, Miranda Zhang, and Ethan Perez.
\newblock Towards understanding sycophancy in language models.
\newblock In \emph{The Twelfth International Conference on Learning Representations}, 2024{\natexlab{b}}.
\newblock URL \url{https://openreview.net/forum?id=tvhaxkMKAn}.

\bibitem[Shazeer et~al.(2017)Shazeer, Mirhoseini, Maziarz, Davis, Le, Hinton, and Dean]{shazeer2017outrageouslylargeneuralnetworks}
Noam Shazeer, *Azalia Mirhoseini, *Krzysztof Maziarz, Andy Davis, Quoc Le, Geoffrey Hinton, and Jeff Dean.
\newblock Outrageously large neural networks: The sparsely-gated mixture-of-experts layer.
\newblock In \emph{International Conference on Learning Representations}, 2017.
\newblock URL \url{https://openreview.net/forum?id=B1ckMDqlg}.

\bibitem[Sherrington(1906)]{sherrington1906observations}
C.~S. Sherrington.
\newblock Observations on the scratch-reflex in the spinal dog.
\newblock \emph{The Journal of Physiology}, 34\penalty0 (1-2):\penalty0 1--50, 3 1906.
\newblock \doi{10.1113/jphysiol.1906.sp001139}.
\newblock URL \url{https://physoc.onlinelibrary.wiley.com/doi/10.1113/jphysiol.1906.sp001139}.
\newblock 185 citations as of [insert current date].

\bibitem[Shevlane et~al.(2023)Shevlane, Farquhar, Garfinkel, Phuong, Whittlestone, Leung, Kokotajlo, Marchal, Anderljung, Kolt, Ho, Siddarth, Avin, Hawkins, Kim, Gabriel, Bolina, Clark, Bengio, Christiano, and Dafoe]{shevlane2023modelevaluationextremerisks}
Toby Shevlane, Sebastian Farquhar, Ben Garfinkel, Mary Phuong, Jess Whittlestone, Jade Leung, Daniel Kokotajlo, Nahema Marchal, Markus Anderljung, Noam Kolt, Lewis Ho, Divya Siddarth, Shahar Avin, Will Hawkins, Been Kim, Iason Gabriel, Vijay Bolina, Jack Clark, Yoshua Bengio, Paul Christiano, and Allan Dafoe.
\newblock Model evaluation for extreme risks, 2023.
\newblock URL \url{https://arxiv.org/abs/2305.15324}.

\bibitem[Shi et~al.(2024)Shi, Beltran-Velez, Nazaret, Zheng, Garriga-Alonso, Jesson, Makar, and Blei]{shi2024hypothesistestingcircuithypothesis}
Claudia Shi, Nicolas Beltran-Velez, Achille Nazaret, Carolina Zheng, Adrià Garriga-Alonso, Andrew Jesson, Maggie Makar, and David~M. Blei.
\newblock Hypothesis testing the circuit hypothesis in llms, 2024.
\newblock URL \url{https://arxiv.org/abs/2410.13032}.

\bibitem[Shlegeris et~al.(2024)Shlegeris, Roger, Chan, and McLean]{shlegeris2024languagemodelsbetterhumans}
Buck Shlegeris, Fabien Roger, Lawrence Chan, and Euan McLean.
\newblock Language models are better than humans at next-token prediction, 2024.
\newblock URL \url{https://arxiv.org/abs/2212.11281}.

\bibitem[Simonyan et~al.(2014{\natexlab{a}})Simonyan, Vedaldi, and Zisserman]{Simonyan_2014_BackpropVisualisation}
Karen Simonyan, Andrea Vedaldi, and Andrew Zisserman.
\newblock Deep inside convolutional networks: Visualising image classification models and saliency maps, 2014{\natexlab{a}}.
\newblock URL \url{https://arxiv.org/abs/1312.6034}.

\bibitem[Simonyan et~al.(2014{\natexlab{b}})Simonyan, Vedaldi, and Zisserman]{simonyan2014deepinsideconvolutionalnetworks}
Karen Simonyan, Andrea Vedaldi, and Andrew Zisserman.
\newblock Deep inside convolutional networks: Visualising image classification models and saliency maps, 2014{\natexlab{b}}.
\newblock URL \url{https://arxiv.org/abs/1312.6034}.

\bibitem[Slack et~al.(2020)Slack, Hilgard, Jia, Singh, and Lakkaraju]{Slack_2020_foolinglime}
Dylan Slack, Sophie Hilgard, Emily Jia, Sameer Singh, and Himabindu Lakkaraju.
\newblock Fooling lime and shap: Adversarial attacks on post hoc explanation methods.
\newblock In \emph{Proceedings of the AAAI/ACM Conference on AI, Ethics, and Society}, AIES '20, pp.\  180--186, New York, NY, USA, 2020. Association for Computing Machinery.
\newblock ISBN 9781450371100.
\newblock \doi{10.1145/3375627.3375830}.
\newblock URL \url{https://doi.org/10.1145/3375627.3375830}.

\bibitem[Slack et~al.(2021)Slack, Hilgard, Lakkaraju, and Singh]{Slac_2021_advances}
Dylan Slack, Anna Hilgard, Himabindu Lakkaraju, and Sameer Singh.
\newblock Counterfactual explanations can be manipulated.
\newblock In M.~Ranzato, A.~Beygelzimer, Y.~Dauphin, P.S. Liang, and J.~Wortman Vaughan (eds.), \emph{Advances in Neural Information Processing Systems}, volume~34, pp.\  62--75. Curran Associates, Inc., 2021.
\newblock URL \url{https://proceedings.neurips.cc/paper_files/paper/2021/file/009c434cab57de48a31f6b669e7ba266-Paper.pdf}.

\bibitem[Smart \& Kasirzadeh(2024)Smart and Kasirzadeh]{Smart_2024}
Andrew Smart and Atoosa Kasirzadeh.
\newblock Beyond model interpretability: socio-structural explanations in machine learning.
\newblock \emph{AI and SOCIETY}, September 2024.
\newblock ISSN 1435-5655.
\newblock \doi{10.1007/s00146-024-02056-1}.
\newblock URL \url{http://dx.doi.org/10.1007/s00146-024-02056-1}.

\bibitem[Smith(2024{\natexlab{a}})]{Smith_2024_strongfeature}
Lewis Smith.
\newblock The “strong” feature hypothesis could be wrong, Aug 2024{\natexlab{a}}.
\newblock URL \url{https://www.lesswrong.com/posts/tojtPCCRpKLSHBdpn/the-strong-feature-hypothesis-could-be-wrong}.

\bibitem[Smith(2024{\natexlab{b}})]{smith2024latent}
Lewis Smith.
\newblock Feature is overloaded terminology.
\newblock LessWrong, 2024{\natexlab{b}}.
\newblock Available at: \url{https://www.lesswrong.com/posts/9Nkb389gidsozY9Tf/lewis-smith-s-shortform?commentId=fd64ALuWK8rXdLKz6}.

\bibitem[Smolensky(1986)]{Smolensky_1986}
P.~Smolensky.
\newblock \emph{Neural and conceptual interpretation of PDP models}, pp.\  390–431.
\newblock MIT Press, Cambridge, MA, USA, 1986.
\newblock ISBN 0262631105.

\bibitem[Sohl-Dickstein et~al.(2015)Sohl-Dickstein, Weiss, Maheswaranathan, and Ganguli]{pmlr-v37-sohl-dickstein15}
Jascha Sohl-Dickstein, Eric Weiss, Niru Maheswaranathan, and Surya Ganguli.
\newblock Deep unsupervised learning using nonequilibrium thermodynamics.
\newblock In Francis Bach and David Blei (eds.), \emph{Proceedings of the 32nd International Conference on Machine Learning}, volume~37 of \emph{Proceedings of Machine Learning Research}, pp.\  2256--2265, Lille, France, 07--09 Jul 2015. PMLR.
\newblock URL \url{https://proceedings.mlr.press/v37/sohl-dickstein15.html}.

\bibitem[Srivastava et~al.(2014)Srivastava, Hinton, Krizhevsky, Sutskever, and Salakhutdinov]{srivastava_2014}
Nitish Srivastava, Geoffrey Hinton, Alex Krizhevsky, Ilya Sutskever, and Ruslan Salakhutdinov.
\newblock Dropout: A simple way to prevent neural networks from overfitting.
\newblock \emph{Journal of Machine Learning Research}, 15\penalty0 (56):\penalty0 1929--1958, 2014.
\newblock URL \url{http://jmlr.org/papers/v15/srivastava14a.html}.

\bibitem[Stander et~al.(2025)Stander, Yu, Fan, and Biderman]{stander2024grokkinggroupmultiplicationcosets}
Dashiell Stander, Qinan Yu, Honglu Fan, and Stella Biderman.
\newblock Grokking group multiplication with cosets.
\newblock In \emph{Proceedings of the 41st International Conference on Machine Learning}, ICML'24. JMLR.org, 2025.

\bibitem[Sundararajan et~al.(2017)Sundararajan, Taly, and Yan]{Sundarajan_2017_IntegratedGradients}
Mukund Sundararajan, Ankur Taly, and Qiqi Yan.
\newblock Axiomatic attribution for deep networks.
\newblock In \emph{ICML}, ICML'17, pp.\  3319--3328. JMLR.org, 2017.

\bibitem[Syed et~al.(2024)Syed, Rager, and Conmy]{syed2023attributionpatchingoutperformsautomated}
Aaquib Syed, Can Rager, and Arthur Conmy.
\newblock Attribution patching outperforms automated circuit discovery.
\newblock In Yonatan Belinkov, Najoung Kim, Jaap Jumelet, Hosein Mohebbi, Aaron Mueller, and Hanjie Chen (eds.), \emph{Proceedings of the 7th BlackboxNLP Workshop: Analyzing and Interpreting Neural Networks for NLP}, pp.\  407--416, Miami, Florida, US, November 2024. Association for Computational Linguistics.
\newblock \doi{10.18653/v1/2024.blackboxnlp-1.25}.
\newblock URL \url{https://aclanthology.org/2024.blackboxnlp-1.25/}.

\bibitem[Szegedy et~al.(2014)Szegedy, Zaremba, Sutskever, Bruna, Erhan, Goodfellow, and Fergus]{szegedy2014intriguingpropertiesneuralnetworks}
Christian Szegedy, Wojciech Zaremba, Ilya Sutskever, Joan Bruna, Dumitru Erhan, Ian Goodfellow, and Rob Fergus.
\newblock Intriguing properties of neural networks, 2014.
\newblock URL \url{https://arxiv.org/abs/1312.6199}.

\bibitem[Szegedy et~al.(2015)Szegedy, Liu, Jia, Sermanet, Reed, Anguelov, Erhan, Vanhoucke, and Rabinovich]{szegedy2015inception}
Christian Szegedy, Wei Liu, Yangqing Jia, Pierre Sermanet, Scott Reed, Dragomir Anguelov, Dumitru Erhan, Vincent Vanhoucke, and Andrew Rabinovich.
\newblock Going deeper with convolutions.
\newblock In \emph{2015 IEEE Conference on Computer Vision and Pattern Recognition (CVPR)}, pp.\  1--9, 2015.
\newblock \doi{10.1109/CVPR.2015.7298594}.

\bibitem[Tamkin et~al.(2025)Tamkin, Taufeeque, and Goodman]{tamkin2023codebookfeaturessparsediscrete}
Alex Tamkin, Mohammad Taufeeque, and Noah~D. Goodman.
\newblock Codebook features: sparse and discrete interpretability for neural networks.
\newblock In \emph{Proceedings of the 41st International Conference on Machine Learning}, ICML'24. JMLR.org, 2025.

\bibitem[Team et~al.(2024)Team, Anil, Borgeaud, Alayrac, Yu, Soricut, Schalkwyk, Dai, Hauth, Millican, Silver, Johnson, Antonoglou, Schrittwieser, Glaese, Chen, Pitler, Lillicrap, Lazaridou, Firat, Molloy, Isard, Barham, Hennigan, Lee, Viola, Reynolds, Xu, Doherty, Collins, Meyer, Rutherford, Moreira, Ayoub, Goel, Krawczyk, Du, Chi, Cheng, Ni, Shah, Kane, Chan, Faruqui, Severyn, Lin, Li, Cheng, Ittycheriah, Mahdieh, Chen, Sun, Tran, Bagri, Lakshminarayanan, Liu, Orban, Güra, Zhou, Song, Boffy, Ganapathy, Zheng, Choe, Ágoston Weisz, Zhu, Lu, Gopal, Kahn, Kula, Pitman, Shah, Taropa, Merey, Baeuml, Chen, Shafey, Zhang, Sercinoglu, Tucker, Piqueras, Krikun, Barr, Savinov, Danihelka, Roelofs, White, Andreassen, von Glehn, Yagati, Kazemi, Gonzalez, Khalman, Sygnowski, Frechette, Smith, Culp, Proleev, Luan, Chen, Lottes, Schucher, Lebron, Rrustemi, Clay, Crone, Kocisky, Zhao, Perz, Yu, Howard, Bloniarz, Rae, Lu, Sifre, Maggioni, Alcober, Garrette, Barnes, Thakoor, Austin, Barth-Maron, Wong, Joshi, Chaabouni,
  Fatiha, Ahuja, Tomar, Senter, Chadwick, Kornakov, Attaluri, Iturrate, Liu, Li, Cogan, Chen, Jia, Gu, Zhang, Grimstad, Hartman, Garcia, Pillai, Devlin, Laskin, de~Las~Casas, Valter, Tao, Blanco, Badia, Reitter, Chen, Brennan, Rivera, Brin, Iqbal, Surita, Labanowski, Rao, Winkler, Parisotto, Gu, Olszewska, Addanki, Miech, Louis, Teplyashin, Brown, Catt, Balaguer, Xiang, Wang, Ashwood, Briukhov, Webson, Ganapathy, Sanghavi, Kannan, Chang, Stjerngren, Djolonga, Sun, Bapna, Aitchison, Pejman, Michalewski, Yu, Wang, Love, Ahn, Bloxwich, Han, Humphreys, Sellam, Bradbury, Godbole, Samangooei, Damoc, Kaskasoli, Arnold, Vasudevan, Agrawal, Riesa, Lepikhin, Tanburn, Srinivasan, Lim, Hodkinson, Shyam, Ferret, Hand, Garg, Paine, Li, Li, Giang, Neitz, Abbas, York, Reid, Cole, Chowdhery, Das, Rogozińska, Nikolaev, Sprechmann, Nado, Zilka, Prost, He, Monteiro, Mishra, Welty, Newlan, Jia, Allamanis, Hu, de~Liedekerke, Gilmer, Saroufim, Rijhwani, Hou, Shrivastava, Baddepudi, Goldin, Ozturel, Cassirer, Xu, Sohn, Sachan,
  Amplayo, Swanson, Petrova, Narayan, Guez, Brahma, Landon, Patel, Zhao, Villela, Wang, Jia, Rahtz, Giménez, Yeung, Keeling, Georgiev, Mincu, Wu, Haykal, Saputro, Vodrahalli, Qin, Cankara, Sharma, Fernando, Hawkins, Neyshabur, Kim, Hutter, Agrawal, Castro-Ros, van~den Driessche, Wang, Yang, yiin Chang, Komarek, McIlroy, Lučić, Zhang, Farhan, Sharman, Natsev, Michel, Bansal, Qiao, Cao, Shakeri, Butterfield, Chung, Rubenstein, Agrawal, Mensch, Soparkar, Lenc, Chung, Pope, Maggiore, Kay, Jhakra, Wang, Maynez, Phuong, Tobin, Tacchetti, Trebacz, Robinson, Katariya, Riedel, Bailey, Xiao, Ghelani, Aroyo, Slone, Houlsby, Xiong, Yang, Gribovskaya, Adler, Wirth, Lee, Li, Kagohara, Pavagadhi, Bridgers, Bortsova, Ghemawat, Ahmed, Liu, Powell, Bolina, Iinuma, Zablotskaia, Besley, Chung, Dozat, Comanescu, Si, Greer, Su, Polacek, Kaufman, Tokumine, Hu, Buchatskaya, Miao, Elhawaty, Siddhant, Tomasev, Xing, Greer, Miller, Ashraf, Roy, Zhang, Ma, Filos, Besta, Blevins, Klimenko, Yeh, Changpinyo, Mu, Chang, Pajarskas, Muir,
  Cohen, Lan, Haridasan, Marathe, Hansen, Douglas, Samuel, Wang, Austin, Lan, Jiang, Chiu, Lorenzo, Sjösund, Cevey, Gleicher, Avrahami, Boral, Srinivasan, Selo, May, Aisopos, Hussenot, Soares, Baumli, Chang, Recasens, Caine, Pritzel, Pavetic, Pardo, Gergely, Frye, Ramasesh, Horgan, Badola, Kassner, Roy, Dyer, Campos, Tomala, Tang, Badawy, White, Mustafa, Lang, Jindal, Vikram, Gong, Caelles, Hemsley, Thornton, Feng, Stokowiec, Zheng, Thacker, Çağlar Ünlü, Zhang, Saleh, Svensson, Bileschi, Patil, Anand, Ring, Tsihlas, Vezer, Selvi, Shevlane, Rodriguez, Kwiatkowski, Daruki, Rong, Dafoe, FitzGerald, Gu-Lemberg, Khan, Hendricks, Pellat, Feinberg, Cobon-Kerr, Sainath, Rauh, Hashemi, Ives, Hasson, Noland, Cao, Byrd, Hou, Wang, Sottiaux, Paganini, Lespiau, Moufarek, Hassan, Shivakumar, van Amersfoort, Mandhane, Joshi, Goyal, Tung, Brock, Sheahan, Misra, Li, Rakićević, Dehghani, Liu, Mittal, Oh, Noury, Sezener, Huot, Lamm, Cao, Chen, Mudgal, Stella, Brooks, Vasudevan, Liu, Chain, Melinkeri, Cohen, Wang,
  Seymore, Zubkov, Goel, Yue, Krishnakumaran, Albert, Hurley, Sano, Mohananey, Joughin, Filonov, Kępa, Eldawy, Lim, Rishi, Badiezadegan, Bos, Chang, Jain, Padmanabhan, Puttagunta, Krishna, Baker, Kalb, Bedapudi, Kurzrok, Lei, Yu, Litvin, Zhou, Wu, Sobell, Siciliano, Papir, Neale, Bragagnolo, Toor, Chen, Anklin, Wang, Feng, Gholami, Ling, Liu, Walter, Moghaddam, Kishore, Adamek, Mercado, Mallinson, Wandekar, Cagle, Ofek, Garrido, Lombriser, Mukha, Sun, Mohammad, Matak, Qian, Peswani, Janus, Yuan, Schelin, David, Garg, He, Duzhyi, Älgmyr, Lottaz, Li, Yadav, Xu, Chinien, Shivanna, Chuklin, Li, Spadine, Wolfe, Mohamed, Das, Dai, He, von Dincklage, Upadhyay, Maurya, Chi, Krause, Salama, Rabinovitch, M, Selvan, Dektiarev, Ghiasi, Guven, Gupta, Liu, Sharma, Shtacher, Paul, Akerlund, Aubet, Huang, Zhu, Zhu, Teixeira, Fritze, Bertolini, Marinescu, Bölle, Paulus, Gupta, Latkar, Chang, Sanders, Wilson, Wu, Tan, Thiet, Doshi, Lall, Mishra, Chen, Luong, Benjamin, Lee, Andrejczuk, Rabiej, Ranjan, Styrc, Yin, Simon,
  Harriott, Bansal, Robsky, Bacon, Greene, Mirylenka, Zhou, Sarvana, Goyal, Andermatt, Siegler, Horn, Israel, Pongetti, Chen, Selvatici, Silva, Wang, Tolins, Guu, Yogev, Cai, Agostini, Shah, Nguyen, Donnaile, Pereira, Friso, Stambler, Kurzrok, Kuang, Romanikhin, Geller, Yan, Jang, Lee, Fica, Malmi, Tan, Banica, Balle, Pham, Huang, Avram, Shi, Singh, Hidey, Ahuja, Saxena, Dooley, Potharaju, O'Neill, Gokulchandran, Foley, Zhao, Dusenberry, Liu, Mehta, Kotikalapudi, Safranek-Shrader, Goodman, Kessinger, Globen, Kolhar, Gorgolewski, Ibrahim, Song, Eichenbaum, Brovelli, Potluri, Lahoti, Baetu, Ghorbani, Chen, Crawford, Pal, Sridhar, Gurita, Mujika, Petrovski, Cedoz, Li, Chen, Santo, Goyal, Punjabi, Kappaganthu, Kwak, LV, Velury, Choudhury, Hall, Shah, Figueira, Thomas, Lu, Zhou, Kumar, Jurdi, Chikkerur, Ma, Yu, Kwak, Ähdel, Rajayogam, Choma, Liu, Barua, Ji, Park, Hellendoorn, Bailey, Bilal, Zhou, Khatir, Sutton, Rzadkowski, Macintosh, Shagin, Medina, Liang, Zhou, Shah, Bi, Dankovics, Banga, Lehmann, Bredesen,
  Lin, Hoffmann, Lai, Chung, Yang, Balani, Bražinskas, Sozanschi, Hayes, Alcalde, Makarov, Chen, Stella, Snijders, Mandl, Kärrman, Nowak, Wu, Dyck, Vaidyanathan, R, Mallet, Rudominer, Johnston, Mittal, Udathu, Christensen, Verma, Irving, Santucci, Elsayed, Davoodi, Georgiev, Tenney, Hua, Cideron, Leurent, Alnahlawi, Georgescu, Wei, Zheng, Scandinaro, Jiang, Snoek, Sundararajan, Wang, Ontiveros, Karo, Cole, Rajashekhar, Tumeh, Ben-David, Jain, Uesato, Datta, Bunyan, Wu, Zhang, Stanczyk, Zhang, Steiner, Naskar, Azzam, Johnson, Paszke, Chiu, Elias, Mohiuddin, Muhammad, Miao, Lee, Vieillard, Park, Zhang, Stanway, Garmon, Karmarkar, Dong, Lee, Kumar, Zhou, Evens, Isaac, Irving, Loper, Fink, Arkatkar, Chen, Shafran, Petrychenko, Chen, Jia, Levskaya, Zhu, Grabowski, Mao, Magni, Yao, Snaider, Casagrande, Palmer, Suganthan, Castaño, Giannoumis, Kim, Rybiński, Sreevatsa, Prendki, Soergel, Goedeckemeyer, Gierke, Jafari, Gaba, Wiesner, Wright, Wei, Vashisht, Kulizhskaya, Hoover, Le, Li, Iwuanyanwu, Liu, Ramirez,
  Khorlin, Cui, LIN, Wu, Aguilar, Pallo, Chakladar, Perng, Abellan, Zhang, Dasgupta, Kushman, Penchev, Repina, Wu, van~der Weide, Ponnapalli, Kaplan, Simsa, Li, Dousse, Yang, Piper, Ie, Pasumarthi, Lintz, Vijayakumar, Andor, Valenzuela, Lui, Paduraru, Peng, Lee, Zhang, Greene, Nguyen, Kurylowicz, Hardin, Dixon, Janzer, Choo, Feng, Zhang, Singhal, Du, McKinnon, Antropova, Bolukbasi, Keller, Reid, Finchelstein, Raad, Crocker, Hawkins, Dadashi, Gaffney, Franko, Bulanova, Leblond, Chung, Askham, Cobo, Xu, Fischer, Xu, Sorokin, Alberti, Lin, Evans, Dimitriev, Forbes, Banarse, Tung, Omernick, Bishop, Sterneck, Jain, Xia, Amid, Piccinno, Wang, Banzal, Mankowitz, Polozov, Krakovna, Brown, Bateni, Duan, Firoiu, Thotakuri, Natan, Geist, tan Girgin, Li, Ye, Roval, Tojo, Kwong, Lee-Thorp, Yew, Sinopalnikov, Ramos, Mellor, Sharma, Wu, Miller, Sonnerat, Vnukov, Greig, Beattie, Caveness, Bai, Eisenschlos, Korchemniy, Tsai, Jasarevic, Kong, Dao, Zheng, Liu, Yang, Zhu, Teh, Sanmiya, Gladchenko, Trdin, Toyama, Rosen, Tavakkol,
  Xue, Elkind, Woodman, Carpenter, Papamakarios, Kemp, Kafle, Grunina, Sinha, Talbert, Wu, Owusu-Afriyie, Du, Thornton, Pont-Tuset, Narayana, Li, Fatehi, Wieting, Ajmeri, Uria, Ko, Knight, Héliou, Niu, Gu, Pang, Li, Levine, Stolovich, Santamaria-Fernandez, Goenka, Yustalim, Strudel, Elqursh, Deck, Lee, Li, Levin, Hoffmann, Holtmann-Rice, Bachem, Arora, Koh, Yeganeh, Põder, Tariq, Sun, Ionita, Seyedhosseini, Tafti, Liu, Gulati, Liu, Ye, Chrzaszcz, Wang, Sethi, Li, Brown, Singh, Fan, Parisi, Stanton, Koverkathu, Choquette-Choo, Li, Lu, Ittycheriah, Shroff, Varadarajan, Bahargam, Willoughby, Gaddy, Desjardins, Cornero, Robenek, Mittal, Albrecht, Shenoy, Moiseev, Jacobsson, Ghaffarkhah, Rivière, Walton, Crepy, Parrish, Zhou, Farabet, Radebaugh, Srinivasan, van~der Salm, Fidjeland, Scellato, Latorre-Chimoto, Klimczak-Plucińska, Bridson, de~Cesare, Hudson, Mendolicchio, Walker, Morris, Mauger, Guseynov, Reid, Odoom, Loher, Cotruta, Yenugula, Grewe, Petrushkina, Duerig, Sanchez, Yadlowsky, Shen, Globerson, Webb,
  Dua, Li, Bhupatiraju, Hurt, Qureshi, Agarwal, Shani, Eyal, Khare, Belle, Wang, Tekur, Kale, Wei, Sang, Saeta, Liechty, Sun, Zhao, Lee, Nayak, Fritz, Vuyyuru, Aslanides, Vyas, Wicke, Ma, Eltyshev, Martin, Cate, Manyika, Amiri, Kim, Xiong, Kang, Luisier, Tripuraneni, Madras, Guo, Waters, Wang, Ainslie, Baldridge, Zhang, Pruthi, Bauer, Yang, Mansour, Gelman, Xu, Polovets, Liu, Cai, Chen, Sheng, Xue, Ozair, Angermueller, Li, Sinha, Wang, Wiesinger, Koukoumidis, Tian, Iyer, Gurumurthy, Goldenson, Shah, Blake, Yu, Urbanowicz, Palomaki, Fernando, Durden, Mehta, Momchev, Rahimtoroghi, Georgaki, Raul, Ruder, Redshaw, Lee, Zhou, Jalan, Li, Hechtman, Schuh, Nasr, Milan, Mikulik, Franco, Green, Nguyen, Kelley, Mahendru, Hu, Howland, Vargas, Hui, Bansal, Rao, Ghiya, Wang, Ye, Sarr, Preston, Elish, Li, Kaku, Gupta, Pasupat, Juan, Someswar, M., Chen, Amini, Fabrikant, Chu, Dong, Muthal, Buthpitiya, Jauhari, Hua, Khandelwal, Hitron, Ren, Rinaldi, Drath, Dabush, Jiang, Godhia, Sachs, Chen, Fan, Taitelbaum, Noga, Dai, Wang,
  Liang, Hamer, Ferng, Elkind, Atias, Lee, Listík, Carlen, van~de Kerkhof, Pikus, Zaher, Müller, Zykova, Stefanec, Gatsko, Hirnschall, Sethi, Xu, Ahuja, Tsai, Stefanoiu, Feng, Dhandhania, Katyal, Gupta, Parulekar, Pitta, Zhao, Bhatia, Bhavnani, Alhadlaq, Li, Danenberg, Tu, Pine, Filippova, Ghosh, Limonchik, Urala, Lanka, Clive, Sun, Li, Wu, Hongtongsak, Li, Thakkar, Omarov, Majmundar, Alverson, Kucharski, Patel, Jain, Zabelin, Pelagatti, Kohli, Kumar, Kim, Sankar, Shah, Ramachandruni, Zeng, Bariach, Weidinger, Vu, Andreev, He, Hui, Kashem, Subramanya, Hsiao, Hassabis, Kavukcuoglu, Sadovsky, Le, Strohman, Wu, Petrov, Dean, and Vinyals]{geminiteam2024geminifamilyhighlycapable}
Gemini Team, Rohan Anil, Sebastian Borgeaud, Jean-Baptiste Alayrac, Jiahui Yu, Radu Soricut, Johan Schalkwyk, Andrew~M. Dai, Anja Hauth, Katie Millican, David Silver, Melvin Johnson, Ioannis Antonoglou, Julian Schrittwieser, Amelia Glaese, Jilin Chen, Emily Pitler, Timothy Lillicrap, Angeliki Lazaridou, Orhan Firat, James Molloy, Michael Isard, Paul~R. Barham, Tom Hennigan, Benjamin Lee, Fabio Viola, Malcolm Reynolds, Yuanzhong Xu, Ryan Doherty, Eli Collins, Clemens Meyer, Eliza Rutherford, Erica Moreira, Kareem Ayoub, Megha Goel, Jack Krawczyk, Cosmo Du, Ed~Chi, Heng-Tze Cheng, Eric Ni, Purvi Shah, Patrick Kane, Betty Chan, Manaal Faruqui, Aliaksei Severyn, Hanzhao Lin, YaGuang Li, Yong Cheng, Abe Ittycheriah, Mahdis Mahdieh, Mia Chen, Pei Sun, Dustin Tran, Sumit Bagri, Balaji Lakshminarayanan, Jeremiah Liu, Andras Orban, Fabian Güra, Hao Zhou, Xinying Song, Aurelien Boffy, Harish Ganapathy, Steven Zheng, HyunJeong Choe, Ágoston Weisz, Tao Zhu, Yifeng Lu, Siddharth Gopal, Jarrod Kahn, Maciej Kula, Jeff
  Pitman, Rushin Shah, Emanuel Taropa, Majd~Al Merey, Martin Baeuml, Zhifeng Chen, Laurent~El Shafey, Yujing Zhang, Olcan Sercinoglu, George Tucker, Enrique Piqueras, Maxim Krikun, Iain Barr, Nikolay Savinov, Ivo Danihelka, Becca Roelofs, Anaïs White, Anders Andreassen, Tamara von Glehn, Lakshman Yagati, Mehran Kazemi, Lucas Gonzalez, Misha Khalman, Jakub Sygnowski, Alexandre Frechette, Charlotte Smith, Laura Culp, Lev Proleev, Yi~Luan, Xi~Chen, James Lottes, Nathan Schucher, Federico Lebron, Alban Rrustemi, Natalie Clay, Phil Crone, Tomas Kocisky, Jeffrey Zhao, Bartek Perz, Dian Yu, Heidi Howard, Adam Bloniarz, Jack~W. Rae, Han Lu, Laurent Sifre, Marcello Maggioni, Fred Alcober, Dan Garrette, Megan Barnes, Shantanu Thakoor, Jacob Austin, Gabriel Barth-Maron, William Wong, Rishabh Joshi, Rahma Chaabouni, Deeni Fatiha, Arun Ahuja, Gaurav~Singh Tomar, Evan Senter, Martin Chadwick, Ilya Kornakov, Nithya Attaluri, Iñaki Iturrate, Ruibo Liu, Yunxuan Li, Sarah Cogan, Jeremy Chen, Chao Jia, Chenjie Gu, Qiao Zhang,
  Jordan Grimstad, Ale~Jakse Hartman, Xavier Garcia, Thanumalayan~Sankaranarayana Pillai, Jacob Devlin, Michael Laskin, Diego de~Las~Casas, Dasha Valter, Connie Tao, Lorenzo Blanco, Adrià~Puigdomènech Badia, David Reitter, Mianna Chen, Jenny Brennan, Clara Rivera, Sergey Brin, Shariq Iqbal, Gabriela Surita, Jane Labanowski, Abhi Rao, Stephanie Winkler, Emilio Parisotto, Yiming Gu, Kate Olszewska, Ravi Addanki, Antoine Miech, Annie Louis, Denis Teplyashin, Geoff Brown, Elliot Catt, Jan Balaguer, Jackie Xiang, Pidong Wang, Zoe Ashwood, Anton Briukhov, Albert Webson, Sanjay Ganapathy, Smit Sanghavi, Ajay Kannan, Ming-Wei Chang, Axel Stjerngren, Josip Djolonga, Yuting Sun, Ankur Bapna, Matthew Aitchison, Pedram Pejman, Henryk Michalewski, Tianhe Yu, Cindy Wang, Juliette Love, Junwhan Ahn, Dawn Bloxwich, Kehang Han, Peter Humphreys, Thibault Sellam, James Bradbury, Varun Godbole, Sina Samangooei, Bogdan Damoc, Alex Kaskasoli, Sébastien M.~R. Arnold, Vijay Vasudevan, Shubham Agrawal, Jason Riesa, Dmitry
  Lepikhin, Richard Tanburn, Srivatsan Srinivasan, Hyeontaek Lim, Sarah Hodkinson, Pranav Shyam, Johan Ferret, Steven Hand, Ankush Garg, Tom~Le Paine, Jian Li, Yujia Li, Minh Giang, Alexander Neitz, Zaheer Abbas, Sarah York, Machel Reid, Elizabeth Cole, Aakanksha Chowdhery, Dipanjan Das, Dominika Rogozińska, Vitaliy Nikolaev, Pablo Sprechmann, Zachary Nado, Lukas Zilka, Flavien Prost, Luheng He, Marianne Monteiro, Gaurav Mishra, Chris Welty, Josh Newlan, Dawei Jia, Miltiadis Allamanis, Clara~Huiyi Hu, Raoul de~Liedekerke, Justin Gilmer, Carl Saroufim, Shruti Rijhwani, Shaobo Hou, Disha Shrivastava, Anirudh Baddepudi, Alex Goldin, Adnan Ozturel, Albin Cassirer, Yunhan Xu, Daniel Sohn, Devendra Sachan, Reinald~Kim Amplayo, Craig Swanson, Dessie Petrova, Shashi Narayan, Arthur Guez, Siddhartha Brahma, Jessica Landon, Miteyan Patel, Ruizhe Zhao, Kevin Villela, Luyu Wang, Wenhao Jia, Matthew Rahtz, Mai Giménez, Legg Yeung, James Keeling, Petko Georgiev, Diana Mincu, Boxi Wu, Salem Haykal, Rachel Saputro, Kiran
  Vodrahalli, James Qin, Zeynep Cankara, Abhanshu Sharma, Nick Fernando, Will Hawkins, Behnam Neyshabur, Solomon Kim, Adrian Hutter, Priyanka Agrawal, Alex Castro-Ros, George van~den Driessche, Tao Wang, Fan Yang, Shuo yiin Chang, Paul Komarek, Ross McIlroy, Mario Lučić, Guodong Zhang, Wael Farhan, Michael Sharman, Paul Natsev, Paul Michel, Yamini Bansal, Siyuan Qiao, Kris Cao, Siamak Shakeri, Christina Butterfield, Justin Chung, Paul~Kishan Rubenstein, Shivani Agrawal, Arthur Mensch, Kedar Soparkar, Karel Lenc, Timothy Chung, Aedan Pope, Loren Maggiore, Jackie Kay, Priya Jhakra, Shibo Wang, Joshua Maynez, Mary Phuong, Taylor Tobin, Andrea Tacchetti, Maja Trebacz, Kevin Robinson, Yash Katariya, Sebastian Riedel, Paige Bailey, Kefan Xiao, Nimesh Ghelani, Lora Aroyo, Ambrose Slone, Neil Houlsby, Xuehan Xiong, Zhen Yang, Elena Gribovskaya, Jonas Adler, Mateo Wirth, Lisa Lee, Music Li, Thais Kagohara, Jay Pavagadhi, Sophie Bridgers, Anna Bortsova, Sanjay Ghemawat, Zafarali Ahmed, Tianqi Liu, Richard Powell,
  Vijay Bolina, Mariko Iinuma, Polina Zablotskaia, James Besley, Da-Woon Chung, Timothy Dozat, Ramona Comanescu, Xiance Si, Jeremy Greer, Guolong Su, Martin Polacek, Raphaël~Lopez Kaufman, Simon Tokumine, Hexiang Hu, Elena Buchatskaya, Yingjie Miao, Mohamed Elhawaty, Aditya Siddhant, Nenad Tomasev, Jinwei Xing, Christina Greer, Helen Miller, Shereen Ashraf, Aurko Roy, Zizhao Zhang, Ada Ma, Angelos Filos, Milos Besta, Rory Blevins, Ted Klimenko, Chih-Kuan Yeh, Soravit Changpinyo, Jiaqi Mu, Oscar Chang, Mantas Pajarskas, Carrie Muir, Vered Cohen, Charline~Le Lan, Krishna Haridasan, Amit Marathe, Steven Hansen, Sholto Douglas, Rajkumar Samuel, Mingqiu Wang, Sophia Austin, Chang Lan, Jiepu Jiang, Justin Chiu, Jaime~Alonso Lorenzo, Lars~Lowe Sjösund, Sébastien Cevey, Zach Gleicher, Thi Avrahami, Anudhyan Boral, Hansa Srinivasan, Vittorio Selo, Rhys May, Konstantinos Aisopos, Léonard Hussenot, Livio~Baldini Soares, Kate Baumli, Michael~B. Chang, Adrià Recasens, Ben Caine, Alexander Pritzel, Filip Pavetic,
  Fabio Pardo, Anita Gergely, Justin Frye, Vinay Ramasesh, Dan Horgan, Kartikeya Badola, Nora Kassner, Subhrajit Roy, Ethan Dyer, Víctor~Campos Campos, Alex Tomala, Yunhao Tang, Dalia~El Badawy, Elspeth White, Basil Mustafa, Oran Lang, Abhishek Jindal, Sharad Vikram, Zhitao Gong, Sergi Caelles, Ross Hemsley, Gregory Thornton, Fangxiaoyu Feng, Wojciech Stokowiec, Ce~Zheng, Phoebe Thacker, Çağlar Ünlü, Zhishuai Zhang, Mohammad Saleh, James Svensson, Max Bileschi, Piyush Patil, Ankesh Anand, Roman Ring, Katerina Tsihlas, Arpi Vezer, Marco Selvi, Toby Shevlane, Mikel Rodriguez, Tom Kwiatkowski, Samira Daruki, Keran Rong, Allan Dafoe, Nicholas FitzGerald, Keren Gu-Lemberg, Mina Khan, Lisa~Anne Hendricks, Marie Pellat, Vladimir Feinberg, James Cobon-Kerr, Tara Sainath, Maribeth Rauh, Sayed~Hadi Hashemi, Richard Ives, Yana Hasson, Eric Noland, Yuan Cao, Nathan Byrd, Le~Hou, Qingze Wang, Thibault Sottiaux, Michela Paganini, Jean-Baptiste Lespiau, Alexandre Moufarek, Samer Hassan, Kaushik Shivakumar, Joost van
  Amersfoort, Amol Mandhane, Pratik Joshi, Anirudh Goyal, Matthew Tung, Andrew Brock, Hannah Sheahan, Vedant Misra, Cheng Li, Nemanja Rakićević, Mostafa Dehghani, Fangyu Liu, Sid Mittal, Junhyuk Oh, Seb Noury, Eren Sezener, Fantine Huot, Matthew Lamm, Nicola~De Cao, Charlie Chen, Sidharth Mudgal, Romina Stella, Kevin Brooks, Gautam Vasudevan, Chenxi Liu, Mainak Chain, Nivedita Melinkeri, Aaron Cohen, Venus Wang, Kristie Seymore, Sergey Zubkov, Rahul Goel, Summer Yue, Sai Krishnakumaran, Brian Albert, Nate Hurley, Motoki Sano, Anhad Mohananey, Jonah Joughin, Egor Filonov, Tomasz Kępa, Yomna Eldawy, Jiawern Lim, Rahul Rishi, Shirin Badiezadegan, Taylor Bos, Jerry Chang, Sanil Jain, Sri Gayatri~Sundara Padmanabhan, Subha Puttagunta, Kalpesh Krishna, Leslie Baker, Norbert Kalb, Vamsi Bedapudi, Adam Kurzrok, Shuntong Lei, Anthony Yu, Oren Litvin, Xiang Zhou, Zhichun Wu, Sam Sobell, Andrea Siciliano, Alan Papir, Robby Neale, Jonas Bragagnolo, Tej Toor, Tina Chen, Valentin Anklin, Feiran Wang, Richie Feng, Milad
  Gholami, Kevin Ling, Lijuan Liu, Jules Walter, Hamid Moghaddam, Arun Kishore, Jakub Adamek, Tyler Mercado, Jonathan Mallinson, Siddhinita Wandekar, Stephen Cagle, Eran Ofek, Guillermo Garrido, Clemens Lombriser, Maksim Mukha, Botu Sun, Hafeezul~Rahman Mohammad, Josip Matak, Yadi Qian, Vikas Peswani, Pawel Janus, Quan Yuan, Leif Schelin, Oana David, Ankur Garg, Yifan He, Oleksii Duzhyi, Anton Älgmyr, Timothée Lottaz, Qi~Li, Vikas Yadav, Luyao Xu, Alex Chinien, Rakesh Shivanna, Aleksandr Chuklin, Josie Li, Carrie Spadine, Travis Wolfe, Kareem Mohamed, Subhabrata Das, Zihang Dai, Kyle He, Daniel von Dincklage, Shyam Upadhyay, Akanksha Maurya, Luyan Chi, Sebastian Krause, Khalid Salama, Pam~G Rabinovitch, Pavan Kumar~Reddy M, Aarush Selvan, Mikhail Dektiarev, Golnaz Ghiasi, Erdem Guven, Himanshu Gupta, Boyi Liu, Deepak Sharma, Idan~Heimlich Shtacher, Shachi Paul, Oscar Akerlund, François-Xavier Aubet, Terry Huang, Chen Zhu, Eric Zhu, Elico Teixeira, Matthew Fritze, Francesco Bertolini, Liana-Eleonora
  Marinescu, Martin Bölle, Dominik Paulus, Khyatti Gupta, Tejasi Latkar, Max Chang, Jason Sanders, Roopa Wilson, Xuewei Wu, Yi-Xuan Tan, Lam~Nguyen Thiet, Tulsee Doshi, Sid Lall, Swaroop Mishra, Wanming Chen, Thang Luong, Seth Benjamin, Jasmine Lee, Ewa Andrejczuk, Dominik Rabiej, Vipul Ranjan, Krzysztof Styrc, Pengcheng Yin, Jon Simon, Malcolm~Rose Harriott, Mudit Bansal, Alexei Robsky, Geoff Bacon, David Greene, Daniil Mirylenka, Chen Zhou, Obaid Sarvana, Abhimanyu Goyal, Samuel Andermatt, Patrick Siegler, Ben Horn, Assaf Israel, Francesco Pongetti, Chih-Wei~"Louis" Chen, Marco Selvatici, Pedro Silva, Kathie Wang, Jackson Tolins, Kelvin Guu, Roey Yogev, Xiaochen Cai, Alessandro Agostini, Maulik Shah, Hung Nguyen, Noah~Ó Donnaile, Sébastien Pereira, Linda Friso, Adam Stambler, Adam Kurzrok, Chenkai Kuang, Yan Romanikhin, Mark Geller, ZJ~Yan, Kane Jang, Cheng-Chun Lee, Wojciech Fica, Eric Malmi, Qijun Tan, Dan Banica, Daniel Balle, Ryan Pham, Yanping Huang, Diana Avram, Hongzhi Shi, Jasjot Singh, Chris
  Hidey, Niharika Ahuja, Pranab Saxena, Dan Dooley, Srividya~Pranavi Potharaju, Eileen O'Neill, Anand Gokulchandran, Ryan Foley, Kai Zhao, Mike Dusenberry, Yuan Liu, Pulkit Mehta, Ragha Kotikalapudi, Chalence Safranek-Shrader, Andrew Goodman, Joshua Kessinger, Eran Globen, Prateek Kolhar, Chris Gorgolewski, Ali Ibrahim, Yang Song, Ali Eichenbaum, Thomas Brovelli, Sahitya Potluri, Preethi Lahoti, Cip Baetu, Ali Ghorbani, Charles Chen, Andy Crawford, Shalini Pal, Mukund Sridhar, Petru Gurita, Asier Mujika, Igor Petrovski, Pierre-Louis Cedoz, Chenmei Li, Shiyuan Chen, Niccolò~Dal Santo, Siddharth Goyal, Jitesh Punjabi, Karthik Kappaganthu, Chester Kwak, Pallavi LV, Sarmishta Velury, Himadri Choudhury, Jamie Hall, Premal Shah, Ricardo Figueira, Matt Thomas, Minjie Lu, Ting Zhou, Chintu Kumar, Thomas Jurdi, Sharat Chikkerur, Yenai Ma, Adams Yu, Soo Kwak, Victor Ähdel, Sujeevan Rajayogam, Travis Choma, Fei Liu, Aditya Barua, Colin Ji, Ji~Ho Park, Vincent Hellendoorn, Alex Bailey, Taylan Bilal, Huanjie Zhou,
  Mehrdad Khatir, Charles Sutton, Wojciech Rzadkowski, Fiona Macintosh, Konstantin Shagin, Paul Medina, Chen Liang, Jinjing Zhou, Pararth Shah, Yingying Bi, Attila Dankovics, Shipra Banga, Sabine Lehmann, Marissa Bredesen, Zifan Lin, John~Eric Hoffmann, Jonathan Lai, Raynald Chung, Kai Yang, Nihal Balani, Arthur Bražinskas, Andrei Sozanschi, Matthew Hayes, Héctor~Fernández Alcalde, Peter Makarov, Will Chen, Antonio Stella, Liselotte Snijders, Michael Mandl, Ante Kärrman, Paweł Nowak, Xinyi Wu, Alex Dyck, Krishnan Vaidyanathan, Raghavender R, Jessica Mallet, Mitch Rudominer, Eric Johnston, Sushil Mittal, Akhil Udathu, Janara Christensen, Vishal Verma, Zach Irving, Andreas Santucci, Gamaleldin Elsayed, Elnaz Davoodi, Marin Georgiev, Ian Tenney, Nan Hua, Geoffrey Cideron, Edouard Leurent, Mahmoud Alnahlawi, Ionut Georgescu, Nan Wei, Ivy Zheng, Dylan Scandinaro, Heinrich Jiang, Jasper Snoek, Mukund Sundararajan, Xuezhi Wang, Zack Ontiveros, Itay Karo, Jeremy Cole, Vinu Rajashekhar, Lara Tumeh, Eyal
  Ben-David, Rishub Jain, Jonathan Uesato, Romina Datta, Oskar Bunyan, Shimu Wu, John Zhang, Piotr Stanczyk, Ye~Zhang, David Steiner, Subhajit Naskar, Michael Azzam, Matthew Johnson, Adam Paszke, Chung-Cheng Chiu, Jaume~Sanchez Elias, Afroz Mohiuddin, Faizan Muhammad, Jin Miao, Andrew Lee, Nino Vieillard, Jane Park, Jiageng Zhang, Jeff Stanway, Drew Garmon, Abhijit Karmarkar, Zhe Dong, Jong Lee, Aviral Kumar, Luowei Zhou, Jonathan Evens, William Isaac, Geoffrey Irving, Edward Loper, Michael Fink, Isha Arkatkar, Nanxin Chen, Izhak Shafran, Ivan Petrychenko, Zhe Chen, Johnson Jia, Anselm Levskaya, Zhenkai Zhu, Peter Grabowski, Yu~Mao, Alberto Magni, Kaisheng Yao, Javier Snaider, Norman Casagrande, Evan Palmer, Paul Suganthan, Alfonso Castaño, Irene Giannoumis, Wooyeol Kim, Mikołaj Rybiński, Ashwin Sreevatsa, Jennifer Prendki, David Soergel, Adrian Goedeckemeyer, Willi Gierke, Mohsen Jafari, Meenu Gaba, Jeremy Wiesner, Diana~Gage Wright, Yawen Wei, Harsha Vashisht, Yana Kulizhskaya, Jay Hoover, Maigo Le,
  Lu~Li, Chimezie Iwuanyanwu, Lu~Liu, Kevin Ramirez, Andrey Khorlin, Albert Cui, Tian LIN, Marcus Wu, Ricardo Aguilar, Keith Pallo, Abhishek Chakladar, Ginger Perng, Elena~Allica Abellan, Mingyang Zhang, Ishita Dasgupta, Nate Kushman, Ivo Penchev, Alena Repina, Xihui Wu, Tom van~der Weide, Priya Ponnapalli, Caroline Kaplan, Jiri Simsa, Shuangfeng Li, Olivier Dousse, Fan Yang, Jeff Piper, Nathan Ie, Rama Pasumarthi, Nathan Lintz, Anitha Vijayakumar, Daniel Andor, Pedro Valenzuela, Minnie Lui, Cosmin Paduraru, Daiyi Peng, Katherine Lee, Shuyuan Zhang, Somer Greene, Duc~Dung Nguyen, Paula Kurylowicz, Cassidy Hardin, Lucas Dixon, Lili Janzer, Kiam Choo, Ziqiang Feng, Biao Zhang, Achintya Singhal, Dayou Du, Dan McKinnon, Natasha Antropova, Tolga Bolukbasi, Orgad Keller, David Reid, Daniel Finchelstein, Maria~Abi Raad, Remi Crocker, Peter Hawkins, Robert Dadashi, Colin Gaffney, Ken Franko, Anna Bulanova, Rémi Leblond, Shirley Chung, Harry Askham, Luis~C. Cobo, Kelvin Xu, Felix Fischer, Jun Xu, Christina Sorokin,
  Chris Alberti, Chu-Cheng Lin, Colin Evans, Alek Dimitriev, Hannah Forbes, Dylan Banarse, Zora Tung, Mark Omernick, Colton Bishop, Rachel Sterneck, Rohan Jain, Jiawei Xia, Ehsan Amid, Francesco Piccinno, Xingyu Wang, Praseem Banzal, Daniel~J. Mankowitz, Alex Polozov, Victoria Krakovna, Sasha Brown, MohammadHossein Bateni, Dennis Duan, Vlad Firoiu, Meghana Thotakuri, Tom Natan, Matthieu Geist, Ser tan Girgin, Hui Li, Jiayu Ye, Ofir Roval, Reiko Tojo, Michael Kwong, James Lee-Thorp, Christopher Yew, Danila Sinopalnikov, Sabela Ramos, John Mellor, Abhishek Sharma, Kathy Wu, David Miller, Nicolas Sonnerat, Denis Vnukov, Rory Greig, Jennifer Beattie, Emily Caveness, Libin Bai, Julian Eisenschlos, Alex Korchemniy, Tomy Tsai, Mimi Jasarevic, Weize Kong, Phuong Dao, Zeyu Zheng, Frederick Liu, Fan Yang, Rui Zhu, Tian~Huey Teh, Jason Sanmiya, Evgeny Gladchenko, Nejc Trdin, Daniel Toyama, Evan Rosen, Sasan Tavakkol, Linting Xue, Chen Elkind, Oliver Woodman, John Carpenter, George Papamakarios, Rupert Kemp, Sushant
  Kafle, Tanya Grunina, Rishika Sinha, Alice Talbert, Diane Wu, Denese Owusu-Afriyie, Cosmo Du, Chloe Thornton, Jordi Pont-Tuset, Pradyumna Narayana, Jing Li, Saaber Fatehi, John Wieting, Omar Ajmeri, Benigno Uria, Yeongil Ko, Laura Knight, Amélie Héliou, Ning Niu, Shane Gu, Chenxi Pang, Yeqing Li, Nir Levine, Ariel Stolovich, Rebeca Santamaria-Fernandez, Sonam Goenka, Wenny Yustalim, Robin Strudel, Ali Elqursh, Charlie Deck, Hyo Lee, Zonglin Li, Kyle Levin, Raphael Hoffmann, Dan Holtmann-Rice, Olivier Bachem, Sho Arora, Christy Koh, Soheil~Hassas Yeganeh, Siim Põder, Mukarram Tariq, Yanhua Sun, Lucian Ionita, Mojtaba Seyedhosseini, Pouya Tafti, Zhiyu Liu, Anmol Gulati, Jasmine Liu, Xinyu Ye, Bart Chrzaszcz, Lily Wang, Nikhil Sethi, Tianrun Li, Ben Brown, Shreya Singh, Wei Fan, Aaron Parisi, Joe Stanton, Vinod Koverkathu, Christopher~A. Choquette-Choo, Yunjie Li, TJ~Lu, Abe Ittycheriah, Prakash Shroff, Mani Varadarajan, Sanaz Bahargam, Rob Willoughby, David Gaddy, Guillaume Desjardins, Marco Cornero, Brona
  Robenek, Bhavishya Mittal, Ben Albrecht, Ashish Shenoy, Fedor Moiseev, Henrik Jacobsson, Alireza Ghaffarkhah, Morgane Rivière, Alanna Walton, Clément Crepy, Alicia Parrish, Zongwei Zhou, Clement Farabet, Carey Radebaugh, Praveen Srinivasan, Claudia van~der Salm, Andreas Fidjeland, Salvatore Scellato, Eri Latorre-Chimoto, Hanna Klimczak-Plucińska, David Bridson, Dario de~Cesare, Tom Hudson, Piermaria Mendolicchio, Lexi Walker, Alex Morris, Matthew Mauger, Alexey Guseynov, Alison Reid, Seth Odoom, Lucia Loher, Victor Cotruta, Madhavi Yenugula, Dominik Grewe, Anastasia Petrushkina, Tom Duerig, Antonio Sanchez, Steve Yadlowsky, Amy Shen, Amir Globerson, Lynette Webb, Sahil Dua, Dong Li, Surya Bhupatiraju, Dan Hurt, Haroon Qureshi, Ananth Agarwal, Tomer Shani, Matan Eyal, Anuj Khare, Shreyas~Rammohan Belle, Lei Wang, Chetan Tekur, Mihir~Sanjay Kale, Jinliang Wei, Ruoxin Sang, Brennan Saeta, Tyler Liechty, Yi~Sun, Yao Zhao, Stephan Lee, Pandu Nayak, Doug Fritz, Manish~Reddy Vuyyuru, John Aslanides, Nidhi Vyas,
  Martin Wicke, Xiao Ma, Evgenii Eltyshev, Nina Martin, Hardie Cate, James Manyika, Keyvan Amiri, Yelin Kim, Xi~Xiong, Kai Kang, Florian Luisier, Nilesh Tripuraneni, David Madras, Mandy Guo, Austin Waters, Oliver Wang, Joshua Ainslie, Jason Baldridge, Han Zhang, Garima Pruthi, Jakob Bauer, Feng Yang, Riham Mansour, Jason Gelman, Yang Xu, George Polovets, Ji~Liu, Honglong Cai, Warren Chen, XiangHai Sheng, Emily Xue, Sherjil Ozair, Christof Angermueller, Xiaowei Li, Anoop Sinha, Weiren Wang, Julia Wiesinger, Emmanouil Koukoumidis, Yuan Tian, Anand Iyer, Madhu Gurumurthy, Mark Goldenson, Parashar Shah, MK~Blake, Hongkun Yu, Anthony Urbanowicz, Jennimaria Palomaki, Chrisantha Fernando, Ken Durden, Harsh Mehta, Nikola Momchev, Elahe Rahimtoroghi, Maria Georgaki, Amit Raul, Sebastian Ruder, Morgan Redshaw, Jinhyuk Lee, Denny Zhou, Komal Jalan, Dinghua Li, Blake Hechtman, Parker Schuh, Milad Nasr, Kieran Milan, Vladimir Mikulik, Juliana Franco, Tim Green, Nam Nguyen, Joe Kelley, Aroma Mahendru, Andrea Hu, Joshua
  Howland, Ben Vargas, Jeffrey Hui, Kshitij Bansal, Vikram Rao, Rakesh Ghiya, Emma Wang, Ke~Ye, Jean~Michel Sarr, Melanie~Moranski Preston, Madeleine Elish, Steve Li, Aakash Kaku, Jigar Gupta, Ice Pasupat, Da-Cheng Juan, Milan Someswar, Tejvi M., Xinyun Chen, Aida Amini, Alex Fabrikant, Eric Chu, Xuanyi Dong, Amruta Muthal, Senaka Buthpitiya, Sarthak Jauhari, Nan Hua, Urvashi Khandelwal, Ayal Hitron, Jie Ren, Larissa Rinaldi, Shahar Drath, Avigail Dabush, Nan-Jiang Jiang, Harshal Godhia, Uli Sachs, Anthony Chen, Yicheng Fan, Hagai Taitelbaum, Hila Noga, Zhuyun Dai, James Wang, Chen Liang, Jenny Hamer, Chun-Sung Ferng, Chenel Elkind, Aviel Atias, Paulina Lee, Vít Listík, Mathias Carlen, Jan van~de Kerkhof, Marcin Pikus, Krunoslav Zaher, Paul Müller, Sasha Zykova, Richard Stefanec, Vitaly Gatsko, Christoph Hirnschall, Ashwin Sethi, Xingyu~Federico Xu, Chetan Ahuja, Beth Tsai, Anca Stefanoiu, Bo~Feng, Keshav Dhandhania, Manish Katyal, Akshay Gupta, Atharva Parulekar, Divya Pitta, Jing Zhao, Vivaan Bhatia,
  Yashodha Bhavnani, Omar Alhadlaq, Xiaolin Li, Peter Danenberg, Dennis Tu, Alex Pine, Vera Filippova, Abhipso Ghosh, Ben Limonchik, Bhargava Urala, Chaitanya~Krishna Lanka, Derik Clive, Yi~Sun, Edward Li, Hao Wu, Kevin Hongtongsak, Ianna Li, Kalind Thakkar, Kuanysh Omarov, Kushal Majmundar, Michael Alverson, Michael Kucharski, Mohak Patel, Mudit Jain, Maksim Zabelin, Paolo Pelagatti, Rohan Kohli, Saurabh Kumar, Joseph Kim, Swetha Sankar, Vineet Shah, Lakshmi Ramachandruni, Xiangkai Zeng, Ben Bariach, Laura Weidinger, Tu~Vu, Alek Andreev, Antoine He, Kevin Hui, Sheleem Kashem, Amar Subramanya, Sissie Hsiao, Demis Hassabis, Koray Kavukcuoglu, Adam Sadovsky, Quoc Le, Trevor Strohman, Yonghui Wu, Slav Petrov, Jeffrey Dean, and Oriol Vinyals.
\newblock Gemini: A family of highly capable multimodal models, 2024.
\newblock URL \url{https://arxiv.org/abs/2312.11805}.

\bibitem[Tegmark \& Omohundro(2023)Tegmark and Omohundro]{tegmark2023provablysafesystemspath}
Max Tegmark and Steve Omohundro.
\newblock Provably safe systems: the only path to controllable agi, 2023.
\newblock URL \url{https://arxiv.org/abs/2309.01933}.

\bibitem[Templeton et~al.(2024)Templeton, Conerly, Marcus, Lindsey, Bricken, Chen, Pearce, Citro, Ameisen, Jones, Cunningham, Turner, McDougall, MacDiarmid, Freeman, Sumers, Rees, Batson, Jermyn, Carter, Olah, and Henighan]{templeton2024scaling}
Adly Templeton, Tom Conerly, Jonathan Marcus, Jack Lindsey, Trenton Bricken, Brian Chen, Adam Pearce, Craig Citro, Emmanuel Ameisen, Andy Jones, Hoagy Cunningham, Nicholas~L Turner, Callum McDougall, Monte MacDiarmid, C.~Daniel Freeman, Theodore~R. Sumers, Edward Rees, Joshua Batson, Adam Jermyn, Shan Carter, Chris Olah, and Tom Henighan.
\newblock Scaling monosemanticity: Extracting interpretable features from claude 3 sonnet.
\newblock \emph{Transformer Circuits Thread}, 2024.
\newblock URL \url{https://transformer-circuits.pub/2024/scaling-monosemanticity/index.html}.

\bibitem[Tenney et~al.(2019)Tenney, Das, and Pavlick]{tenney2019bertrediscoversclassicalnlp}
Ian Tenney, Dipanjan Das, and Ellie Pavlick.
\newblock {BERT} rediscovers the classical {NLP} pipeline.
\newblock In Anna Korhonen, David Traum, and Llu{\'i}s M{\`a}rquez (eds.), \emph{Proceedings of the 57th Annual Meeting of the Association for Computational Linguistics}, pp.\  4593--4601, Florence, Italy, July 2019. Association for Computational Linguistics.
\newblock \doi{10.18653/v1/P19-1452}.
\newblock URL \url{https://aclanthology.org/P19-1452/}.

\bibitem[Thibodeau(2022)]{Thibodeau_2022}
Jacques Thibodeau.
\newblock But is it really in rome? an investigation of the rome model editing technique - ai alignment forum, Dec 2022.
\newblock URL \url{https://www.alignmentforum.org/posts/QL7J9wmS6W2fWpofd/but-is-it-really-in-rome-an-investigation-of-the-rome-model}.

\bibitem[Thurnherr \& Scheurer(2024)Thurnherr and Scheurer]{thurnherr2024tracrbench}
Hannes Thurnherr and J{\'e}r{\'e}my Scheurer.
\newblock Tracrbench: Generating interpretability testbeds with large language models.
\newblock In \emph{ICML 2024 Workshop on Mechanistic Interpretability}, 2024.
\newblock URL \url{https://openreview.net/forum?id=vNubZ5zK8h}.

\bibitem[Till(2024)]{till2024sparse}
Demian Till.
\newblock Do sparse autoencoders find ``true features''?
\newblock \emph{LessWrong}, 2024.
\newblock URL \url{https://www.lesswrong.com/posts/QoR8noAB3Mp2KBA4B/do-sparse-autoencoders-find-true-features}.

\bibitem[Todd et~al.(2024)Todd, Li, Sharma, Mueller, Wallace, and Bau]{Todd_2024_Probe}
Eric Todd, Millicent~L. Li, Arnab~Sen Sharma, Aaron Mueller, Byron~C. Wallace, and David Bau.
\newblock Function vectors in large language models.
\newblock In \emph{Proceedings of the 2024 International Conference on Learning Representations}, 2024.
\newblock URL \url{https://openreview.net/forum?id=AwyxtyMwaG}.
\newblock arXiv:2310.15213.

\bibitem[Tong et~al.(2024{\natexlab{a}})Tong, Jones, and Steinhardt]{tong2024massproducingfailuresmultimodalsystems}
Shengbang Tong, Erik Jones, and Jacob Steinhardt.
\newblock Mass-producing failures of multimodal systems with language models.
\newblock In \emph{Proceedings of the 37th International Conference on Neural Information Processing Systems}, NIPS '23, Red Hook, NY, USA, 2024{\natexlab{a}}. Curran Associates Inc.

\bibitem[Tong et~al.(2024{\natexlab{b}})Tong, Mao, Huang, et~al.]{tong2024automating}
Siyuan Tong, Kaiwen Mao, Zhe Huang, et~al.
\newblock Automating psychological hypothesis generation with ai: when large language models meet causal graph.
\newblock \emph{Humanities and Social Sciences Communications}, 11:\penalty0 896, 2024{\natexlab{b}}.
\newblock \doi{10.1057/s41599-024-03407-5}.
\newblock URL \url{https://doi.org/10.1057/s41599-024-03407-5}.

\bibitem[Turner et~al.(2024)Turner, Thiergart, Leech, Udell, Vazquez, Mini, and MacDiarmid]{turner2024steeringlanguagemodelsactivation}
Alexander~Matt Turner, Lisa Thiergart, Gavin Leech, David Udell, Juan~J. Vazquez, Ulisse Mini, and Monte MacDiarmid.
\newblock Steering language models with activation engineering, 2024.
\newblock URL \url{https://arxiv.org/abs/2308.10248}.

\bibitem[Turpin et~al.(2023)Turpin, Michael, Perez, and Bowman]{Turpin2023_cot}
Miles Turpin, Julian Michael, Ethan Perez, and Samuel Bowman.
\newblock Language models don\textquotesingle t always say what they think: Unfaithful explanations in chain-of-thought prompting.
\newblock In A.~Oh, T.~Naumann, A.~Globerson, K.~Saenko, M.~Hardt, and S.~Levine (eds.), \emph{Advances in Neural Information Processing Systems}, volume~36, pp.\  74952--74965. Curran Associates, Inc., 2023.
\newblock URL \url{https://proceedings.neurips.cc/paper_files/paper/2023/file/ed3fea9033a80fea1376299fa7863f4a-Paper-Conference.pdf}.

\bibitem[{UK Government}(2022)]{ukgov2022ai}
{UK Government}.
\newblock National {AI} strategy.
\newblock Government strategy, {Government of the United Kingdom}, 2022.
\newblock URL \url{https://www.gov.uk/government/publications/national-ai-strategy/national-ai-strategy-html-version}.

\bibitem[{United Kingdom AI Safety Institute}(2024)]{ukAISI2024early}
{United Kingdom AI Safety Institute}.
\newblock Early lessons from evaluating frontier {AI} systems.
\newblock Technical report, {United Kingdom AI Safety Institute}, 2024.
\newblock URL \url{https://www.aisi.gov.uk/work/early-lessons-from-evaluating-frontier-ai-systems}.

\bibitem[{United States Congress}()]{USCopyright2022}
{United States Congress}.
\newblock Copyright law of the united states (title 17) and related laws contained in title 17 of the united states code.
\newblock URL \url{https://www.copyright.gov/title17/}.
\newblock This publication contains the text of Title 17 of the United States Code, including all amendments enacted by Congress through December 23, 2022. It includes the Copyright Act of 1976 and all subsequent amendments to copyright law; the Semiconductor Chip Protection Act of 1984, as amended; and the Vessel Hull Design Protection Act, as amended.

\bibitem[Valle-Perez et~al.(2019)Valle-Perez, Camargo, and Louis]{vallepérez2019deeplearninggeneralizesparameterfunction}
Guillermo Valle-Perez, Chico~Q. Camargo, and Ard~A. Louis.
\newblock Deep learning generalizes because the parameter-function map is biased towards simple functions.
\newblock In \emph{International Conference on Learning Representations}, 2019.
\newblock URL \url{https://openreview.net/forum?id=rye4g3AqFm}.

\bibitem[van~der Weij et~al.(2024)van~der Weij, Hofstätter, Jaffe, Brown, and Ward]{vanderweij2024aisandbagginglanguagemodels}
Teun van~der Weij, Felix Hofstätter, Ollie Jaffe, Samuel~F. Brown, and Francis~Rhys Ward.
\newblock Ai sandbagging: Language models can strategically underperform on evaluations, 2024.
\newblock URL \url{https://arxiv.org/abs/2406.07358}.

\bibitem[Vig et~al.(2020)Vig, Gehrmann, Belinkov, Qian, Nevo, Singer, and Shieber]{vig2020causalmediationanalysisinterpreting}
Jesse Vig, Sebastian Gehrmann, Yonatan Belinkov, Sharon Qian, Daniel Nevo, Yaron Singer, and Stuart Shieber.
\newblock Investigating gender bias in language models using causal mediation analysis.
\newblock In H.~Larochelle, M.~Ranzato, R.~Hadsell, M.F. Balcan, and H.~Lin (eds.), \emph{Advances in Neural Information Processing Systems}, volume~33, pp.\  12388--12401. Curran Associates, Inc., 2020.
\newblock URL \url{https://proceedings.neurips.cc/paper_files/paper/2020/file/92650b2e92217715fe312e6fa7b90d82-Paper.pdf}.

\bibitem[Viégas \& Wattenberg(2023)Viégas and Wattenberg]{viégas2023modelusermodelexploring}
Fernanda Viégas and Martin Wattenberg.
\newblock The system model and the user model: Exploring ai dashboard design, 2023.
\newblock URL \url{https://arxiv.org/abs/2305.02469}.

\bibitem[Voita \& Titov(2020)Voita and Titov]{voita-titov-2020-information}
Elena Voita and Ivan Titov.
\newblock Information-theoretic probing with minimum description length.
\newblock In Bonnie Webber, Trevor Cohn, Yulan He, and Yang Liu (eds.), \emph{Proceedings of the 2020 Conference on Empirical Methods in Natural Language Processing (EMNLP)}, pp.\  183--196, Online, November 2020. Association for Computational Linguistics.
\newblock \doi{10.18653/v1/2020.emnlp-main.14}.
\newblock URL \url{https://aclanthology.org/2020.emnlp-main.14}.

\bibitem[Voita et~al.(2019)Voita, Talbot, Moiseev, Sennrich, and Titov]{voita-2019-analyzing}
Elena Voita, David Talbot, Fedor Moiseev, Rico Sennrich, and Ivan Titov.
\newblock Analyzing multi-head self-attention: Specialized heads do the heavy lifting, the rest can be pruned.
\newblock In Anna Korhonen, David Traum, and Llu{\'\i}s M{\`a}rquez (eds.), \emph{Proceedings of the 57th Annual Meeting of the Association for Computational Linguistics}, pp.\  5797--5808, Florence, Italy, July 2019. Association for Computational Linguistics.
\newblock \doi{10.18653/v1/P19-1580}.
\newblock URL \url{https://aclanthology.org/P19-1580}.

\bibitem[Voita et~al.(2024)Voita, Ferrando, and Nalmpantis]{voita-etal-2024-neurons}
Elena Voita, Javier Ferrando, and Christoforos Nalmpantis.
\newblock Neurons in large language models: Dead, n-gram, positional.
\newblock In Lun-Wei Ku, Andre Martins, and Vivek Srikumar (eds.), \emph{Findings of the Association for Computational Linguistics ACL 2024}, pp.\  1288--1301, Bangkok, Thailand and virtual meeting, August 2024. Association for Computational Linguistics.
\newblock URL \url{https://aclanthology.org/2024.findings-acl.75}.

\bibitem[Voss et~al.(2021)Voss, Cammarata, Goh, Petrov, Schubert, Egan, Lim, and Olah]{voss2021visualizing}
Chelsea Voss, Nick Cammarata, Gabriel Goh, Michael Petrov, Ludwig Schubert, Ben Egan, Swee~Kiat Lim, and Chris Olah.
\newblock Visualizing weights.
\newblock \emph{Distill}, 2021.
\newblock \doi{10.23915/distill.00024.007}.
\newblock https://distill.pub/2020/circuits/visualizing-weights.

\bibitem[Wang et~al.(2024{\natexlab{a}})Wang, Yue, Su, and Sun]{wang2024grokkedtransformersimplicitreasoners}
Boshi Wang, Xiang Yue, Yu~Su, and Huan Sun.
\newblock Grokked transformers are implicit reasoners: A mechanistic journey to the edge of generalization.
\newblock In \emph{ICML 2024 Workshop on Mechanistic Interpretability}, 2024{\natexlab{a}}.
\newblock URL \url{https://openreview.net/forum?id=ns8IH5Sn5y}.

\bibitem[Wang et~al.(2024{\natexlab{b}})Wang, Farrugia-Roberts, Hoogland, Carroll, Wei, and Murfet]{wang2024loss}
George Wang, Matthew Farrugia-Roberts, Jesse Hoogland, Liam Carroll, Susan Wei, and Daniel Murfet.
\newblock Loss landscape geometry reveals stagewise development of transformers.
\newblock In \emph{High-dimensional Learning Dynamics 2024: The Emergence of Structure and Reasoning}, 2024{\natexlab{b}}.
\newblock URL \url{https://openreview.net/forum?id=2JabyZjM5H}.

\bibitem[Wang et~al.(2023)Wang, Variengien, Conmy, Shlegeris, and Steinhardt]{Wang_2022_Circuits}
Kevin~Ro Wang, Alexandre Variengien, Arthur Conmy, Buck Shlegeris, and Jacob Steinhardt.
\newblock Interpretability in the wild: a circuit for indirect object identification in {GPT}-2 small.
\newblock In \emph{The Eleventh International Conference on Learning Representations}, 2023.
\newblock URL \url{https://openreview.net/forum?id=NpsVSN6o4ul}.

\bibitem[Wang et~al.(2024{\natexlab{c}})Wang, Zhu, Liu, Zheng, Chen, and Li]{wang2023knowledgeeditinglargelanguage}
Song Wang, Yaochen Zhu, Haochen Liu, Zaiyi Zheng, Chen Chen, and Jundong Li.
\newblock Knowledge editing for large language models: A survey.
\newblock \emph{ACM Comput. Surv.}, 57\penalty0 (3), November 2024{\natexlab{c}}.
\newblock ISSN 0360-0300.
\newblock \doi{10.1145/3698590}.
\newblock URL \url{https://doi.org/10.1145/3698590}.

\bibitem[Wang et~al.(2024{\natexlab{d}})Wang, Zhang, Guo, and Shen]{wang2024gradientbasedfeatureattribution}
Yongjie Wang, Tong Zhang, Xu~Guo, and Zhiqi Shen.
\newblock Gradient based feature attribution in explainable ai: A technical review, 2024{\natexlab{d}}.
\newblock URL \url{https://arxiv.org/abs/2403.10415}.

\bibitem[Ward et~al.(2023)Ward, Toni, Belardinelli, and Everitt]{ward2023honesty}
Francis Ward, Francesca Toni, Francesco Belardinelli, and Tom Everitt.
\newblock Honesty is the best policy: Defining and mitigating ai deception.
\newblock In A.~Oh, T.~Naumann, A.~Globerson, K.~Saenko, M.~Hardt, and S.~Levine (eds.), \emph{Advances in Neural Information Processing Systems}, volume~36, pp.\  2313--2341. Curran Associates, Inc., 2023.
\newblock URL \url{https://proceedings.neurips.cc/paper_files/paper/2023/file/06fc7ae4a11a7eb5e20fe018db6c036f-Paper-Conference.pdf}.

\bibitem[Warstadt et~al.(2019)Warstadt, Singh, and Bowman]{warstadt2019neuralnetworkacceptabilityjudgments}
Alex Warstadt, Amanpreet Singh, and Samuel~R. Bowman.
\newblock Neural network acceptability judgments.
\newblock \emph{Transactions of the Association for Computational Linguistics}, 7:\penalty0 625--641, 2019.
\newblock \doi{10.1162/tacl_a_00290}.
\newblock URL \url{https://aclanthology.org/Q19-1040/}.

\bibitem[Watanabe(2009)]{Watanabe_2009}
Sumio Watanabe.
\newblock \emph{Algebraic Geometry and Statistical Learning Theory}.
\newblock Cambridge Monographs on Applied and Computational Mathematics. Cambridge University Press, 2009.

\bibitem[Watkins(2023)]{Mwatkins_2023}
Matthew Watkins.
\newblock Mapping the semantic void: Strange goings-on in gpt embedding spaces, Dec 2023.
\newblock URL \url{https://www.lesswrong.com/posts/c6uTNm5erRrmyJvvD/mapping-the-semantic-void-strange-goings-on-in-gpt-embedding}.

\bibitem[Watson(2022)]{watson2022conceptual}
David~S. Watson.
\newblock Conceptual challenges for interpretable machine learning.
\newblock \emph{Synthese}, 200:\penalty0 65, 2022.
\newblock \doi{10.1007/s11229-022-03485-5}.
\newblock URL \url{https://doi.org/10.1007/s11229-022-03485-5}.

\bibitem[Wei et~al.(2015)Wei, Zhou, Torrabla, and Freeman]{wei2015understandingintraclassknowledgeinside}
Donglai Wei, Bolei Zhou, Antonio Torrabla, and William Freeman.
\newblock Understanding intra-class knowledge inside cnn, 2015.
\newblock URL \url{https://arxiv.org/abs/1507.02379}.

\bibitem[Wei et~al.(2022)Wei, Tay, Bommasani, Raffel, Zoph, Borgeaud, Yogatama, Bosma, Zhou, Metzler, Chi, Hashimoto, Vinyals, Liang, Dean, and Fedus]{wei2022emergentabilitieslargelanguage}
Jason Wei, Yi~Tay, Rishi Bommasani, Colin Raffel, Barret Zoph, Sebastian Borgeaud, Dani Yogatama, Maarten Bosma, Denny Zhou, Donald Metzler, Ed~H. Chi, Tatsunori Hashimoto, Oriol Vinyals, Percy Liang, Jeff Dean, and William Fedus.
\newblock Emergent abilities of large language models.
\newblock \emph{Transactions on Machine Learning Research}, 2022.
\newblock ISSN 2835-8856.
\newblock URL \url{https://openreview.net/forum?id=yzkSU5zdwD}.
\newblock Survey Certification.

\bibitem[Wei et~al.(2024)Wei, Wang, Schuurmans, Bosma, Ichter, Xia, Chi, Le, and Zhou]{wei2023chainofthoughtpromptingelicitsreasoning}
Jason Wei, Xuezhi Wang, Dale Schuurmans, Maarten Bosma, Brian Ichter, Fei Xia, Ed~H. Chi, Quoc~V. Le, and Denny Zhou.
\newblock Chain-of-thought prompting elicits reasoning in large language models.
\newblock In \emph{Proceedings of the 36th International Conference on Neural Information Processing Systems}, NIPS '22, Red Hook, NY, USA, 2024. Curran Associates Inc.
\newblock ISBN 9781713871088.

\bibitem[Wei et~al.(2023)Wei, Murfet, Gong, Li, Gell-Redman, and Quella]{Wei_2023_deeplearningsingular}
Susan Wei, Daniel Murfet, Mingming Gong, Hui Li, Jesse Gell-Redman, and Thomas Quella.
\newblock Deep learning is singular, and that’s good.
\newblock \emph{IEEE Transactions on Neural Networks and Learning Systems}, 34\penalty0 (12):\penalty0 10473--10486, 2023.
\newblock \doi{10.1109/TNNLS.2022.3167409}.

\bibitem[Weiss et~al.(2021)Weiss, Goldberg, and Yahav]{Weiss_2021_transformer}
Gail Weiss, Yoav Goldberg, and Eran Yahav.
\newblock Thinking like transformers.
\newblock In Marina Meila and Tong Zhang (eds.), \emph{Proceedings of the 38th International Conference on Machine Learning}, volume 139 of \emph{Proceedings of Machine Learning Research}, pp.\  11080--11090. PMLR, 18--24 Jul 2021.
\newblock URL \url{https://proceedings.mlr.press/v139/weiss21a.html}.

\bibitem[Wongpiromsarn et~al.(2023)Wongpiromsarn, Ghasemi, Cubuktepe, Bakirtzis, Carr, Karabag, Neary, Gohari, and Topcu]{Wongpiromsarn_2023}
Tichakorn Wongpiromsarn, Mahsa Ghasemi, Murat Cubuktepe, Georgios Bakirtzis, Steven Carr, Mustafa~O. Karabag, Cyrus Neary, Parham Gohari, and Ufuk Topcu.
\newblock Formal methods for autonomous systems.
\newblock \emph{Foundations and Trends® in Systems and Control}, 10\penalty0 (3–4):\penalty0 180–407, 2023.
\newblock ISSN 2325-6826.
\newblock \doi{10.1561/2600000029}.
\newblock URL \url{http://dx.doi.org/10.1561/2600000029}.

\bibitem[Wortsman et~al.(2022)Wortsman, Ilharco, Gadre, Roelofs, Gontijo-Lopes, Morcos, Namkoong, Farhadi, Carmon, Kornblith, and Schmidt]{wortsman_2022_ModelSoups}
Mitchell Wortsman, Gabriel Ilharco, Samir~Ya Gadre, Rebecca Roelofs, Raphael Gontijo-Lopes, Ari~S Morcos, Hongseok Namkoong, Ali Farhadi, Yair Carmon, Simon Kornblith, and Ludwig Schmidt.
\newblock Model soups: averaging weights of multiple fine-tuned models improves accuracy without increasing inference time.
\newblock In Kamalika Chaudhuri, Stefanie Jegelka, Le~Song, Csaba Szepesvari, Gang Niu, and Sivan Sabato (eds.), \emph{Proceedings of the 39th International Conference on Machine Learning}, volume 162 of \emph{Proceedings of Machine Learning Research}, pp.\  23965--23998. PMLR, 17--23 Jul 2022.
\newblock URL \url{https://proceedings.mlr.press/v162/wortsman22a.html}.

\bibitem[Wu et~al.(2024{\natexlab{a}})Wu, Jaburi, Drori, and Gross]{wu2024unifyingverifyingmechanisticinterpretations}
Wilson Wu, Louis Jaburi, Jacob Drori, and Jason Gross.
\newblock Unifying and verifying mechanistic interpretations: A case study with group operations, 2024{\natexlab{a}}.
\newblock URL \url{https://arxiv.org/abs/2410.07476}.

\bibitem[Wu et~al.(2023)Wu, Geiger, Icard, Potts, and Goodman]{wu2024interpretabilityscaleidentifyingcausal}
Zhengxuan Wu, Atticus Geiger, Thomas Icard, Christopher Potts, and Noah Goodman.
\newblock Interpretability at scale: Identifying causal mechanisms in alpaca.
\newblock In \emph{Thirty-seventh Conference on Neural Information Processing Systems}, 2023.
\newblock URL \url{https://openreview.net/forum?id=nRfClnMhVX}.

\bibitem[Wu et~al.(2024{\natexlab{b}})Wu, Arora, Wang, Geiger, Jurafsky, Manning, and Potts]{wu2024reftrepresentationfinetuninglanguage}
Zhengxuan Wu, Aryaman Arora, Zheng Wang, Atticus Geiger, Dan Jurafsky, Christopher~D Manning, and Christopher Potts.
\newblock Re{FT}: Representation finetuning for language models.
\newblock In \emph{The Thirty-eighth Annual Conference on Neural Information Processing Systems}, 2024{\natexlab{b}}.
\newblock URL \url{https://openreview.net/forum?id=fykjplMc0V}.

\bibitem[Wu et~al.(2024{\natexlab{c}})Wu, Geiger, Huang, Arora, Icard, Potts, and Goodman]{wu2024replymakelovetal}
Zhengxuan Wu, Atticus Geiger, Jing Huang, Aryaman Arora, Thomas Icard, Christopher Potts, and Noah~D. Goodman.
\newblock A reply to makelov et al. (2023)'s "interpretability illusion" arguments, 2024{\natexlab{c}}.
\newblock URL \url{https://arxiv.org/abs/2401.12631}.

\bibitem[Wynroe \& Sharkey(2024)Wynroe and Sharkey]{wynroe2024decomposing}
Keith Wynroe and Lee Sharkey.
\newblock Decomposing the {QK} circuit with bilinear sparse dictionary learning.
\newblock \emph{AI Alignment Forum}, jul 2024.
\newblock URL \url{https://www.lesswrong.com/posts/2ep6FGjTQoGDRnhrq/decomposing-the-qk-circuit-with-bilinear-sparse-dictionary}.

\bibitem[Xin et~al.(2022)Xin, Zhong, Chen, Takagi, Seltzer, and Rudin]{Xin_2022_SparseDecisionTrees}
Rui Xin, Chudi Zhong, Zhi Chen, Takuya Takagi, Margo Seltzer, and Cynthia Rudin.
\newblock Exploring the whole rashomon set of sparse decision trees.
\newblock In Alice~H. Oh, Alekh Agarwal, Danielle Belgrave, and Kyunghyun Cho (eds.), \emph{Advances in Neural Information Processing Systems}, 2022.
\newblock URL \url{https://openreview.net/forum?id=WHqVVk3UHr}.

\bibitem[Ye \& Durrett(2022)Ye and Durrett]{Ye_2022_unreliability}
Xi~Ye and Greg Durrett.
\newblock The unreliability of explanations in few-shot prompting for textual reasoning.
\newblock In S.~Koyejo, S.~Mohamed, A.~Agarwal, D.~Belgrave, K.~Cho, and A.~Oh (eds.), \emph{Advances in Neural Information Processing Systems}, volume~35, pp.\  30378--30392. Curran Associates, Inc., 2022.
\newblock URL \url{https://proceedings.neurips.cc/paper_files/paper/2022/file/c402501846f9fe03e2cac015b3f0e6b1-Paper-Conference.pdf}.

\bibitem[Yedidia(2023)]{Yedidia_2023}
Adam Yedidia.
\newblock Gpt-2’s positional embedding matrix is a helix, July 2023.
\newblock URL \url{https://www.lesswrong.com/posts/qvWP3aBDBaqXvPNhS/gpt-2-s-positional-embedding-matrix-is-a-helix}.

\bibitem[Yom~Din et~al.(2024)Yom~Din, Karidi, Choshen, and Geva]{yom-din-etal-2024-jump}
Alexander Yom~Din, Taelin Karidi, Leshem Choshen, and Mor Geva.
\newblock Jump to conclusions: Short-cutting transformers with linear transformations.
\newblock In Nicoletta Calzolari, Min-Yen Kan, Veronique Hoste, Alessandro Lenci, Sakriani Sakti, and Nianwen Xue (eds.), \emph{Proceedings of the 2024 Joint International Conference on Computational Linguistics, Language Resources and Evaluation (LREC-COLING 2024)}, pp.\  9615--9625, Torino, Italia, May 2024. ELRA and ICCL.
\newblock URL \url{https://aclanthology.org/2024.lrec-main.840}.

\bibitem[Yosinski et~al.(2015)Yosinski, Clune, Nguyen, Fuchs, and Lipson]{yosinski2015understandingneuralnetworksdeep}
Jason Yosinski, Jeff Clune, Anh~Mai Nguyen, Thomas~J. Fuchs, and Hod Lipson.
\newblock Understanding neural networks through deep visualization.
\newblock \emph{CoRR}, abs/1506.06579, 2015.
\newblock URL \url{http://arxiv.org/abs/1506.06579}.

\bibitem[Yu et~al.(2024)Yu, Cao, Cheung, and Dong]{yu2024mechanisticunderstandingmitigationlanguage}
Lei Yu, Meng Cao, Jackie~CK Cheung, and Yue Dong.
\newblock Mechanistic understanding and mitigation of language model non-factual hallucinations.
\newblock In Yaser Al-Onaizan, Mohit Bansal, and Yun-Nung Chen (eds.), \emph{Findings of the Association for Computational Linguistics: EMNLP 2024}, pp.\  7943--7956, Miami, Florida, USA, November 2024. Association for Computational Linguistics.
\newblock \doi{10.18653/v1/2024.findings-emnlp.466}.
\newblock URL \url{https://aclanthology.org/2024.findings-emnlp.466/}.

\bibitem[Yu et~al.(2023)Yu, Merullo, and Pavlick]{yu-etal-2023-characterizing}
Qinan Yu, Jack Merullo, and Ellie Pavlick.
\newblock Characterizing mechanisms for factual recall in language models.
\newblock In Houda Bouamor, Juan Pino, and Kalika Bali (eds.), \emph{Proceedings of the 2023 Conference on Empirical Methods in Natural Language Processing}, pp.\  9924--9959, Singapore, December 2023. Association for Computational Linguistics.
\newblock \doi{10.18653/v1/2023.emnlp-main.615}.
\newblock URL \url{https://aclanthology.org/2023.emnlp-main.615}.

\bibitem[Yun et~al.(2021)Yun, Chen, Olshausen, and LeCun]{yun2023transformervisualizationdictionarylearning}
Zeyu Yun, Yubei Chen, Bruno Olshausen, and Yann LeCun.
\newblock Transformer visualization via dictionary learning: contextualized embedding as a linear superposition of transformer factors.
\newblock In Eneko Agirre, Marianna Apidianaki, and Ivan Vuli{\'c} (eds.), \emph{Proceedings of Deep Learning Inside Out (DeeLIO): The 2nd Workshop on Knowledge Extraction and Integration for Deep Learning Architectures}, pp.\  1--10, Online, June 2021. Association for Computational Linguistics.
\newblock \doi{10.18653/v1/2021.deelio-1.1}.
\newblock URL \url{https://aclanthology.org/2021.deelio-1.1/}.

\bibitem[Zeiler \& Fergus(2014)Zeiler and Fergus]{Zeiler_2013_VisualizingConvolutionalNetworks}
Matthew~D. Zeiler and Rob Fergus.
\newblock Visualizing and understanding convolutional networks.
\newblock In David Fleet, Tomas Pajdla, Bernt Schiele, and Tinne Tuytelaars (eds.), \emph{Computer Vision -- ECCV 2014}, pp.\  818--833, Cham, 2014. Springer International Publishing.
\newblock ISBN 978-3-319-10590-1.

\bibitem[Zhang et~al.(2024)Zhang, Lepori, and Pavlick]{zhang2024instillinginductivebiasessubnetworks}
Enyan Zhang, Michael~A. Lepori, and Ellie Pavlick.
\newblock Instilling inductive biases with subnetworks, 2024.
\newblock URL \url{https://arxiv.org/abs/2310.10899}.

\bibitem[Zhang et~al.(2020)Zhang, Wang, Shen, Ji, Luo, and Wang]{zhang2019interpretabledeeplearning}
Xinyang Zhang, Ningfei Wang, Hua Shen, Shouling Ji, Xiapu Luo, and Ting Wang.
\newblock Interpretable deep learning under fire.
\newblock In \emph{Proceedings of the USENIX Security Symposium (Security)}, 2020.

\bibitem[Zhou et~al.(2015)Zhou, Khosla, Àgata Lapedriza, Oliva, and Torralba]{zhou2015objectdetectorsemergedeep}
Bolei Zhou, Aditya Khosla, Àgata Lapedriza, Aude Oliva, and Antonio Torralba.
\newblock Object detectors emerge in deep scene cnns.
\newblock In \emph{ICLR}, 2015.
\newblock URL \url{http://arxiv.org/abs/1412.6856}.

\bibitem[Zhou et~al.(2022)Zhou, Booth, Ribeiro, and Shah]{Zhou_2022_featureattribution}
Yilun Zhou, Serena Booth, Marco~Tulio Ribeiro, and Julie Shah.
\newblock Do feature attribution methods correctly attribute features?
\newblock \emph{Proceedings of the AAAI Conference on Artificial Intelligence}, 36\penalty0 (9):\penalty0 9623--9633, Jun. 2022.
\newblock \doi{10.1609/aaai.v36i9.21196}.
\newblock URL \url{https://ojs.aaai.org/index.php/AAAI/article/view/21196}.

\bibitem[Zimmermann et~al.(2021)Zimmermann, Borowski, Geirhos, Bethge, Wallis, and Brendel]{zimmermann2021how}
Roland~Simon Zimmermann, Judy Borowski, Robert Geirhos, Matthias Bethge, Thomas S.~A. Wallis, and Wieland Brendel.
\newblock How well do feature visualizations support causal understanding of {CNN} activations?
\newblock In A.~Beygelzimer, Y.~Dauphin, P.~Liang, and J.~Wortman Vaughan (eds.), \emph{Advances in Neural Information Processing Systems}, 2021.
\newblock URL \url{https://openreview.net/forum?id=vLPqnPf9k0}.

\bibitem[Zou et~al.(2023{\natexlab{a}})Zou, Phan, Chen, Campbell, Guo, Ren, Pan, Yin, Mazeika, Dombrowski, Goel, Li, Byun, Wang, Mallen, Basart, Koyejo, Song, Fredrikson, Kolter, and Hendrycks]{zou2023representationengineeringtopdownapproach}
Andy Zou, Long Phan, Sarah Chen, James Campbell, Phillip Guo, Richard Ren, Alexander Pan, Xuwang Yin, Mantas Mazeika, Ann-Kathrin Dombrowski, Shashwat Goel, Nathaniel Li, Michael~J. Byun, Zifan Wang, Alex Mallen, Steven Basart, Sanmi Koyejo, Dawn Song, Matt Fredrikson, J.~Zico Kolter, and Dan Hendrycks.
\newblock Representation engineering: A top-down approach to ai transparency, 2023{\natexlab{a}}.
\newblock URL \url{https://arxiv.org/abs/2310.01405}.

\bibitem[Zou et~al.(2023{\natexlab{b}})Zou, Wang, Carlini, Nasr, Kolter, and Fredrikson]{zou2023universaltransferableadversarialattacks}
Andy Zou, Zifan Wang, Nicholas Carlini, Milad Nasr, J.~Zico Kolter, and Matt Fredrikson.
\newblock Universal and transferable adversarial attacks on aligned language models, 2023{\natexlab{b}}.
\newblock URL \url{https://arxiv.org/abs/2307.15043}.

\bibitem[Zou et~al.(2024)Zou, Phan, Wang, Duenas, Lin, Andriushchenko, Wang, Kolter, Fredrikson, and Hendrycks]{zou2024improvingalignmentrobustnesscircuit}
Andy Zou, Long Phan, Justin Wang, Derek Duenas, Maxwell Lin, Maksym Andriushchenko, Rowan Wang, Zico Kolter, Matt Fredrikson, and Dan Hendrycks.
\newblock Improving alignment and robustness with circuit breakers, 2024.
\newblock URL \url{https://arxiv.org/abs/2406.04313}.

\end{thebibliography}

\addtocontents{toc}{\protect\setcounter{tocdepth}{-1}}
\appendix
\section{Summary of open questions}
\label{sec:summary-of-open-questions}

\subsection{Open problems in mechanistic interpretability methods and foundations}

\subsubsection{Reverse engineering: Identifying the roles of network components}

\parasection{Reverse engineering step 1: Neural network decomposition}

\begin{enumerate}
    \item How should we decompose networks into more interpretable constituent parts?
    \begin{enumerate}[label=\alph*.]
        \item What isomorphism or what approximation of a neural network (or parts of it) is the best way to express it for the purposes of interpreting it?
        \item How should we coarse grain neural networks?
        \item How should we build higher level abstractions on top of low-level network components?
    \end{enumerate}
    
    \item How true is the linear representation hypothesis?
    \begin{enumerate}[label=\alph*.]
        \item To what extent do models encode concepts linearly in their representations?
        \item How should we characterize representations that are not linearly represented in neural networks?
        \item What properties of a concept, or of the training distribution, result in a particular concept becoming encoded linearly (or not)?
    \end{enumerate}
    
    \item Is the combination of the linear representation hypothesis and superposition the right frame for thinking about computation in neural networks?
    \begin{enumerate}[label=\alph*.]
        \item Can we fully determine the causes of feature superposition and polysemanticity within neural networks?
        \item How should we understand superposition in attention blocks?
        \item How should we understand cross-layer superposition?
        \item What new theoretical insights can be gleaned from considering how networks perform computation natively in superposition, rather than treating superposition purely as a compression strategy?
    \end{enumerate}
    
    \item Can the problems with SDL be overcome?
    \begin{enumerate}[label=\alph*.]
        \item What lies in SDL reconstruction errors? Will the errors converge to zero with methodological progress?
        \item Is sparsity the correct proxy for interpretability?
        \item Can the approach be scaled to the largest models?
        \item Does SDL make sense if we don't believe in the linear representation hypothesis?
        \item Is sparsity the best possible proxy for interpretability?
        \item Is it correct to think of SDL features as `bags of features', or is there important information contained within the geometry of representation space?
        \item If SDL finds compositions of the “true” features, is this a problem?
        \item Can SDL be applied to all architectural components to fully decompose networks?
        \item How can we better measure the success of SDL techniques?
        \item How should we connect sparsely activating features into circuits? Will this be practically possible, and the best possible description of network mechanisms?
        \item Can we develop new methods that address conceptual and practical issues with SDL?
    \end{enumerate}
    
    \item How important is the geometry of activation space for explaining neural network behavior?
    \begin{enumerate}[label=\alph*.]
        \item How can we identify the underlying functional structure of networks (which defines why activations are located in particular geometric arrangements in activation space)?
        \item Must we understand global feature geometry or only local feature geometry in order to understand computation in neural networks?
    \end{enumerate}
    
    \item Can we connect theories for how neural networks generalize to interpretability?
    \begin{enumerate}[label=\alph*.]
        \item Can we distinguish parts of networks that underlie generalization from parts that underlie memorization?
        \item What mechanisms underlie the relationship between interpretability and generalization?
        \item Are there connections between adversarial robustness and superposition?
        \item Can we connect interpretability to theories of deep learning like SLT?
    \end{enumerate}
    
    \item Can we build intrinsically more interpretable models at low performance cost? How helpful is this?
    \begin{enumerate}[label=\alph*.]
        \item Interpretability training: Can we train networks that are interpretable by default at low performance cost?
        \item Interpretable inference: Can we convert already-trained models into forms that are much easier to completely interpret at little performance cost?
        \item How can we train large-scale models such that the concepts they use are naturally understandable to humans?
        \item How can we design training objectives so that the model is incentivized to use known specific abstractions?
        \item How can we localize concepts we want to control (e.g., long-term plans) during training?
    \end{enumerate}
\end{enumerate}

\parasection{Reverse engineering step 2: Identifying the functional role of components}
\begin{enumerate}
    \item Can we improve on max-activating input data set examples for understanding the causes of network component activations?
    \begin{enumerate}[label=\alph*.]
        \item How can we avoid imposing human bias to explanations?
        \item Can we progress toward deeper descriptions based on internal mechanisms?
        \item How might we develop interpretation methods that can recognize and work with unfamiliar concepts - computational patterns that don't map cleanly to human intuitions?
    \end{enumerate}
    
    \item How can we develop attribution methods that faithfully and efficiently compute which network components are important for some downstream metric?
    \begin{enumerate}[label=\alph*.]
        \item How can we develop attribution methods that capture higher-order effects beyond first-order approximations of model behavior?
        \item Is it possible to create perturbation-based methods that don't force models to operate outside their training distribution?
        \item Can we develop hybrid approaches that combine the strengths of different attribution methods while mitigating their individual weaknesses?
    \end{enumerate}
    
    \item How can we better measure the downstream effects of model components?
    \begin{enumerate}[label=\alph*.]
        \item How can we reliably distinguish between true causal pathways and compensatory effects like the “Hydra effect” when performing interventions?
    \end{enumerate}
\end{enumerate}

\parasection{Reverse engineering step 3: Validation of descriptions}
\begin{enumerate}
    \item Can we improve our ability to validate mechanistic explanations for model behavior in ways that do not depend on researcher intuition and are computationally tractable to use?
    \begin{enumerate}[label=\alph*.]
        \item Can we improve on methodologies for evaluating hypotheses through their predictive power on activations of network components?
        \item Can we develop methodologies for evaluating hypotheses through their predictive power for model behavior (e.g. unusual failures or adv examples)?
        \item Can we develop methodologies for handcoding weights that faithfully represent some hypotheses, as a drop in replacement for subnetworks we claim to understand?
        \item Can we develop a wider suite of networks with known ground truth explanations to validate techniques against?
        \item Can we use mechanistic explanations to achieve engineering goals?
        \item Can we use mechanistic explanations to achieve engineering goals in a way that improves upon black box baselines?
    \end{enumerate}
    
    \item Can we develop “model organisms” as a community, which are understood deeply, and seen as a test-bed for new unproven interpretability methodologies to be tested?
    
    \item Can we establish standardized baselines and benchmarks for comparing different interpretability approaches on real-world, non-cherry-picked tasks, where the ground truth is known?
    
    \item What would constitute a comprehensive set of “stress tests” for interpretability hypotheses that could reliably detect interpretability illusions?
    
    \item How might we design evaluation frameworks that assess interpretability methods on their average case and worst-case performance rather than just best-case scenarios?
    
    \item How can we ensure that our understanding of internals generalizes to out-of-distribution inputs?
\end{enumerate}

\subsubsection{Concept-based interpretability: Identifying network components for given roles}
\begin{enumerate}
    \item How can we reliably distinguish causal from merely correlated features when probing neural networks?
    
    \item Can we develop automated systems to generate high-quality probing data sets, reducing the current heavy reliance on human effort?
    
    \item What regularization and validation techniques can be used to prevent spurious correlations while ensuring probes find generalizable features?
    
    \item How can we improve probing for concepts that may not have clear positive/negative examples?
\end{enumerate}

\subsubsection{Proceduralizing mechanistic interpretability into circuit discovery pipelines}
\begin{enumerate}
    \item Can we develop techniques that build on lower level methods that provide deeper or more complete insights about neural networks?
    \item How much can we learn from further work in the existing circuit discovery paradigm?
    \begin{enumerate}[label=\alph*.]
        \item Should we expect circuit discovery to benefit from further methodological progress in decomposing neural networks? Will faithfulness go up and explanation description length go down?
        \item Can we remove the constraint that task definition for circuit discovery is inherently concept-based, which may be complicating mechanistic analysis?
        \item Can we get around the practical issues of negative and backup behavior?
        \item Will circuit discovery provide insights into arbitrary tasks, or will it only be helpful in cases where we are able to crisply define tasks?
    \end{enumerate}
\end{enumerate}

\subsubsection{Automating steps in mechanistic interpretability research}
\begin{enumerate}
    \item Can we improve on AI automated feature description and validation methods?
    \begin{enumerate}[label=\alph*.]
        \item Through automating the generation and testing of arbitrary hypotheses?
        \item Through describing differences between features?
        \item Through descriptions of how components interact?
    \end{enumerate}
    
    \item Can we improve on ACDC-like circuit discovery methods?
    
    \item Can we automate other parts of the mechanistic interpretability pipeline?
    \begin{enumerate}[label=\alph*.]
        \item Conceptual interpretability research?
        \item Decomposition method discovery?
        \item More ad hoc validation of hypotheses?
    \end{enumerate}
    
    \item Should we take steps to mitigate potentially misaligned AI systems sabotaging AI automated interpretability?
\end{enumerate}

\subsection{Open problems in applications of mechanistic interpretability}

\subsubsection{Using mechanistic interpretability for better monitoring and auditing of AI systems for potentially unsafe cognition}
\begin{enumerate}
    \item Can we effectively use interpretability for safety evaluations?
    \begin{enumerate}[label=\alph*.]
        \item Can we develop robust “white box” evaluations that detect concerning internal patterns without needing to understand the entire network?
        \item Can we reliably distinguish between features that merely recognize deceptive behavior versus mechanisms that generate deceptive behavior?
        \item How can we validate that learned features capture all concerning patterns of reasoning?
        \item Can we reliably identify which features are appropriately versus spuriously relevant to a given task?
    \end{enumerate}
    
    \item Can we leverage interpretability to enhance red-teaming and system testing?
    \begin{enumerate}[label=\alph*.]
        \item Can we use interpretability insights to make red-teaming more efficient than current methods?
        \item How can we best use feature attribution to help human red-teamers identify problematic input patterns?
    \end{enumerate}
    
    \item Can we develop effective test-time monitoring systems based on interpretability?
    \begin{enumerate}[label=\alph*.]
        \item Can we get mechanistic anomaly detection to work?
        \item Can we create passive monitoring systems based on model internals that effectively flag concerning internal patterns during deployment?
        \item Can we develop monitoring systems that work with only feature-level understanding rather than requiring deep mechanical insights?
    \end{enumerate}
\end{enumerate}

\subsubsection{Using mechanistic interpretability for better control of AI system behavior}
\begin{enumerate}
    \item Can we improve steering methods through interpretability?
    \begin{enumerate}[label=\alph*.]
        \item How can we make activation steering more precise and reduce its side effects?
        \item Can we develop methods to steer entire mechanisms rather than just single features?
    \end{enumerate}
    
    \item Can we achieve reliable model unlearning and editing?
    \begin{enumerate}[label=\alph*.]
        \item Will carving the network at its true joints help us improve on model unlearning and editing?
        \item Can mechanistic interpretability help us develop better methods for evaluating unlearning efficacy?
        \item Can mechanistic interpretability help us determine which classes of model edit are even possible, without damaging generalization in undesirable ways?
    \end{enumerate}
    
    \item Can we better understand and improve finetuning through interpretability?
    \begin{enumerate}[label=\alph*.]
        \item Can we make finetuning more sample-efficient by targeting specific parameters?
        \item Can we develop better tools for analyzing feature-level or mechanism-level differences between model versions?
    \end{enumerate}
\end{enumerate}

\subsubsection{Using mechanistic interpretability for better predictions about AI systems}
\begin{enumerate}
    \item Can we predict model behavior in novel situations outside of the distribution of inputs we have access to with mechanistic understanding?
    \begin{enumerate}[label=\alph*.]
        \item Can we reliably predict when and how jailbreaking or safety bypasses might occur?
        \item How can we identify internal signatures that predict specific failure modes like hallucination?
        \item Can we develop methods to predict model behavior without requiring behavioral evaluations?
        \item Can we find “values” or “goals” in systems that might be indicative of behavior in more generality?
        \item Is it possible to prove the absence of specific dangerous capabilities through mechanistic analysis?
    \end{enumerate}
    
    \item Can we develop formal verification methods for AI systems?
    \begin{enumerate}[label=\alph*.]
        \item Can current toy model verification approaches scale to frontier systems?
        \item How much of neural computation can be reduced to verifiable symbolic operations?
        \item Can we create formal guarantees about system behavior in complex, non-formalizable environments?
        \item What level of mechanistic understanding is necessary for meaningful formal verification?
    \end{enumerate}
    
    \item Can we make rigorous claims about model safety?
    \begin{enumerate}[label=\alph*.]
        \item Can we definitively prove the absence of specific dangerous mechanisms?
        \item How can we verify claims about model values and goals in a rigorous way?
        \item What types of safety claims are possible with current interpretability methods?
        \item Can we develop “enumerative safety” approaches that reliably identify all relevant mechanisms?
    \end{enumerate}
    
    \item Can we better predict AI capability development through interpretability?
    \begin{enumerate}[label=\alph*.]
        \item Can we identify early signatures that predict emergent capabilities?
        \item How do model mechanisms evolve dynamically during training?
        \item Can we map the connection between small-scale circuits and large-scale capabilities?
        \item How does the loss landscape's structure relate to capability emergence?
    \end{enumerate}
    
    \item Can we understand the relationship between training data and capabilities?
    \begin{enumerate}[label=\alph*.]
        \item How do specific training examples influence the development of model mechanisms?
        \item Can we predict model limitations based on training data composition?
        \item Can we design training data sets to reliably produce specific desired capabilities?
        \item How does data set structure affect the balance between in-context and weights-based learning?
    \end{enumerate}
    
    \item Can we predict latent or maskable capabilities?
    \begin{enumerate}[label=\alph*.]
        \item How can we identify capabilities that could be `unlocked' through prompting or finetuning?
        \item Can we detect when finetuning has masked rather than removed capabilities?
        \item How do we analyze mechanisms that span multiple timesteps or sequential behaviors?
        \item Can we predict which model capabilities are fundamental versus superficially trained?
    \end{enumerate}
\end{enumerate}

\subsubsection{Using mechanistic interpretability to improve our ability to perform inference, improve training and make use of learned representations}
\begin{enumerate}
    \item Can we use interpretability to make inference more efficient?
    \begin{enumerate}[label=\alph*.]
        \item How can we identify skippable computations without affecting outputs?
        \item Can we create more effective distillation methods through mechanistic understanding?
        \item How can we optimize model architecture based on component function analysis?
        \item Can we identify and optimize critical computational pathways?
    \end{enumerate}
    
    \item Can we improve training through mechanistic insights?
    \begin{enumerate}[label=\alph*.]
        \item Can we better select training data by understanding example influence?
        \item How can we monitor and optimize capability emergence during training?
        \item Can we develop more parameter-efficient training methods through component analysis?
        \item Can we create better architectures through component understanding?
        \item Can we identify and enhance components with specific functionalities?
    \end{enumerate}
    
    \item Can we instill capabilities directly into networks?
    \begin{enumerate}[label=\alph*.]
        \item Can we design better inductive biases based on mechanistic insights?
        \item Is it possible to create modular architectures with swappable components?
        \item Can we develop reliable methods for combining model parameters?
        \item Is it possible to transfer specific capabilities between models?
    \end{enumerate}
\end{enumerate}

\subsubsection{Using mechanistic interpretability for 'microscope AI'}
\begin{enumerate}
    \item Can we leverage AI models for scientific discovery?
    \begin{enumerate}[label=\alph*.]
        \item How can we extract novel patterns and predictors that models have found?
        \item Can we make microscope AI techniques accessible to domain experts?
        \item How do we validate scientific insights derived from model interpretability?
        \item Can we extend microscope AI beyond current simple correlational discoveries?
    \end{enumerate}
    
    \item Can we develop better knowledge extraction methods?
    \begin{enumerate}[label=\alph*.]
        \item How can we detect when models have found genuinely novel patterns?
        \item Can we automate the process of finding scientific insights in model weights?
        \item How do we bridge the gap between model features and scientific concepts?
        \item Can we make these techniques usable without deep machine learning expertise?
    \end{enumerate}
\end{enumerate}

\subsubsection{Mechanistic interpretability on a broader range of models and model families}
\begin{enumerate}
    \item Can interpretability methods generalize across architectures?
    \begin{enumerate}[label=\alph*.]
        \item Do current interpretability methods (SDL, circuit analysis) transfer to SSMs? Or, like the transition from CNNs to transformers, are new approaches necessary?
        \item Which insights are model-specific versus universal?
        \item How can we adapt methods for multimodal models?
    \end{enumerate}
    
    \item How do different models trained on similar data compare mechanistically?
    \begin{enumerate}[label=\alph*.]
        \item Is the “universality hypothesis” true across models? To what extent do neural networks learn similar features and circuits to each other (and to humans?)
        \item Do different architectures learn fundamentally different features?
        \item How do mechanisms of particular tasks differ between transformers, CNNs, and SSMs?
        \item Are there insights we can gain from comparing architectures?
    \end{enumerate}
    
    \item Can we future-proof interpretability research?
    \begin{enumerate}[label=\alph*.]
        \item How can we prepare for interpreting novel architectures?
        \item Should we focus on architecture-specific or general methods?
        \item Can we identify truly fundamental interpretability principles?
        \item Will current methods work on future frontier models?
    \end{enumerate}
\end{enumerate}

\subsubsection{Human computer interaction with model internals}
\begin{enumerate}
    \item Can we create interfaces that use mechanistic understanding to enhance human-neural network interaction?
    \begin{enumerate}[label=\alph*.]
        \item How can we visualize model internals in an intuitive way?
        \item Can we develop real-time interpretability dashboards?
        \item What's the right balance between simplicity and depth in these interfaces?
        \item How do we make complex model mechanisms understandable to non-experts?
    \end{enumerate}
    
    \item Can we develop interpretability tools to help auditors?
    \begin{enumerate}[label=\alph*.]
        \item How can we help auditors find potential failure modes directly?
        \item Can we develop tools to detect bias at the mechanism level?
        \item What interfaces would make auditing more efficient and thorough?
        \item How can we present technical findings to policy makers?
    \end{enumerate}
    
    \item Can we improve end-user interaction with AI?
    \begin{enumerate}[label=\alph*.]
        \item How can transparency features help users calibrate trust?
        \item Can we create intuitive controls based on model mechanisms?
        \item Can we create intuitive ways to steer model behavior?
    \end{enumerate}
\end{enumerate}

\subsubsection{Governance}
\begin{enumerate}
    \item Can mechanistic analysis help identify and prevent failures?
    \begin{enumerate}[label=\alph*.]
        \item Can we identify specific mechanisms that caused AI failures?
        \item How do we map the causal chain of mechanisms leading to incidents?
        \item Can we detect when similar mechanisms are about to activate?
        \item Is it possible to isolate and modify failure-causing mechanisms?
    \end{enumerate}
    
    \item Can we study mechanism patterns related to governance?
    \begin{enumerate}[label=\alph*.]
        \item Can we identify mechanisms responsible for specific dangerous capabilities?
        \item How do we detect deceptive or evasive mechanisms?
        \item Can we map the mechanisms involved in model decision-making?
        \item Is it possible to verify the absence of specific harmful mechanisms?
    \end{enumerate}
    
    \item Can mechanistic insights verify compliance?
    \begin{enumerate}[label=\alph*.]
        \item How can we trace decision mechanisms to explain model outputs?
        \item Can we identify mechanisms that process copyrighted content?
        \item Is it possible to detect mechanisms that encode specific knowledge?
        \item How do we verify modifications to problematic mechanisms?
    \end{enumerate}
\end{enumerate}

\subsubsection{Open socio-technical problems in mechanistic interpretability}

\parasection{Translating technical progress in mechanistic interpretability into levers for AI policy and governance}
\begin{enumerate}
    \item Can we use a mechanistic understanding to better evaluate AI capabilities?
    \begin{enumerate}[label=\alph*.]
        \item How can we use interpretability to improve capability elicitation?
        \item Can we use interpretability to reliably detect when models are strategically underperforming capabilities evaluations?
    \end{enumerate}
    
    \item Can we use a mechanistic understanding to improve our ability to forecast when or whether new capabilities will arise ahead of time?

    \item How can we use interpretability to better estimate the likelihood of different threat models?

    \item Can we use interpretability to prevent AI incidents?
    \begin{enumerate}[label=\alph*.]
        \item Can we use interpretability to construct reliable test-time monitors to detect AI incidents?
        \item Can we use reliably prevent similar incidents in the future, by using interpretability to design new evaluation tasks on incident scenarios?
    \end{enumerate}
    \item Can interpretability help verify which workloads GPUs are being used for?
    \item How should interpretability inform copyright law?
    \item How can mechanistic understanding help resolve copyright challenges in generative AI, particularly regarding the detection and removal of memorized copyrighted works?
\end{enumerate}

\parasection{Social and philosophical challenges in mechanistic interpretability}
\begin{enumerate}
    \item What is interpretability?
    \begin{enumerate}[label=\alph*.]
        \item What are the goals of the field?
        \item How should success be graded?
        \item Should we treat interpretability as a science or an engineering discipline? What implications does this have on what research should be done?
    \end{enumerate}
    
    \item How can we mitigate downside risks of interpretability research?
    \begin{enumerate}[label=\alph*.]
        \item How can we communicate the results of our research such that the risk of their misuse is minimized?
    \end{enumerate}
\end{enumerate}

\end{document}